%% file: main.tex
\PassOptionsToPackage{table}{xcolor}
\documentclass[sigconf]{acmart}
\newif\ifarxiv
\arxivtrue

\usepackage{balance} 

\AtBeginDocument{%
  }

\copyrightyear{2026}
\acmYear{2026}
\setcopyright{cc}
\setcctype{by}
\acmConference[KDD 2026] {Proceedings of the 32nd ACM SIGKDD Conference on Knowledge Discovery and Data Mining V.2}{August 9--13, 2026}{Jeju Island, Republic of Korea.}
\acmBooktitle{Proceedings of the 32nd ACM SIGKDD Conference on Knowledge Discovery and Data Mining V.2 (KDD 2026), August 9--13, 2026, Jeju Island, Republic of Korea}
\acmISBN{979-8-4007-2259-2/2026/08}
\acmDOI{10.1145/3770855.3817551}
\acmSubmissionID{v2dtb413}
\usepackage[utf8]{inputenc} 
\usepackage{amsfonts}       
\usepackage{nicefrac}       
\usepackage{array}

\usepackage{wrapfig}
\usepackage{multirow}
\usepackage{makecell}
\usepackage{enumitem}       
\usepackage{subcaption}
\usepackage{caption}
\usepackage{mathtools}
\usepackage{rotating}
\usepackage{caption}
\usepackage{subcaption}
\usepackage{bm}
\usepackage{appendix}
\usepackage{perpage} 
\MakePerPage{footnote} 
\usepackage{algorithm}
\usepackage{algorithmic}

\input{dfn}

\renewcommand{\rv}[1]{{{\textcolor{black}{#1}}}}

\newenvironment{rve}{\color{black}\bfseries}{}

\title{Beyond Holistic Models: Systematic Component-level Benchmarking of Deep Multivariate Time-Series Forecasting}

\author{Shuang Liang}
\authornote{Equal contribution.}
\email{liangs1104@stu.sufe.edu.cn}
\affiliation{%
  \institution{Shanghai University of Finance and Economics}
  \city{Shanghai}
  \country{China}
}

\author{Chaochuan Hou}
\authornotemark[1]
\email{houchaochuan@foxmail.com}
\affiliation{%
  \institution{Shanghai University of Finance and Economics}
  \city{Shanghai}
  \country{China}
}

\author{Xu Yao}
\authornotemark[1]
\email{yaoxu@stu.sufe.edu.cn}
\affiliation{%
  \institution{Shanghai University of Finance and Economics}
  \city{Shanghai}
  \country{China}
}

\author{Shiping Wang}
\authornotemark[1]
\email{shiping.wsp@antgroup.com}
\affiliation{
  \institution{Ant Group}
  \city{Shanghai}
  \country{China}
}

\author{Hailiang Huang}
\authornotemark[2]
\email{hlhuang@shufe.edu.cn}
\affiliation{%
  \department{Key Laboratory of Interdisciplinary Research of Computation and Economics}
  \institution{Shanghai University of Finance and Economics}
  \city{Shanghai}
  \country{China}
}

\author{Songqiao Han}
\authornotemark[2]
\email{han.songqiao@shufe.edu.cn}
\affiliation{%
  \department{Key Laboratory of Interdisciplinary Research of Computation and Economics}
  \institution{Shanghai University of Finance and Economics}
  \city{Shanghai}
  \country{China}
}

\author{Minqi Jiang}
\authornote{Corresponding author.}
\email{jiangmq95@163.com}
\affiliation{%
  \department{Key Laboratory of Interdisciplinary Research of Computation and Economics}
  \institution{Shanghai University of Finance and Economics}
  \city{Shanghai}
  \country{China}
}

\renewcommand{\shortauthors}{Shuang Liang et al.}

\begin{abstract}

\input{0abstract}
\end{abstract}

\begin{document}

\ccsdesc[500]{Computing methodologies~Machine learning}

\keywords{Component-level Analysis; Benchmark; Time Series Forecasting}

\maketitle

\section{Introduction}
\label{sec:intro}

\input{1intro}

\section{Related Work}
\label{sec:related}
\input{2related}


\vspace{-9pt}
\section{\system: \textit{Benchmarking} and \textit{Automating} Deconstructed Components in Deep MTSF}
\label{sec: setting}
\input{3setting}


\section{Experiments}
\label{sec: exp}
\input{4exp}
\section{Conclusions and Future Work}
\label{sec: conclu}
\input{5discussion}

\begin{acks}
  \label{sec.acknowledgments}
  \noindent This work was supported by the National Natural Science Foundation of China (Nos. 72271151, 72342009, 72442024, 72172085) and Ant Group.
  We acknowledge AI tools for assisting with LaTeX formatting, English grammar polishing, and literature search. Literature sorting and verification, as well as review of all AI-assisted content, were conducted manually.
\end{acks}


\bibliographystyle{ACM-Reference-Format}
\balance
\bibliography{ref}

\appendix

\input{6appendix.tex}

\end{document}

%% file: dfn.tex
\usepackage{xspace}
\newcommand{\system}{{TSCOMP}\xspace}
\newcommand{\ndatasets}{{13}\xspace}
\newcommand{\ndatasetsall}{{14}\xspace}

\newcommand{\bal}{
\begin{align}}
  \newcommand{\ean}{
\end{align}}
\newcommand{\bit}{
\begin{itemize}}
    \newcommand{\eit}{
\end{itemize}}
\newcommand{\ben}{
\begin{enumerate}}
    \newcommand{\een}{
\end{enumerate}}
\newcommand{\beq}{
\begin{equation}}
  \newcommand{\eeq}{
\end{equation}}

\newcommand{\add}[1] {\textcolor{black}{#1}}

\newcommand{\rv}[1] {\textcolor{blue}{#1}}

\definecolor{aliceblue}{rgb}{0.91796875, 0.91796875, 0.999}
\definecolor{aliceyellow}{rgb}{0.999, 0.96875, 0.91796875}
\definecolor{alicegreen}{rgb}{0.91796875, 0.95703125, 0.91796875}
\definecolor{blue}{rgb}{0,0,1}

\newcommand{\tcr}[1]{\textcolor{red}{#1}}
\newcommand{\tcb}[1]{\textcolor{blue}{#1}}

%% file: 0abstract.tex
While previous research in multivariate time series forecasting has focused on developing complex holistic models, this work advocates for a shift toward a granular, component-level understanding of their impacts. We propose \system, the first large-scale benchmark that systematically deconstructs deep forecasting methods into their core, fine-grained components—spanning series preprocessing, encoding strategies, network architectures including specific and large time-series models, and optimization methods. Using constrained orthogonal experimental design and extensive evaluations, we conduct multi-view analyses that reveal component effectiveness across different backbones, data characteristics, and their interactions. Beyond providing insights, this benchmark establishes a fine-grained performance corpus comprising over 20,000 model-dataset evaluations, which supports the learning of automated component selection, enabling zero-shot model construction on new datasets. Our experiments demonstrate that the corpus-driven approach, despite its simplicity, consistently outperforms state-of-the-art methods, validating the soundness of our evaluation design and confirming that systematic component selection surpasses manually designed complex architectures. All code and the performance corpus are publicly available at \url{https://github.com/SUFE-AILAB/TSCOMP}.

%% file: 1intro.tex
\begin{figure*}[htbp]
  \centering
  \includegraphics[width=0.9\textwidth]{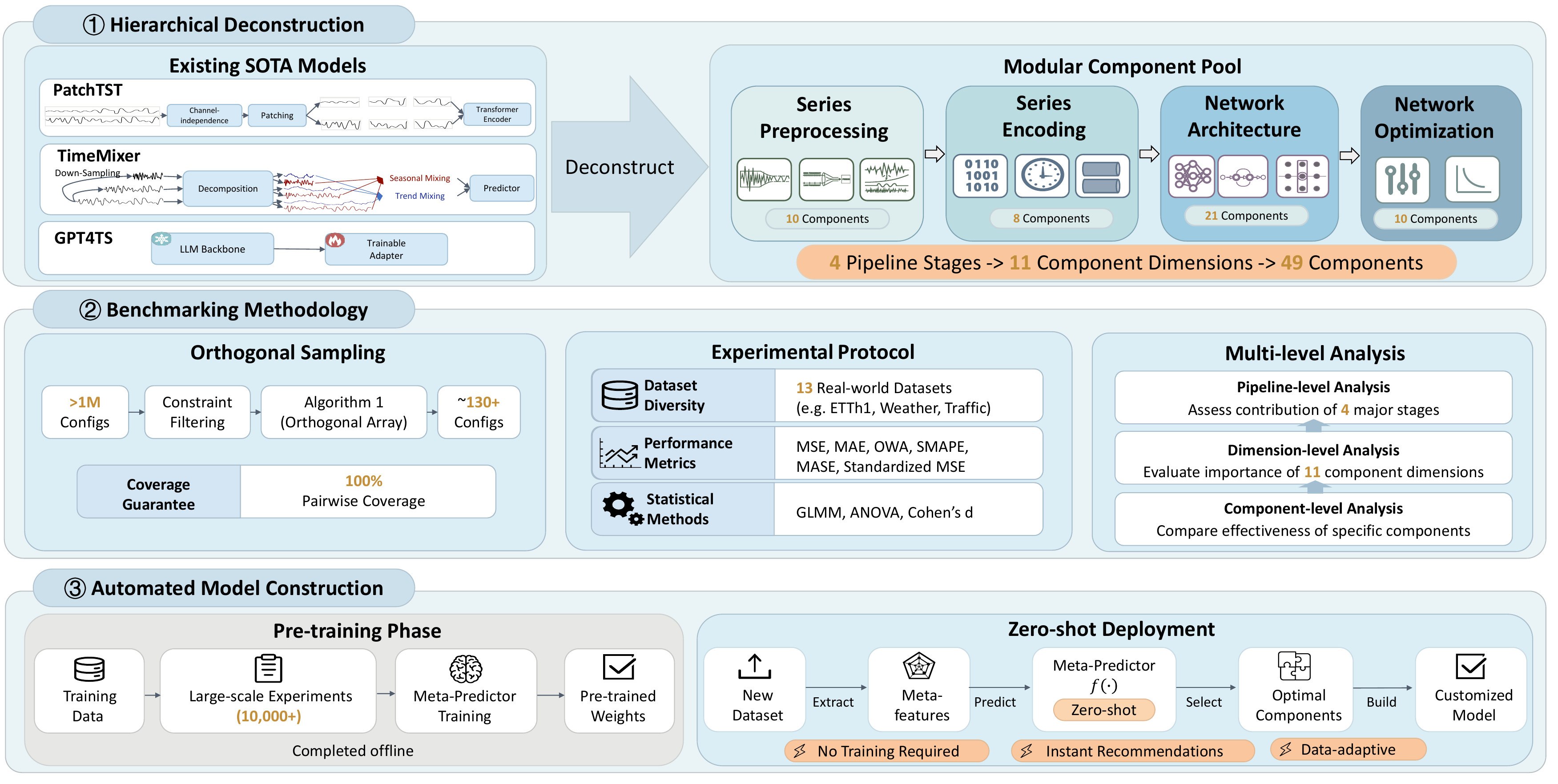}
  \caption{Overview of the proposed \system framework. \system deconstructs existing SOTA models into a modular component pool. Through large-scale experimental analysis, \system conducts bottom-up evaluation from component-level comparisons to dimension-level and pipeline-level importance ranking. The resulting performance corpus enables automated model construction via a pre-trained meta-predictor that delivers zero-shot, data-adaptive component selection.}
  \label{fig:pipeline}
\end{figure*}

Multivariate time series refers to time series data involving multiple interdependent variables, which are widely present in various fields such as finance~\cite{sezer2020financial}, energy~\cite{alvarez2010energy,deb2017review}, traffic~\cite{cirstea2022towards,yin2016forecasting}, and health~\cite{bui2018time,kaushik2020ai}. Among the numerous analysis tasks, multivariate time series forecasting (MTSF) attracts substantial attention from the research community due to its significant practical applications.
Traditional approaches to MTSF are largely based on statistical methods~\cite{abraham2009statistical,zhang2003time} and machine learning techniques~\cite{hartanto2023stock,masini2023machine}. In recent years, deep learning (DL) has become the most active area of research for MTSF, driven by its ability to handle complex patterns and large-scale datasets effectively~\cite{wang2024deep}.

Early academic efforts of deep MTSF methods like RNN-type methods~\cite{yamak2019comparison} struggle with capturing long-term temporal dependencies due to their inherent limitations of gradient vanishing or exploding problems~\cite{zhou2021informer,zhou2022fedformer}. To address these issues, Transformer shows significant potential by effectively modeling temporal correlations via attention variants~\cite{li2019enhancing,zhou2022fedformer}. Although simpler MLP-based structures~\cite{zeng2023dlinear} later challenged this paradigm, innovations like patching and channel-independence strategies~\cite{nie2023PatchTST} further enhanced its performance. Alongside these architectural advances, critical modular studies have emerged, focusing on variable dependency~\cite{zhang2023crossformer}, normalization~\cite{Liu2022NonstationaryTR}, and decomposition~\cite{liu2023koopa}. This evolution has recently expanded to Large Language Models (LLMs)~\cite{jin2024timellm, zhou2023one} and Time Series Foundation Models (TSFMs)~\cite{liu2024timer, goswami2024moment}.

As the field of MTSF continues to diversify, existing studies typically address concerns about methodological effectiveness by conducting large-scale benchmarks~\cite{wang2024deep,shao2024exploring,qiu2024tfb}.
These studies consistently indicate that no single approach---whether a specific deep forecasting model (e.g., MLP, Transformer) or large time-series models---dominates across all scenarios~\cite{liu2025breaking}.
This variability suggests the need to investigate effective MTSF design at finer granularities.
Specifically, the MTSF formulation involves a multi-stage modeling pipeline, where each stage (e.g., series preprocessing) comprises distinct component dimensions (e.g., series normalization) instantiated by specific components (e.g., RevIN).
However, existing benchmarks typically evaluate models holistically, failing to analyze this multi-level hierarchy.
Consequently, contributions of internal mechanisms remain obscured.
This ambiguity isolates effective designs within specific methods, hindering the combination of these strengths into superior solutions.

To bridge these gaps, we propose \system, a comprehensive framework designed to systematically deconstruct and benchmark deep MTSF methods. Instead of viewing models as indivisible black boxes, \system performs a hierarchical deconstruction across three levels: the \textit{Pipeline}, \textit{Component Dimensions}, and \textit{Deconstructed Components} (see Fig.~\ref{fig:pipeline}). To ensure rigorous evaluation, we employ a constrained orthogonal experimental protocol that isolates the contribution of individual components. This enables a multi-view analysis that extends beyond general performance rankings: we investigate component efficacy under different backbones, their distinct adaptability to diverse data characteristics and domains, and the complex interactions between deconstructed components.

Building upon the proposed benchmark, we establish a fine-grained performance corpus that not only validates prevailing claims but also serves as a robust foundation for automated model construction. Based on this valuable corpus, \system learns how components adapt to different data characteristics and adaptively assembles optimal components tailored to specific datasets. This approach consistently surpasses state-of-the-art methods. We summarize the key contributions of \system as follows:

\begin{itemize}[leftmargin=0.8em, itemsep=3pt, topsep=2pt]
  \item \textbf{Comprehensive Benchmark via Hierarchical Deconstruction}.
    We propose \system, the first large-scale benchmark that systematically deconstructs deep MTSF methods. \system examines the MTSF workflow through a hierarchical design space, spanning from the overall modeling pipeline to fine-grained specific components. To rigorously assess these elements, we design a constrained orthogonal evaluation protocol that isolates the core mechanisms driving forecasting performance.

  \item \textbf{Multi-View Analysis and Insights}.
    We conduct a large-scale analysis that provides both overall and conditional insights. Beyond evaluating general component effectiveness, we extensively investigate performance variations across different backbones (including specific models and emerging LLMs/TSFMs), diverse data domains, and data characteristics. Furthermore, we explore the intricate interaction effects among deconstructed components, verifying community claims with rigorous experimental evidence.

  \item \textbf{Open-Sourced Corpus and Automated Construction}.
    We open-source the resulting fine-grained performance corpus and validate its utility for model design. This corpus facilitates automated construction of MTSF methods that are adaptively tailored to different forecasting scenarios, consistently achieving better results than state-of-the-art methods.
\end{itemize}

%% file: 2related.tex
\subsection{Deep Learning-based MTSF}

MTSF evolves from traditional statistical methods like ARIMA and Gaussian processes
to modern deep learning approaches. While legacy RNNs struggle with long-term dependencies, Transformers revolutionize temporal modeling through attention mechanisms \cite{wu2021autoformer, nie2023PatchTST}.
Alternatively, MLP-based models regain prominence for their simplicity and effectiveness. DLinear \cite{zeng2023dlinear} demonstrates that linear mappings with decomposition often surpass complex Transformers. Advanced variants like TimeMixer \cite{wang2024timemixer} and OLinear \cite{yue2025olinear} leverage multi-scale analysis and orthogonal decomposition. Leveraging foundation models represents a paradigm shift. Adaptation methods like GPT4TS \cite{zhou2023one} transfer language model knowledge via prompt engineering \cite{jin2024timellm} or fine-tuning \cite{chang2023llm4ts}. Native TSFMs like Timer \cite{liu2024timer} and Time-MOE \cite{shi2025timemoe} target zero-shot generalization. In this work, \system systematically deconstructs these diverse methodologies into atomic components across the entire forecasting pipeline.

Convergent design directions emerge across these approaches. Preprocessing addresses non-stationarity via adaptive normalization \cite{kim2021RevIN, fan2023dish} or decomposition \cite{zeng2023dlinear, liu2023koopa}, while temporal modeling focuses on multi-scale dependency capture \cite{wang2024timemixer, zhou2022fedformer}. Architecturally, strategies balance robust channel-independent processing \cite{nie2023PatchTST} against correlation-aware channel-dependent modeling \cite{liu2024itransformer, chen2023tsmixer}. Tokenization spans point-wise to series-wise representations \cite{zhou2021informer, nie2023PatchTST, liu2024itransformer}, coupled with dependency mechanisms like recurrence, convolution, and attention \cite{gu2024mamba, bai2018tcn}. These modular innovations underpin \system's component decomposition framework (Table~\ref{tab:design space}).
Driven by the rapid evolution of MTSF research, \system deconstructs models into atomic components to explore their real contributions and enable flexible model structure selection and configuration.
\vspace{-0.1in}

\subsection{Benchmarks for Time Series Forecasting}
Recent time series forecasting benchmark studies~\cite{wang2024deep,shao2024exploring,qiu2024tfb,liu2024timer} have conducted large-scale experiments across a diverse range of datasets. However, most of these works treat current models as monolithic entities.
TSlib~\cite{wang2024deep}, one of the most popular repositories for time series analysis, provides a comprehensive survey and evaluates recent time series models across various time series analysis tasks.
From the perspective of time series characteristics, BasicTS~\cite{shao2024exploring} analyzes model architectures and the strategy of treating channels (or variables) independently.
With a more extensive experimental setup, TFB~\cite{qiu2024tfb} additionally includes machine learning and statistical forecasting methods, and covers datasets from a broader range of domains. More recently,
OpenLTM~\cite{liu2024timer} provides a system to evaluate Time Series Foundation Models as well as Large Language Models for time series methods.
Although some surveys and benchmarks analyze the fine-grained components of time series models, their scope is often limited. Wen et al.~\cite{wen2020time} discuss various time series data augmentation techniques and evaluate their effectiveness. Another survey~\cite{wen2022transformers} systematically reviews fine-grained components within Transformer-based architectures, but lacks broader coverage of model structures and evaluations.

\rv{These limitations prevent the aforementioned studies from comprehensively and meticulously evaluating the entire MTSF pipeline, spanning from sequence preprocessing to model parameter optimization.}
To the best of our knowledge, \system is the first benchmark that not only provides component-level fine-grained analysis, but also conducts large-scale empirical evaluations.

\subsection{AutoML for Time Series Forecasting}

Current automated MTSF methods primarily utilize ensemble~\cite{shchur2023autogluon} or meta-learning~\cite{abdallah2022autoforecast,fischer2024autoxpcr} strategies.
Ensemble approaches integrate models from a predefined pool but incur substantial computational costs.
Meta-learning methods select optimal models using dataset-level meta-features.
However, these approaches operate at the coarse model level, limiting performance gains to existing architectural bounds.
Recently, TimeFuse~\cite{liu2025breaking} advanced this paradigm via adaptive sample-level fusion.
It dynamically weights pre-trained forecasters by analyzing instance-specific statistical and spectral patterns.
This strategy leverages complementary model strengths to handle diverse temporal dynamics.
In contrast, \system pioneers fine-grained component-level automation for MTSF.
We extend the search space beyond fixed architectures to comprehensive component dimensions.
This enables constructing novel pipelines that surpass the capabilities of rigid SOTA models.

%% file: 3setting.tex
\subsection{Overview}

While existing MTSF benchmarks evaluate complete models as unified entities, \system introduces a fine-grained evaluation paradigm that systematically deconstructs deep forecasting methods into modular components.
As illustrated in Fig.~\ref{fig:pipeline}, \system adopts a hierarchical benchmark framework across the \textbf{pipeline}, \textbf{dimension}, and \textbf{component} levels, spanning from holistic workflows to core modules in MTSF tasks.
To ensure comprehensive evaluation, we employ a constrained orthogonal experimental protocol that systematically assess individual components, their interactions, and adaptability to different data characteristics.

This paper further demonstrates that our benchmark results provide a valuable corpus for automated model construction.
Through large-scale evaluation spanning diverse datasets and component combinations, we obtain systematic insights into component effectiveness under varying data characteristics.
We show that these component-level insights enable data-adaptive selection and assembly of optimal forecasting pipelines, consistently achieving superior performance compared to SOTA methods (Sec.~\ref{subsec:TSGym_overview}).

\subsection{Hierarchical Deconstruction}
\label{subsec:hierarchical_deconstruction}

\subsubsection{Problem Definition for MTSF}
We focus on multivariate time series forecasting (MTSF) with $C$ variates.
Given historical data $\mathbf{\chi}=\left\{\boldsymbol{x}_{1}^{t},\ldots, \boldsymbol{x}_{C}^{t}\right\}^{L}_{t=1}$, where $L$ denotes the look-back sequence length and $\boldsymbol{x}_{i}^{t}$ represents the $i$-th variate, the task predicts the $T$-step future sequence $\mathbf{\hat{\chi}}=\left\{\boldsymbol{\hat{x}}_{1}^{t},\ldots, \boldsymbol{x}_{C}^{t}\right\}^{L+T}_{t=L+1}$.
Following~\cite{zhou2021informer}, we directly predict all future steps to avoid error accumulation when $T>1$.

\subsubsection{Design Space Construction}
Existing MTSF models often concentrate design innovations on specific modules---for instance, iTransformer~\cite{liu2024itransformer} innovates with inverted encoding mechanisms, and TimeMixer~\cite{wang2024timemixer} introduces multi-scale mixing strategies.
However, these innovations are tightly coupled with other deconstructed components, making it difficult to isolate their individual contributions.
To enable systematic component-level analysis, we deconstruct SOTA models into modular components along the standard MTSF workflow.
This component-level deconstruction enables systematic benchmarking to identify core elements that drive forecasting improvements.

\input{Tables/Components_full}

We organize the design space hierarchically across three levels as illustrated in Fig.~\ref{fig:pipeline}.
At the \textbf{pipeline level}, we model the standard MTSF workflow as a sequence: \textit{Series Preprocessing} $\rightarrow$ \textit{Series Encoding} $\rightarrow$ \textit{Network Architecture} $\rightarrow$ \textit{Network Optimization}.
At the \textbf{dimension level}, each pipeline stage comprises multiple component dimensions (e.g., normalization methods, tokenization strategies, attention mechanisms).
At the \textbf{component level}, each dimension instantiates multiple concrete implementations extracted from SOTA methods (e.g., RevIN normalization, series patching, sparse attention).
This hierarchical deconstruction yields a structured design space covering diverse modeling strategies.

Formally,
We define $k$ dimensions $\mathcal{DD}=\{DD_1,...,DD_k\}$ to describe the MTSF modeling pipeline.
Each dimension $DD_{i}$ contains multiple deconstructed components $DC$.
The Cartesian product yields all valid model combinations:
$\mathcal{M}=DD_1 \times DD_2 \times \cdots \times DD_k = \{(DC_1, DC_2, \ldots, DC_k) \mid DC_i \in DD_i\}$.
Table~\ref{tab:design space} presents the complete design space across 4 pipeline stages, covering 11 dimensions and 49 deconstructed components.
This deconstruction process involves deep examination of model papers and source code to extract core innovations.
For instance, \system includes TimeMixer's multi-scale mixing~\cite{wang2024timemixer}, iTransformer's inverted encoding~\cite{liu2024itransformer}, and PatchTST's channel-independent patching~\cite{nie2023PatchTST}.
We also incorporate diverse attention mechanisms~\cite{wen2022transformers} and emerging LLM/TSFM architectures.
\ifarxiv Comprehensive descriptions of all deconstructed components appear in Appx.~\ref{appx:design_choices_details}. \fi

\subsection{Benchmarking Methodology}
To ensure systematic and fair evaluation, we employ a constrained orthogonal experimental design to achieve comprehensive coverage of component interactions while maintaining a manageable scale.
Standardized experimental protocols are further established, encompassing datasets, evaluation metrics, and training configurations, to ensure reproducibility and fair comparison.

\subsubsection{Constrained Orthogonal Experimental Design}
\mbox{}\\
\noindent \textbf{Design Space Complexity.} The Cartesian product of component dimensions yields over $10^6$ theoretical configurations.
However, fundamental mechanisms render many combinations incompatible.
For instance, inverted encoding inherently conflicts with channel-independent strategies.
Pre-trained backbones also require specific attention protocols.
We strictly exclude invalid combinations to ensure architectural soundness.
Even after filtering, the remaining pool consists of thousands of models.
This scale remains computationally intractable for multi-dataset evaluation.
These structural constraints necessitate a more efficient sampling strategy.

\input{Tables/orthogonal_pool_algorithm}

\noindent \textbf{Pairwise Coverage Criterion.} To enable systematic analysis, we employ a constrained orthogonal experimental design.
This strategy targets pairwise coverage of components in the mapping $f$.
Pairwise coverage balances interaction analysis with computational tractability.
Exhaustive $k$-way coverage ($k \geq 3$) yields an impractically large pool.
Conversely, single-component analysis fails to reveal critical interaction effects on performance $\mathcal{L}$.
Algorithm~\ref{alg:orthogonal_pool} adopts a greedy strategy to construct the pool.
It iteratively selects configurations to cover every valid pairwise interaction.
This approach reduces the set to approximately 136 models per horizon.
This tractable size ensures a rigorous basis for evaluating component effectiveness.

\subsubsection{Experimental Protocol}
\mbox{}\\
\noindent \textbf{Datasets.}
We conduct extensive evaluations on \ndatasets standard long-term forecasting benchmarks, including ETT variants, Electricity, Traffic, Weather, Exchange, ILI, NYSE, NASDAQ, FRED-MD, and Covid-19, following established protocols~\cite{wu2021autoformer,jin2024timellm}.
Detailed specifications are provided in Appx.~\ref{appx:data}.

\vspace{5pt}
\noindent \textbf{Evaluation Metrics.}
We adopt Mean Squared Error (MSE) as the primary accuracy metric.
To facilitate aggregating results across diverse datasets and prediction horizons, we employ Standardized MSE to eliminate scale discrepancies.
Comprehensive results using MAE, SMAPE, and MASE are provided in Appx.~\ref{appx:metrics}.

\vspace{5pt}
\noindent \textbf{Statistical Analysis Framework.}
To rigorously quantify component effectiveness, we employ a three-tiered statistical framework:
(i) \textit{Marginal Contribution Analysis} uses Generalized Linear Mixed Models (GLMM) to estimate the independent effect of each component while controlling for dataset and horizon variability.
(ii) \textit{Variance Contribution Analysis} employs Analysis of Variance (ANOVA) to quantify the proportion of performance variance explained by each component dimension.
(iii) \textit{Effect Size Analysis} utilizes Cohen's $d$ to measure the magnitude of performance differences across data characteristics, ensuring robustness beyond mere statistical significance.

\subsection{Automated Model Construction}
\label{subsec:TSGym_overview}

\textbf{Performance Corpus via \system}.
Our systematic benchmarking yields a comprehensive performance corpus by evaluating $m$ constraint-validated configurations $\mathcal{M}$ from the Constrained Orthogonal Pool across $n$ training datasets $\bm{\mathcal{D}}_{\text{train}}$ under identical conditions.
This produces a performance matrix $\bm{P}\in \mathbb{R}^{n \times m}$ capturing fine-grained component-data interactions.
This corpus enables zero-shot automated model construction, predicting effectiveness on unseen datasets without exhaustive experimentation.

\noindent \textbf{Automated Model Construction}.
We construct a meta-dataset $\mathcal{D}_{meta} = \left\{(\mathcal{D}_i, M_j, R_{i,j})\right\}$ from the performance matrix $\bm{P}$.
To ensure fair learning across datasets with varying difficulty scales, we convert raw MSE values $\bm{P}_{i,j}$ to normalized rankings $R_{i,j}=\mathrm{rank}(\bm{P}_{i,j})/m \in [0,1]$, where smaller values indicate better performance.
Each configuration $M_j$ is decomposed into component indices $\{c_{j,1}, \dots, c_{j,k}\}$ and embedded via a learnable codebook $\phi(\cdot)$.
The meta-predictor $f_{\theta}$ learns to map dataset meta-features $\mathbf{E}^{meta}_i$ (extracted by the pre-trained tabular model TabPFN~\cite{hollmann2022tabpfn}\ifarxiv; detailed in Appx.~\ref{appx:meta-features}\fi) and component embeddings $\oplus_{t=1}^{k} \phi(c_{j,t})$ to predicted rankings (Eq.~\eqref{eq:meta-predictor}).%
\begin{equation}
  f\left(\mathcal{D}_i, M_j\right)=\bm{R}_{i,j},
  f \;: \underbracket{\mathbf{E}_{i}^{meta}}_{\text{meta features}},
  \underbracket{\mathbf{E}_{j}^{comp}}_{\text{component embed.}}
  \mapsto \bm{R}_{i,j}
  \label{eq:meta-predictor}
\end{equation}
where $i\in \{1,\ldots,n\}$ and $j\in \{1,\ldots,m\}$.
We implement the meta-predictor as a two-layer MLP trained via regression on the benchmark corpus.
At test time, we extract meta-features from a new dataset's training split, obtain predicted rankings using trained $f(\cdot)$, and select the top-$k$ components to construct MTSF models.
This procedure requires no neural network training on $\mathbf{X}_{\text{test}}$, enabling users to obtain model recommendations by simply providing their dataset without running extensive experiments. \ifarxiv Details of the meta-predictor are provided in Appx.~\ref{appx:meta-details}. \fi

%% file: Tables/Components_full.tex
\begin{table}[!h]
  \small
  \centering
  \caption{
  \system supports comprehensive deconstructed components for deep time-series forecasting methods.}
  \resizebox{0.98\linewidth}{!}{
    \renewcommand{\arraystretch}{1.3}
    \begin{tabular}{lll}
      \toprule
      \textbf{Pipeline} & \textbf{Component Dimensions} & \textbf{Deconstructed Components} \\
      \midrule

      \multirow{4}{*}{\makecell{\textit{Series} \\ \textit{Preprocessing}}}
      &    \textbf{Series Normalization }   & w/o Norm, Stat, RevIN~\cite{kim2021reversible}, DishTS \cite{fan2023dish} \\ \cmidrule(l){2-3}
      &    \textbf{Series Decomposition }   & \makecell[l]{w/o Decomp, Moving Average (MA), \\ MoEMA \cite{zhou2022fedformer}, DFT \cite{wang2024timemixer}} \\ \cmidrule(l){2-3}
      &    \textbf{Series Sampling/Mixing }   & w/o Mixing, w/ Mixing \cite{wang2024timemixer} \\
      \midrule
      \multirow{4}{*}{\makecell{\textit{Series} \\ \textit{Encoding}}}
      &   \textbf{Channel Independent }   & Channel Depen, Channel Indepen \\ \cmidrule(l){2-3}
      &   \textbf{Series Tokenization }    & \makecell[l]{Point Encoding, Series Patching \cite{nie2023PatchTST}, \\ Inverted Encoding \cite{liu2024itransformer}, Ortho Encoding \cite{yue2025olinear}} \\ \cmidrule(l){2-3}
      &   \textbf{Timestamp Embedding }    & w/o Embedding, w/ Embedding \\
      \midrule
      \multirow{10}{*}[30pt]{\makecell{\textit{Network} \\ \textit{Architecture}}}
      & \makecell[l]{\textbf{Network Backbone}} & \makecell[l]{\textbf{MLP}: DNN, NormLin \cite{yue2025olinear}; \\
        \textbf{RNN}: GRU, xLSTM \cite{kraus2025xlstmmixer} \\
        \textbf{Transformer}: w/o Attn, SelfAttn, \\ AutoCorr \cite{wu2021autoformer}, SparseAttn \cite{zhou2021informer}, \\ FrequencyAttn \cite{zhou2022fedformer}, DestationaryAttn \cite{Liu2022NonstationaryTR} \\
        \textbf{LLM}: GPT4TS \cite{zhou2023one}, TimeLLM \cite{jin2024timellm}; \\
      \textbf{TSFM}: Timer \cite{liu2024timer}, Moment \cite{goswami2024moment}, \\ TimeMoE \cite{shi2025timemoe}, Chronos \cite{ansari2024chronos}} \\ \cmidrule(l){2-3}
      & \textbf{Feature Attention } & w/o Attn, SelfAttn, SparseAttn \\ \cmidrule(l){2-3}
      & \textbf{Retrieval Augmented (RAG) } & w/o RAG, w/ RAG \cite{han2025retrieval} \\
      \midrule
      \multirow{3}{*}{\makecell{\textit{Network} \\ \textit{Optimization}}}
      & \textbf{Sequence Length} & 48, 96, 192, 512 \\ \cmidrule(l){2-3}
      & \textbf{Loss Function} & \makecell[l]{MSE, MAE, HUBER, DBLoss \cite{qiu2024dbloss}, \\ PSLoss \cite{kudrat2025patchwise}, FreDFLoss \cite{wang2025fredf}} \\
      \bottomrule
    \end{tabular}%
  }
  \label{tab:design space}%
\end{table}

%% file: Tables/orthogonal_pool_algorithm.tex
\begin{algorithm}[h!]
  \caption{Constrained Orthogonal Pool Generation.}
  \label{alg:orthogonal_pool}
  \begin{algorithmic}[1]
    \small
    \STATE {\bfseries Input:} Component Space $\mathcal{DD}$, Constraints $\textsc{IsValid}(\cdot)$, Initial Pool $\mathcal{P}_{init}$.
    \STATE \textit{// Phase 1: Initialization}
    \STATE Initialize $\mathcal{R}$ with all valid pairwise component interactions derived from $\mathcal{DD}$.
    \STATE Set $\mathcal{M}_s \leftarrow \mathcal{P}_{init}$.
    \STATE Remove interactions from $\mathcal{R}$ that are already covered by $\mathcal{M}_s$.
    \STATE \textit{// Phase 2: Greedy Search}
    \WHILE{$\mathcal{R} \neq \emptyset$}
    \STATE Generate a batch $\mathcal{S}_{cand}$ of valid random models using $\textsc{IsValid}(\cdot)$.
    \IF{$\mathcal{S}_{cand} = \emptyset$} \STATE \textbf{break} \ENDIF
    \STATE Select $M^* \in \mathcal{S}_{cand}$ that covers the most remaining interactions in $\mathcal{R}$.
    \IF{$M^*$ covers new interactions}
    \STATE Add $M^*$ to pool: $\mathcal{M}_s \leftarrow \mathcal{M}_s \cup \{M^*\}$.
    \STATE Update $\mathcal{R}$ by removing interactions covered by $M^*$.
    \ELSE
    \STATE \textbf{break} \textit{// Stop if no progress}
    \ENDIF
    \ENDWHILE
    \STATE {\bfseries Output:} Sampled model pool $\mathcal{M}_s$.
  \end{algorithmic}
\end{algorithm}

%% file: 4exp.tex
\label{exp:benchmark}

We systematically evaluate decoupled MTSF design space through multi-level analysis to identify generally effective designs (\ref{exp:overall}), examine how effectiveness varies across network architectures (\ref{exp:architecture-specific}) and data characteristics (\ref{exp:data-specific}), and demonstrate how the extensive benchmark corpus enables automated model construction (\ref{exp:automated}). \ifarxiv Comprehensive visualizations, including effect ranges, pipeline importance, and detailed component performance, are provided in Appx.~\ref{appx:complete_results}. \fi

\subsection{Overall Analysis}
\label{exp:overall}
\subsubsection{Component-Level}
\label{exp:component-level}

We perform a fine-grained analysis of individual components across all component dimensions using a generalized linear mixed model (GLMM) to isolate the marginal contribution of each component. The results quantify the impact of each component on forecasting performance (standardized MSE). Specifically, we standardize MSE across datasets and forecast horizons to ensure rigorous cross-task evaluation.

\begin{figure}[t!]
  \centering
  \begin{subfigure}[t]{0.32\columnwidth}
    \centering
    \includegraphics[width=\textwidth]{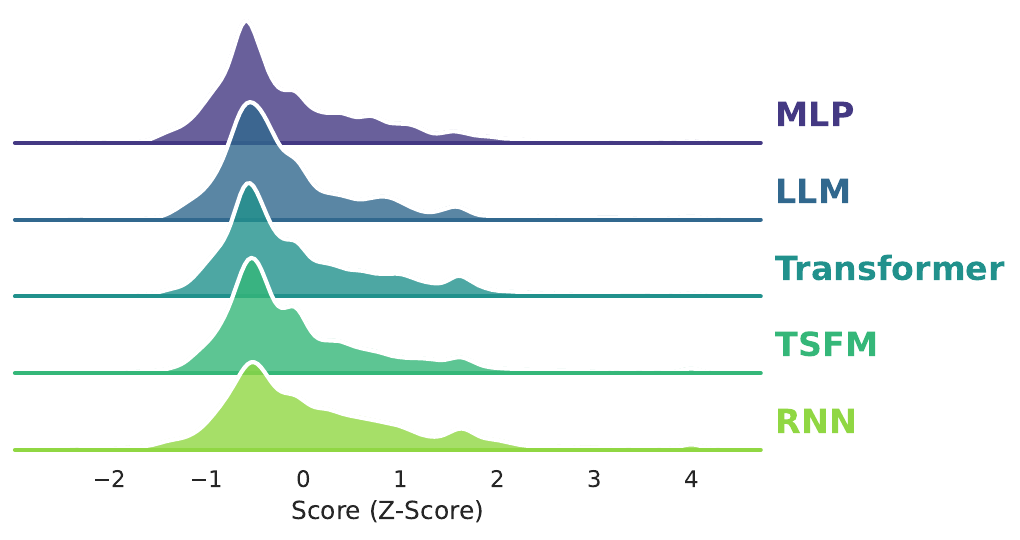}
    \caption{\scriptsize Network Architecture}
    \label{fig:exp-Backbone}
  \end{subfigure}
  \hfill
  \begin{subfigure}[t]{0.32\columnwidth}
    \centering
    \includegraphics[width=\textwidth]{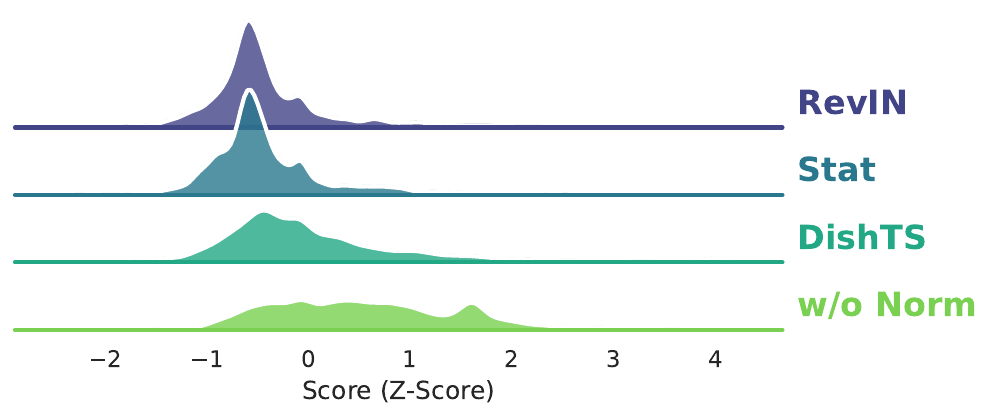}
    \caption{\scriptsize Series Normalization}
    \label{fig:exp-Normalization}
  \end{subfigure}
  \hfill
  \begin{subfigure}[t]{0.32\columnwidth}
    \centering
    \includegraphics[width=\textwidth]{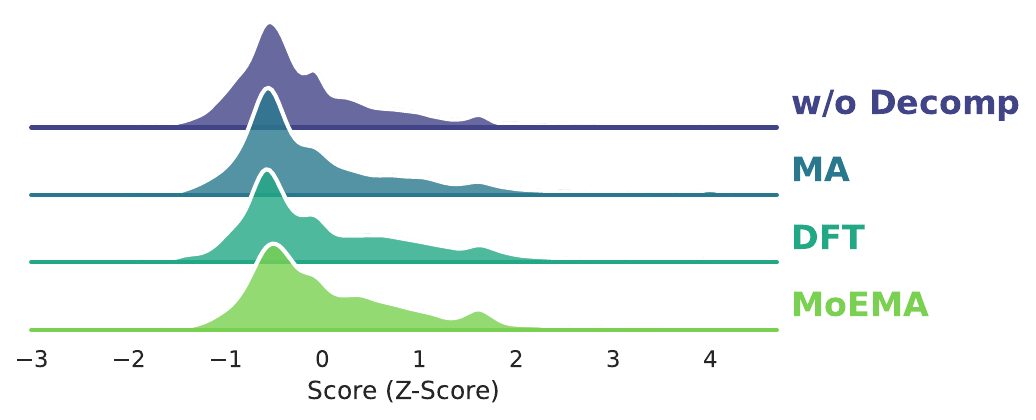}
    \caption{\scriptsize Series Decomposition}
    \label{fig:exp-Decomposition}
  \end{subfigure}

  \begin{subfigure}[t]{0.32\columnwidth}
    \centering
    \includegraphics[width=\textwidth]{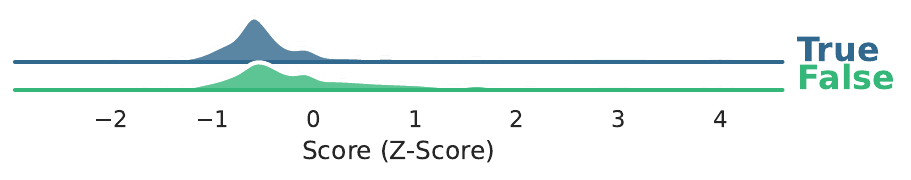}
    \caption{\scriptsize Channel Independent}
    \label{fig:exp-CI}
  \end{subfigure}
  \hfill
  \begin{subfigure}[t]{0.32\columnwidth}
    \centering
    \includegraphics[width=\textwidth]{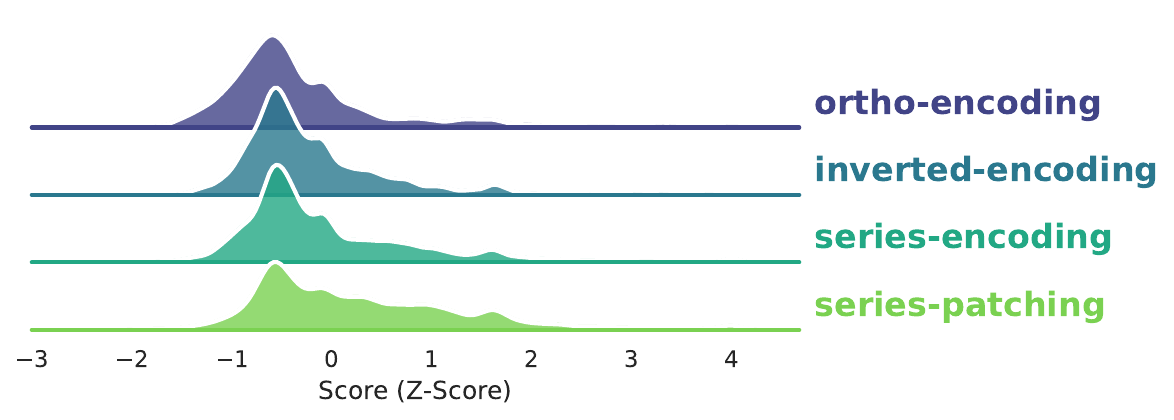}
    \caption{\scriptsize Series Tokenization}
    \label{fig:exp-Tokenization}
  \end{subfigure}
  \hfill
  \begin{subfigure}[t]{0.32\columnwidth}
    \centering
    \includegraphics[width=\textwidth]{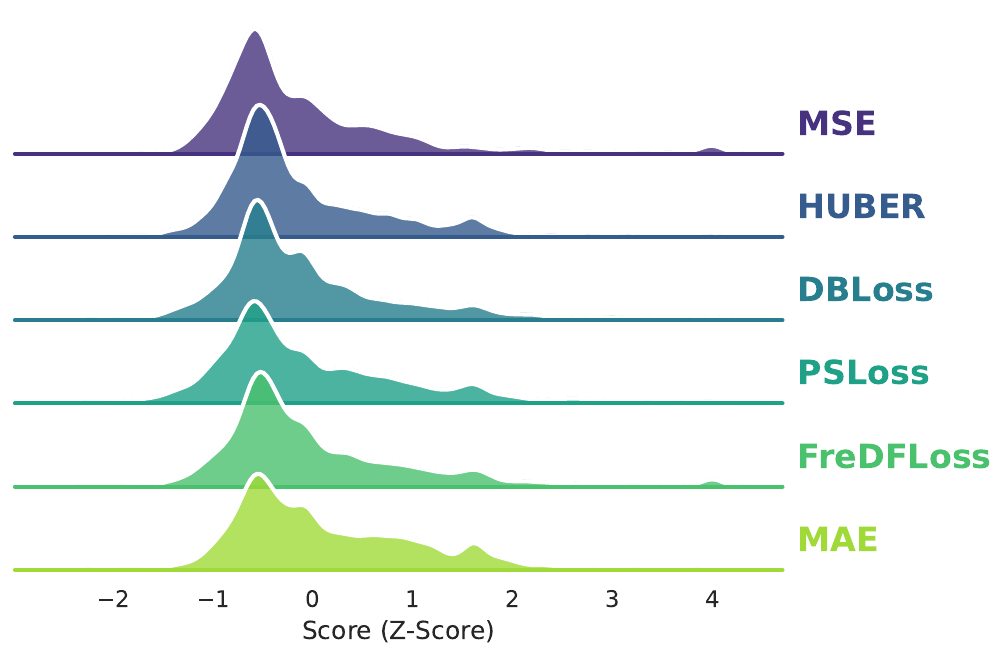}
    \caption{\scriptsize Loss Function}
    \label{fig:exp-loss}
  \end{subfigure}
  \caption{Component performance distributions (standardized MSE). Lower values indicate better performance.}
  \label{fig:exp-distribution}
\end{figure}

Our component-level analysis follows the MTSF pipeline stages, comparing component effectiveness within component dimensions (Table~\ref{tab:coef-comparison}, `General' column; and Fig.~\ref{fig:exp-distribution}).
Series normalization yields substantial performance improvements, with \textit{RevIN} and \textit{Stationary} achieving the strongest MSE reductions and effectively stabilizing non-stationary dynamics.
In contrast, \textit{Series Decomposition} exhibits mixed effectiveness—while decomposition methods increase MSE on average.
For series encoding, \textit{Channel Independence} delivers strong performance gains, confirming independent modeling is generally superior; tokenization strategies also demonstrate robustness, with \textit{Inverted and Ortho} significantly outperforming \textit{Point-wise Encoding}.
In network optimization, loss functions \textit{HUBER} and \textit{MAE} significantly outperform \textit{MSE}, providing viable alternatives depending on error distribution.

\begin{figure}[h!]
  \centering
  \begin{minipage}{0.68\columnwidth}
    \centering
    \begin{subfigure}[t]{0.49\linewidth}
      \centering
      \includegraphics[width=\textwidth]{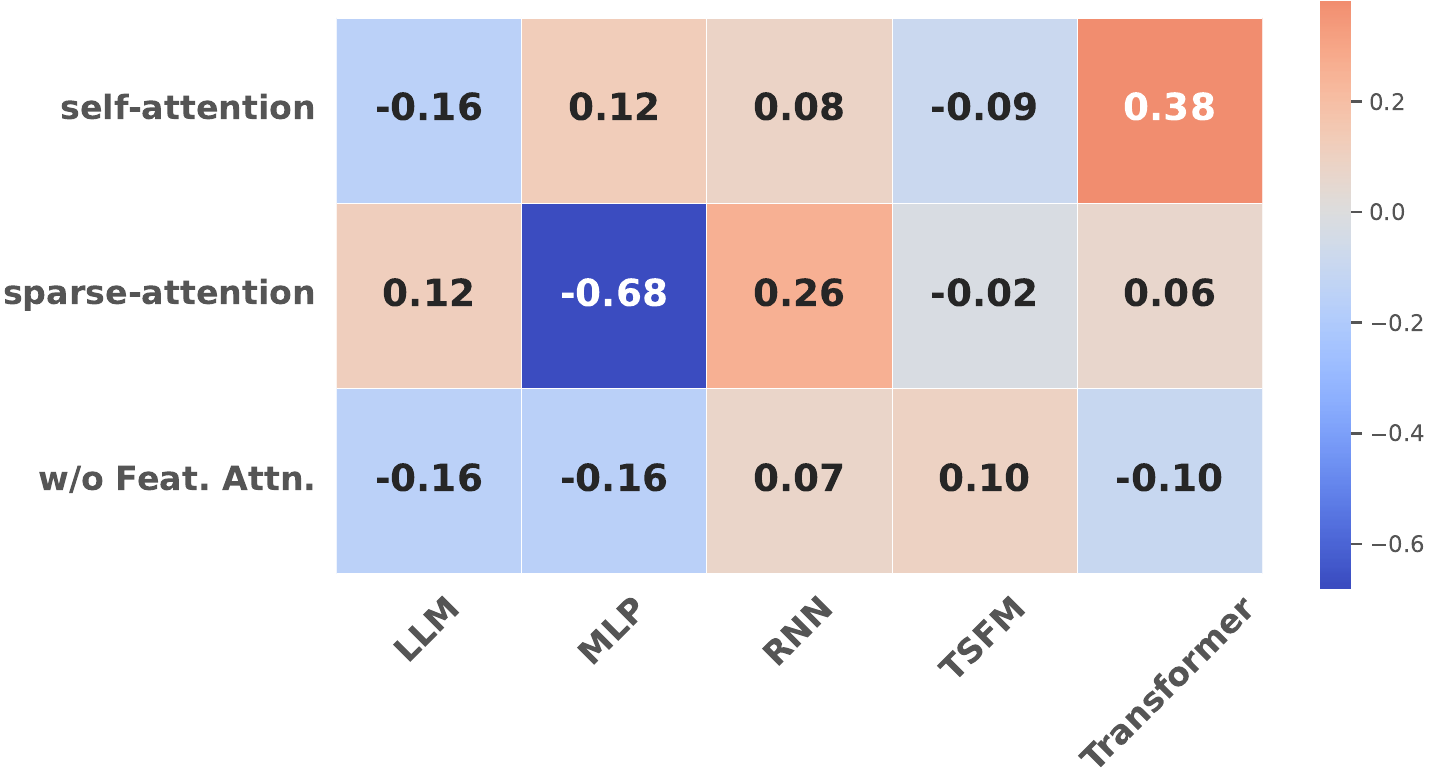}
      \caption{Arch vs. Attn}
      \label{fig:interaction-arch}
    \end{subfigure}
    \hfill
    \begin{subfigure}[t]{0.49\linewidth}
      \centering
      \includegraphics[width=\textwidth]{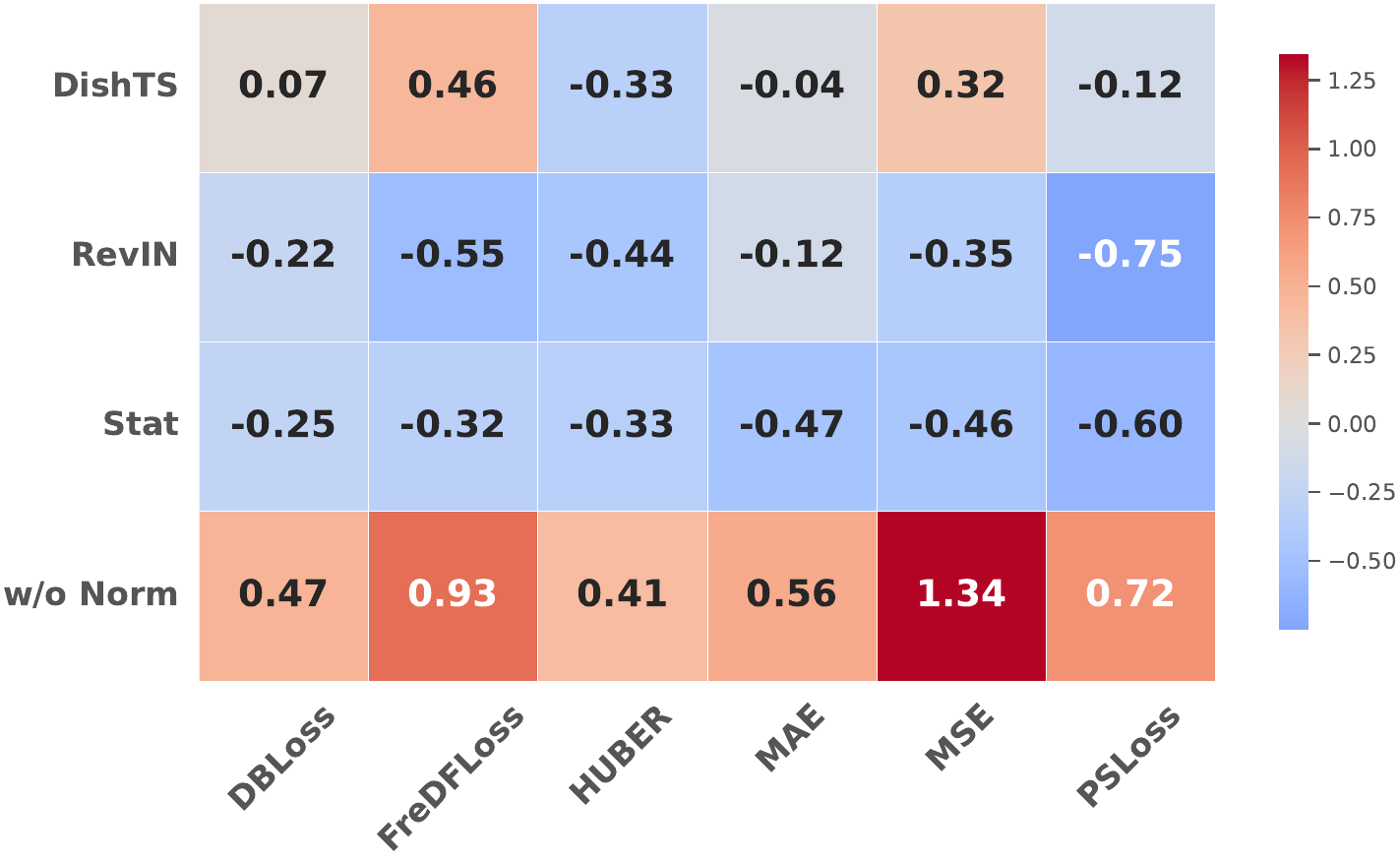}
      \caption{Loss vs. Norm}
      \label{fig:interaction-loss}
    \end{subfigure}
  \end{minipage}\hfill
  \begin{minipage}{0.31\columnwidth}
    \caption{Component interaction analysis. Blue indicates superior performance.}
    \label{fig:component-interaction}
  \end{minipage}
\end{figure}

Beyond individual component performance, we investigate pairwise interactions to identify synergistic or conflicting effects. Fig.~\ref{fig:component-interaction} visualizes the interaction between key components, using the average performance of combinations containing both components. Surprisingly, the combination of a simple \textit{MLP} with \textit{Sparse Feature Attention} yields superior performance (Fig.~\ref{fig:interaction-arch}), highlighting the efficacy of lightweight structures augmented with explicit feature correlation modeling. Conversely, Fig.~\ref{fig:interaction-loss} reveals that the standard \textit{MSE} loss performs poorly \textit{without Series Normalization}, demonstrating its sensitivity to distribution shifts and lack of robustness. \ifarxiv\add{Beyond these visual pairwise synergies, we explicitly model and quantify higher-order interactions in Appx.~\ref{appx:higher_order} to validate the additivity assumption underlying our main effect estimates.}\fi

\subsubsection{Dimension-Level}
\label{exp:dimension-level}
We quantify the relative importance of component dimensions via ANOVA methods, as illustrated in Table~\ref{tab:dimension-anova}.
\add{\textit{Series Normalization} emerges as the primary driver of performance within the contemporary design space, explaining 63.0\% of total performance variance. This substantially exceeds all other dimensions, suggesting proper normalization as a foundational element of effective forecasting.}
Secondly, series encoding dimensions (\textit{Channel Independence}: 11.1\%, \textit{Series Tokenization}: 7.1\%) collectively contribute 18.2\%, while network-related dimensions exhibit surprisingly limited aggregate impact—feature attention, RAG, loss function, and sequence length—despite being focal points in many MTSF studies.
\vspace{-0.5em}
\input{latex_table_useinpaper/main/tab_dimension_anova}
\vspace{-0.5em}

\subsubsection{Pipeline-Level}
\label{exp:pipeline-level}

We aggregate variance contributions across pipeline stages, as shown in Table~\ref{tab:dimension-anova}.
\add{Within the defined representative search space, \textit{Series Preprocessing} emerges as the most influential phase, accounting for 66.6\% of total explained variance—over eight times the contribution of \textit{Network Architecture} (8.0\%).}
\textit{Series Encoding} (18.3\%) plays a substantial role, explaining more than double the variance of \textit{Network Architecture}. Interestingly, \textit{Network Optimization} (7.1\%) and \textit{Network Architecture} exhibit comparable but limited influence, suggesting that optimization and architectural tuning yield diminishing returns once preprocessing and encoding are properly configured. \ifarxiv\add{Furthermore, Appx.~\ref{appx:preprocessing_dominance_metrics} confirms this preprocessing dominance is metric-invariant (persisting across MAE, RMSE, MASE) and scenario-robust, proving it is not an MSE artifact.}\fi

\subsection{Architecture-Specific Analysis}
\label{exp:architecture-specific}
\input{latex_table_useinpaper/main/tab_coef_comparison}

While the overall analysis illuminates global trends, different model architectures exhibit distinct preferences for pipeline components. We dissect these specific sensitivities in Table~\ref{tab:coef-comparison}, revealing that optimal configurations diverge significantly across model families.

\subsubsection{MLP-Based Architectures}
For MLP-based models, \textit{Channel Independence} demonstrates a substantial performance gain, suggesting that independent variable modeling simplifies the learning task. Regarding tokenization, while globally ineffective, structured projections like \textit{Orthogonal Encoding} show potential benefits for MLPs, implying that explicit feature construction might compensate for the lack of sequential processing capabilities. \textit{Series Normalization} dominates performance variance, as MLPs lack sequential modeling capabilities and thus rely heavily on proper series preprocessing to stabilize inputs.

\subsubsection{RNN-Based Architectures}
Sequential processing renders RNNs uniquely susceptible to inter-variable interference. Consequently, \textit{Channel Independence} yields disproportionate benefits, nearly doubling the global average, confirming that channel isolation prevents error propagation across variables in recurrent steps. Conversely, standard decomposition methods like \textit{Moving Average} significantly degrade performance, suggesting that smoothing operations remove critical short-term fluctuations that recurrent cells rely on for step-wise updates. \textit{Sequence Length} demonstrates substantial variance contribution, as longer horizons exacerbate gradient vanishing and error accumulation inherent in step-wise recurrent processing.

\subsubsection{Transformer-Based Architectures}
For Transformer-based models, three critical patterns emerge. First, \textit{Series Decomposition} methods significantly degrade performance, suggesting that frequency-domain or smoothing operations disrupt the attention mechanism's ability to capture temporal patterns. Second, \textit{Orthogonal Encoding} proves highly effective, indicating that structured tokenization enhances representation quality. Third, advanced \textit{Loss Functions} demonstrate substantial benefits, confirming that carefully designed objectives are critical for Transformer optimization. \textit{Loss Function} design dominates performance variance, as Transformers' complex attention mechanisms require carefully designed objectives to effectively learn temporal patterns.

\subsubsection{Large Time Series Models}
LLM-based models uniquely benefit from \textit{Moving Average Decomposition} and \textit{Multi-scale Mixing}, achieving coefficients of $-0.23$* (vs General $+0.25$*) and $-0.17$ (vs General $-0.09$*) respectively, despite mixing strategies originally designed for MLP architectures (e.g., TimeMixer). Unlike other architectures where \textit{Channel Independence} plays a significant role, LLMs are remarkably insensitive to it. However, they exhibit the highest sensitivity to \textit{Series Decomposition} among all model families. TSFMs exhibit extreme sensitivity to \textit{Series Tokenization} strategies. Critically, despite being commonly paired, \textit{Series Patching} and \textit{Channel Independence} exhibit opposite effects for TSFMs: patching degrades performance despite being the encoding used during pre-training, while channel independence proves effective, indicating these design choices should be decoupled to maximize pre-trained model adaptation.
\vspace{-0.5em}
\input{latex_table_useinpaper/main/tab_arch_combined}
\vspace{-0.5em}

\subsubsection{Intra-Backbone and Pipeline-Level Analysis}
Table~\ref{tab:arch-backbone-effectiveness} reveals that within-family performance varies substantially: \textit{MLP} and \textit{RNN} variants do not outperform their baselines, while \textit{Frequency Enhanced Attention} significantly improves \textit{Transformer} forecasting. Among large models, compared to \textit{GPT4TS}, other \textit{LLM} and \textit{TSFM} variants do not exhibit advantages in full-shot scenarios. Table~\ref{tab:arch-pipeline-variance} shifts focus to pipeline stages, demonstrating that architectural families prioritize different design phases: MLP relies heavily on \textit{Series Preprocessing}, TSFMs uniquely emphasize \textit{Series Encoding}, while Transformers and LLM-based methods heavily rely on \textit{Network Architecture} design.

\subsection{Data-Specific Analysis.}
\label{exp:data-specific}
\begin{figure}[h!]
  \centering
  \begin{subfigure}[t]{0.24\columnwidth}
    \centering
    \includegraphics[width=\textwidth]{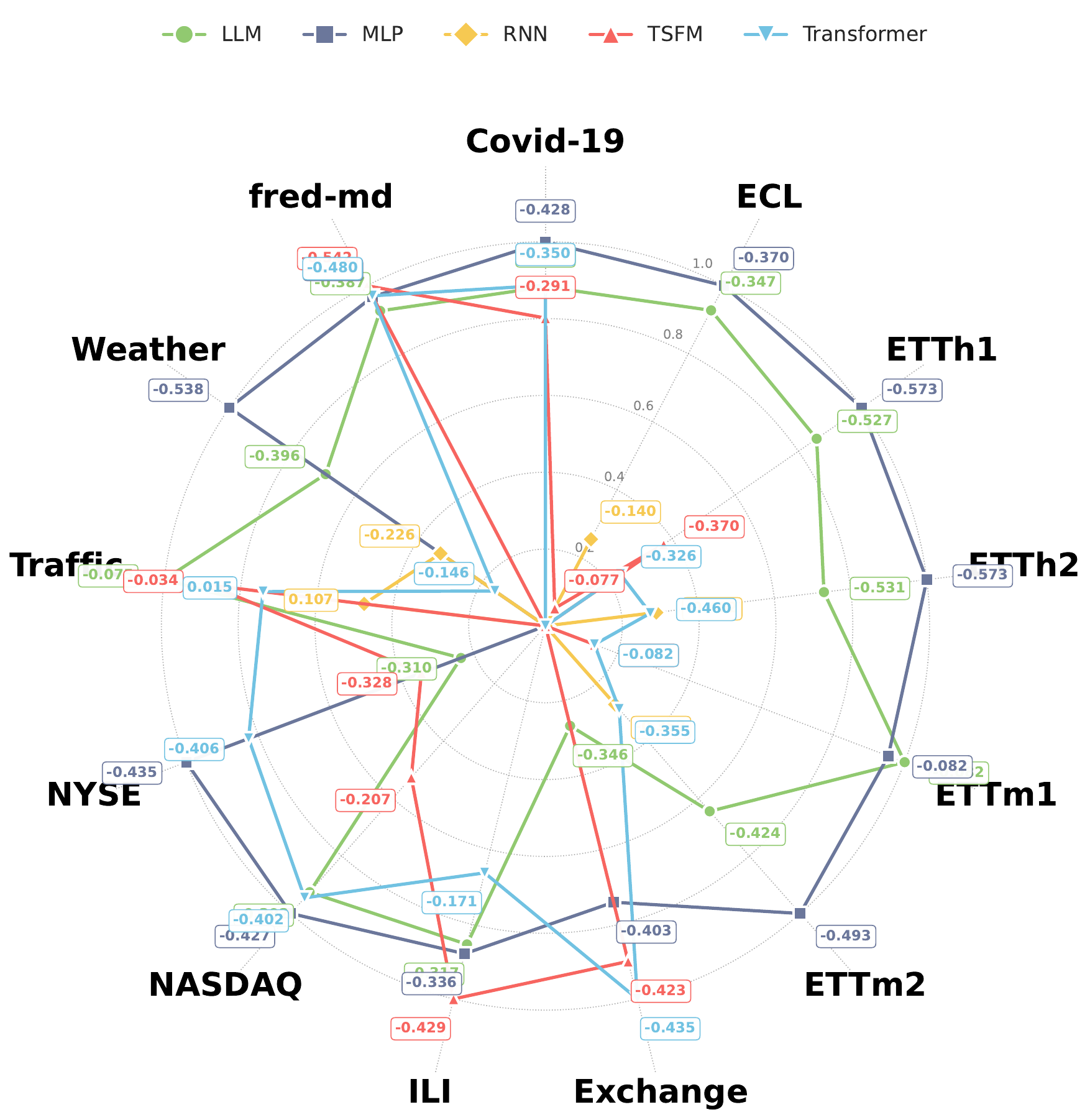}
    \caption{\scriptsize \shortstack[c]{Samples vs. \\ Backbone}}
    \label{fig:char-samples}
  \end{subfigure}
  \begin{subfigure}[t]{0.24\columnwidth}
    \centering
    \includegraphics[width=\textwidth]{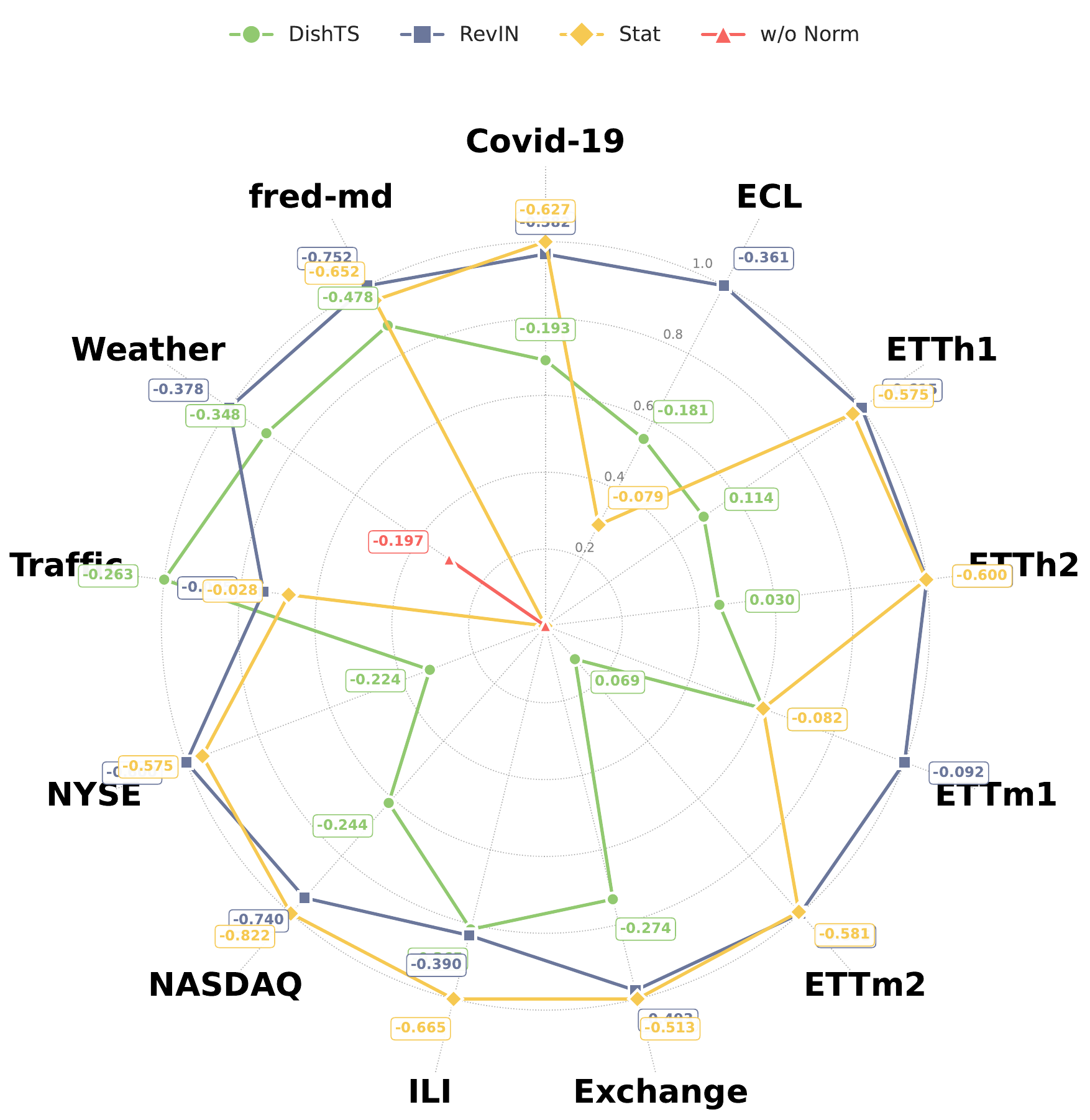}
    \caption{\scriptsize \shortstack[c]{Shift vs. \\ Normalization}}
    \label{fig:char-norm}
  \end{subfigure}
  \begin{subfigure}[t]{0.24\columnwidth}
    \centering
    \includegraphics[width=\textwidth]{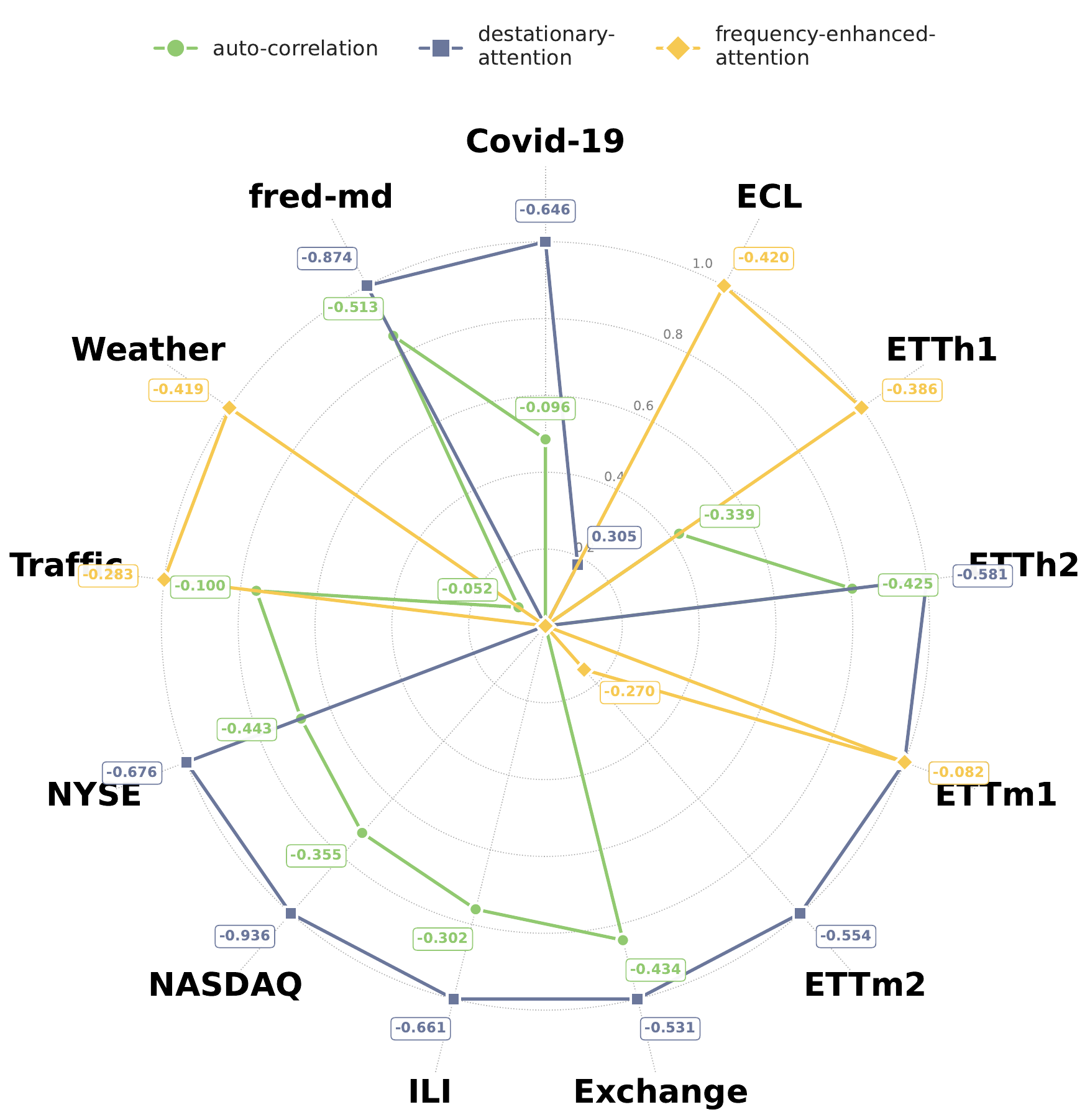}
    \caption{\scriptsize \shortstack[c]{Dynamics vs. \\  Attention}}
    \label{fig:char-attn}
  \end{subfigure}
  \begin{subfigure}[t]{0.24\columnwidth}
    \centering
    \includegraphics[width=\textwidth]{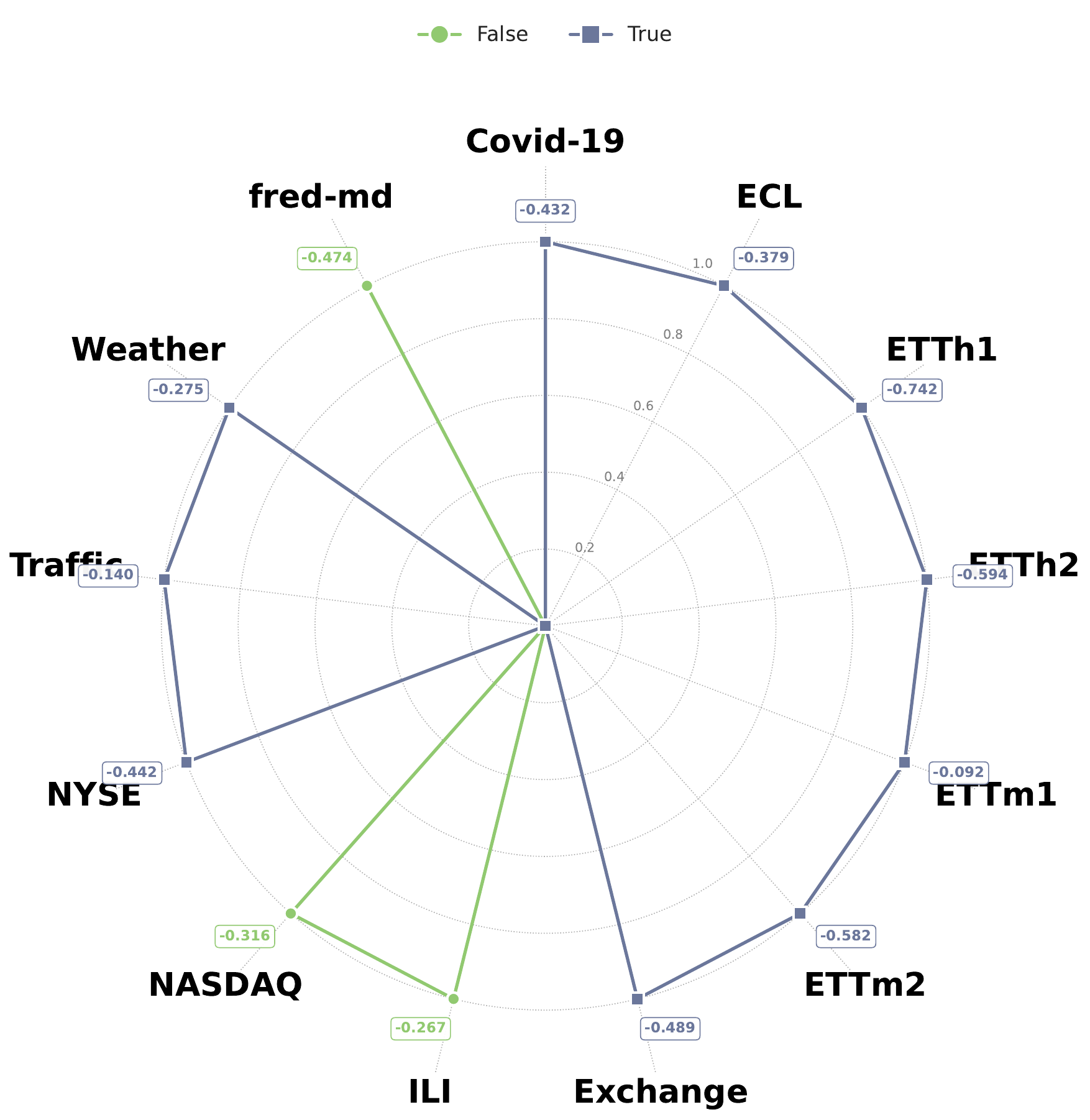}
    \caption{\scriptsize \shortstack[c]{Correlation vs. \\ Independence}}
    \label{fig:char-ci}
  \end{subfigure}
  \caption{Component adaptability to data characteristics. Larger coverage areas indicate better performance.}
  \label{fig:data-specific-radar}
\end{figure}

Beyond architecture-specific patterns, we examine how component effectiveness varies with dataset characteristics, specifically across five data properties~\cite{qiu2024tfb}: sample size, distribution shift, temporal dynamics, multivariate correlation, and stationarity\ifarxiv~ (Table~\ref{tab:meta_data_tfb})\fi.
By comparing top-$3$ and bottom-$3$ datasets for each property via mean difference tests with Cohen's d effect sizes, we observe distinct component preferences, as illustrated in Table~\ref{tab:data-specific-analysis} and Fig.~\ref{fig:data-specific-radar}.

\input{latex_table_useinpaper/main/tab_char_dataspec}
\vspace{-1.5em}
\vspace{-0.7em}

\noindent (i) \textbf{Simple models excel with sufficient samples} (Fig.~\ref{fig:char-samples}, Tab.~\ref{tab:data-specific-samples}): MLPs gain substantial performance on longer datasets as increased sample availability allows task-specific convergence to reliable patterns, whereas large models struggle, suggesting that excessive downstream adaptation may override pre-trained knowledge.

\noindent (ii) \textbf{Distribution shift demands stationarity} (Fig.~\ref{fig:char-norm}, Tab.~\ref{tab:data-specific-norm}): Standard instance normalization (RevIN) fails under high shift, necessitating stationarity-inducing methods like Stationary Norm to effectively mitigate non-stationary dynamics.

\noindent (iii) \textbf{Mechanism efficacy is context-dependent} (Fig.~\ref{fig:char-attn}, Tab.~\ref{tab:data-specific-attn}): Auto-correlation and Destationary Attention demonstrate relative performance improvements on datasets with high autocorrelation and strong non-stationarity, respectively. This confirms that attention mechanisms designed with specific prior assumptions effectively address their targeted problems.

\noindent (iv) \textbf{Multivariate dependency dictates channel strategy} (Fig.~\ref{fig:char-ci}, Tab.~\ref{tab:data-specific-ci}): Channel Independence (CI) significantly degrades performance on highly correlated datasets, suggesting that CI is not a universal solution and channel strategy should align with datasets.

\subsection{Automated Model Construction}
\label{exp:automated}

The preceding analyses (Sec.~\ref{exp:overall}-\ref{exp:data-specific}) substantiate that component efficacy is intrinsically coupled with architectural constraints and data characteristics. Leveraging these insights, we investigate whether the benchmark corpus generated by \system facilitates effective automated model construction. Specifically, we demonstrate the system's capability to synthesize optimal forecasting pipelines for unseen datasets in a zero-shot manner.

\subsubsection{Experimental Setup}
We constrain the search space to \textit{MLP-based} configurations, which Sec.~\ref{exp:architecture-specific} identifies as robust baselines. \ifarxiv\add{As Appx.~\ref{appx:backbone_expansion} details, incorporating RNNs and Transformers yields negligible gains since MLPs consistently perform optimally. Thus, restricting the search space to MLPs reduces computational overhead without sacrificing performance.}\fi Consequently, we sample $500$ candidate configurations for each of the $8$ evaluation datasets ($7$ long-term: ETT, Traffic, Weather, ECL; $1$ short-term: M4). The meta-predictor, trained on the performance corpus from historical datasets (introduced in Sec.~3.4), ranks these candidates to select optimal model combinations with minimal retraining. We evaluate long-term forecasting using MSE and MAE, and short-term forecasting using SMAPE, MASE, and OWA. The forecast horizons for long-term tasks are \( \{96, 192, 336, 720\} \), and for short-term tasks are \( \{6, 8, 13, 14, 18, 48\} \). Detailed information about the datasets and metrics can be found in Appx.~\ref{appx:data}.

\begin{figure}[t]
  \centering
  \begin{minipage}{0.5\columnwidth}
    \centering
    \includegraphics[width=\linewidth]{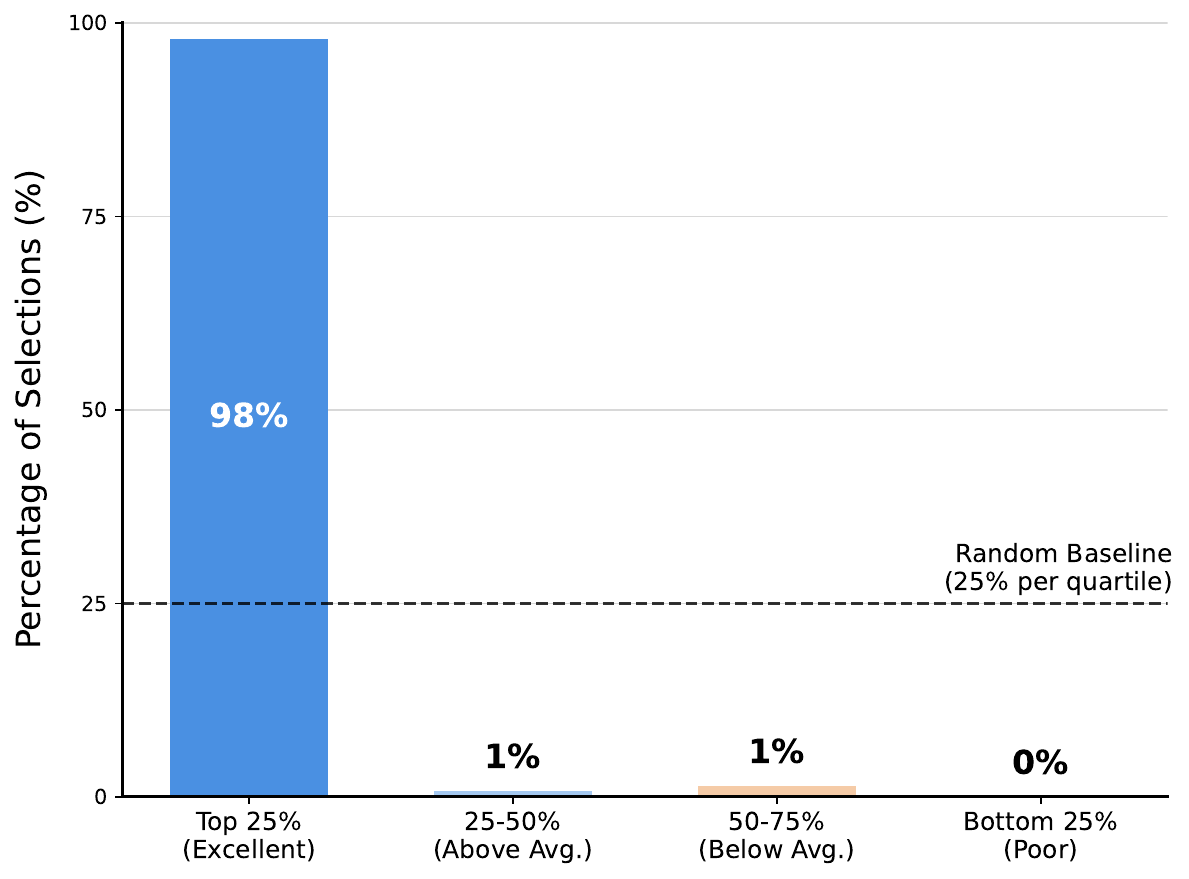}
  \end{minipage}%
  \hfill
  \begin{minipage}{0.45\columnwidth}
    \caption{Selection quality distribution of the meta-predictor's top-5 recommendations across all evaluation tasks.}
    \label{fig:selection-quality}
  \end{minipage}
\end{figure}

\subsubsection{Selection Quality Analysis}

We assess the meta-predictor's ranking capability based on its top-$5$ recommendations across test tasks, as shown in Fig.~\ref{fig:selection-quality}.
The selections are highly concentrated: 98\% fall within the top quartile, and over 99\% within the top half.
This substantially exceeds the random choice baseline, confirming that the benchmark corpus contains rich, learnable patterns—even a simple meta-predictor can achieve strong selection quality.

\subsubsection{Performance Comparison with SOTA}
\add{To validate the effectiveness of \system's automated model construction, we comprehensively compare it against deep MTSF models, AutoML methods, and large time-series models (LTSM). Due to space limitations, we present representative SOTA methods from recent research. These include deep MTSF models such as OLinear~\cite{yue2025olinear}, RAFT~\cite{han2025retrieval}, DUET~\cite{qiu2025DUET}, TimeMixer~\cite{wang2024timemixer}, TimeXer~\cite{wang2024timexer}, PAttn~\cite{liu2022pyraformer}, and iTransformer~\cite{liu2024itransformer}. We also evaluate AutoML methods (AutoGluon~\cite{shchur2023autogluon}, AutoTS~\cite{AutoTS_github}, TimeFuse~\cite{liu2025breaking}) and Large Time-series Models (GPT4TS~\cite{zhou2023one}, Timer~\cite{liu2024timer}, Moment~\cite{goswami2024moment}). \ifarxiv Complete results are provided in Appx.~\ref{appx:sota_complete_results}.\fi}

\input{Table_Exp/Exp_SOTA_Merged}

We compare \system against the existing SOTA deep MTSF models using the ensemble of top-$5$ selected model combinations, as is shown in Table~\ref{tab:tsgym_vs_sota_merged}.
\system dominates the M4 short-term forecasting task (Table~\ref{tab:tsgym_vs_sota_short}) and achieves state-of-the-art results on 10 out of 14 in long-term forecasting tasks (Table~\ref{tab:tsgym_vs_sota}).
Notably, these gains are achieved with simple MLPs, demonstrating that precise component selection matters more than architectural complexity.

\input{Table_Exp/Comparison_TimeFuse}
\add{Comparisons with automated baselines and large time-series models further underscore the advantages of \system. As illustrated in Table~\ref{tab:simplified_comparison}, \system outperforms the adaptive fusion framework TimeFuse by up to \textbf{10.4\%} in MSE, while also surpassing leading LTSMs like GPT4TS. Crucially, while these large models are evaluated in a full-shot setting, \system achieves these results through zero-shot component recommendation followed by fitting a lightweight MLP-based model. Consequently, the computational cost of our pipeline is significantly lower than the heavy overhead of full-shot fine-tuning required for large-scale models. These findings demonstrate that precise component selection can consistently outperform increasing model scale.}

Collectively, these results demonstrate that systematic component selection via our benchmark corpus enables competitive zero-shot performance without manual tuning, substantially lowering the barrier to effective time series forecasting.

%% file: latex_table_useinpaper/main/tab_dimension_anova.tex
\begin{table}[h!]
  \centering
  \caption{Dimension-Level ANOVA Analysis: Variance Explained by Design Dimensions.}
  \vspace{-0.5em}
  \label{tab:dimension-anova}
  \resizebox{\columnwidth}{!}{
    \begin{tabular}{lcc | lcc}
      \toprule
      \textbf{Design Dimension} & \textbf{Variance (\%)} & \textbf{p-val} & \textbf{Design Dimension} & \textbf{Variance (\%)} & \textbf{p-val} \\
      \midrule
      \multicolumn{3}{c|}{\cellcolor{gray!15}\textit{Series Preprocessing (Total: 66.6\%)}} & \multicolumn{3}{c}{\cellcolor{gray!15}\textit{Series Encoding (Total: 18.3\%)}} \\
      \textbf{Series Normalization} & $63.0^{***}$ & 0.000 & \textbf{Channel Independence} & $11.1^{***}$ & 0.000 \\
      \textbf{Series Decomposition} & $3.2^{***}$ & 0.000 & \textbf{Series Tokenization} & $7.1^{***}$ & 0.000 \\
      \textbf{Series Sampling/Mixing} & $0.4^{***}$ & 0.000 & \textbf{Timestamp Embedding} & $0.1$ & 0.051 \\
      \multicolumn{3}{c|}{\cellcolor{gray!15}\textit{Network Architecture (Total: 8.0\%)}} & \multicolumn{3}{c}{\cellcolor{gray!15}\textit{Network Optimization (Total: 7.1\%)}} \\
      \textbf{Feature Attention} & $3.2^{***}$ & 0.000 & \textbf{Sequence Length} & $1.6^{***}$ & 0.000 \\
      \textbf{Retrieval Augmented (RAG)} & $2.1^{***}$ & 0.000 & \textbf{Loss Function} & $5.4^{***}$ & 0.000 \\
      \bottomrule
    \end{tabular}
  }
\end{table}

%% file: latex_table_useinpaper/main/tab_coef_comparison.tex
\begin{table}[t] 
  \centering
  \caption{Coefficient Analysis Across Architectures.}
  \vspace{-0.5em}
  \label{tab:coef-comparison}
  \scriptsize
  \setlength{\tabcolsep}{2pt}
  \begin{tabular}{llcccccc}
    \toprule
    \textbf{Target} & \textbf{Metric} & \textbf{General} & \textbf{MLP} & \textbf{RNN} & \textbf{Transformer} & \textbf{LLM} & \textbf{TSFM} \\
    \midrule
    \cellcolor{gray!15}\textbf{Series Normalization} & \cellcolor{gray!15}Variance (\%) & \cellcolor{gray!15}63.0\rlap{*} & \cellcolor{gray!15}53.9\rlap{*} & \cellcolor{gray!15}39.0\rlap{*} & \cellcolor{gray!15}45.2\rlap{*} & \cellcolor{gray!15}45.1\rlap{*} & \cellcolor{gray!15}42.4\rlap{*} \\
    \hspace{1em}Stat & \multirow{3}{*}{Coef} & -1.16\rlap{*} & -0.93 & -0.97\rlap{*} & -1.10\rlap{*} & -1.06\rlap{*} & -1.01\rlap{*} \\
    \hspace{1em}RevIN & & -1.18\rlap{*} & -0.85\rlap{*} & -0.78\rlap{*} & -1.31\rlap{*} & -1.49\rlap{*} & -0.89\rlap{*} \\
    \hspace{1em}DishTS & & -0.63\rlap{*} & 0.13 & -0.85\rlap{*} & -0.44\rlap{*} & -1.11\rlap{*} & -0.45\rlap{*} \\
    \midrule
    \cellcolor{gray!15}\textbf{Series Decomposition} & \cellcolor{gray!15}Variance (\%) & \cellcolor{gray!15}3.2\rlap{*} & \cellcolor{gray!15}6.0 & \cellcolor{gray!15}3.8\rlap{*} & \cellcolor{gray!15}11.6\rlap{*} & \cellcolor{gray!15}12.1\rlap{*} & \cellcolor{gray!15}1.5\rlap{*} \\
    \hspace{1em}MA & \multirow{3}{*}{Coef} & 0.25\rlap{*} & 0.05 & 0.46\rlap{*} & 0.12 & -0.23\rlap{*} & 0.02 \\
    \hspace{1em}MoEMA & & 0.13\rlap{*} & -0.10 & 0.07 & 0.46\rlap{*} & 0.25 & -0.14\rlap{*} \\
    \hspace{1em}DFT & & 0.23\rlap{*} & 0.45 & -0.05 & 0.53\rlap{*} & 1.07\rlap{*} & 0.10 \\
    \midrule
    \cellcolor{gray!15}\textbf{Series Sampling/Mixing} & \cellcolor{gray!15}Variance (\%) & \cellcolor{gray!15}0.4\rlap{*} & \cellcolor{gray!15}1.9 & \cellcolor{gray!15}0.0 & \cellcolor{gray!15}0.1 & \cellcolor{gray!15}0.1 & \cellcolor{gray!15}0.9\rlap{*} \\
    \hspace{1em}w/ Mixing & Coef & -0.09\rlap{*} & -0.31 & 0.00 & 0.04 & -0.17 & -0.16\rlap{*} \\
    \midrule
    \cellcolor{gray!15}\textbf{Channel Independence} & \cellcolor{gray!15}Variance (\%) & \cellcolor{gray!15}11.1\rlap{*} & \cellcolor{gray!15}8.2\rlap{*} & \cellcolor{gray!15}20.3\rlap{*} & \cellcolor{gray!15}4.3\rlap{*} & \cellcolor{gray!15}0.2 & \cellcolor{gray!15}17.9\rlap{*} \\
    \hspace{1em}Channel Indepen & Coef & -0.52\rlap{*} & -0.78\rlap{*} & -0.83\rlap{*} & -0.44\rlap{*} & -0.48 & -0.85\rlap{*} \\
    \midrule
    \cellcolor{gray!15}\textbf{Series Tokenization} & \cellcolor{gray!15}Variance (\%) & \cellcolor{gray!15}7.1\rlap{*} & \cellcolor{gray!15}7.2 & \cellcolor{gray!15}0.6 & \cellcolor{gray!15}1.3 & \cellcolor{gray!15}13.5\rlap{*} & \cellcolor{gray!15}24.7\rlap{*} \\
    \hspace{1em}Series Patching & \multirow{3}{*}{Coef} & 0.17\rlap{*} & -0.44 & 0.12 & -0.07 & 0.70\rlap{*} & 0.64\rlap{*} \\
    \hspace{1em}Inverted Encoding & & -0.25\rlap{*} & - & - & -0.21 & -0.15 & -0.28\rlap{*} \\
    \hspace{1em}Ortho Encoding & & -0.29\rlap{*} & -1.21 & -0.14 & -0.31\rlap{*} & 0.36 & -0.25\rlap{*} \\
    \midrule
    \cellcolor{gray!15}\textbf{Timestamp Embedding} & \cellcolor{gray!15}Variance (\%) & \cellcolor{gray!15}0.1 & \cellcolor{gray!15}3.0 & \cellcolor{gray!15}0.4 & \cellcolor{gray!15}0.3 & \cellcolor{gray!15}0.2 & \cellcolor{gray!15}0.6 \\
    \hspace{1em}w/ Embedding & Coef & -0.05 & -0.36 & -0.05 & -0.08 & 0.06 & 0.11 \\
    \midrule
    \cellcolor{gray!15}\textbf{Feature Attention} & \cellcolor{gray!15}Variance (\%) & \cellcolor{gray!15}3.2\rlap{*} & \cellcolor{gray!15}1.9 & \cellcolor{gray!15}4.7\rlap{*} & \cellcolor{gray!15}1.0 & \cellcolor{gray!15}3.9\rlap{*} & \cellcolor{gray!15}4.1\rlap{*} \\
    \hspace{1em}SelfAttn & \multirow{2}{*}{Coef} & -0.17\rlap{*} & -0.34 & -0.14 & -0.09 & -0.26 & -0.21\rlap{*} \\
    \hspace{1em}SparseAttn & & -0.24\rlap{*} & -0.20 & -0.42\rlap{*} & -0.18\rlap{*} & -0.49\rlap{*} & -0.28\rlap{*} \\
    \midrule
    \cellcolor{gray!15}\textbf{Retrieval Augmented} & \cellcolor{gray!15}Variance (\%) & \cellcolor{gray!15}2.1\rlap{*} & \cellcolor{gray!15}1.2 & \cellcolor{gray!15}0.2 & \cellcolor{gray!15}8.8\rlap{*} & \cellcolor{gray!15}0.3 & \cellcolor{gray!15}0.9\rlap{*} \\
    \hspace{1em}w/ RAG & Coef & 0.20\rlap{*} & 0.47 & -0.09 & 0.48\rlap{*} & 0.41 & 0.14\rlap{*} \\
    \midrule
    \cellcolor{gray!15}\textbf{Sequence Length} & \cellcolor{gray!15}Variance (\%) & \cellcolor{gray!15}1.6\rlap{*} & \cellcolor{gray!15}6.9 & \cellcolor{gray!15}14.1\rlap{*} & \cellcolor{gray!15}1.1 & \cellcolor{gray!15}9.6\rlap{*} & \cellcolor{gray!15}1.6\rlap{*} \\
    \hspace{1em}96 & \multirow{3}{*}{Coef} & -0.12\rlap{*} & -0.37 & 0.51\rlap{*} & -0.14\rlap{*} & 0.61\rlap{*} & -0.20\rlap{*} \\
    \hspace{1em}192 & & -0.00 & -1.06 & 0.46\rlap{*} & -0.09 & 1.19\rlap{*} & -0.12 \\
    \hspace{1em}512 & & 0.07\rlap{*} & -0.52 & 0.58\rlap{*} & -0.02 & 0.47\rlap{*} & -0.06 \\
    \midrule
    \cellcolor{gray!15}\textbf{Loss Function} & \cellcolor{gray!15}Variance (\%) & \cellcolor{gray!15}5.4\rlap{*} & \cellcolor{gray!15}7.4 & \cellcolor{gray!15}15.4\rlap{*} & \cellcolor{gray!15}20.9\rlap{*} & \cellcolor{gray!15}14.9\rlap{*} & \cellcolor{gray!15}5.3\rlap{*} \\
    \hspace{1em}MAE & \multirow{5}{*}{Coef} & -0.33\rlap{*} & -0.37 & 0.50\rlap{*} & -0.39\rlap{*} & -0.52\rlap{*} & -0.07 \\
    \hspace{1em}HUBER & & -0.36\rlap{*} & 0.21 & -0.47 & -0.69\rlap{*} & -0.23\rlap{*} & -0.31\rlap{*} \\
    \hspace{1em}DBLoss & & -0.17\rlap{*} & -0.43 & 0.26\rlap{*} & -0.40\rlap{*} & 0.17 & -0.26\rlap{*} \\
    \hspace{1em}PSLoss & & -0.24\rlap{*} & -0.08 & -0.06 & -0.67\rlap{*} & -1.32\rlap{*} & 0.08 \\
    \hspace{1em}FreDFLoss & & -0.11\rlap{*} & -0.35 & 0.22 & -0.61\rlap{*} & -0.62 & -0.01 \\
    \midrule
    \textbf{Backbone Choice} & Coef & - & 0.00 & 0.06\rlap{*} & 0.14\rlap{*} & 0.04 & 0.06 \\
    \bottomrule
  \end{tabular}
  \\
  \begin{minipage}{\linewidth}
    \raggedright
    \footnotesize
    \textit{Note:} * indicates p < 0.05. General = General Coefficient.
  \end{minipage}
\end{table}


%% file: latex_table_useinpaper/main/tab_arch_combined.tex
\begin{table}[h!]
  \centering
  \caption{Architecture-Specific Analysis: Pipeline Stage Importance and Backbone Variant Effectiveness.}
  \label{tab:arch-combined}
  \vspace{-1em}
  \begin{minipage}[t]{\linewidth}
    \centering
    \subcaption{Effectiveness of Specific Backbone Variants (GLMM Coef)}
    \label{tab:arch-backbone-effectiveness}
    \vspace{-0.5em}
    \resizebox{0.9\linewidth}{!}{
      \scriptsize
      \begin{tabular}{lc | lc | lc}
        \toprule
        \textbf{Variant} & \textbf{Coef} & \textbf{Variant} & \textbf{Coef} & \textbf{Variant} & \textbf{Coef} \\
        \midrule
        \multicolumn{2}{c|}{\cellcolor{gray!10}\textit{\textbf{MLP-Based}}} & \multicolumn{2}{c|}{\cellcolor{gray!10}\textit{\textbf{Transformer}}} & \multicolumn{2}{c}{\cellcolor{gray!10}\textit{\textbf{Large Models}}} \\
        DNN & 0.00 & SelfAttn & 0.00 & GPT4TS & 0.00 \\
        NormLin & 0.30 & AutoCorr & 0.08 & TimeLLM & 0.34\rlap{*} \\
        \multicolumn{2}{c|}{\cellcolor{gray!10}\textit{\textbf{RNN-Based}}} & SparseAttn & -0.11 & TimeMoE & 0.29\rlap{*} \\
        GRU & 0.00 & FreqAttn & -0.25\rlap{*} & Chronos & 0.03 \\
        xLSTM & 0.29 & Destationary & -0.01 & Timer & 0.03 \\
        & & & & Moment & 0.21\rlap{*} \\
        \bottomrule
      \end{tabular}
    }
  \end{minipage}

  \vspace{0em}

  \begin{minipage}[h!]{\linewidth}
    \centering
    \subcaption{Pipeline Stage Variance Contribution (\%)}
    \label{tab:arch-pipeline-variance}
    \vspace{-0.5em}
    \resizebox{0.95\linewidth}{!}{
      \scriptsize
      \begin{tabular}{lccccc}
        \toprule
        \textbf{Pipeline Stage} & \textbf{MLP} & \textbf{RNN} & \textbf{Trans} & \textbf{LLM} & \textbf{TSFM} \\
        \midrule
        \textit{Series Preprocessing} & 61.7 & 43.2 & 56.8 & 57.4 & 44.8 \\
        \textit{Series Encoding} & 18.3 & 20.3 & 5.9 & 13.9 & 43.2 \\
        \textit{Network Architecture} & 5.6 & 5.9 & 15.4 & 4.3 & 5.1 \\
        \textit{Network Optimization} & 14.3 & 30.5 & 21.9 & 24.5 & 6.9 \\
        \bottomrule
      \end{tabular}
    }
  \end{minipage}

  \begin{minipage}{\linewidth}
    \vspace{0.5em}
    \scriptsize
    \textit{Note:} Top panel: Performance coefficients of specific backbone variants relative to the category baseline. Bottom panel: Variance explained by pipeline stages for each architecture family.
  \end{minipage}
\end{table}

%% file: latex_table_useinpaper/main/tab_char_dataspec.tex

\begin{table}[h!]
  \centering
  \caption{Data-Specific Component Analysis: Effect of Data Characteristics on Performance (Mean Difference Test).}
  \label{tab:data-specific-analysis}
  \small
  \vspace{-1em}
  \begin{minipage}[t]{0.49\columnwidth}
    \centering
    \subcaption{\scriptsize Network Backbone vs. Sample Size}
    \label{tab:data-specific-samples}
    \resizebox{\linewidth}{!}{
      \begin{tabular}{lrrr}
        \toprule
        \textbf{Backbone} & \textbf{Diff} & \textbf{d} & \textbf{p-val} \\
        \midrule
        MLP & -0.17 & -0.19 & 0.017\rlap{*} \\
        RNN & -0.16 & -0.14 & 0.076 \\
        Transformer & +0.03 & 0.03 & 0.575 \\
        LLM & -0.12 & -0.11 & 0.220 \\
        TSFM & +0.21 & 0.21 & <.001\rlap{*} \\
        \bottomrule
      \end{tabular}
    }
  \end{minipage}
  \hfill
  \begin{minipage}[t]{0.49\columnwidth}
    \centering
    \subcaption{\scriptsize Series Norm. vs. Distribution Shift}
    \label{tab:data-specific-norm}
    \resizebox{\linewidth}{!}{
      \begin{tabular}{lrrr}
        \toprule
        \textbf{Norm Method} & \textbf{Diff} & \textbf{d} & \textbf{p-val} \\
        \midrule
        w/o Norm & +0.31 & 0.31 & <.001\rlap{*} \\
        RevIN & +0.11 & 0.12 & 0.068 \\
        DishTS & -0.09 & -0.09 & 0.106 \\
        Stationary & -0.24 & -0.30 & <.001\rlap{*} \\
        \bottomrule
      \end{tabular}
    }
  \end{minipage}

  \vspace{0.5em}

  \begin{minipage}[t]{0.49\columnwidth}
    \centering
    \subcaption{\scriptsize Attention Mechanism vs. Dynamics}
    \label{tab:data-specific-attn}
    \resizebox{\linewidth}{!}{
      \begin{tabular}{llrrr}
        \toprule
        \textbf{Mechanism} & \textbf{Char.} & \textbf{Diff} & \textbf{d} & \textbf{p-val} \\
        \midrule
        Auto-Corr & Trans. & -0.30 & -0.31 & 0.008\rlap{*} \\
        Destationary & Unstat. & -0.74 & -1.29 & <.001\rlap{*} \\
        \bottomrule
      \end{tabular}
    }
  \end{minipage}
  \hfill
  \begin{minipage}[t]{0.49\columnwidth}
    \centering
    \subcaption{\scriptsize Channel Independence vs. Correlation}
    \label{tab:data-specific-ci}
    \resizebox{\linewidth}{!}{
      \begin{tabular}{lrrr}
        \toprule
        \textbf{Mode} & \textbf{Diff} & \textbf{d} & \textbf{p-val} \\
        \midrule
        Independent & +0.22 & 0.31 & <.001\rlap{*} \\
        Mixing & -0.07 & -0.07 & 0.053 \\
        \bottomrule
      \end{tabular}
    }
  \end{minipage}

  \vspace{0.2em}
  \begin{minipage}{\columnwidth}
    \scriptsize
    \textit{Note:} \textbf{Diff} = Mean(High Characteristic) - Mean(Low Characteristic). Positive Diff indicates performance degradation on datasets with high characteristic intensity; Negative Diff indicates improvement. \textbf{d} = Cohen's d effect size. * indicates p < 0.05.
  \end{minipage}
\end{table}

%% file: Table_Exp/Exp_SOTA_Merged.tex
\begin{table}[h!]
  \centering
  \caption{Comparison with state-of-the-art deep MTSF models.}
  \label{tab:tsgym_vs_sota_merged}
  \vspace{-1em}
  \begin{subtable}{\columnwidth}
    \centering
    \caption{Short-term forecasting performance}
    \vspace{-0.3em}
    \label{tab:tsgym_vs_sota_short}
    \resizebox{\columnwidth}{!}{
      \setlength{\tabcolsep}{1.5pt}
      \begin{tabular}{ccccccccc}
        \toprule
        \multicolumn{1}{c}{\textbf{Models}} & \multicolumn{1}{c}{\textbf{\system (Ours)}} & \multicolumn{1}{c}{\textbf{OLinear}} & \multicolumn{1}{c}{\textbf{RAFT}} & \multicolumn{1}{c}{\textbf{DUET}} & \multicolumn{1}{c}{\textbf{TimeMixer}} & \multicolumn{1}{c}{\textbf{TimeXer}} & \multicolumn{1}{c}{\textbf{PAttn}} & \multicolumn{1}{c}{\textbf{iTransformer}} \\
        \midrule
        OWA & \textbf{\textcolor{red}{0.869}} & 1.651 & 1.257 & 0.958 & \underline{\textcolor{blue}{0.875}} & 0.919 & 0.916 & 0.959 \\
        SMAPE & \underline{\textcolor{blue}{12.004}} & 21.972 & 12.784 & 12.816 & \textbf{\textcolor{red}{11.880}} & 12.289 & 12.367 & 12.793 \\
        MASE & \textbf{\textcolor{red}{1.574}} & 3.122 & 3.250 & 1.750 & \underline{\textcolor{blue}{1.604}} & 1.677 & 1.683 & 1.757 \\
        \bottomrule
    \end{tabular}}
    \par\vspace{2pt}
    \parbox{\linewidth}{\footnotesize Results are averaged across diverse sampling intervals. We highlight the \textcolor{red}{\textbf{1st}} and \underline{\textcolor{blue}{2nd}} best results. \ifarxiv See Table~\ref{tab:TSGym_vs_Sota_short_full} for full details. \fi}
  \end{subtable}

  \par\vspace{5pt}
  \begin{subtable}{\columnwidth}
    \centering
    \caption{Long-term forecasting performance.}
    \vspace{-0.3em}
    \label{tab:tsgym_vs_sota}
    \resizebox{\columnwidth}{!}{
      \setlength{\tabcolsep}{1.5pt}
      \begin{tabular}{c|r@{\hspace{3pt}}r|r@{\hspace{3pt}}r|r@{\hspace{3pt}}r|r@{\hspace{3pt}}r|r@{\hspace{3pt}}r|r@{\hspace{3pt}}r|r@{\hspace{3pt}}r|r@{\hspace{3pt}}r}
        \toprule
        \multicolumn{1}{c}{\textbf{Models}} & \multicolumn{2}{c}{\scalebox{0.85}{\textbf{\system (Ours)}}} & \multicolumn{2}{c}{\scalebox{0.85}{\textbf{OLinear}}} & \multicolumn{2}{c}{\scalebox{0.85}{\textbf{RAFT}}} & \multicolumn{2}{c}{\scalebox{0.85}{\textbf{DUET}}} & \multicolumn{2}{c}{\scalebox{0.85}{\textbf{TimeMixer}}} & \multicolumn{2}{c}{\scalebox{0.85}{\textbf{TimeXer}}} & \multicolumn{2}{c}{\scalebox{0.85}{\textbf{PAttn}}} & \multicolumn{2}{c}{\scalebox{0.85}{\textbf{iTransformer}}} \\
        \midrule
        Metric & MSE & MAE & MSE & MAE & MSE & MAE & MSE & MAE & MSE & MAE & MSE & MAE & MSE & MAE & MSE & MAE \\
        \midrule
        \textbf{ETTh1}& \textbf{\textcolor{red}{0.407}} & \textbf{\textcolor{red}{0.424}} & 0.426 & \underline{\textcolor{blue}{0.426}} & \underline{\textcolor{blue}{0.421}} & 0.436 & 0.438 & 0.443 & 0.444 & 0.436 & 0.458 & 0.448 & 0.472 & 0.454 & 0.451 & 0.446 \\
        \textbf{ETTh2}& \textbf{\textcolor{red}{0.336}} & \textbf{\textcolor{red}{0.384}} & 0.367 & \underline{\textcolor{blue}{0.388}} & 0.362 & 0.409 & \underline{\textcolor{blue}{0.358}} & 0.393 & 0.387 & 0.408 & 0.376 & 0.402 & 0.387 & 0.412 & 0.382 & 0.407 \\
        \textbf{ETTm1}& \textbf{\textcolor{red}{0.341}} & \textbf{\textcolor{red}{0.371}} & 0.374 & \underline{\textcolor{blue}{0.377}} & \underline{\textcolor{blue}{0.348}} & 0.378 & 0.352 & 0.381 & 0.382 & 0.397 & 0.383 & 0.398 & 0.385 & 0.400 & 0.418 & 0.416 \\
        \textbf{ETTm2}& \textbf{\textcolor{red}{0.246}} & \textbf{\textcolor{red}{0.306}} & 0.271 & \underline{\textcolor{blue}{0.314}} & \underline{\textcolor{blue}{0.256}} & 0.320 & 0.259 & 0.316 & 0.281 & 0.327 & 0.280 & 0.325 & 0.289 & 0.335 & 0.289 & 0.331 \\
        \textbf{Weather}& \textbf{\textcolor{red}{0.222}} & \textbf{\textcolor{red}{0.256}} & 0.238 & \underline{\textcolor{blue}{0.260}} & 0.240 & 0.286 & \underline{\textcolor{blue}{0.232}} & 0.261 & 0.244 & 0.274 & 0.242 & 0.272 & 0.257 & 0.280 & 0.261 & 0.281 \\
        \textbf{ECL}& 0.161 & 0.253 & 0.159 & \underline{\textcolor{blue}{0.248}} & \textbf{\textcolor{red}{0.156}} & 0.253 & \underline{\textcolor{blue}{0.156}} & \textbf{\textcolor{red}{0.247}} & 0.185 & 0.274 & 0.172 & 0.270 & 0.205 & 0.286 & 0.175 & 0.266 \\
        \textbf{Traffic}& 0.405 & 0.271 & 0.448 & \textbf{\textcolor{red}{0.247}} & \underline{\textcolor{blue}{0.401}} & 0.281 & \textbf{\textcolor{red}{0.396}} & \underline{\textcolor{blue}{0.256}} & 0.502 & 0.307 & 0.467 & 0.288 & 0.513 & 0.328 & 0.422 & 0.282 \\
        \midrule
        \multicolumn{1}{c}{\textbf{$1^{\text{st}}$ Count}} & \multicolumn{2}{c}{\textbf{\textcolor{red}{10}}} & \multicolumn{2}{c}{1} & \multicolumn{2}{c}{1} & \multicolumn{2}{c}{\underline{\textcolor{blue}{2}}} & \multicolumn{2}{c}{0} & \multicolumn{2}{c}{0} & \multicolumn{2}{c}{0} & \multicolumn{2}{c}{0} \\
        \bottomrule
      \end{tabular}
    }
    \par\vspace{3pt}
    \parbox{\linewidth}{\footnotesize Results are averaged across four prediction horizons. We highlight the \textcolor{red}{\textbf{1st}} and \underline{\textcolor{blue}{2nd}} best results. \ifarxiv Refer to Table~\ref{tab:TSGym_vs_Sota_fullAvg} for complete results. \fi}
  \end{subtable}
\end{table}

%% file: Table_Exp/Comparison_TimeFuse.tex
\begin{table}[h!]
  \centering
  \caption{\add{Comparison with AutoML and LTSM.}}
  \vspace{-0.8em}
  \label{tab:simplified_comparison}
  \resizebox{\columnwidth}{!}{
    \setlength{\tabcolsep}{1.0pt}
    \begin{tabular}{c|r@{\hspace{2pt}}r|r@{\hspace{2pt}}r|r@{\hspace{2pt}}r|r@{\hspace{2pt}}r|r@{\hspace{2pt}}r|r@{\hspace{2pt}}r|r@{\hspace{2pt}}r|r@{\hspace{2pt}}r}
      \toprule
      \multicolumn{1}{c}{\multirow{2}{*}{\textbf{Models}}} & \multicolumn{10}{c}{\textbf{AutoML}} & \multicolumn{6}{c}{\textbf{\add{LTSM}}} \\
      \cmidrule(lr){2-11} \cmidrule(lr){12-17}
      \multicolumn{1}{c}{} & \multicolumn{2}{c}{\scalebox{0.75}{\textbf{\system (Ours)}}} & \multicolumn{2}{c}{\scalebox{0.75}{\textbf{TimeFuse (Zero)}}} & \multicolumn{2}{c}{\scalebox{0.75}{\textbf{TimeFuse (Few)}}} & \multicolumn{2}{c}{\scalebox{0.75}{\textbf{AutoGluon}}} & \multicolumn{2}{c}{\scalebox{0.75}{\textbf{AutoTS}}} & \multicolumn{2}{c}{\scalebox{0.75}{\textbf{GPT4TS}}} & \multicolumn{2}{c}{\scalebox{0.75}{\textbf{Timer}}} & \multicolumn{2}{c}{\scalebox{0.75}{\textbf{Moment}}} \\
      \midrule
      Metric & MSE & MAE & MSE & MAE & MSE & MAE & MSE & MAE & MSE & MAE & MSE & MAE & MSE & MAE & MSE & MAE \\
      \midrule
      \textbf{ETTh1} & \textbf{\textcolor{red}{0.407}} & \textbf{\textcolor{red}{0.424}} & \underline{\textcolor{blue}{0.427}} & 0.434 & 0.439 & 0.437 & 0.503 & 0.473 & 0.981 & 0.610 & 0.428 & \underline{\textcolor{blue}{0.426}} & 0.472 & 0.480 & 0.649 & 0.547 \\
      \textbf{ETTh2} & \textbf{\textcolor{red}{0.336}} & \textbf{\textcolor{red}{0.384}} & 0.386 & 0.415 & 0.380 & 0.408 & 0.419 & 0.430 & 0.589 & 0.488 & \underline{\textcolor{blue}{0.354}} & \underline{\textcolor{blue}{0.395}} & 0.381 & 0.425 & 0.572 & 0.531 \\
      \textbf{ETTm1} & \textbf{\textcolor{red}{0.341}} & \textbf{\textcolor{red}{0.371}} & 0.363 & 0.386 & 0.370 & 0.391 & 0.482 & 0.408 & 0.744 & 0.546 & \underline{\textcolor{blue}{0.352}} & \underline{\textcolor{blue}{0.383}} & 0.372 & 0.395 & 0.403 & 0.411 \\
      \textbf{ETTm2} & \textbf{\textcolor{red}{0.246}} & \textbf{\textcolor{red}{0.306}} & 0.277 & 0.325 & 0.272 & \underline{\textcolor{blue}{0.323}} & 0.273 & 0.337 & 0.392 & 0.389 & \underline{\textcolor{blue}{0.267}} & 0.326 & 0.274 & 0.335 & 0.320 & 0.361 \\
      \textbf{ECL} & \textbf{\textcolor{red}{0.161}} & \textbf{\textcolor{red}{0.253}} & 0.169 & 0.270 & 0.182 & 0.272 & 0.265 & 0.328 & 0.327 & 0.355 & \underline{\textcolor{blue}{0.167}} & \underline{\textcolor{blue}{0.263}} & 0.231 & 0.317 & 0.171 & 0.270 \\
      \textbf{Traffic} & \textbf{\textcolor{red}{0.405}} & \textbf{\textcolor{red}{0.271}} & 0.471 & 0.296 & 0.501 & 0.306 & 0.555 & 0.325 & 0.739 & 0.311 & \underline{\textcolor{blue}{0.414}} & 0.294 & 0.644 & 0.400 & \underline{\textcolor{blue}{0.414}} & \underline{\textcolor{blue}{0.289}} \\
      \textbf{Weather} & \textbf{\textcolor{red}{0.222}} & \textbf{\textcolor{red}{0.256}} & 0.233 & 0.270 & 0.236 & 0.270 & 0.236 & 0.270 & 0.519 & 0.372 & 0.236 & 0.271 & 0.335 & 0.365 & \underline{\textcolor{blue}{0.228}} & \underline{\textcolor{blue}{0.268}} \\
      \midrule
      \multicolumn{1}{c}{\textbf{{$1^{\text{st}}$ Count}}} & \multicolumn{2}{c}{\textbf{\textcolor{red}{14}}} & \multicolumn{2}{c}{0} & \multicolumn{2}{c}{0} & \multicolumn{2}{c}{0} & \multicolumn{2}{c}{0} & \multicolumn{2}{c}{0} & \multicolumn{2}{c}{0} & \multicolumn{2}{c}{0} \\
      \bottomrule
    \end{tabular}
  }
\end{table}

%% file: 5discussion.tex
To advance beyond holistic evaluations in multivariate time-series forecasting (MTSF), this paper introduced \system, a novel framework centered on fine-grained component analysis and the automated construction of specialized forecasting models. Through systematic decomposition of MTSF pipelines into component dimensions and design choices, \system uncovers crucial insights into component-level performance analysis and facilitates automated construction of customized models. Extensive experimental results indicate that the MTSF models constructed by the
proposed \system significantly outperform current MTSF SOTA solutions, demonstrating the advantage of adaptively customizing models according to distinct data characteristics. Our results show that \system is highly effective, even without exhaustively covering all SOTA components, and we release our benchmark code, results, and performance corpus to benefit the MTSF community.
\add{Our future work will establish an LLM-agent-powered workflow to prevent the performance corpus from becoming an outdated static snapshot. This workflow will automatically construct a continuous knowledge base from emerging papers and systematically decompose their novel architectures into reusable components, thereby enabling self-evolving model synthesis.}

%% file: 6appendix.tex
\section{Datasets}\label{appx:data}
We conduct extensive evaluations on \ndatasets standard long-term forecasting benchmarks: four ETT variants (ETTh1, ETTh2, ETTm1, ETTm2), Electricity (abbreviated as ECL), Traffic, Weather, Exchange, ILI, FRED-MD, NASDAQ, NYSE, and Covid-19, complemented by the M4 dataset for short-term forecasting tasks, with complete dataset specifications provided in Table~\ref{tab:datasets}. Furthermore, Table~\ref{tab:meta_data_tfb}~\cite{qiu2024tfb} details the meta-data characteristics of these datasets, such as trend, seasonality, and stationarity metrics.

The forecast horizon \( L \) is set to \( \{96, 192, 336, 720\} \) for standard long-term tasks, while datasets with limited samples (ILI, NYSE, NASDAQ, Fred-MD, Covid-19) adopt \( \{24, 36, 48, 60\} \).
For M4, the horizons are \( \{6, 8, 13, 14, 18, 48\} \).

\input{Tables/datasets}

\input{latex_table_useinpaper/appendix/datasets_char_fromTFB}

\section{Metrics Mathematical Formula}\label{appx:metrics}
\noindent \textbf{Evaluation Metrics}. We follow the experimental setup of most prior works, using Mean Squared Error (MSE) and Mean Absolute Error (MAE) as evaluation metrics for long-term forecasting tasks, and using Symmetric Mean Absolute Percentage Error (SMAPE), Mean Absolute Scaled Error (MASE), and Overall Weighted Average (OWA) as metrics for short-term forecasting tasks. The mathematical formulas for these evaluation metrics can be calculated as follows\cite{wu2023timesnet}:
\begin{align}
  \text{MSE} &= \frac{1}{H} \sum_{i=1}^H (\mathbf{X}_{i} - \widehat{\mathbf{X}}_{i})^2, \label{equ:mse} \\
  \text{MAE} &= \frac{1}{H} \sum_{i=1}^H |\mathbf{X}_{i} - \widehat{\mathbf{X}}_{i}|, \label{equ:mae} \\
  \text{SMAPE} &= \frac{200}{H} \sum_{i=1}^H \frac{|\mathbf{X}_{i} - \widehat{\mathbf{X}}_{i}|}{|\mathbf{X}_{i}| + |\widehat{\mathbf{X}}_{i}|}, \label{equ:smape} \\
  \text{MAPE} &= \frac{100}{H} \sum_{i=1}^H \frac{|\mathbf{X}_{i} - \widehat{\mathbf{X}}_{i}|}{|\mathbf{X}_{i}|}, \label{equ:mape} \\
  \text{MASE} &= \frac{1}{H} \sum_{i=1}^H \frac{|\mathbf{X}_{i} - \widehat{\mathbf{X}}_{i}|}{\frac{1}{H-m}\sum_{j=m+1}^{H}|\mathbf{X}_j - \mathbf{X}_{j-m}|}, \label{equ:mase} \\
  \text{OWA} &= \frac{1}{2} \left[ \frac{\text{SMAPE}}{\text{SMAPE}_{\textrm{Naïve2}}}  + \frac{\text{MASE}}{\text{MASE}_{\textrm{Naïve2}}}  \right]. \label{equ:owa}
\end{align}
where $m$ is the periodicity of the data. $\mathbf{X},\widehat{\mathbf{X}}\in\mathbb{R}^{H\times C}$ are the ground truth and prediction results of the future with $H$ time points and $C$ dimensions. $\mathbf{X}_{i}$ denotes the $i$-th future time point.

\section{System Configuration}\label{appx:system_config}
We conducted all experiments in the same experimental environment, which includes four NVIDIA A100 GPUs with 80GB and eight 40GB of memory. We saved overall experimental time by running experiments in parallel.

\ifarxiv
\input{6appendix_DE.tex}

\fi

%% file: Tables/datasets.tex
\begin{table*}[!h]
  \centering
  \footnotesize
  \caption{Data description of the \ndatasetsall datasets included in \system.} 
  \footnotesize
  \resizebox{1\textwidth}{!}{
    \begin{tabular}{@{}c l l l r r l @{}}
      \toprule
      Task & Dataset      & Domain      & Frequency & Lengths & Dim  & Description\\ \midrule

      \multirow{13}{*}{LTF} & ETTh1        & Electricity & 1 hour    & 14,400      & 7         & Power transformer 1, comprising seven indicators such as oil temperature and useful load\\
      & ETTh2        & Electricity & 1 hour    & 14,400      & 7         & Power transformer 2, comprising seven indicators such as oil temperature and useful load\\
      & ETTm1        & Electricity & 15 mins   & 57,600      & 7         & Power transformer 1, comprising seven indicators such as oil temperature and useful load\\
      & ETTm2        & Electricity & 15 mins   & 57,600      & 7         & Power transformer 2, comprising seven indicators such as oil temperature and useful load\\
      & ECL          & Electricity & 1 hour    & 26,304      & 321       & Electricity records the electricity consumption in kWh every 1 hour from 2012 to 2014\\
      & Traffic      & Traffic     & 1 hour    & 17,544      & 862       & Road occupancy rates measured by 862 sensors on San Francisco Bay area freeways\\
      & Weather      & Environment & 10 mins   & 52,696      & 21        & Recorded every for the whole year 2020, which contains 21 meteorological indicators\\
      & FRED-MD      & Economic    & 1 month   & 728         & 107       & Time series showing a set of macroeconomic indicators from the Federal Reserve Bank\\
      & Exchange     & Economic    & 1 day     & 7,588       & 8         & ExchangeRate collects the daily exchange rates of eight countries\\
      & NASDAQ       & Stock       & 1 day     & 1,244       & 5         & Records opening price, closing price, trading volume, lowest price, and highest price\\
      & NYSE         & Stock       & 1 day     & 1,243       & 5         & Records opening price, closing price, trading volume, lowest price, and highest price\\
      & ILI          & Health      & 1 week    & 966         & 7         & Recorded indicators of patients data from Centers for Disease Control and Prevention\\
      & Covid-19     & Health      & 1 day     & 1,392       & 948       & Provide opportunities for researchers to investigate the dynamics of COVID-19\\

      \midrule
      \multirow{6}{*}{STF} & \multirow{6}{*}{M4}    & \multirow{6}{2cm}{Demographic, Finance, Industry, Macro, Micro and Other} & Yearly   & \multirow{6}{*}{19-9933} & \multirow{6}{*}{100000}    & \multirow{6}{10cm}{M4 competition dataset containing 100,000 unaligned time series with varying lengths and time periods} \\
      &  &  & Quarterly    &   &  & \\
      &  &  & Monthly    &   &  & \\
      &  &  & Weakly    &   &   & \\
      &  &  & Daily    &   &   & \\
      &  &  & Hourly    &   &   & \\
      \bottomrule

    \end{tabular}

  }
  \label{tab:datasets} 
\end{table*}

%% file: latex_table_useinpaper/appendix/datasets_char_fromTFB.tex
\begin{table}[htb]
  \centering
  \caption{Datasets characteristics from TFB~\cite{qiu2024tfb}. Trend and Seasonal properties are indicated by \checkmark (True) or $\times$ (False).}
  \label{tab:meta_data_tfb}
  \resizebox{\columnwidth}{!}{%
    \begin{tabular}{lccccccc}
      \toprule
      \textbf{Dataset Name} & \textbf{Length} & \textbf{Trend} & \textbf{Seasonal} & \textbf{Unstationary} & \textbf{Transition} & \textbf{Shifting} & \textbf{Correlation} \\
      \midrule
      Covid-19 & 1392 & \checkmark & $\times$ & 0.3225 & 0.1259 & 0.2363 & 0.6040 \\
      ECL & 26304 & $\times$ & \checkmark & 0.0051 & 0.0105 & 0.0749 & 0.8025 \\
      ETTh1 & 14400 & \checkmark & $\times$ & 0.0012 & 0.0198 & 0.0614 & 0.6302 \\
      ETTh2 & 14400 & \checkmark & $\times$ & 0.0218 & 0.0420 & 0.4038 & 0.5090 \\
      ETTm1 & 57600 & $\times$ & $\times$ & 9.73e-05 & 0.0269 & 0.0630 & 0.6124 \\
      ETTm2 & 57600 & \checkmark & $\times$ & 0.0030 & 0.0377 & 0.4056 & 0.5036 \\
      Exchange & 7588 & \checkmark & $\times$ & 0.3598 & 0.0623 & 0.3253 & 0.5655 \\
      fred-md & 728 & \checkmark & $\times$ & 0.5735 & 0.1143 & 0.3943 & 0.6600 \\
      NASDAQ & 1244 & \checkmark & $\times$ & 0.1693 & 0.0741 & 0.9318 & 0.5636 \\
      ILI & 966 & $\times$ & $\times$ & 0.1692 & 0.0378 & 0.7211 & 0.6742 \\
      NYSE & 1243 & \checkmark & $\times$ & 0.6794 & 0.1667 & 0.6200 & 0.6129 \\
      Traffic & 17544 & $\times$ & $\times$ & 3.71e-08 & 0.0109 & 0.0670 & 0.8135 \\
      Weather & 52696 & $\times$ & $\times$ & 1.04e-08 & 0.0368 & 0.2136 & 0.6942 \\
      \bottomrule
    \end{tabular}
  }
\end{table}

%% file: 6appendix_DE.tex
\section{Details of \system} \label{appx:adgym_details}
In this section, we introduce detailed descriptions of the deconstructed components, extracted meta-features and the trained meta-predictors.

\subsection{More Details of Deconstructed Components in \system.}
\label{appx:design_choices_details}

\system unifies MTSF design into a four-stage pipeline: \textbf{\textit{Series Preprocessing}} $\rightarrow$ \textbf{\textit{Series Encoding}} $\rightarrow$ \textbf{\textit{Network Architecture}} $\rightarrow$ \textbf{\textit{Network Optimization}} (Figure~\ref{fig:pipeline}).
We analyze the design and efficacy of specialized modules adopted in state-of-the-art models by decomposing them into 11 distinct component dimensions.

\vspace{5pt}
\noindent\textbf{Stage 1: Series Preprocessing}. This stage handles input data characteristics.
(1) \textit{Normalization Strategies} mitigate distribution shifts through adaptive statistical alignment (e.g., RevIN~\cite{kim2021RevIN}, DishTS~\cite{fan2023dish}, Non-Stationary Transformer~\cite{liu2022nonstationary}).
(2) \textit{Decomposition Methods} break series into trend and seasonality components via time-domain moving averages (e.g., DLinear~\cite{zeng2023dlinear}) or frequency-domain DFT partitions (e.g., Koopa~\cite{liu2023koopa}).
(3) \textit{Series Sampling/Mixing} addresses temporal hierarchy through pyramidal attention (Pyraformer~\cite{liu2022pyraformer}), mixed experts (FEDformer~\cite{zhou2022fedformer}), or bidirectional mixing (TimeMixer~\cite{wang2024timemixer}).
These preprocessing steps align the input with the target label space $\mathbf{Y}$ for subsequent transformation.

\vspace{5pt}
\noindent\textbf{Stage 2: Series Encoding}. This stage focuses on transforming raw values into learnable representations.
(4) \textit{Channel Independence (CI)} vs. \textit{Channel Dependence (CD)} strategies determine inter-variable modeling paradigms; CI ensures robustness (e.g., PatchTST~\cite{nie2023PatchTST}) while CD explicitly captures multivariate dependencies (e.g., iTransformer~\cite{liu2024itransformer}, TSMixer~\cite{chen2023tsmixer}).
(5) \textit{Tokenization} varies by granularity from point-wise (Informer~\cite{zhou2021informer}) and patch-based (PatchTST~\cite{nie2023PatchTST}, Pathformer~\cite{chen2024pathformer}) to series-wise (iTransformer~\cite{liu2024itransformer}) encodings.
(6) \textit{Timestamp Embeddings} capture temporal context.
The resulting encodings define the feature space mapping $f$ processed by the architectural backbones.

\vspace{5pt}
\noindent\textbf{Stage 3: Network Architecture}. This core stage determines the representation learning mechanism through diverse architectural mechanisms.
We explicitly deconstruct it into: (7) \textit{Backbones}, including MLPs, RNNs (e.g., GRU, xLSTM~\cite{kraus2025xlstmmixer}), TCNs~\cite{bai2018tcn}, and Transformers with diverse attention variants (e.g., sparse~\cite{zhou2021informer}, frequency-domain~\cite{zhou2022fedformer}, Mamba hybrids~\cite{gu2024mamba}).
(8) \textit{Feature Attention} for capturing strictly inter-token dependencies.
(9) \textit{Retrieval-Augmented Generation (RAG)}~\cite{han2025retrieval} for leveraging external knowledge.
Large models like LLMs~\cite{zhou2023one} and TSFMs~\cite{liu2024timer} are also integrated as backbones in our system to enhance predictive capacity.
These architectural choices determine the structural inductive bias, which is further refined through targeted optimization strategies.

\vspace{5pt}
\noindent\textbf{Stage 4: Network Optimization}. The final stage governs the training process and objective $\mathcal{L}$.
It comprises: (10) \textit{Sequence Length Configuration}.
(11) \textit{Specialized Loss Functions} designed for time series characteristics, such as distribution-balanced DBLoss~\cite{qiu2024dbloss}, shape-aware PSLoss~\cite{kudrat2025patchwise}, and frequency-domain FreDFLoss~\cite{wang2025fredf}.
These optimization choices define the final training objective $\mathcal{L}$ for the assembled pipeline.

\begin{figure}[htb]
  \centering
  \includegraphics[width=0.45\textwidth]{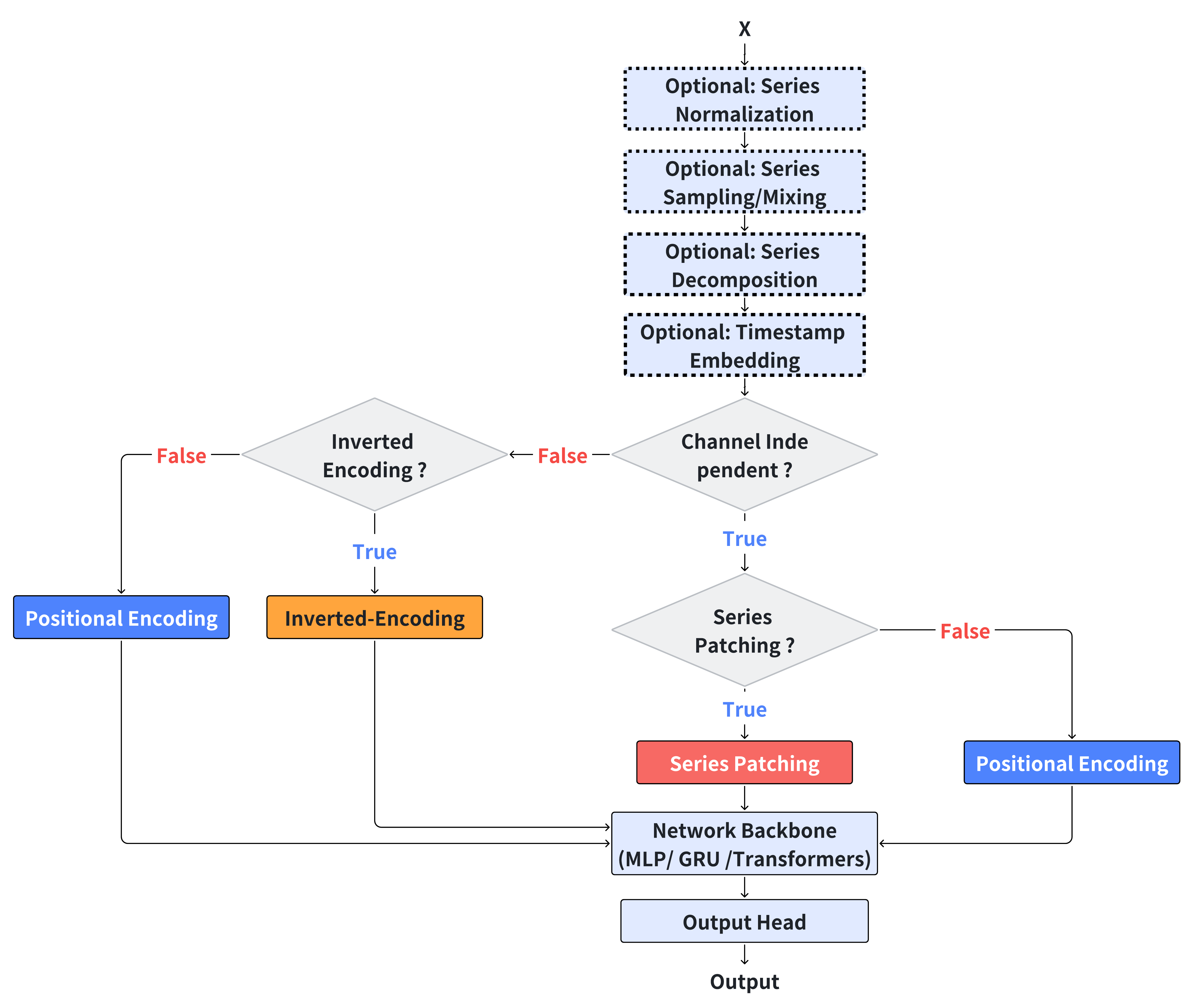}
  \caption{\rv{\system pipeline framework. This diagram shows the component-based design of \system. The final \system structure is formed by combining different component options.}}
  \label{fig:tsgym_pipeline}
\end{figure}

\noindent We prioritize competitive components from state-of-the-art models.
While individual modules show efficacy in isolation, their combined interactions remain underexplored.
\system systematically evaluates these synergies.
Crucially, we enforce architectural constraints to filter invalid combinations (e.g., pairing MLPs with series attention).
See Figure~\ref{fig:tsgym_pipeline} for the detailed workflow.

\subsection{Meta-Features and Meta-Predictors}
\label{appx:meta-details}

\subsubsection{Details of Meta-Features}
\label{appx:meta-features}

A critical challenge in automated forecasting is effectively characterizing the target dataset to inform adaptive model selection. Conventional approaches typically rely on handcrafted statistical metrics—such as skewness, kurtosis, and entropy—which characterize the static marginal distribution of the data but often fail to explicitly capture the conditional dependencies between historical observations and future targets. To bridge this gap, \system introduces a novel extraction method designed to encode the intrinsic predictive logic of the dataset.

\add{Our method leverages the In-Context Learning (ICL) capabilities of TabPFN~\cite{hollmann2022tabpfn}, a pre-trained Transformer model for tabular data. Rather than computing fixed statistics, we formulate a proxy classification task to probe the dataset's underlying dynamics. Given a multivariate time series data $\mathbf{D} \in \mathbb{R}^{N \times C}$ (with sequence length $N$ and $C$ channels), we construct a tabular proxy dataset by sampling $M$ instances. Specifically, for the $i$-th instance, we randomly select a channel $c \in \{1, \dots, C\}$ and a time step $t \in \{L, \dots, N-1\}$. We extract a historical subsequence of length $L$ to serve as the input feature vector $X_i$, and the observation at the next step $t+1$ as the continuous target $v_i$:}
\add{
  \begin{align}
    X_i &= \mathbf{D}_{t-L+1 : t, \ c} \in \mathbb{R}^L \\
    v_i &= \mathbf{D}_{t+1, \ c} \in \mathbb{R}
  \end{align}
}

\add{Then, the continuous target $v_i$ is discretized into a categorical label $Y_i$ via a bucketing function $\mathcal{B}(\cdot)$ with $K$ bins:}
\begin{equation}
  \add{Y_i = \mathcal{B}(v_i) \in \{1, 2, \dots, K\}}
\end{equation}

\add{By repeating this procedure, we construct a tabular dataset $\mathcal{T} = \{(X_i, Y_i)\}_{i=1}^M$. This dataset $\mathcal{T}$ is then fed into the pre-trained tabular foundation model. The dataset-level meta-feature $\mathbf{m}$ is extracted by aggregating the intermediate representations (e.g., via mean pooling) from the pretrained tabular foundation $f_{\text{Encoder}}$:}
\begin{equation}
  \add{\mathbf{m} = \text{Aggregate}\Big( f_{\text{Encoder}}(\mathcal{T}) \Big) \in \mathbb{R}^{d}}
\end{equation}

\add{By formulating the proxy task as a mapping from historical windows ($X_i$) to future states ($Y_i$), we shift the focus from marginal data distributions to conditional predictive relationships ($P(Y|X)$). Consequently, the resulting TabPFN embeddings capture the underlying transition laws (i.e., temporal dynamics) of the series, rather than just static statistical summaries.}

\add{Empirically, this representation strategy yields meta-features that exhibit remarkably high consistency with actual model performance. As shown in Table~\ref{tab:meta_consistency}, we observe a significant negative correlation between dataset distances in the meta-feature space and their performance rank consistencies. This negative correlation is highly desirable as it implies that datasets closer in our meta-feature space exhibit more similar model performances, confirming the semantic consistency of our meta-features. Our results substantially surpass both traditional statistical baselines (e.g., TSFEL~\cite{barandas2020tsfel}) and the recently proposed TimeFuse~\cite{liu2025breaking}.}

\input{Table_Exp/appx_meta_consistency.tex}
\add{Furthermore, our meta-features naturally follow a normal-like distribution---a property inherited from TabPFN's pre-training on synthetic priors and its internal Layer Normalization. This provides greater numerical stability during meta-predictor optimization compared to the often skewed distributions of handcrafted statistics.}

\subsubsection{Details of Meta-Predictor}
\label{subsec:meta-predictor}

\mbox{}\\
\textbf{Overview}. Unlike traditional methods selecting off-the-shelf models, \system customizes models for MTSF tasks via fine-grained component selection.
Given a constraint-validated model set $\mathcal{M}=\{M_1,...,M_m\}$, we learn the mapping from configurations to performance, enabling zero-shot selection on new tasks.

\vspace{5pt}
\noindent \textbf{Performance Corpus Generation}.
We evaluate configurations from two sources: the Constrained Orthogonal Pool ($m_{orth} \approx 130$) tested across all $n=13$ training datasets $\bm{\mathcal{D}}_{\text{train}}=\{\mathcal{D}_1,\ldots,\mathcal{D}_n\}$, and the Random Sampling Pool ($m_{rand} = 500$) tested on a subset of 7 training datasets, both across 4 prediction horizons.
This extensive evaluation yields a total performance corpus of approximately 20,760 entries.
Based on these results, we construct a performance matrix $\bm{P}\in \mathbb{R}^{n \times m}$.
To account for varying dataset difficulty, we convert MSE values $\bm{P}_{i,j}$ to normalized rankings $R_{i,j}=\mathrm{rank}(\bm{P}_{i,j})/m \in [0,1]$.
This ensures that the meta-predictor learns relative model quality rather than dataset-specific error scales.

\vspace{5pt}
\noindent \textbf{Meta-Predictor Formulation}.
The meta-dataset $\mathcal{D}_{meta}$ consists of tuples $(\mathcal{D}_i, M_j, R_{i,j})$.
Each configuration $M_j \in \mathcal{M}$ is decomposed into component indices $\{c_{j,1}, \dots, c_{j,k}\}$.
For each deconstructed component, we first use the LabelEncoder class from scikit-learn to convert it into a numerical class index.
This index is then transformed into dense embeddings $\mathbf{E}_{j}^{comp} = \oplus_{t=1}^{k} \phi(c_{j,t})$ via a learnable codebook $\phi$ (implemented as an $nn.Embedding$ layer).
Similarly, we extract meta-features $\mathbf{E}_{i}^{meta}$ from the training split of $\mathcal{D}_i$ using TabPFN~\cite{hollmann2022tabpfn} (see Appx.~\ref{appx:meta-features}).
The meta-predictor $f$ learns the mapping formulated as:
\begin{equation}
  f\left(\mathcal{D}_i, M_j\right)=\bm{R}_{i,j},
  f \;: \underbracket{\mathbf{E}_{i}^{meta}}_{\text{meta features}},
  \underbracket{\mathbf{E}_{j}^{comp}}_{\text{component embed.}}
  \mapsto \bm{R}_{i,j}
  \label{eq:meta-predictor-appx}
\end{equation}
where $i\in \{1,\ldots,n\}$ and $j\in \{1,\ldots,m\}$.
The meta-predictor is optimized using Pearson loss to learn the relative performance ranks, thereby emphasizing the linear correlation between predicted and actual rankings.

\vspace{5pt}
\noindent \textbf{Zero-Shot Implementation}.
The meta-predictor is implemented as a two-layer MLP.
At inference time, we extract meta-features $\mathbf{X}_{\text{test}}$ from a new dataset without any model training.
The trained $f$ provides predicted rankings across potential configurations, allowing users to select the optimal top-$k$ component combinations instantly.
This procedure eliminates the need for exhaustive local experimentation on target datasets.
The meta-predictor is pretrained on extensive benchmarking results from training datasets, allowing immediate deployment on new forecasting tasks.

\section{Additional Experimental Results}
\label{appx:complete_results}

\subsection{Analysis of Higher-Order Component Interactions}
\label{appx:higher_order}

\add{To assess the impact of higher-order interactions, we conducted a rigorous Type III ANOVA with treatment contrasts. We modeled all 54 feasible pairwise terms and 66 estimable three-way combinations (via nested F-tests due to rank deficiency), using partial $\eta^2$ with FDR correction to quantify effect sizes.}

\add{As summarized in Table~\ref{tab:higher_order_summary}, analysis confirms higher-order interactions are statistically prevalent: 30/54 pairwise and 58/66 three-way combinations are significant under FDR correction. Notably, specific component synergies---such as the pairing of Attention Type $\times$ Loss Function or Normalization $\times$ Backbone---play a statistically significant role in model dynamics, as highlighted in Table~\ref{tab:higher_order_top}.}

\add{Main effects overwhelmingly dominate the variance in performance. While pairwise interactions are statistically significant, they only increment the total $R^2$ by 5.27\% from 27.29\% to 32.56\% as detailed in Table~\ref{tab:higher_order_summary}. Thus, the main effects alone account for 83.8\% (27.29\% / 32.56\%) of the explainable performance variance. Furthermore, the maximum individual interaction effect size ($\eta^2$) peaks at 0.043, which is marginal compared to the primary components. Therefore, using main effects to estimate component contributions serves as a robust, highly pragmatic proxy for automated search and analytical ranking, allowing us to avoid the combinatorial explosion of modeling all $n$-way interactions.}
\begin{table}[h]
  \centering
  \caption{Summary of Higher-Order Interaction Analysis}\label{tab:higher_order_summary}
  \resizebox{\columnwidth}{!}{
    \begin{tabular}{lcc}
      \toprule
      \textbf{Metric} & \textbf{Pairwise} & \textbf{Three-Way} \\
      \midrule
      Estimable configurations & 54 & 66 \\
      Significant (FDR-corrected) & 30 & 58 \\
      Global F-test & \makecell{F = 6.47,\\ p = 2.35$\times 10^{-88}$} & -- \\
      Base Model $R^2$ & 27.29\% & -- \\
      Pairwise Interaction Model $R^2$ & 32.56\% & -- \\
      $R^2$ Increment & 5.27\% & -- \\
      Maximum $\eta^2$ & 0.014 & 0.043 \\
      \bottomrule
    \end{tabular}
  }
\end{table}

\begin{table}[h]
  \centering
  \caption{Top Significant Interactions}\label{tab:higher_order_top}
  \resizebox{\columnwidth}{!}{
    \begin{tabular}{lc}
      \toprule
      \textbf{Interaction} & $\eta^2$ \\
      \midrule
      \multicolumn{2}{l}{\textbf{Pairwise}} \\
      Attention Type $\times$ Loss Function & 0.014*** \\
      Normalization $\times$ Backbone & 0.013*** \\
      Decomposition $\times$ Backbone & 0.012*** \\
      \midrule
      \multicolumn{2}{l}{\textbf{Three-Way}} \\
      Timestamp $\times$ Decomposition $\times$ Sequence Length & 0.034*** \\
      Timestamp $\times$ Feature Attention $\times$ Sequence Length & 0.026*** \\
      Decomposition $\times$ Feature Attention $\times$ Retrieval-Augmented & 0.023*** \\
      \bottomrule
      \multicolumn{2}{l}{\small \textit{Note: Significance levels: *** $p<0.001$}}
    \end{tabular}
  }
\end{table}

\subsection{Robustness and Generalization of Preprocessing Dominance}
\label{appx:preprocessing_dominance_metrics}

\add{To rigorously examine whether preprocessing dominance is a metric-induced artifact, we extended our evaluation to four metrics: scale-sensitive metrics (MAE, RMSE) and scale-independent metric (MASE). MASE (Eq.\eqref{equ:mase}) neutralizes numerical scales by normalizing model MAE with a naive forecast's MAE on the training set.}

\begin{table}[h]
  \centering
  \caption{Component Contribution Across Metrics (\% variance explained)}\label{tab:metric_robustness}
  \vspace{-0.5em}
  \resizebox{\columnwidth}{!}{
    \begin{tabular}{lcccc}
      \toprule
      \textbf{Design Dimension} & \textbf{MSE} & \textbf{MAE} & \textbf{RMSE} & \textbf{MASE} \\
      \midrule
      \rowcolor{gray!15} \textit{Series Preprocessing} & \textbf{66.6} & \textbf{66.1} & \textbf{67.5} & \textbf{58.7} \\
      \textbf{Series Normalization} & $63.0^{***}$ & $63.1^{***}$ & $64.6^{***}$ & $55.0^{***}$ \\
      \textbf{Series Decomposition} & $3.2^{***}$ & $2.5^{***}$ & $2.6^{***}$ & $3.6^{***}$ \\
      \textbf{Series Sampling} & $0.4^{***}$ & $0.4^{***}$ & $0.4^{***}$ & $0.1^{*}$ \\
      \midrule
      \rowcolor{gray!15} \textit{Series Encoding} & \textbf{18.3} & \textbf{19.7} & \textbf{19.2} & \textbf{23.2} \\
      \textbf{Channel Independence} & $11.1^{***}$ & $12.6^{***}$ & $12.1^{***}$ & $17.7^{***}$ \\
      \textbf{Series Tokenization} & $7.1^{***}$ & $6.9^{***}$ & $7.0^{***}$ & $5.4^{***}$ \\
      \textbf{Timestamp Embedding} & $0.1$ & $0.2^{**}$ & $0.1^{*}$ & $0.0$ \\
      \midrule
      \rowcolor{gray!15} \textit{Network Architecture} & \textbf{8.0} & \textbf{7.8} & \textbf{7.7} & \textbf{5.2} \\
      \textbf{Backbone} & $0.9^{***}$ & $0.5^{***}$ & $0.8^{***}$ & $0.9^{***}$ \\
      \textbf{Attention Type} & $1.8^{***}$ & $1.2^{***}$ & $1.6^{***}$ & $1.9^{***}$ \\
      \textbf{Feature Attention} & $3.2^{***}$ & $2.9^{***}$ & $3.0^{***}$ & $1.6^{***}$ \\
      \textbf{RAG} & $2.1^{***}$ & $3.0^{***}$ & $2.4^{***}$ & $0.9^{***}$ \\
      \midrule
      \rowcolor{gray!15} \textit{Network Optimization} & \textbf{7.1} & \textbf{6.5} & \textbf{5.5} & \textbf{12.9} \\
      \textbf{Loss Function} & $5.4^{***}$ & $4.9^{***}$ & $4.4^{***}$ & $10.4^{***}$ \\
      \textbf{Sequence Length} & $1.6^{***}$ & $1.6^{***}$ & $1.1^{***}$ & $2.5^{***}$ \\
      \bottomrule
      \multicolumn{5}{l}{\small \textit{Note: Significance levels: *** $p<0.001$, ** $p<0.01$, * $p<0.05$.}}
    \end{tabular}
  }
\end{table}

\add{As shown in Table~\ref{tab:metric_robustness}, under MSE, Series Preprocessing explains 66.6\% of the variance (specifically, Series Normalization: 63.0\%) while Network Architecture explains only 8.0\%. Under the scale-independent MASE, Series Preprocessing still accounts for a dominant 58.7\%, whereas Network Architecture's contribution further drops to 5.2\%. Notably, the importance ratio of preprocessing-to-architecture actually increases from 8.3 under MSE to 11.3 under MASE. These results confirm that Series Preprocessing ranks first across all scale-sensitive and scale-independent metrics.}

\add{Beyond metric sensitivity, we further quantify how reliably a combination generalizes across domains using a Performance-to-Volatility Ratio. Within each scenario (dataset $\times$ prediction horizon), we rank all $N$ combinations by MSE and convert them into a performance score: $S = 1 - \frac{\text{Rank}}{N}$. For each combination, we calculate the mean ($\mu$) and standard deviation ($\sigma$) of $S$ across all 52 scenarios. The ratio ($\mu / \sigma$) rewards consistently high-ranking configurations by penalizing fluctuations, serving as a statistically rigorous proxy for robustness and transferability.}

\add{As detailed in Table~\ref{tab:robustness_dimension}, when evaluated under this rigorous metric, our analysis firmly corroborates our prior findings: \textit{Series Preprocessing} remains the most critical pipeline stage, contributing 55.5\% to overall model robustness. Interestingly, when performance volatility is considered, the importance of the other three stages aligns at a comparable level ($\sim$14--16\%).}

\begin{table}[h]
  \centering
  \caption{Pipeline and Component Importance (Robustness Dimension, \%)}\label{tab:robustness_dimension}
  \vspace{-0.5em}
  \resizebox{\columnwidth}{!}{
    \begin{tabular}{lc | lc}
      \toprule
      \textbf{Component} & \textbf{Score (\%)} & \textbf{Component} & \textbf{Score (\%)} \\
      \midrule
      \multicolumn{2}{c|}{\cellcolor{gray!15}\textit{Series Preprocessing (Total: 55.5\%)}} & \multicolumn{2}{c}{\cellcolor{gray!15}\textit{Network Architecture (Total: 15.0\%)}} \\
      \textbf{Series Normalization} & $54.5^{***}$ & \textbf{Feature Attention} & $6.2^{**}$ \\
      \textbf{Series Decomposition} & $1.0$ & \textbf{RAG} & $3.9^{**}$ \\
      \textbf{Series Sampling} & $0.0$ & \textbf{Attention Type} & $3.2$ \\
      \multicolumn{2}{c|}{\cellcolor{gray!15}\textit{Series Encoding (Total: 15.6\%)}} & \textbf{Backbone} & $1.6$ \\
      \textbf{Channel Independence} & $8.1^{***}$ & \multicolumn{2}{c}{\cellcolor{gray!15}\textit{Network Optimization (Total: 13.9\%)}} \\
      \textbf{Series Tokenization} & $7.4^{**}$ & \textbf{Sequence Length} & $9.6^{***}$ \\
      \textbf{Timestamp Embedding} & $0.1$ & \textbf{Loss Function} & $4.2$ \\
      \bottomrule
      \multicolumn{4}{l}{\small \textit{Note: Significance levels: *** $p<0.001$, ** $p<0.01$, * $p<0.05$.}}
    \end{tabular}
  }
\end{table}

\add{Collectively, these metric-invariant and scenario-robust evaluations suggest that preprocessing dominance is an inherent MTSF property rather than an evaluation artifact. The consistent superiority of preprocessing underscores that handling non-stationarity and distribution shift is more fundamental than architectural complexity.}

\subsection{Justification for Backbone Corpus Design}
\label{appx:backbone_expansion}

\add{Restricting the automated construction to MLP-based configurations reduces computational costs. However, it raises concerns about whether the framework leverages insights from complex architectures. To address this, Table~\ref{tab:expanded_backbone} compares the default MLP-only corpus against an expanded space incorporating RNN and Transformer architectures. Our experiments reveal that these complex architectures do not yield significant performance improvements over the MLP-only configuration. To investigate this, we analyzed the backbone distributions of the Top-$K$ (10, 20, 50) optimal configurations. The results reveal an absolute prevalence of MLP backbones. They ranked first in 164 out of 168 evaluation scenarios (spanning 7 datasets, 4 prediction lengths, 3 Top-$K$ settings, and both mean/median MSE values).}

\add{This dominance explains the negligible gains from backbone expansion and aligns with recent studies and our evaluations in Table~\ref{tab:coef-comparison} and Fig.~\ref{fig:exp-Backbone}. Furthermore, the minimal performance gap between the two settings highlights the robustness of our automation framework. It consistently identifies optimal configurations even as the search complexity increases. This ensures stable performance regardless of the backbone variety. Thus, while the identified optimal configurations are MLP-based, the underlying meta-learning methodology remains architecture-agnostic.}
\input{Table_Exp/appx_ExpandedBackboneCorpus.tex}

\subsection{Experimental Cost Analysis}
\label{appx:experiments_cost}

\add{Table~\ref{tab:experimental_cost} provides a fair comparison of offline preparation time, online processing time, and predictive performance. We evaluate these metrics on the ETTh1 dataset. This evaluation uses a prediction length of 96 and runs on 8 A800 GPUs. We acknowledge that building the algorithm corpus involves a significant offline computational investment. However, similar to the pre-training phase of many large models, this cost is front-loaded and strictly decoupled from the practical deployment phase. Once this one-time preparation is complete, \system is much faster and more cost-effective when applied to new datasets. For instance, our \system-fast variant achieves superior accuracy while being nearly $7\times$ faster than AutoGluon, effectively delivering high performance with minimal online overhead.}

\add{This efficiency advantage is driven by our meta-learning strategy. \system performs zero-shot recommendations on new datasets, instantly identifying an optimal lightweight MLP-based configuration. Because these recommended models are inherently efficient, they ensure both rapid model fitting and inference. In contrast, baselines incur heavy online costs. For instance, AutoGluon requires iterative searching for every new task, while TimeFuse relies on ensembling for subsequences.}

\input{Table_Exp/appx_experiments_cost.tex}

\subsection{Comprehensive Results of \system Against State-of-the-Art Methods}\label{appx:sota_complete_results}
Due to space limitations in the main text, here we provide complete experimental comparisons for both long-term and short-term forecasting tasks. Table \ref{tab:TSGym_vs_Sota_fullAvg} details the full long-term forecasting performance across all prediction horizons, while Table \ref{tab:TSGym_vs_Sota_short_full} presents the comprehensive short-term forecasting results. Following standard benchmarking conventions, we highlight top-performing methods in \textcolor{red}{\textbf{red}} and second-best results with \underline{\textcolor{blue}{underlined}} formatting. These extensive evaluations consistently validate \system's competitive performance across diverse temporal prediction scenarios.
\input{Table_Exp/Exp_SOTA_LongTerm_Full_icml}
\input{Table_Exp/Exp_SOTA_ShortTerm_Full_icml}

\input{generated_figures}

%% file: Table_Exp/appx_meta_consistency.tex
\begin{table}[H]
  \centering
  \caption{\add{Consistency between meta-features and model performance}}\label{tab:meta_consistency}
  \vspace{-0.8em}
  \begin{tabular}{l|ccc}
    \toprule
    \textbf{Metric} & \textbf{Ours} & \textbf{TimeFuse} & \textbf{TSFEL} \\
    \midrule
    Pearson $r$ & \textbf{-0.549$^\ast$} & -0.291 & -0.055 \\
    Pearson $p$ & 0.010 & 0.200 & 0.813 \\
    \midrule
    Spearman $r$ & \textbf{-0.610$^{\ast\ast}$} & -0.353 & -0.329 \\
    Spearman $p$ & 0.003 & 0.116 & 0.146 \\
    \bottomrule
  \end{tabular}
\end{table}

%% file: Table_Exp/appx_ExpandedBackboneCorpus.tex
\begin{table}[H]
  \centering
  \small
  \caption{\add{Performance Comparison: MLP-only vs. Expanded Backbone Corpus}}\label{tab:expanded_backbone}
  \vspace{-0.8em}
  \begin{tabular}{c|rr|rr}
    \toprule
    \multicolumn{1}{c}{\textbf{Models}} & \multicolumn{4}{c}{\textbf{TSCOMP}} \\
    \cmidrule(lr){1-5}
    Corpus & \multicolumn{2}{c}{\textbf{MLP-only}} & \multicolumn{2}{c}{\textbf{+RNN+Transformer}} \\
    \midrule
    Metric & MSE & MAE & MSE & MAE \\
    \midrule
    \textbf{ETTh1} & 0.407 & \textbf{\textcolor{red}{0.424}} & \textbf{\textcolor{red}{0.404}} & \textbf{\textcolor{red}{0.424}} \\
    \textbf{ETTh2} & \textbf{\textcolor{red}{0.336}} & 0.384 & 0.340 & \textbf{\textcolor{red}{0.383}} \\
    \textbf{ETTm1} & \textbf{\textcolor{red}{0.341}} & 0.371 & 0.343 & \textbf{\textcolor{red}{0.370}} \\
    \textbf{ETTm2} & \textbf{\textcolor{red}{0.246}} & \textbf{\textcolor{red}{0.306}} & 0.256 & 0.319 \\
    \textbf{ECL} & 0.161 & \textbf{\textcolor{red}{0.253}} & \textbf{\textcolor{red}{0.159}} & \textbf{\textcolor{red}{0.253}} \\
    \textbf{Traffic} & \textbf{\textcolor{red}{0.405}} & 0.271 & 0.420 & \textbf{\textcolor{red}{0.268}} \\
    \textbf{Weather} & \textbf{\textcolor{red}{0.222}} & \textbf{\textcolor{red}{0.256}} & 0.223 & \textbf{\textcolor{red}{0.256}} \\
    \midrule
    \multicolumn{1}{c}{\textbf{$1^{\text{st}}$ Count}} & \multicolumn{2}{c}{\textbf{\textcolor{red}{6}}} & \multicolumn{2}{c}{5} \\
    \bottomrule
  \end{tabular}
\end{table}

%% file: Table_Exp/appx_experiments_cost.tex
\begin{table}[H]
  \centering
  \caption{\add{Experimental Cost Summary}}\label{tab:experimental_cost}
  \vspace{-0.8em}
  \resizebox{\columnwidth}{!}{
    \begin{tabular}{l|cccccccc}
      \toprule
      \textbf{Metric} & \textbf{AutoTS} & \textbf{AutoGluon} & \multicolumn{2}{c}{\textbf{Auto-sklearn}~\cite{feurer-neurips15a}} & \multicolumn{2}{c}{\textbf{TimeFuse}~\cite{liu2025breaking}} & \multicolumn{2}{c}{\textbf{\system~(Ours)}} \\
      & \cite{AutoTS_github} & \cite{shchur2023autogluon} & (fast) & (standard) & (Zero-shot) & (Few-shot) & \textbf{fast} & \textbf{standard} \\
      \midrule
      \textbf{Offline Time} & - & - & - & - & 0.35h & 0.35h & \textbf{5.8h} & \textbf{22.5h} \\
      \textbf{Online Time} & 740.9s & 1102.2s & 360.8s & 1795.2s & 1275.9s & 1522.9s & \textbf{163.3s} & \textbf{332.7s} \\
      \midrule
      \textbf{MSE} & 0.604 & 0.410 & 1.601 & 1.652 & 0.369 & 0.378 & \textbf{0.361} & \textbf{0.362} \\
      \textbf{MAE} & 0.507 & 0.419 & 1.001 & 1.015 & 0.397 & 0.397 & \textbf{0.387} & \textbf{0.390} \\
      \bottomrule
      \multicolumn{9}{p{1.7\columnwidth}}{\footnotesize{\textit{*Note: \system-standard denotes the complete framework, whereas \system-fast excludes RAG, Series Sampling, and Series Decomposition. Auto-sklearn-standard is configured with \texttt{time\_left\_for\_this\_task}=1200s and \texttt{per\_run\_time\_limit}=150s, while Auto-sklearn-fast operates under stricter limits (120s and 30s, respectively).}}}
    \end{tabular}
  }
\end{table}

%% file: Table_Exp/Exp_SOTA_LongTerm_Full_icml.tex
\begin{table*}
  \centering
  \caption{Full results for the long-term forecasting task across all prediction horizons (96, 192, 336, 720). Lower MSE and MAE values indicate superior accuracy. We highlight the \textcolor{red}{\textbf{1st}} and \underline{\textcolor{blue}{2nd}} best results.}
  \label{tab:TSGym_vs_Sota_fullAvg}
  \resizebox{\textwidth}{!}{
    \setlength{\tabcolsep}{2pt}
    \begin{tabular}{cc|rr|rr|rr|rr|rr|rr|rr|rr|rr|rr|rr|rr|rr|rr}
      \toprule
      \multicolumn{2}{c}{\multirow{2}{*}{Models}} & \multicolumn{2}{c}{\textbf{\system}} & \multicolumn{2}{c}{OLinear} & \multicolumn{2}{c}{RAFT} & \multicolumn{2}{c}{DUET} & \multicolumn{2}{c}{TimeMixer} & \multicolumn{2}{c}{TimeXer} & \multicolumn{2}{c}{PAttn} & \multicolumn{2}{c}{iTrans.} & \multicolumn{2}{c}{Mamba} & \multicolumn{2}{c}{MICN} & \multicolumn{2}{c}{TimesNet} & \multicolumn{2}{c}{PatchTST} & \multicolumn{2}{c}{DLinear} & \multicolumn{2}{c}{Cross.} \\
      \multicolumn{2}{c}{} & \multicolumn{2}{c}{(\textbf{Ours})} & \multicolumn{2}{c}{\cite{yue2025olinear}} & \multicolumn{2}{c}{\cite{han2025retrieval}} & \multicolumn{2}{c}{\cite{qiu2025DUET}} & \multicolumn{2}{c}{\cite{wang2024timemixer}} & \multicolumn{2}{c}{\cite{wang2024timexer}} & \multicolumn{2}{c}{\cite{tan2024pattn}} & \multicolumn{2}{c}{\cite{liu2024itransformer}} & \multicolumn{2}{c}{\cite{gu2024mamba}} & \multicolumn{2}{c}{\cite{wang2023micn}} & \multicolumn{2}{c}{\cite{wu2023timesnet}} & \multicolumn{2}{c}{\cite{nie2023PatchTST}} & \multicolumn{2}{c}{\cite{zeng2023dlinear}} & \multicolumn{2}{c}{\cite{zhang2023crossformer}} \\
      \cmidrule(l){3-30}
      \multicolumn{2}{c}{Metric} & MSE & MAE & MSE & MAE & MSE & MAE & MSE & MAE & MSE & MAE & MSE & MAE & MSE & MAE & MSE & MAE & MSE & MAE & MSE & MAE & MSE & MAE & MSE & MAE & MSE & MAE & MSE & MAE \\
      \midrule
      \multirow{5}{*}{\rotatebox{90}{\textbf{ETTh1}}}
      & 96 & 0.362 & 0.390 & \tcb{0.362} & \tcr{0.383} & 0.372 & 0.402 & \tcr{0.355} & \tcb{0.387} & 0.379 & 0.398 & 0.385 & 0.404 & 0.390 & 0.404 & 0.392 & 0.409 & 0.487 & 0.454 & 0.415 & 0.432 & 0.421 & 0.435 & 0.381 & 0.399 & 0.396 & 0.411 & 0.410 & 0.428 \\
      & 192 & \tcr{0.396} & \tcr{0.414} & 0.416 & \tcb{0.415} & \tcb{0.403} & 0.419 & 0.418 & 0.425 & 0.435 & 0.431 & 0.436 & 0.439 & 0.458 & 0.438 & 0.440 & 0.435 & 0.564 & 0.508 & 0.517 & 0.487 & 0.476 & 0.465 & 0.431 & 0.434 & 0.446 & 0.441 & 0.457 & 0.462 \\
      & 336 & \tcr{0.418} & \tcr{0.428} & 0.459 & 0.439 & 0.436 & 0.443 & \tcb{0.421} & \tcb{0.434} & 0.487 & 0.449 & 0.481 & 0.451 & 0.506 & 0.471 & 0.487 & 0.459 & 0.519 & 0.487 & 0.627 & 0.567 & 0.481 & 0.461 & 0.472 & 0.459 & 0.496 & 0.474 & 0.557 & 0.518 \\
      & 720 & 0.450 & \tcb{0.464} & 0.468 & 0.466 & 0.473 & 0.480 & 0.558 & 0.526 & 0.475 & 0.467 & 0.529 & 0.498 & 0.535 & 0.501 & 0.486 & 0.479 & 0.605 & 0.565 & 0.816 & 0.673 & 0.517 & 0.493 & 0.562 & 0.520 & 0.521 & 0.517 & 0.735 & 0.647 \\
      & Avg & \tcr{0.407} & \tcr{0.424} & 0.426 & \tcb{0.426} & 0.421 & 0.436 & 0.438 & 0.443 & 0.444 & 0.436 & 0.458 & 0.448 & 0.472 & 0.454 & 0.451 & 0.446 & 0.544 & 0.503 & 0.594 & 0.540 & 0.474 & 0.463 & 0.461 & 0.453 & 0.465 & 0.461 & 0.540 & 0.514 \\
      \midrule
      \multirow{5}{*}{\rotatebox{90}{\textbf{ETTh2}}}
      & 96 & \tcr{0.270} & \tcb{0.335} & 0.286 & \tcr{0.330} & 0.285 & 0.344 & \tcb{0.281} & 0.342 & 0.293 & 0.343 & 0.287 & 0.339 & 0.305 & 0.356 & 0.301 & 0.351 & 0.357 & 0.382 & 0.381 & 0.420 & 0.322 & 0.363 & 0.297 & 0.347 & 0.344 & 0.398 & 0.683 & 0.602 \\
      & 192 & \tcr{0.329} & \tcr{0.373} & 0.366 & 0.381 & 0.356 & 0.397 & \tcb{0.344} & \tcb{0.380} & 0.377 & 0.396 & 0.365 & 0.391 & 0.375 & 0.401 & 0.379 & 0.398 & 0.457 & 0.442 & 0.496 & 0.485 & 0.394 & 0.405 & 0.381 & 0.402 & 0.482 & 0.479 & 0.953 & 0.701 \\
      & 336 & \tcr{0.355} & \tcr{0.397} & 0.404 & 0.413 & 0.379 & 0.425 & \tcb{0.368} & \tcb{0.405} & 0.445 & 0.443 & 0.419 & 0.431 & 0.424 & 0.435 & 0.420 & 0.432 & 0.478 & 0.463 & 0.629 & 0.557 & 0.464 & 0.457 & 0.441 & 0.444 & 0.596 & 0.542 & 1.914 & 1.125 \\
      & 720 & \tcr{0.390} & \tcb{0.431} & \tcb{0.411} & \tcr{0.430} & 0.428 & 0.470 & 0.437 & 0.446 & 0.435 & 0.448 & 0.435 & 0.448 & 0.444 & 0.458 & 0.427 & 0.446 & 0.577 & 0.506 & 0.838 & 0.659 & 0.418 & 0.438 & 0.445 & 0.460 & 0.842 & 0.662 & 3.818 & 1.653 \\
      & Avg & \tcr{0.336} & \tcr{0.384} & 0.367 & \tcb{0.388} & 0.362 & 0.409 & \tcb{0.358} & 0.393 & 0.387 & 0.408 & 0.376 & 0.402 & 0.387 & 0.412 & 0.382 & 0.407 & 0.467 & 0.448 & 0.586 & 0.530 & 0.399 & 0.416 & 0.391 & 0.413 & 0.566 & 0.520 & 1.842 & 1.021 \\
      \midrule
      \multirow{5}{*}{\rotatebox{90}{\textbf{ETTm1}}}
      & 96 & \tcr{0.281} & \tcb{0.335} & 0.302 & \tcr{0.334} & 0.304 & 0.351 & \tcb{0.293} & 0.343 & 0.322 & 0.361 & 0.320 & 0.357 & 0.323 & 0.363 & 0.335 & 0.370 & 0.360 & 0.386 & 0.318 & 0.371 & 0.326 & 0.369 & 0.343 & 0.375 & 0.345 & 0.372 & 0.340 & 0.394 \\
      & 192 & \tcr{0.321} & \tcr{0.361} & 0.356 & \tcb{0.363} & \tcb{0.327} & 0.365 & 0.330 & 0.366 & 0.368 & 0.387 & 0.364 & 0.386 & 0.365 & 0.386 & 0.389 & 0.399 & 0.459 & 0.435 & 0.358 & 0.397 & 0.382 & 0.403 & 0.369 & 0.391 & 0.382 & 0.390 & 0.474 & 0.498 \\
      & 336 & \tcr{0.350} & \tcr{0.379} & 0.389 & 0.387 & \tcb{0.355} & \tcb{0.383} & 0.365 & 0.390 & 0.390 & 0.404 & 0.397 & 0.407 & 0.396 & 0.408 & 0.450 & 0.433 & 0.516 & 0.490 & 0.421 & 0.447 & 0.427 & 0.424 & 0.405 & 0.410 & 0.413 & 0.412 & 0.813 & 0.657 \\
      & 720 & \tcb{0.411} & \tcr{0.410} & 0.451 & 0.425 & \tcr{0.406} & \tcb{0.412} & 0.421 & 0.424 & 0.449 & 0.437 & 0.452 & 0.442 & 0.454 & 0.442 & 0.498 & 0.462 & 0.640 & 0.536 & 0.495 & 0.489 & 0.499 & 0.460 & 0.458 & 0.446 & 0.478 & 0.454 & 0.813 & 0.705 \\
      & Avg & \tcr{0.341} & \tcr{0.371} & 0.374 & \tcb{0.377} & \tcb{0.348} & 0.378 & 0.352 & 0.381 & 0.382 & 0.397 & 0.383 & 0.398 & 0.385 & 0.400 & 0.418 & 0.416 & 0.494 & 0.462 & 0.398 & 0.426 & 0.408 & 0.414 & 0.394 & 0.405 & 0.404 & 0.407 & 0.610 & 0.564 \\
      \midrule
      \multirow{5}{*}{\rotatebox{90}{\textbf{ETTm2}}}
      & 96 & \tcr{0.159} & \tcr{0.248} & 0.170 & \tcb{0.249} & \tcb{0.164} & 0.256 & 0.166 & 0.254 & 0.176 & 0.259 & 0.171 & 0.255 & 0.180 & 0.265 & 0.184 & 0.265 & 0.197 & 0.274 & 0.194 & 0.291 & 0.190 & 0.269 & 0.183 & 0.267 & 0.194 & 0.293 & 0.340 & 0.406 \\
      & 192 & \tcr{0.211} & \tcr{0.282} & 0.233 & \tcb{0.290} & \tcb{0.220} & 0.296 & 0.240 & 0.302 & 0.240 & 0.301 & 0.241 & 0.302 & 0.249 & 0.311 & 0.247 & 0.307 & 0.285 & 0.333 & 0.269 & 0.343 & 0.253 & 0.308 & 0.247 & 0.308 & 0.288 & 0.364 & 0.788 & 0.632 \\
      & 336 & \tcr{0.266} & \tcr{0.320} & 0.291 & 0.328 & 0.275 & 0.335 & \tcb{0.271} & \tcb{0.325} & 0.307 & 0.344 & 0.303 & 0.343 & 0.313 & 0.351 & 0.311 & 0.347 & 0.368 & 0.389 & 0.402 & 0.434 & 0.329 & 0.352 & 0.309 & 0.348 & 0.358 & 0.411 & 1.111 & 0.758 \\
      & 720 & \tcr{0.350} & \tcr{0.373} & 0.389 & 0.387 & 0.366 & 0.395 & 0.360 & \tcb{0.381} & 0.401 & 0.403 & 0.405 & 0.401 & 0.414 & 0.411 & 0.415 & 0.407 & 0.591 & 0.487 & 0.553 & 0.516 & 0.418 & 0.407 & 0.424 & 0.415 & 0.556 & 0.526 & 5.453 & 1.676 \\
      & Avg & \tcr{0.246} & \tcr{0.306} & 0.271 & \tcb{0.314} & \tcb{0.256} & 0.320 & 0.259 & 0.316 & 0.281 & 0.327 & 0.280 & 0.325 & 0.289 & 0.335 & 0.289 & 0.331 & 0.360 & 0.371 & 0.354 & 0.396 & 0.298 & 0.334 & 0.290 & 0.335 & 0.349 & 0.399 & 1.923 & 0.868 \\
      \midrule
      \multirow{5}{*}{\rotatebox{90}{\textbf{Weather}}}
      & 96 & \tcr{0.147} & \tcb{0.192} & \tcb{0.152} & \tcr{0.188} & 0.167 & 0.225 & 0.155 & 0.198 & 0.162 & 0.209 & 0.159 & 0.206 & 0.177 & 0.218 & 0.178 & 0.218 & 0.192 & 0.240 & 0.191 & 0.248 & 0.169 & 0.218 & 0.175 & 0.216 & 0.195 & 0.256 & 0.172 & 0.243 \\
      & 192 & \tcr{0.190} & \tcr{0.235} & 0.202 & \tcb{0.238} & 0.211 & 0.264 & \tcb{0.198} & 0.238 & 0.209 & 0.252 & 0.204 & 0.248 & 0.222 & 0.257 & 0.225 & 0.257 & 0.259 & 0.296 & 0.240 & 0.300 & 0.227 & 0.267 & 0.223 & 0.258 & 0.238 & 0.296 & 0.231 & 0.304 \\
      & 336 & \tcr{0.240} & \tcr{0.272} & 0.258 & 0.280 & 0.258 & 0.301 & 0.251 & \tcb{0.278} & 0.261 & 0.291 & 0.263 & 0.292 & 0.277 & 0.298 & 0.280 & 0.298 & 0.323 & 0.342 & 0.280 & 0.327 & 0.291 & 0.311 & 0.279 & 0.298 & 0.281 & 0.331 & 0.279 & 0.346 \\
      & 720 & \tcb{0.311} & \tcr{0.324} & 0.340 & 0.334 & 0.325 & 0.353 & 0.323 & \tcb{0.329} & 0.345 & 0.345 & 0.342 & 0.343 & 0.353 & 0.347 & 0.359 & 0.351 & 0.392 & 0.383 & 0.351 & 0.387 & 0.358 & 0.353 & 0.357 & 0.347 & 0.347 & 0.384 & 0.362 & 0.404 \\
      & Avg & \tcr{0.222} & \tcr{0.256} & 0.238 & \tcb{0.260} & 0.240 & 0.286 & 0.232 & 0.261 & 0.244 & 0.274 & 0.242 & 0.272 & 0.257 & 0.280 & 0.261 & 0.281 & 0.291 & 0.315 & 0.265 & 0.316 & 0.261 & 0.287 & 0.258 & 0.280 & 0.265 & 0.317 & 0.261 & 0.324 \\
      \midrule
      \multirow{5}{*}{\rotatebox{90}{\textbf{ECL}}}
      & 96 & \tcb{0.128} & 0.222 & 0.131 & \tcb{0.221} & 0.131 & 0.229 & \tcr{0.127} & \tcr{0.218} & 0.156 & 0.247 & 0.141 & 0.243 & 0.183 & 0.266 & 0.148 & 0.239 & 0.190 & 0.292 & 0.170 & 0.283 & 0.166 & 0.269 & 0.180 & 0.273 & 0.210 & 0.301 & 0.149 & 0.250 \\
      & 192 & 0.148 & 0.242 & 0.150 & \tcb{0.239} & \tcb{0.147} & 0.243 & \tcr{0.146} & \tcr{0.237} & 0.169 & 0.259 & 0.158 & 0.256 & 0.188 & 0.271 & 0.166 & 0.257 & 0.209 & 0.313 & 0.181 & 0.293 & 0.186 & 0.287 & 0.187 & 0.279 & 0.210 & 0.305 & 0.167 & 0.264 \\
      & 336 & \tcb{0.160} & 0.255 & 0.165 & \tcr{0.254} & \tcr{0.160} & 0.258 & 0.164 & \tcb{0.255} & 0.187 & 0.278 & 0.174 & 0.273 & 0.204 & 0.287 & 0.177 & 0.269 & 0.199 & 0.305 & 0.191 & 0.303 & 0.198 & 0.298 & 0.204 & 0.296 & 0.223 & 0.319 & 0.190 & 0.288 \\
      & 720 & 0.207 & 0.294 & 0.188 & \tcr{0.277} & \tcr{0.186} & 0.282 & \tcb{0.188} & \tcb{0.278} & 0.227 & 0.312 & 0.216 & 0.309 & 0.245 & 0.320 & 0.209 & 0.299 & 0.240 & 0.339 & 0.211 & 0.321 & 0.222 & 0.318 & 0.246 & 0.328 & 0.258 & 0.350 & 0.263 & 0.353 \\
      & Avg & 0.161 & 0.253 & 0.159 & \tcb{0.248} & \tcr{0.156} & 0.253 & \tcb{0.156} & \tcr{0.247} & 0.185 & 0.274 & 0.172 & 0.270 & 0.205 & 0.286 & 0.175 & 0.266 & 0.209 & 0.312 & 0.188 & 0.300 & 0.193 & 0.293 & 0.204 & 0.294 & 0.225 & 0.319 & 0.192 & 0.289 \\
      \midrule
      \multirow{5}{*}{\rotatebox{90}{\textbf{Traffic}}}
      & 96 & \tcb{0.374} & 0.256 & 0.398 & \tcr{0.227} & 0.376 & 0.269 & \tcr{0.360} & \tcb{0.238} & 0.479 & 0.298 & 0.427 & 0.276 & 0.499 & 0.325 & 0.392 & 0.268 & 0.702 & 0.390 & 0.518 & 0.310 & 0.589 & 0.323 & 0.458 & 0.298 & 0.697 & 0.429 & 0.545 & 0.278 \\
      & 192 & \tcb{0.389} & 0.266 & 0.435 & \tcr{0.241} & 0.391 & 0.277 & \tcr{0.385} & \tcb{0.249} & 0.493 & 0.298 & 0.446 & 0.281 & 0.498 & 0.319 & 0.413 & 0.278 & 0.643 & 0.364 & 0.536 & 0.319 & 0.618 & 0.330 & 0.470 & 0.304 & 0.646 & 0.407 & 0.547 & 0.287 \\
      & 336 & 0.405 & 0.268 & 0.460 & \tcr{0.250} & \tcb{0.402} & 0.281 & \tcr{0.401} & \tcb{0.259} & 0.507 & 0.312 & 0.478 & 0.290 & 0.511 & 0.324 & 0.425 & 0.283 & 0.649 & 0.364 & 0.547 & 0.321 & 0.629 & 0.334 & 0.483 & 0.307 & 0.654 & 0.410 & 0.575 & 0.295 \\
      & 720 & 0.450 & 0.294 & 0.501 & \tcr{0.271} & \tcr{0.434} & 0.297 & \tcb{0.436} & \tcb{0.277} & 0.528 & 0.319 & 0.516 & 0.306 & 0.544 & 0.342 & 0.458 & 0.300 & 0.720 & 0.403 & 0.574 & 0.331 & 0.662 & 0.348 & 0.517 & 0.325 & 0.694 & 0.429 & 0.595 & 0.323 \\
      & Avg & 0.405 & 0.271 & 0.448 & \tcr{0.247} & \tcb{0.401} & 0.281 & \tcr{0.396} & \tcb{0.256} & 0.502 & 0.307 & 0.467 & 0.288 & 0.513 & 0.328 & 0.422 & 0.282 & 0.678 & 0.380 & 0.544 & 0.321 & 0.624 & 0.333 & 0.482 & 0.308 & 0.673 & 0.419 & 0.566 & 0.296 \\
      \midrule
      \multicolumn{2}{c}{\textbf{$1^{\text{st}}$ Count}} & \multicolumn{2}{c}{\textcolor{red}{\textbf{40}}} & \multicolumn{2}{c}{\underline{\textcolor{blue}{12}}} & \multicolumn{2}{c}{5} & \multicolumn{2}{c}{10} & \multicolumn{2}{c}{0} & \multicolumn{2}{c}{0} & \multicolumn{2}{c}{0} & \multicolumn{2}{c}{0} & \multicolumn{2}{c}{0} & \multicolumn{2}{c}{0} & \multicolumn{2}{c}{0} & \multicolumn{2}{c}{0} & \multicolumn{2}{c}{0} & \multicolumn{2}{c}{0} \\
      \bottomrule
    \end{tabular}
  }
  \vspace{2mm}
  \resizebox{\textwidth}{!}{
    \setlength{\tabcolsep}{2pt}
    \begin{tabular}{cc|rr|rr|rr|rr|rr|rr|rr|rr|rr|rr|rr|rr|rr|rr|rr}
      \toprule
      \multicolumn{2}{c}{\multirow{2}{*}{Models}} & \multicolumn{2}{c}{SegRNN} & \multicolumn{2}{c}{Koopa} & \multicolumn{2}{c}{TSMixer} & \multicolumn{2}{c}{FreTS} & \multicolumn{2}{c}{Pyra.} & \multicolumn{2}{c}{Nonsta.} & \multicolumn{2}{c}{ETS.} & \multicolumn{2}{c}{FED.} & \multicolumn{2}{c}{SCINet} & \multicolumn{2}{c}{LightTS} & \multicolumn{2}{c}{Auto.} & \multicolumn{2}{c}{In.} & \multicolumn{2}{c}{Re.} & \multicolumn{2}{c}{Trans.} & \multicolumn{2}{c}{FiLM} \\
      \multicolumn{2}{c}{} & \multicolumn{2}{c}{\cite{lin2023segrnn}} & \multicolumn{2}{c}{\cite{liu2023koopa}} & \multicolumn{2}{c}{\cite{chen2023tsmixer}} & \multicolumn{2}{c}{\cite{yi2023frets}} & \multicolumn{2}{c}{\cite{liu2022pyraformer}} & \multicolumn{2}{c}{\cite{Liu2022NonstationaryTR}} & \multicolumn{2}{c}{\cite{woo2022etsformer}} & \multicolumn{2}{c}{\cite{zhou2022fedformer}} & \multicolumn{2}{c}{\cite{liu2022scinet}} & \multicolumn{2}{c}{\cite{zhang2022LightTS}} & \multicolumn{2}{c}{\cite{wu2021autoformer}} & \multicolumn{2}{c}{\cite{zhou2021informer}} & \multicolumn{2}{c}{\cite{Kitaev2020Reformer}} & \multicolumn{2}{c}{\cite{vaswani2017attention}} & \multicolumn{2}{c}{\cite{zhou2022film}} \\
      \cmidrule(l){3-32}
      \multicolumn{2}{c}{Metric} & MSE & MAE & MSE & MAE & MSE & MAE & MSE & MAE & MSE & MAE & MSE & MAE & MSE & MAE & MSE & MAE & MSE & MAE & MSE & MAE & MSE & MAE & MSE & MAE & MSE & MAE & MSE & MAE & MSE & MAE \\
      \midrule
      \multirow{5}{*}{\rotatebox{90}{\textbf{ETTh1}}}
      & 96 & 0.371 & 0.396 & 0.403 & 0.416 & 0.483 & 0.493 & 0.396 & 0.408 & 0.707 & 0.631 & 0.534 & 0.499 & 0.496 & 0.481 & 0.377 & 0.416 & 0.467 & 0.457 & 0.449 & 0.452 & 0.438 & 0.450 & 0.920 & 0.729 & 0.834 & 0.664 & 0.852 & 0.723 & 0.411 & 0.428 \\
      & 192 & 0.417 & 0.422 & 0.424 & 0.440 & 0.571 & 0.548 & 0.453 & 0.444 & 0.711 & 0.627 & 0.545 & 0.509 & 0.723 & 0.643 & 0.414 & 0.440 & 0.512 & 0.481 & 0.499 & 0.481 & 0.544 & 0.502 & 0.958 & 0.750 & 0.933 & 0.717 & 0.906 & 0.753 & 0.443 & 0.443 \\
      & 336 & 0.451 & 0.439 & 0.468 & 0.470 & 0.666 & 0.608 & 0.499 & 0.470 & 0.990 & 0.797 & 0.743 & 0.627 & 0.895 & 0.740 & 0.456 & 0.466 & 0.548 & 0.498 & 0.552 & 0.512 & 0.484 & 0.478 & 1.157 & 0.839 & 0.954 & 0.736 & 1.116 & 0.844 & 0.461 & 0.453 \\
      & 720 & \tcb{0.445} & \tcr{0.456} & 0.586 & 0.555 & 0.738 & 0.667 & 0.555 & 0.532 & 0.973 & 0.782 & 0.802 & 0.667 & 0.919 & 0.764 & 0.521 & 0.502 & 0.553 & 0.518 & 0.622 & 0.575 & 0.544 & 0.524 & 1.239 & 0.887 & 1.161 & 0.831 & 1.017 & 0.809 & \tcr{0.438} & 0.465 \\
      & Avg & \tcb{0.421} & 0.428 & 0.471 & 0.470 & 0.615 & 0.579 & 0.476 & 0.464 & 0.845 & 0.709 & 0.656 & 0.576 & 0.758 & 0.657 & 0.442 & 0.456 & 0.520 & 0.488 & 0.530 & 0.505 & 0.502 & 0.489 & 1.068 & 0.801 & 0.971 & 0.737 & 0.973 & 0.782 & 0.438 & 0.447 \\
      \midrule
      \multirow{5}{*}{\rotatebox{90}{\textbf{ETTh2}}}
      & 96 & 0.282 & 0.340 & 0.307 & 0.359 & 1.107 & 0.837 & 0.354 & 0.404 & 1.653 & 1.007 & 0.437 & 0.440 & 0.386 & 0.426 & 0.344 & 0.388 & 0.345 & 0.386 & 0.394 & 0.432 & 0.375 & 0.412 & 2.837 & 1.345 & 1.782 & 1.073 & 2.147 & 1.180 & 0.317 & 0.355 \\
      & 192 & 0.371 & 0.396 & 0.370 & 0.401 & 2.569 & 1.379 & 0.485 & 0.481 & 4.623 & 1.709 & 0.510 & 0.479 & 0.514 & 0.501 & 0.426 & 0.443 & 0.424 & 0.431 & 0.515 & 0.498 & 0.613 & 0.546 & 6.436 & 2.112 & 2.612 & 1.307 & 4.170 & 1.637 & 0.391 & 0.401 \\
      & 336 & 0.421 & 0.433 & 0.414 & 0.441 & 2.556 & 1.360 & 0.613 & 0.552 & 5.105 & 1.906 & 0.602 & 0.529 & 0.748 & 0.623 & 0.455 & 0.465 & 0.464 & 0.464 & 0.666 & 0.575 & 0.466 & 0.474 & 4.886 & 1.820 & 2.545 & 1.257 & 3.450 & 1.437 & 0.416 & 0.422 \\
      & 720 & 0.423 & 0.451 & 0.476 & 0.494 & 2.423 & 1.307 & 0.739 & 0.620 & 4.217 & 1.765 & 0.668 & 0.566 & 0.683 & 0.600 & 0.481 & 0.489 & 0.480 & 0.479 & 0.956 & 0.700 & 0.484 & 0.500 & 3.861 & 1.688 & 3.010 & 1.316 & 2.715 & 1.363 & 0.427 & 0.439 \\
      & Avg & 0.374 & 0.405 & 0.392 & 0.424 & 2.164 & 1.221 & 0.548 & 0.514 & 3.900 & 1.597 & 0.554 & 0.504 & 0.583 & 0.538 & 0.427 & 0.446 & 0.428 & 0.440 & 0.633 & 0.551 & 0.485 & 0.483 & 4.505 & 1.741 & 2.487 & 1.238 & 3.121 & 1.404 & 0.388 & 0.405 \\
      \midrule
      \multirow{5}{*}{\rotatebox{90}{\textbf{ETTm1}}}
      & 96 & 0.331 & 0.371 & 0.302 & 0.352 & 0.488 & 0.472 & 0.340 & 0.374 & 0.577 & 0.510 & 0.422 & 0.419 & 0.556 & 0.543 & 0.377 & 0.421 & 0.345 & 0.380 & 0.360 & 0.395 & 0.520 & 0.478 & 0.771 & 0.659 & 0.901 & 0.671 & 0.560 & 0.523 & 0.308 & 0.353 \\
      & 192 & 0.371 & 0.392 & 0.352 & 0.386 & 0.473 & 0.479 & 0.382 & 0.398 & 0.616 & 0.555 & 0.499 & 0.455 & 0.564 & 0.546 & 0.426 & 0.442 & 0.385 & 0.398 & 0.404 & 0.419 & 0.576 & 0.507 & 0.736 & 0.619 & 0.919 & 0.689 & 0.714 & 0.628 & 0.340 & 0.369 \\
      & 336 & 0.398 & 0.412 & 0.380 & 0.405 & 0.539 & 0.530 & 0.419 & 0.423 & 0.789 & 0.652 & 0.548 & 0.490 & 0.682 & 0.627 & 0.445 & 0.455 & 0.418 & 0.417 & 0.446 & 0.451 & 0.630 & 0.534 & 1.061 & 0.789 & 1.029 & 0.744 & 1.076 & 0.800 & 0.369 & 0.386 \\
      & 720 & 0.456 & 0.444 & 0.427 & 0.438 & 0.621 & 0.574 & 0.496 & 0.472 & 1.023 & 0.731 & 0.678 & 0.547 & 0.780 & 0.672 & 0.500 & 0.483 & 0.488 & 0.452 & 0.542 & 0.515 & 0.576 & 0.524 & 1.232 & 0.832 & 1.157 & 0.798 & 1.018 & 0.777 & 0.419 & 0.414 \\
      & Avg & 0.389 & 0.405 & 0.365 & 0.395 & 0.530 & 0.514 & 0.409 & 0.417 & 0.751 & 0.612 & 0.537 & 0.478 & 0.645 & 0.597 & 0.437 & 0.450 & 0.409 & 0.412 & 0.438 & 0.445 & 0.575 & 0.511 & 0.950 & 0.725 & 1.001 & 0.725 & 0.842 & 0.682 & 0.359 & 0.380 \\
      \midrule
      \multirow{5}{*}{\rotatebox{90}{\textbf{ETTm2}}}
      & 96 & 0.173 & 0.255 & 0.178 & 0.259 & 0.252 & 0.364 & 0.192 & 0.283 & 0.419 & 0.476 & 0.231 & 0.301 & 0.265 & 0.376 & 0.193 & 0.282 & 0.183 & 0.268 & 0.225 & 0.320 & 0.250 & 0.323 & 0.495 & 0.559 & 0.772 & 0.654 & 0.437 & 0.483 & 0.176 & 0.265 \\
      & 192 & 0.236 & 0.297 & 0.233 & 0.300 & 0.466 & 0.531 & 0.278 & 0.343 & 0.764 & 0.658 & 0.434 & 0.402 & 0.814 & 0.752 & 0.266 & 0.326 & 0.250 & 0.310 & 0.326 & 0.392 & 0.288 & 0.347 & 0.549 & 0.582 & 1.481 & 0.917 & 0.997 & 0.746 & 0.222 & 0.294 \\
      & 336 & 0.294 & 0.335 & 0.288 & 0.345 & 0.901 & 0.760 & 0.359 & 0.396 & 1.335 & 0.896 & 0.475 & 0.436 & 1.507 & 1.049 & 0.324 & 0.363 & 0.317 & 0.352 & 0.496 & 0.492 & 0.351 & 0.384 & 1.577 & 0.970 & 2.167 & 1.098 & 1.394 & 0.916 & 0.274 & 0.329 \\
      & 720 & 0.387 & 0.401 & \tcb{0.353} & 0.394 & 2.522 & 1.346 & 0.557 & 0.519 & 5.022 & 1.791 & 0.665 & 0.513 & 4.202 & 1.779 & 0.421 & 0.422 & 0.426 & 0.413 & 0.678 & 0.586 & 0.435 & 0.428 & 3.422 & 1.372 & 3.022 & 1.308 & 3.453 & 1.391 & 0.354 & 0.382 \\
      & Avg & 0.273 & 0.322 & 0.263 & 0.325 & 1.035 & 0.750 & 0.347 & 0.385 & 1.885 & 0.955 & 0.451 & 0.413 & 1.697 & 0.989 & 0.301 & 0.348 & 0.294 & 0.335 & 0.431 & 0.448 & 0.331 & 0.371 & 1.511 & 0.870 & 1.861 & 0.994 & 1.570 & 0.884 & 0.257 & 0.317 \\
      \midrule
      \multirow{5}{*}{\rotatebox{90}{\textbf{Weather}}}
      & 96 & 0.166 & 0.227 & 0.167 & 0.218 & 0.179 & 0.249 & 0.184 & 0.239 & 0.206 & 0.289 & 0.183 & 0.228 & 0.209 & 0.298 & 0.216 & 0.296 & 0.168 & 0.215 & 0.170 & 0.232 & 0.319 & 0.366 & 0.339 & 0.396 & 0.363 & 0.393 & 0.369 & 0.409 & 0.195 & 0.236 \\
      & 192 & 0.213 & 0.273 & 0.199 & 0.247 & 0.216 & 0.284 & 0.223 & 0.274 & 0.259 & 0.336 & 0.252 & 0.289 & 0.283 & 0.369 & 0.295 & 0.367 & 0.220 & 0.260 & 0.215 & 0.274 & 0.329 & 0.382 & 0.441 & 0.455 & 0.408 & 0.430 & 0.538 & 0.511 & 0.230 & 0.265 \\
      & 336 & 0.272 & 0.318 & \tcb{0.245} & 0.285 & 0.260 & 0.316 & 0.271 & 0.313 & 0.298 & 0.359 & 0.298 & 0.318 & 0.456 & 0.497 & 0.334 & 0.384 & 0.278 & 0.302 & 0.263 & 0.313 & 0.375 & 0.413 & 0.602 & 0.549 & 0.638 & 0.578 & 0.643 & 0.580 & 0.266 & 0.295 \\
      & 720 & 0.356 & 0.375 & \tcr{0.307} & 0.335 & 0.317 & 0.357 & 0.342 & 0.369 & 0.417 & 0.436 & 0.474 & 0.432 & 0.444 & 0.485 & 0.415 & 0.426 & 0.360 & 0.355 & 0.330 & 0.363 & 0.406 & 0.415 & 1.101 & 0.778 & 0.545 & 0.514 & 0.879 & 0.684 & 0.322 & 0.339 \\
      & Avg & 0.252 & 0.298 & \tcb{0.230} & 0.271 & 0.243 & 0.301 & 0.255 & 0.299 & 0.295 & 0.355 & 0.302 & 0.317 & 0.348 & 0.412 & 0.315 & 0.368 & 0.256 & 0.283 & 0.245 & 0.295 & 0.357 & 0.394 & 0.621 & 0.545 & 0.489 & 0.479 & 0.607 & 0.546 & 0.253 & 0.284 \\
      \midrule
      \multirow{5}{*}{\rotatebox{90}{\textbf{ECL}}}
      & 96 & 0.157 & 0.251 & 0.153 & 0.256 & 0.214 & 0.318 & 0.189 & 0.277 & 0.284 & 0.375 & 0.168 & 0.270 & 0.247 & 0.351 & 0.197 & 0.311 & 0.180 & 0.289 & 0.214 & 0.319 & 0.195 & 0.312 & 0.341 & 0.421 & 0.299 & 0.386 & 0.256 & 0.356 & 0.154 & 0.246 \\
      & 192 & 0.170 & 0.263 & 0.169 & 0.271 & 0.220 & 0.331 & 0.192 & 0.279 & 0.292 & 0.386 & 0.182 & 0.283 & 0.267 & 0.363 & 0.212 & 0.323 & 0.206 & 0.311 & 0.227 & 0.331 & 0.232 & 0.337 & 0.370 & 0.446 & 0.335 & 0.413 & 0.272 & 0.370 & 0.167 & 0.259 \\
      & 336 & 0.186 & 0.281 & 0.229 & 0.328 & 0.241 & 0.352 & 0.207 & 0.296 & 0.305 & 0.399 & 0.195 & 0.300 & 0.281 & 0.375 & 0.221 & 0.336 & 0.234 & 0.337 & 0.248 & 0.351 & 0.220 & 0.332 & 0.388 & 0.460 & 0.346 & 0.420 & 0.287 & 0.380 & 0.189 & 0.284 \\
      & 720 & 0.224 & 0.316 & 0.258 & 0.354 & 0.271 & 0.371 & 0.246 & 0.333 & 0.307 & 0.394 & 0.233 & 0.324 & 0.309 & 0.394 & 0.272 & 0.375 & 0.259 & 0.354 & 0.281 & 0.374 & 0.268 & 0.372 & 0.400 & 0.461 & 0.313 & 0.392 & 0.276 & 0.364 & 0.249 & 0.340 \\
      & Avg & 0.184 & 0.278 & 0.202 & 0.303 & 0.236 & 0.343 & 0.209 & 0.296 & 0.297 & 0.389 & 0.195 & 0.294 & 0.276 & 0.371 & 0.225 & 0.336 & 0.220 & 0.323 & 0.243 & 0.344 & 0.229 & 0.338 & 0.375 & 0.447 & 0.323 & 0.403 & 0.273 & 0.367 & 0.190 & 0.282 \\
      \midrule
      \multirow{5}{*}{\rotatebox{90}{\textbf{Traffic}}}
      & 96 & 0.630 & 0.315 & 0.446 & 0.321 & 0.554 & 0.375 & 0.564 & 0.368 & 0.685 & 0.389 & 0.610 & 0.339 & 0.974 & 0.570 & 0.587 & 0.369 & 0.585 & 0.377 & 0.612 & 0.405 & 0.617 & 0.390 & 0.742 & 0.416 & 0.703 & 0.392 & 0.643 & 0.357 & 0.411 & 0.285 \\
      & 192 & 0.639 & 0.319 & 0.451 & 0.329 & 0.571 & 0.393 & 0.567 & 0.366 & 0.678 & 0.383 & 0.638 & 0.352 & 1.028 & 0.582 & 0.606 & 0.373 & 0.634 & 0.409 & 0.637 & 0.421 & 0.645 & 0.404 & 0.774 & 0.434 & 0.690 & 0.378 & 0.670 & 0.368 & 0.406 & 0.284 \\
      & 336 & 0.659 & 0.326 & 0.599 & 0.419 & 0.563 & 0.381 & 0.598 & 0.374 & 0.692 & 0.389 & 0.664 & 0.365 & 1.049 & 0.587 & 0.632 & 0.391 & 0.671 & 0.431 & 0.660 & 0.432 & 0.610 & 0.379 & 0.832 & 0.468 & 0.693 & 0.377 & 0.684 & 0.373 & 0.425 & 0.297 \\
      & 720 & 0.696 & 0.346 & 0.679 & 0.452 & 0.628 & 0.424 & 0.660 & 0.399 & 0.719 & 0.400 & 0.680 & 0.370 & 1.093 & 0.602 & 0.633 & 0.384 & 0.725 & 0.459 & 0.717 & 0.454 & 0.657 & 0.406 & 0.945 & 0.527 & 0.697 & 0.378 & 0.683 & 0.374 & 0.525 & 0.371 \\
      & Avg & 0.656 & 0.327 & 0.544 & 0.380 & 0.579 & 0.393 & 0.598 & 0.377 & 0.693 & 0.390 & 0.648 & 0.356 & 1.036 & 0.585 & 0.615 & 0.379 & 0.654 & 0.419 & 0.656 & 0.428 & 0.632 & 0.395 & 0.823 & 0.461 & 0.696 & 0.381 & 0.670 & 0.368 & 0.442 & 0.309 \\
      \midrule
      \multicolumn{2}{c}{\textbf{$1^{\text{st}}$ Count}} & \multicolumn{2}{c}{1} & \multicolumn{2}{c}{1} & \multicolumn{2}{c}{0} & \multicolumn{2}{c}{0} & \multicolumn{2}{c}{0} & \multicolumn{2}{c}{0} & \multicolumn{2}{c}{0} & \multicolumn{2}{c}{0} & \multicolumn{2}{c}{0} & \multicolumn{2}{c}{0} & \multicolumn{2}{c}{0} & \multicolumn{2}{c}{0} & \multicolumn{2}{c}{0} & \multicolumn{2}{c}{0} & \multicolumn{2}{c}{1} \\
      \bottomrule
    \end{tabular}
  }
\end{table*}

%% file: Table_Exp/Exp_SOTA_ShortTerm_Full_icml.tex
\begin{table*}[htbp]
  \centering
  \caption{Full results for the short-term forecasting task in the M4 dataset. Lower OWA, SMAPE, and MASE values indicate superior accuracy. We highlight the \textcolor{red}{\textbf{1st}} and \underline{\textcolor{blue}{2nd}} best results.}
  \label{tab:TSGym_vs_Sota_short_full}
  \resizebox{\textwidth}{!}{
    \renewcommand{\arraystretch}{1.1}
    \setlength{\tabcolsep}{1.5pt}
    \begin{tabular}{cc|ccccccccccccccccccccccccccc}
      \toprule
      \multicolumn{2}{c|}{Metric} & \system & OLinear & RAFT & DUET & TimeMixer & TimeXer & PAttn & iTrans. & Mamba & MICN & TimesNet & PatchTST & DLinear & Cross. & SegRNN & TSMixer & FreTS & Pyra. & ETS. & FED. & SCINet & LightTS & Auto. & In. & Re. & Trans. & FiLM \\
      \midrule
      \multirow{3}{*}{\rotatebox{90}{Yearly}} & OWA & 0.795 & 1.661 & 0.842 & 0.845 & \textcolor{red}{\textbf{0.786}} & 0.797 & 0.829 & 0.837 & \textcolor{blue}{\underline{0.787}} & 0.870 & 0.791 & 0.801 & 0.843 & 4.790 & 0.859 & 0.795 & 0.800 & 0.935 & 0.987 & 0.808 & 0.801 & 0.794 & 1.027 & 1.070 & 1.186 & 4.414 & 0.806 \\
      & SMAPE & 13.553 & 28.555 & 14.391 & 14.386 & \textcolor{red}{\textbf{13.322}} & 13.547 & 14.067 & 14.223 & \textcolor{blue}{\underline{13.348}} & 14.549 & 13.411 & 13.631 & 14.402 & 79.570 & 14.336 & 13.541 & 13.563 & 16.064 & 16.128 & 13.669 & 13.578 & 13.440 & 17.457 & 18.428 & 20.135 & 69.678 & 13.986 \\
      & MASE & 3.022 & 6.260 & 3.195 & 3.221 & \textcolor{red}{\textbf{3.007}} & 3.039 & 3.169 & 3.193 & \textcolor{blue}{\underline{3.010}} & 3.379 & 3.027 & 3.051 & 3.198 & 18.721 & 3.339 & 3.027 & 3.058 & 3.529 & 3.926 & 3.099 & 3.068 & 3.045 & 3.916 & 4.020 & 4.533 & 18.139 & \textcolor{red}{\textbf{3.007}} \\
      \midrule
      \multirow{3}{*}{\rotatebox{90}{Quarterly}} & OWA & \textcolor{blue}{\underline{0.886}} & 1.761 & 0.954 & 0.997 & 0.898 & 0.932 & 0.900 & 0.960 & 0.911 & 1.020 & \textcolor{red}{\textbf{0.885}} & 0.969 & 0.928 & 8.195 & 1.009 & 0.924 & 0.909 & 1.008 & 1.311 & 0.938 & 0.922 & 0.888 & 1.289 & 1.110 & 0.993 & 8.191 & 0.960 \\
      & SMAPE & \textcolor{blue}{\underline{10.168}} & 19.089 & 10.668 & 11.138 & 10.182 & 10.460 & 10.200 & 10.800 & 10.305 & 11.384 & \textcolor{red}{\textbf{10.049}} & 10.893 & 10.500 & 74.237 & 11.188 & 10.486 & 10.330 & 11.315 & 13.568 & 10.629 & 10.425 & 10.177 & 14.119 & 12.380 & 11.141 & 73.874 & 10.743 \\
      & MASE & \textcolor{red}{\textbf{1.162}} & 2.454 & 1.288 & 1.348 & 1.195 & 1.254 & 1.198 & 1.287 & 1.214 & 1.380 & 1.176 & 1.302 & 1.238 & 13.231 & 1.374 & 1.227 & 1.208 & 1.355 & 1.906 & 1.248 & 1.229 & \textcolor{blue}{\underline{1.167}} & 1.777 & 1.502 & 1.337 & 13.266 & 1.296 \\
      \midrule
      \multirow{3}{*}{\rotatebox{90}{Monthly}} & OWA & \textcolor{red}{\textbf{0.885}} & 1.630 & 0.940 & 0.983 & \textcolor{blue}{\underline{0.885}} & 0.951 & 0.950 & 0.992 & 0.931 & 0.981 & \textcolor{blue}{\underline{0.885}} & 1.019 & 0.936 & 7.637 & 1.076 & 0.912 & 0.916 & 1.047 & 1.285 & 0.994 & 0.924 & 0.886 & 1.364 & 1.118 & 1.428 & 7.667 & 0.942 \\
      & SMAPE & 12.944 & 21.930 & 13.373 & 13.727 & \textcolor{blue}{\underline{12.747}} & 13.296 & 13.378 & 13.868 & 13.172 & 13.754 & 12.759 & 14.139 & 13.382 & 68.893 & 15.061 & 13.061 & 13.068 & 14.631 & 15.494 & 14.052 & 13.146 & \textcolor{red}{\textbf{12.744}} & 18.161 & 15.453 & 18.721 & 70.067 & 13.351 \\
      & MASE & \textcolor{red}{\textbf{0.928}} & 1.849 & 1.012 & 1.080 & 0.944 & 1.042 & 1.034 & 1.086 & 1.009 & 1.072 & \textcolor{blue}{\underline{0.942}} & 1.125 & 1.003 & 11.163 & 1.178 & 0.977 & 0.985 & 1.148 & 1.591 & 1.078 & 0.996 & 0.944 & 1.563 & 1.239 & 1.657 & 11.142 & 1.019 \\
      \midrule
      \multirow{3}{*}{\rotatebox{90}{Weekly}} & OWA & 1.094 & 1.427 & 1.049 & 1.087 & 1.187 & 1.134 & 1.236 & 1.275 & 1.449 & 1.501 & 1.158 & 1.048 & 1.461 & 28.636 & \textcolor{red}{\textbf{0.997}} & 1.555 & 1.286 & 1.457 & \textcolor{blue}{\underline{1.025}} & 1.097 & 1.438 & 1.340 & 1.557 & 1.343 & 1.382 & 28.094 & 1.280 \\
      & SMAPE & 9.579 & 12.098 & 9.492 & 9.741 & 10.919 & 10.222 & 10.759 & 11.122 & 12.495 & 11.791 & 10.601 & 9.540 & 11.805 & 198.371 & \textcolor{red}{\textbf{9.150}} & 12.661 & 11.394 & 13.023 & \textcolor{blue}{\underline{9.238}} & 9.742 & 12.757 & 12.060 & 12.484 & 12.000 & 11.204 & 191.432 & 11.539 \\
      & MASE & 3.173 & 4.256 & 2.948 & 3.084 & 3.281 & 3.200 & 3.604 & 3.708 & 4.262 & 4.761 & 3.219 & 2.930 & 4.534 & 98.925 & \textcolor{red}{\textbf{2.766}} & 4.799 & 3.688 & 4.145 & \textcolor{blue}{\underline{2.890}} & 3.139 & 4.122 & 3.789 & 4.865 & 3.824 & 4.278 & 98.016 & 3.609 \\
      \midrule
      \multirow{3}{*}{\rotatebox{90}{Daily}} & OWA & \textcolor{red}{\textbf{0.987}} & 1.252 & 1.003 & 1.017 & 1.015 & 1.059 & 1.014 & 1.174 & 1.142 & 1.242 & 1.042 & 1.006 & 1.089 & 48.627 & \textcolor{blue}{\underline{0.987}} & 1.200 & 1.093 & 1.220 & 1.056 & 1.000 & 1.082 & 1.088 & 1.455 & 1.343 & 1.511 & 29.621 & 1.082 \\
      & SMAPE & \textcolor{blue}{\underline{3.026}} & 3.771 & 3.057 & 3.096 & 3.086 & 3.232 & 3.105 & 3.564 & 3.452 & 3.780 & 3.171 & 3.080 & 3.316 & 179.226 & \textcolor{red}{\textbf{3.008}} & 3.661 & 3.313 & 3.691 & 3.200 & 3.075 & 3.288 & 3.317 & 4.354 & 4.046 & 4.553 & 99.710 & 3.267 \\
      & MASE & \textcolor{red}{\textbf{3.211}} & 4.148 & 3.285 & 3.337 & 3.330 & 3.464 & 3.304 & 3.864 & 3.771 & 4.076 & 3.419 & 3.280 & 3.572 & 125.892 & \textcolor{blue}{\underline{3.236}} & 3.927 & 3.598 & 4.027 & 3.476 & 3.247 & 3.552 & 3.562 & 4.855 & 4.454 & 5.005 & 86.874 & 3.580 \\
      \midrule
      \multirow{3}{*}{\rotatebox{90}{Hourly}} & OWA & \textcolor{blue}{\underline{0.868}} & 1.397 & 81.403 & 1.277 & 1.542 & 1.645 & 1.510 & 1.327 & \textcolor{red}{\textbf{0.731}} & 1.528 & 1.109 & 2.180 & 1.040 & 11.691 & 1.726 & 1.567 & 1.373 & 2.829 & 1.164 & 1.058 & 1.368 & 1.190 & 1.703 & 2.945 & 3.108 & 6.496 & 1.444 \\
      & SMAPE & \textcolor{blue}{\underline{17.155}} & 22.655 & 80.152 & 19.110 & 20.247 & 25.842 & 22.273 & 19.809 & \textcolor{red}{\textbf{14.943}} & 24.641 & 18.434 & 27.984 & 17.260 & 128.419 & 34.609 & 22.940 & 20.864 & 29.717 & 19.776 & 18.859 & 24.171 & 21.348 & 25.825 & 35.033 & 33.251 & 99.337 & 21.070 \\
      & MASE & \textcolor{blue}{\underline{1.924}} & 3.740 & 379.485 & 3.627 & 4.749 & 4.514 & 4.330 & 3.777 & \textcolor{red}{\textbf{1.553}} & 4.108 & 2.910 & 6.797 & 2.732 & 39.269 & 3.759 & 4.517 & 3.857 & 9.680 & 3.000 & 2.610 & 3.405 & 2.919 & 4.794 & 9.544 & 10.553 & 18.173 & 4.169 \\
      \midrule
      \multirow{3}{*}{\rotatebox{90}{Average}} & OWA & \textcolor{red}{\textbf{0.869}} & 1.651 & 1.257 & 0.958 & 0.875 & 0.919 & 0.916 & 0.959 & 0.903 & 0.980 & \textcolor{blue}{\underline{0.872}} & 0.961 & 0.921 & 8.941 & 1.009 & 0.905 & 0.898 & 1.028 & 1.212 & 0.939 & 0.906 & 0.877 & 1.274 & 1.123 & 1.278 & 8.041 & 0.924 \\
      & SMAPE & 12.004 & 21.972 & 12.784 & 12.816 & \textcolor{blue}{\underline{11.880}} & 12.289 & 12.367 & 12.793 & 12.118 & 12.984 & \textcolor{red}{\textbf{11.869}} & 12.816 & 12.510 & 78.006 & 13.515 & 12.196 & 12.139 & 13.759 & 14.653 & 12.683 & 12.220 & 11.923 & 16.457 & 14.986 & 16.661 & 72.701 & 12.470 \\
      & MASE & \textcolor{red}{\textbf{1.574}} & 3.122 & 3.250 & 1.750 & 1.604 & 1.677 & 1.683 & 1.757 & 1.649 & 1.829 & \textcolor{blue}{\underline{1.599}} & 1.732 & 1.693 & 18.679 & 1.825 & 1.662 & 1.647 & 1.913 & 2.294 & 1.689 & 1.658 & 1.610 & 2.320 & 2.121 & 2.429 & 16.803 & 1.673 \\
      \bottomrule
    \end{tabular}
  }
\end{table*}

%% file: generated_figures.tex
\subsection{Dimension Importance Analysis}
\begin{figure*}[htbp]
  \centering
  \begin{subfigure}[t]{0.16\textwidth}
    \centering
    \includegraphics[width=\textwidth]{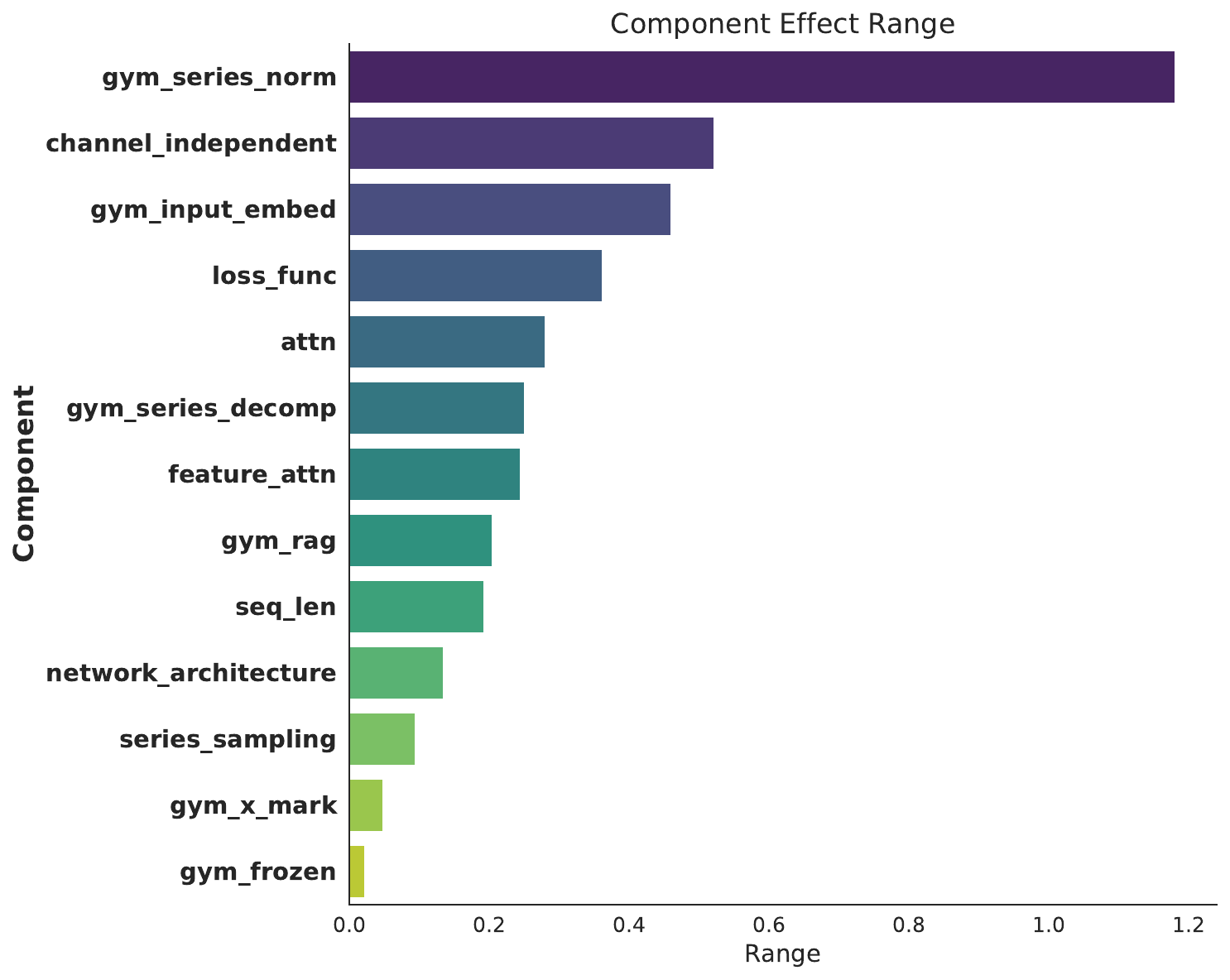}
    \caption{Global}
    \label{fig:appx_effect_range_Global}
  \end{subfigure}
  \hfill
  \begin{subfigure}[t]{0.16\textwidth}
    \centering
    \includegraphics[width=\textwidth]{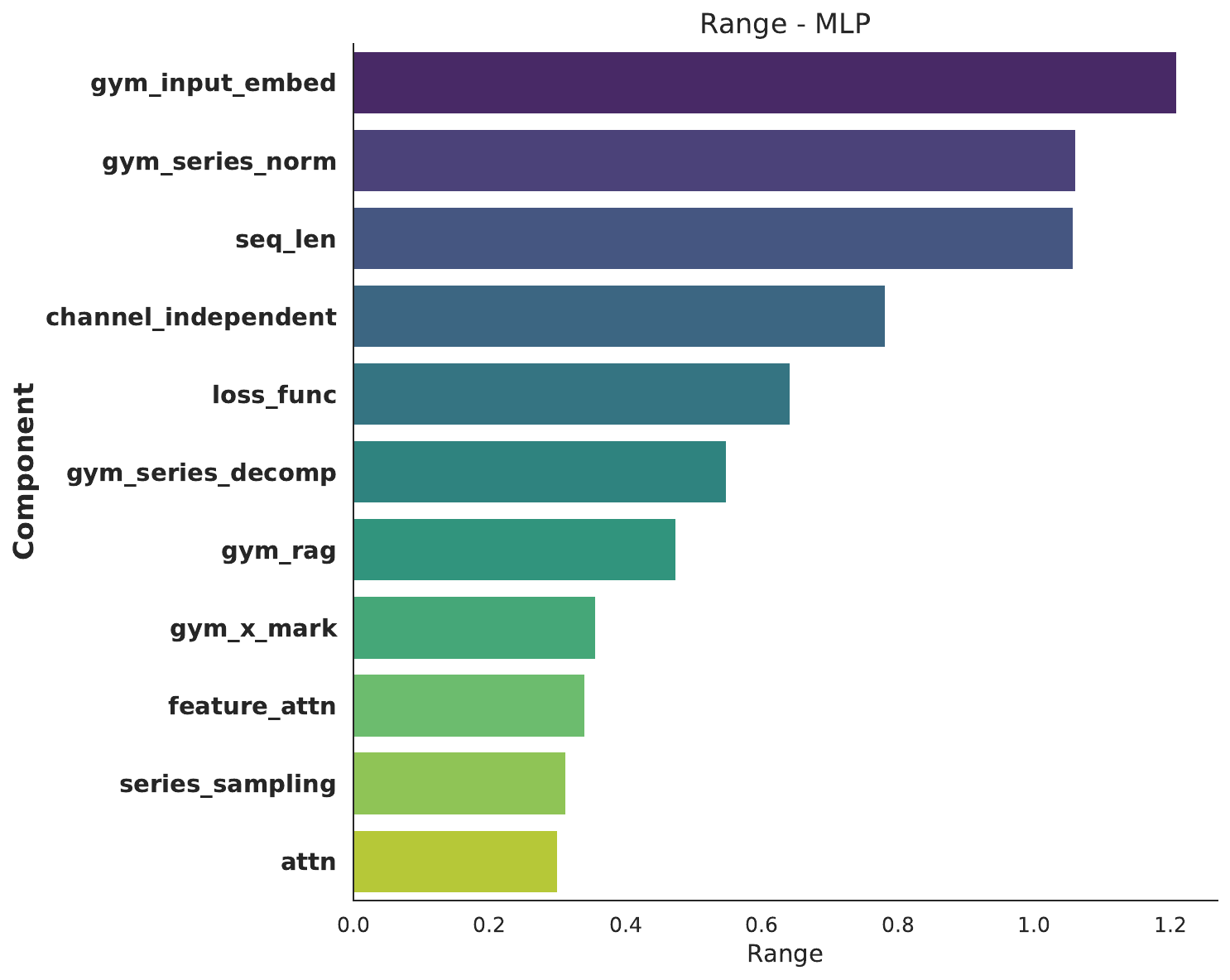}
    \caption{MLP}
    \label{fig:appx_effect_range_MLP}
  \end{subfigure}
  \hfill
  \begin{subfigure}[t]{0.16\textwidth}
    \centering
    \includegraphics[width=\textwidth]{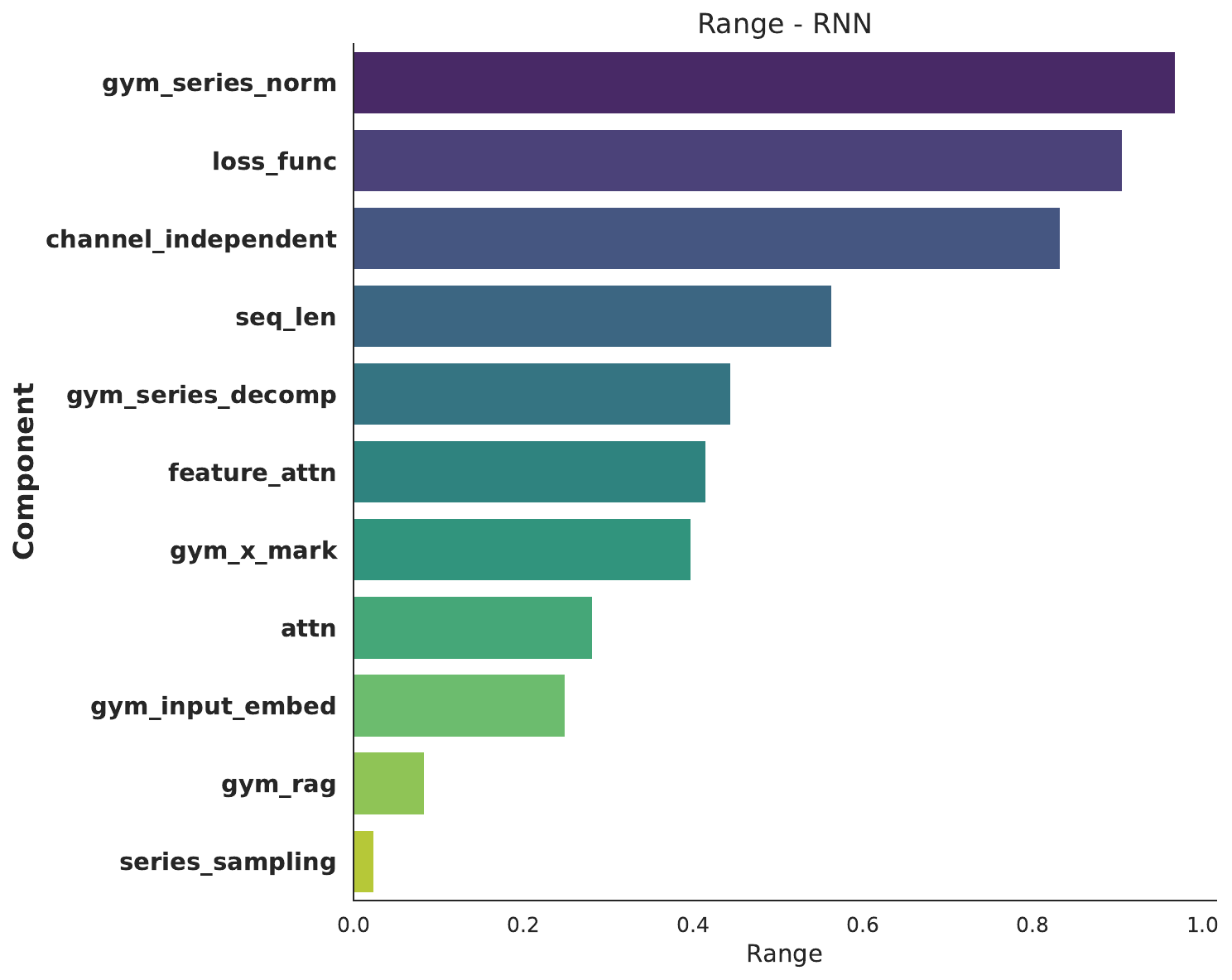}
    \caption{RNN}
    \label{fig:appx_effect_range_RNN}
  \end{subfigure}
  \begin{subfigure}[t]{0.16\textwidth}
    \centering
    \includegraphics[width=\textwidth]{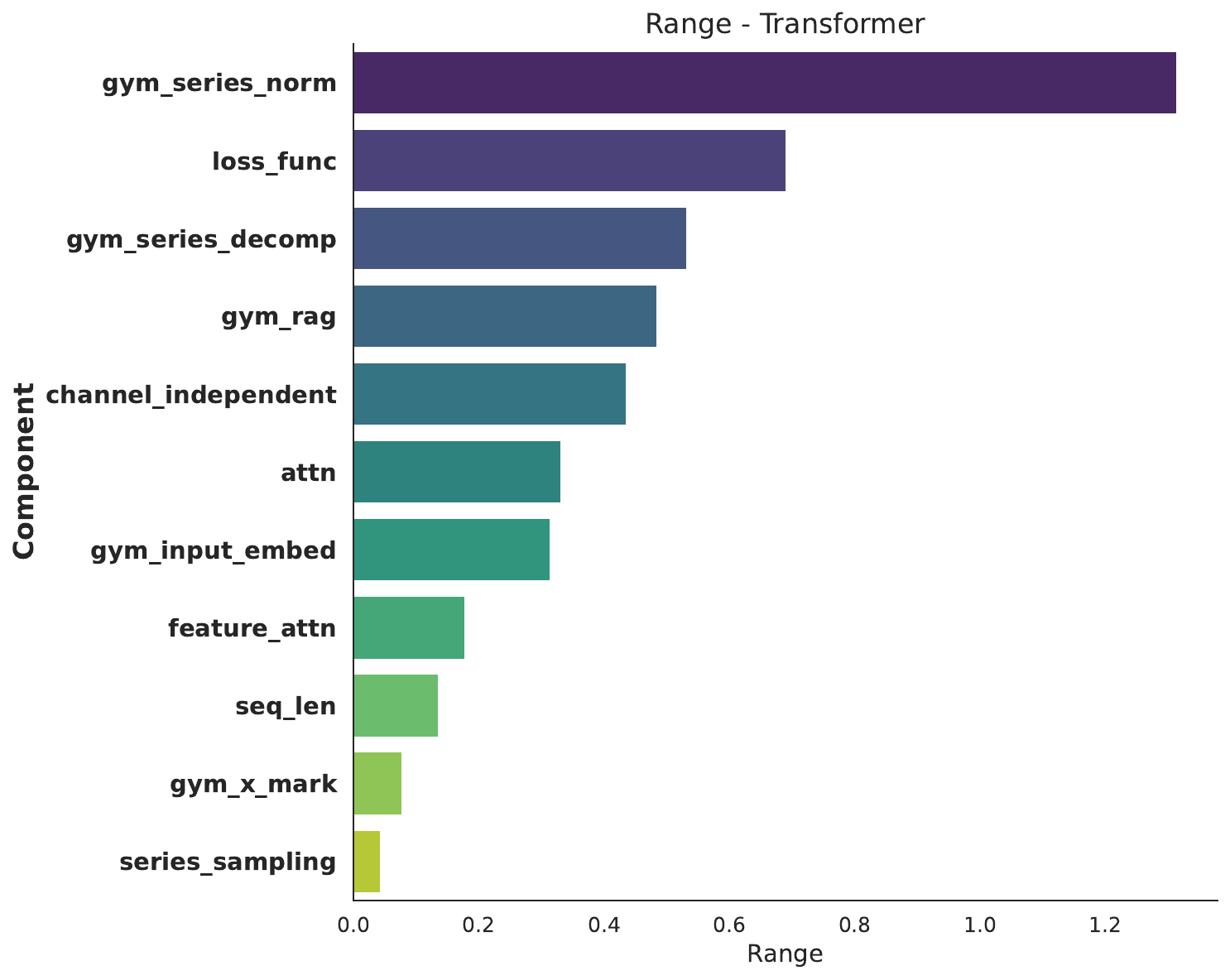}
    \caption{Transformer}
    \label{fig:appx_effect_range_Transformer}
  \end{subfigure}
  \hfill
  \begin{subfigure}[t]{0.16\textwidth}
    \centering
    \includegraphics[width=\textwidth]{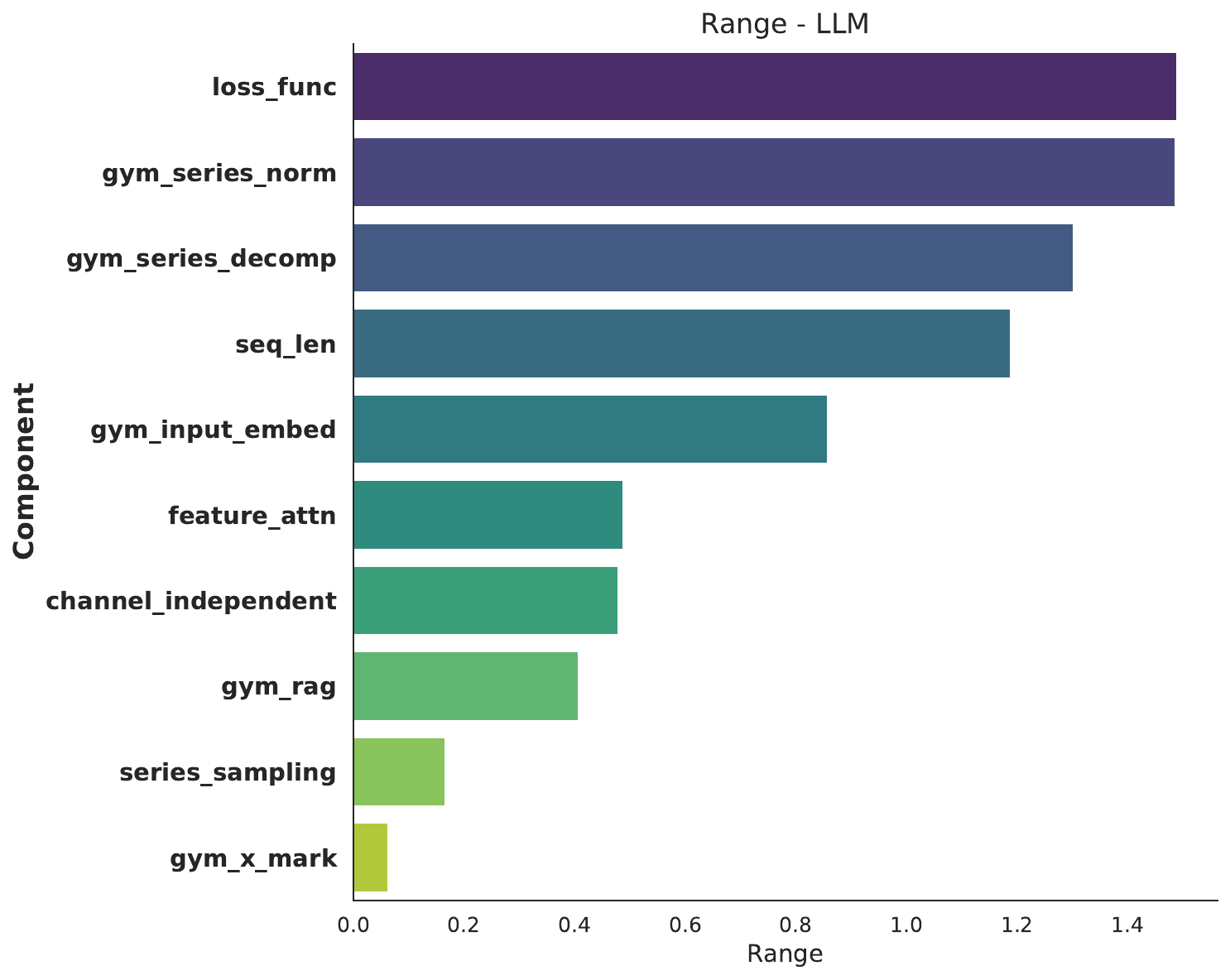}
    \caption{LLM}
    \label{fig:appx_effect_range_LLM}
  \end{subfigure}
  \hfill
  \begin{subfigure}[t]{0.16\textwidth}
    \centering
    \includegraphics[width=\textwidth]{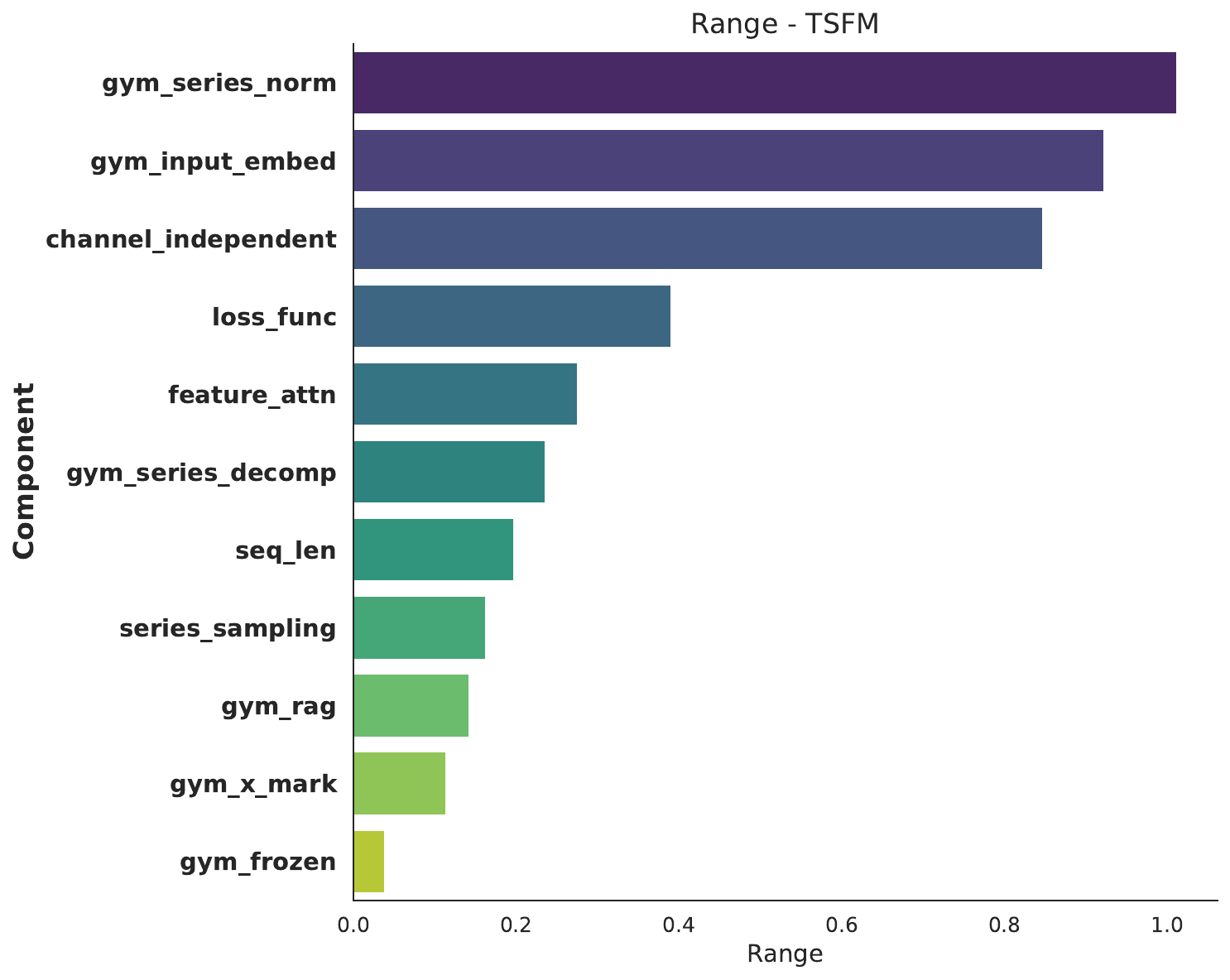}
    \caption{TSFM}
    \label{fig:appx_effect_range_TSFM}
  \end{subfigure}
  \caption{Dimension Importance (Effect Range) (Effect Range Plots). This figure visualizes the performance distributions across different model architectures.}
  \label{fig:appx_effect_range}
\end{figure*}

\begin{figure*}[htbp]
  \centering
  \begin{subfigure}[t]{0.16\textwidth}
    \centering
    \includegraphics[width=\textwidth]{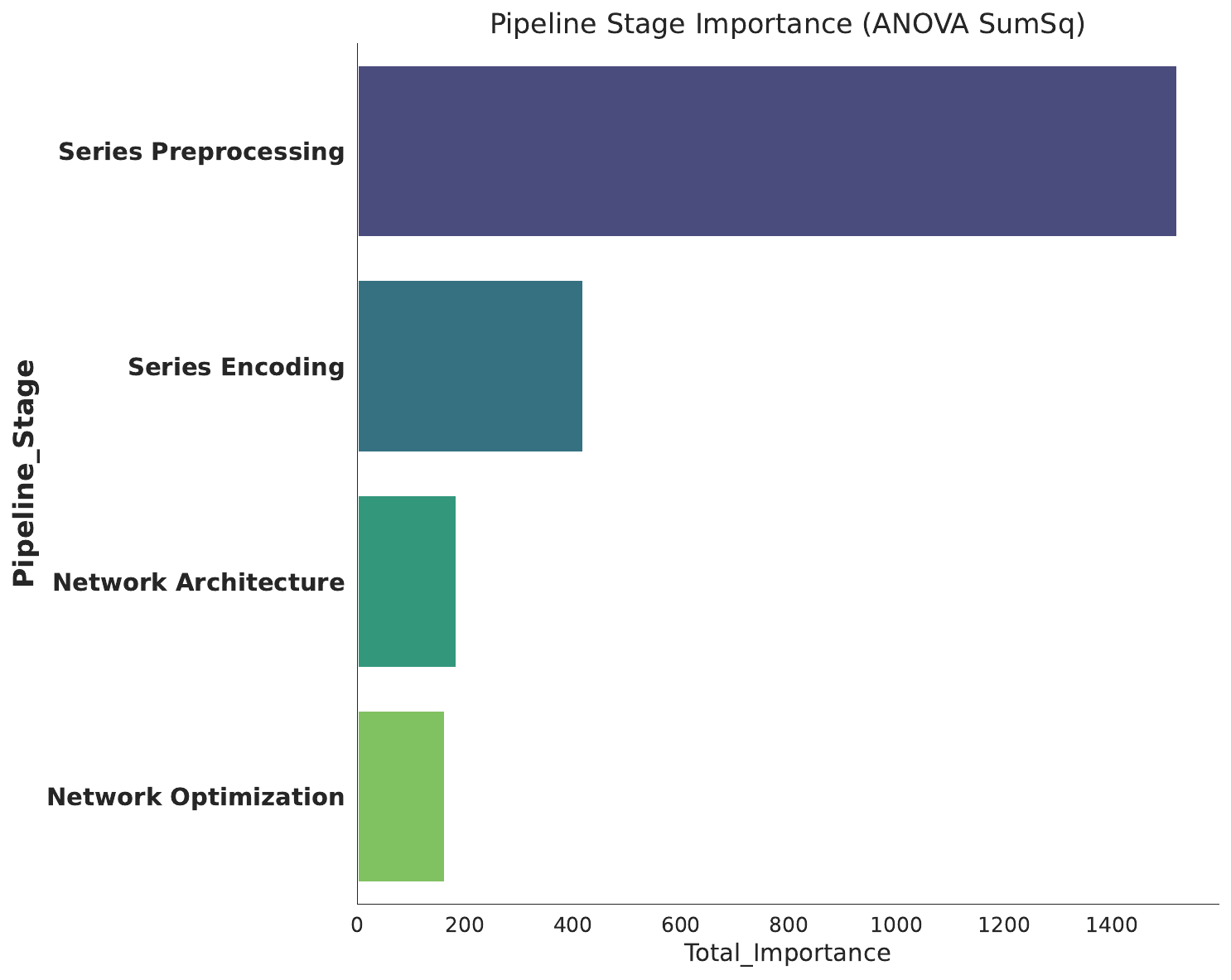}
    \caption{Global}
    \label{fig:appx_pipeline_stage_Global}
  \end{subfigure}
  \hfill
  \begin{subfigure}[t]{0.16\textwidth}
    \centering
    \includegraphics[width=\textwidth]{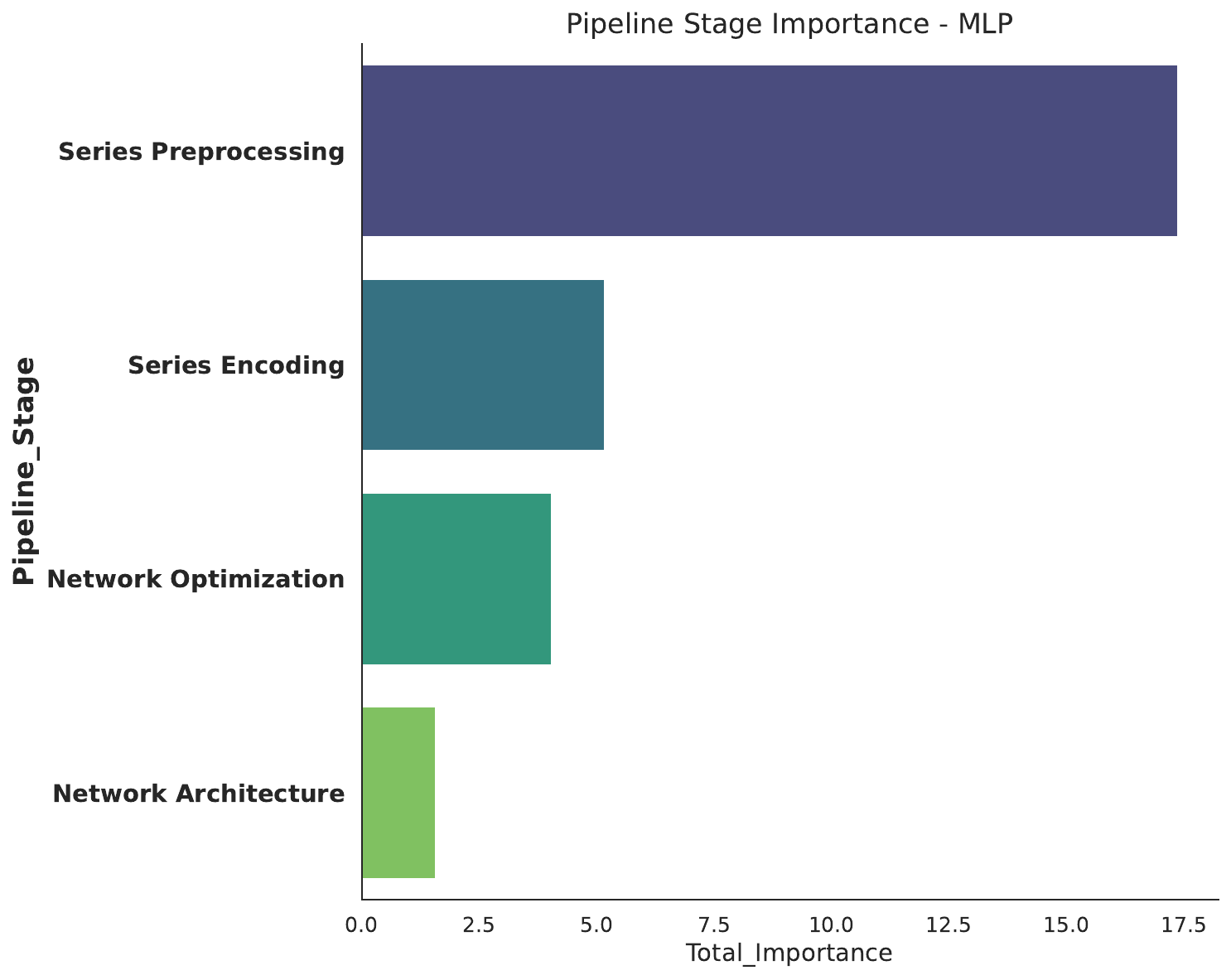}
    \caption{MLP}
    \label{fig:appx_pipeline_stage_MLP}
  \end{subfigure}
  \hfill
  \begin{subfigure}[t]{0.16\textwidth}
    \centering
    \includegraphics[width=\textwidth]{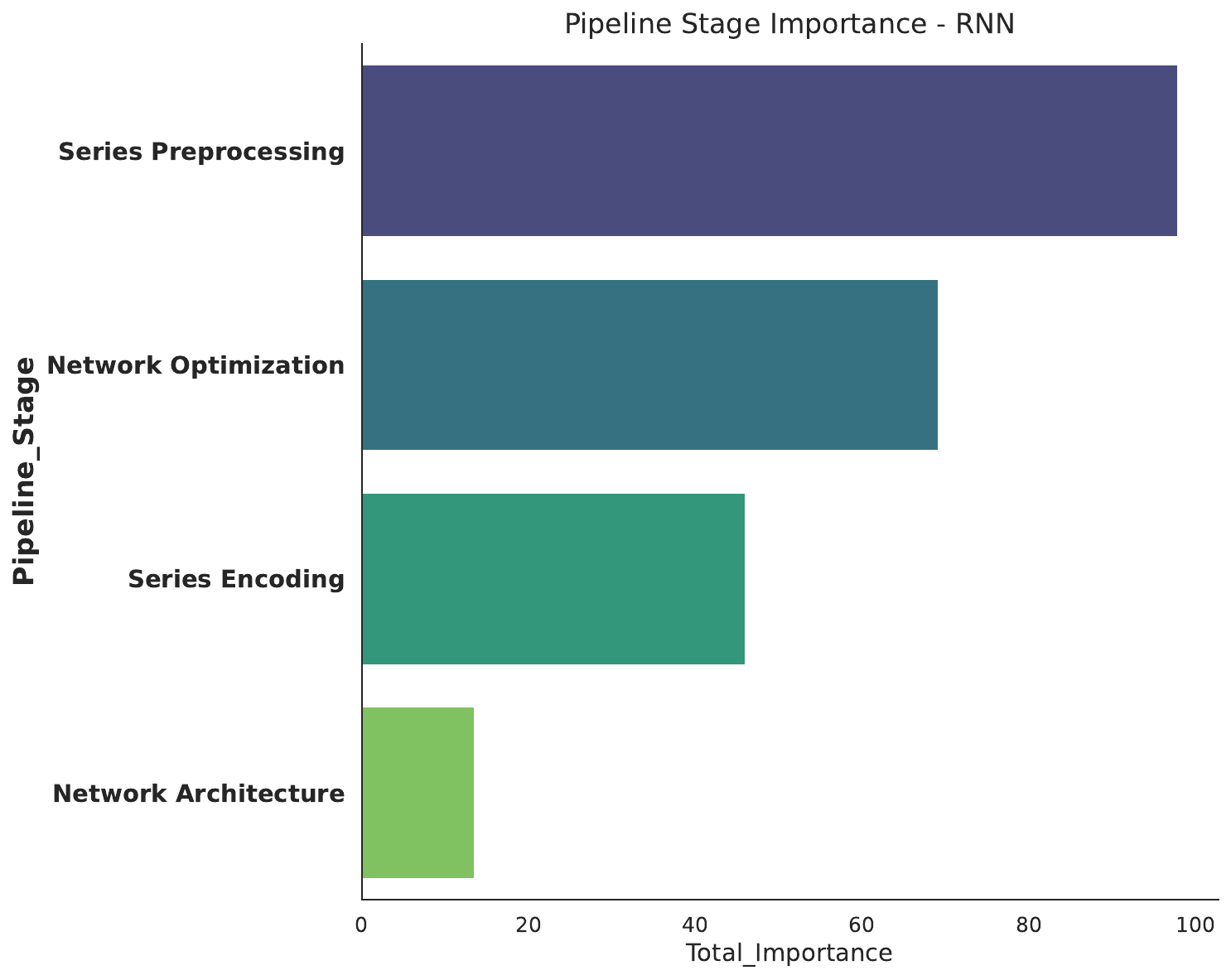}
    \caption{RNN}
    \label{fig:appx_pipeline_stage_RNN}
  \end{subfigure}
  \begin{subfigure}[t]{0.16\textwidth}
    \centering
    \includegraphics[width=\textwidth]{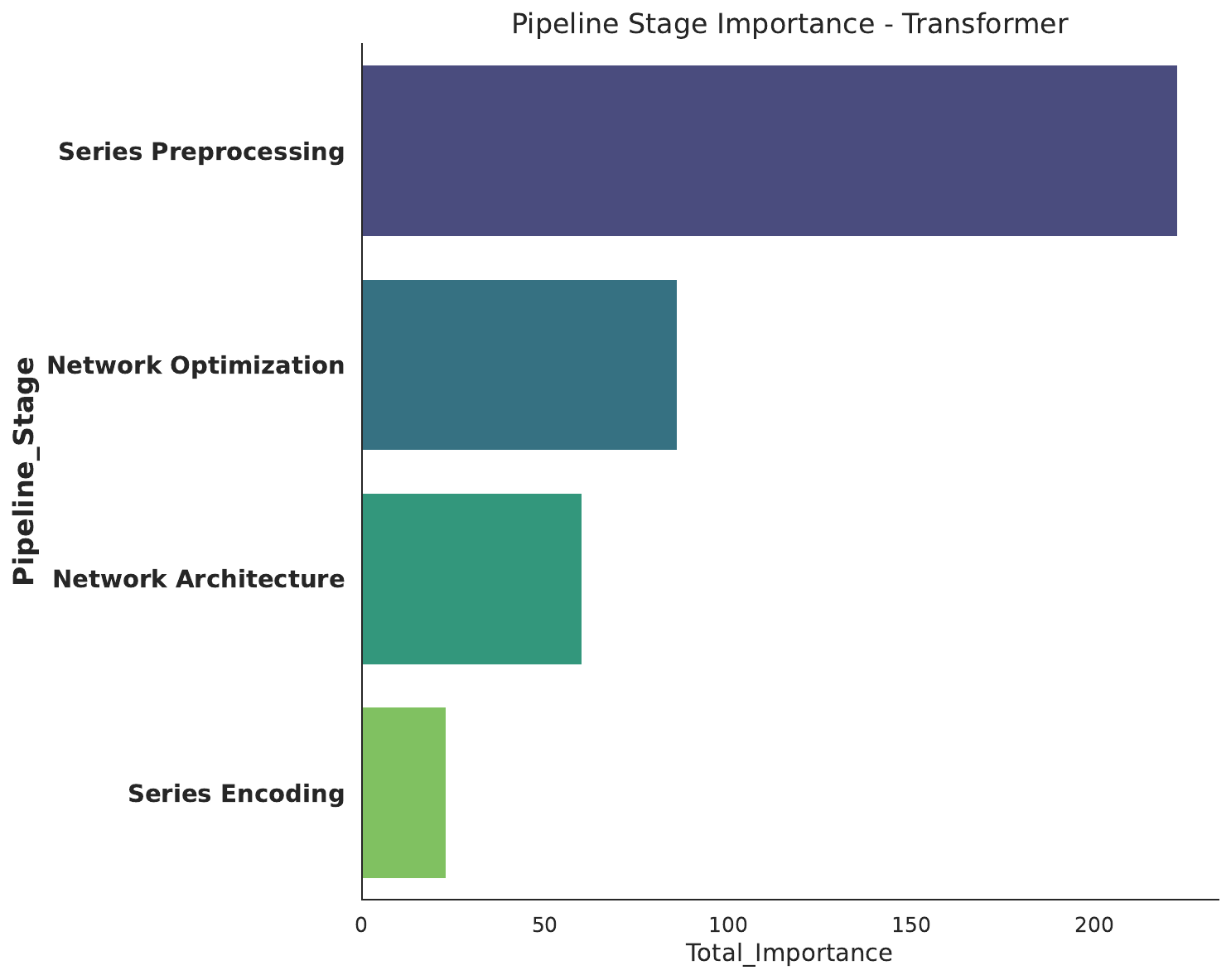}
    \caption{Transformer}
    \label{fig:appx_pipeline_stage_Transformer}
  \end{subfigure}
  \hfill
  \begin{subfigure}[t]{0.16\textwidth}
    \centering
    \includegraphics[width=\textwidth]{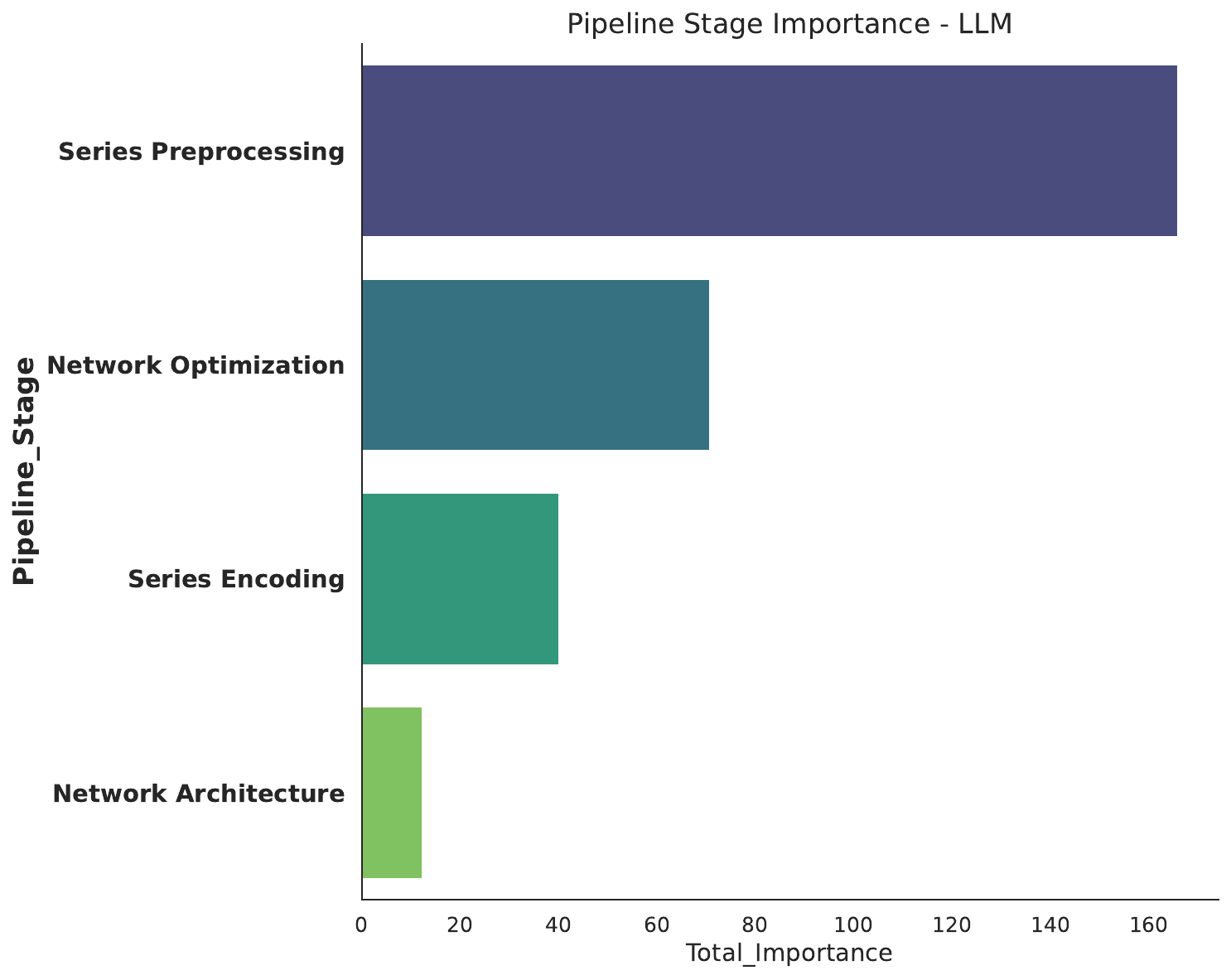}
    \caption{LLM}
    \label{fig:appx_pipeline_stage_LLM}
  \end{subfigure}
  \hfill
  \begin{subfigure}[t]{0.16\textwidth}
    \centering
    \includegraphics[width=\textwidth]{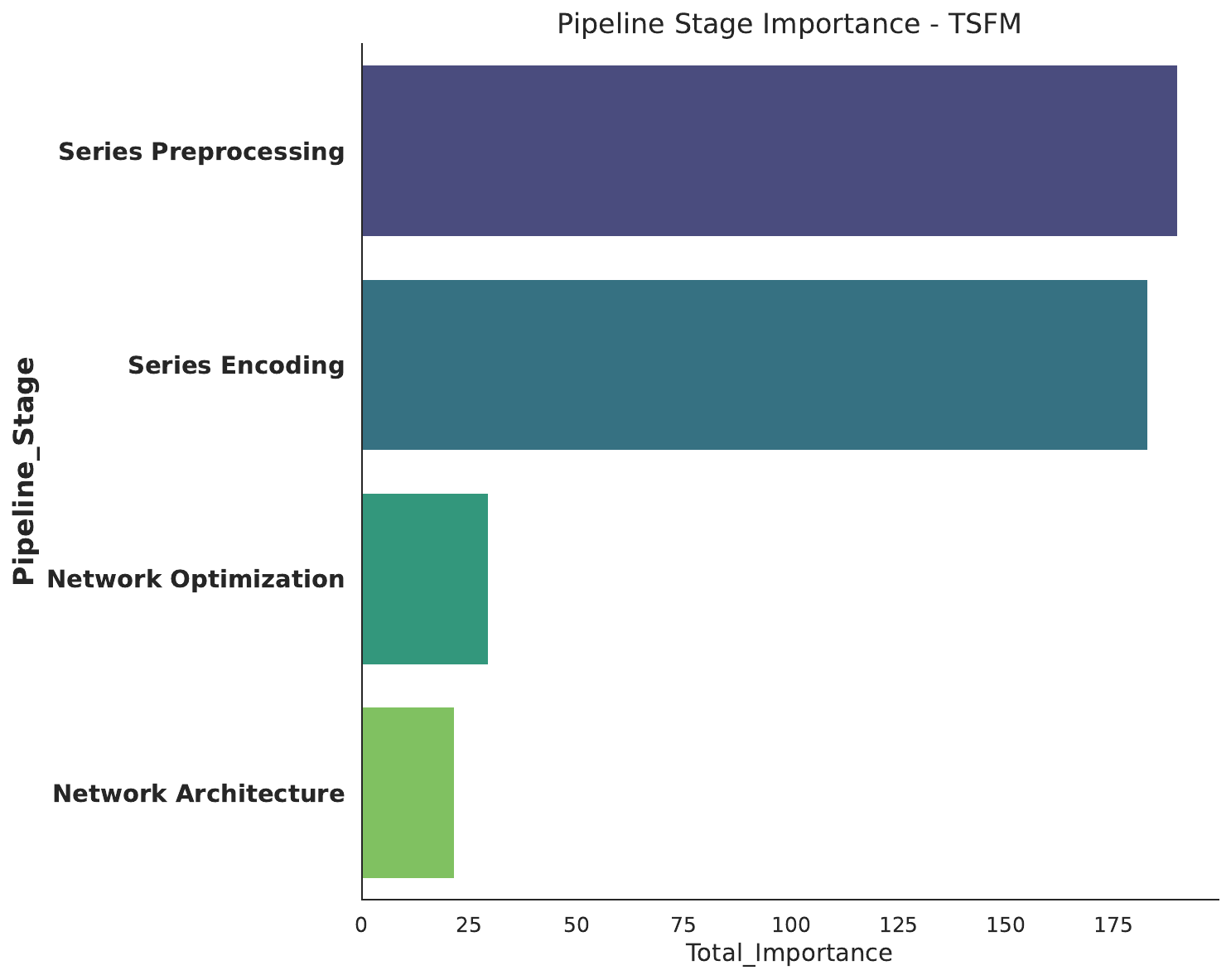}
    \caption{TSFM}
    \label{fig:appx_pipeline_stage_TSFM}
  \end{subfigure}
  \caption{Pipeline Stage Importance (Pipeline Importance Plots). This figure visualizes the performance distributions across different model architectures.}
  \label{fig:appx_pipeline_stage}
\end{figure*}

We analyze the importance of different dimensions (Effect Range) and pipeline stages across various architectures (Fig.~\ref{fig:appx_effect_range} and Fig.~\ref{fig:appx_pipeline_stage}).

\subsection{Detailed Component Analysis}
We provide detailed ridgeline plots (distributions) and radar charts for each component, visualizing performance across datasets and architectures.
\subsubsection{Series Preprocessing}
\begin{figure*}[htbp]
  \centering
  \begin{subfigure}[t]{0.16\textwidth}
    \centering
    \includegraphics[width=\textwidth]{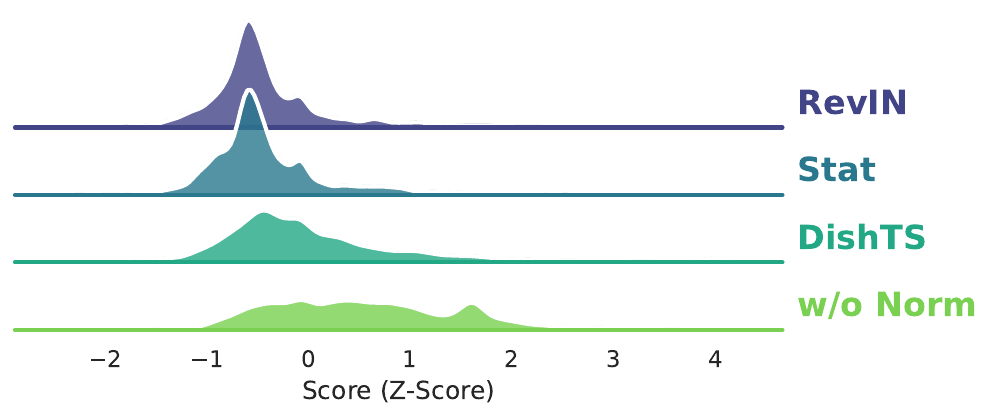}
    \caption{Global}
    \label{fig:appx_dist_gym_series_norm_Global}
  \end{subfigure}
  \hfill
  \begin{subfigure}[t]{0.16\textwidth}
    \centering
    \includegraphics[width=\textwidth]{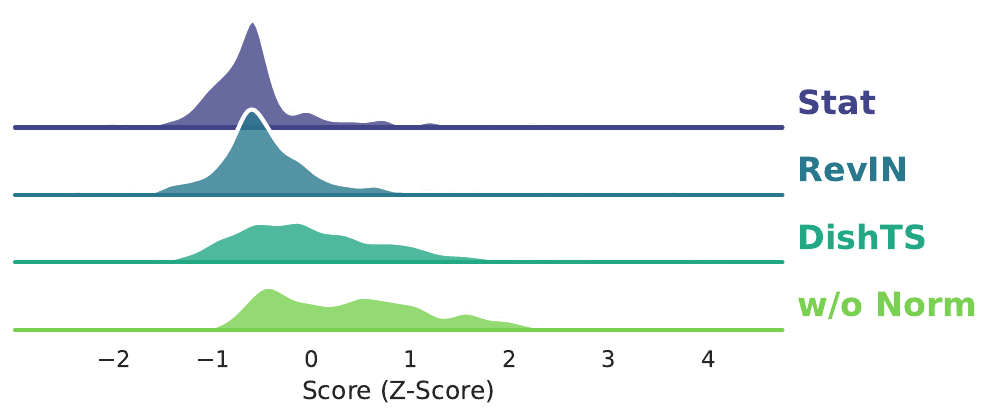}
    \caption{MLP}
    \label{fig:appx_dist_gym_series_norm_MLP}
  \end{subfigure}
  \hfill
  \begin{subfigure}[t]{0.16\textwidth}
    \centering
    \includegraphics[width=\textwidth]{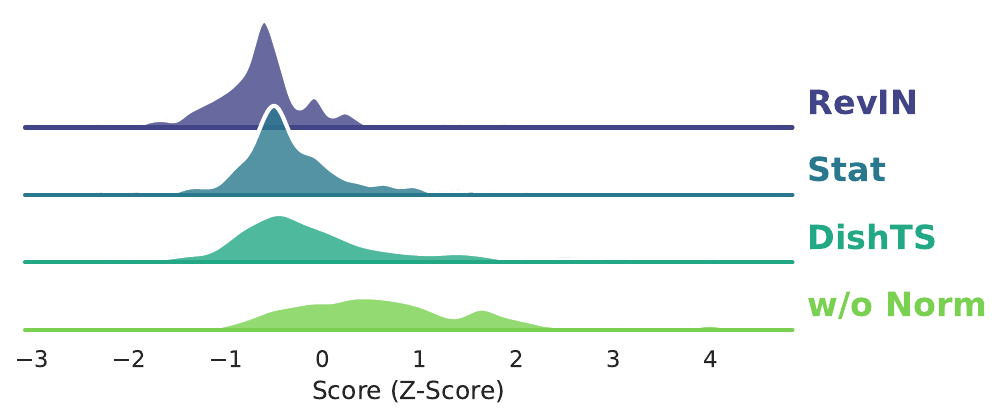}
    \caption{RNN}
    \label{fig:appx_dist_gym_series_norm_RNN}
  \end{subfigure}
  \begin{subfigure}[t]{0.16\textwidth}
    \centering
    \includegraphics[width=\textwidth]{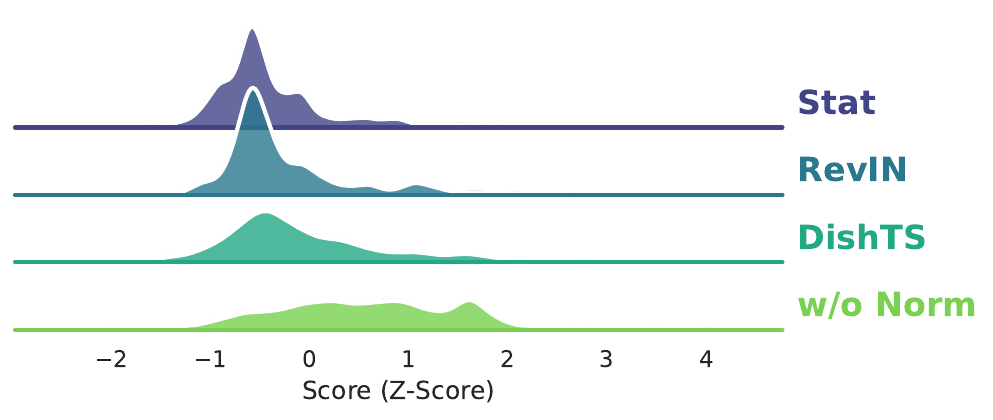}
    \caption{Transformer}
    \label{fig:appx_dist_gym_series_norm_Transformer}
  \end{subfigure}
  \hfill
  \begin{subfigure}[t]{0.16\textwidth}
    \centering
    \includegraphics[width=\textwidth]{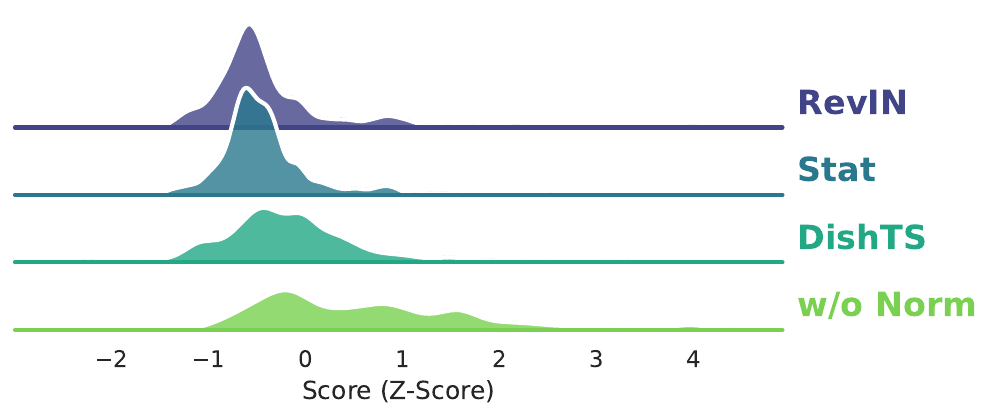}
    \caption{LLM}
    \label{fig:appx_dist_gym_series_norm_LLM}
  \end{subfigure}
  \hfill
  \begin{subfigure}[t]{0.16\textwidth}
    \centering
    \includegraphics[width=\textwidth]{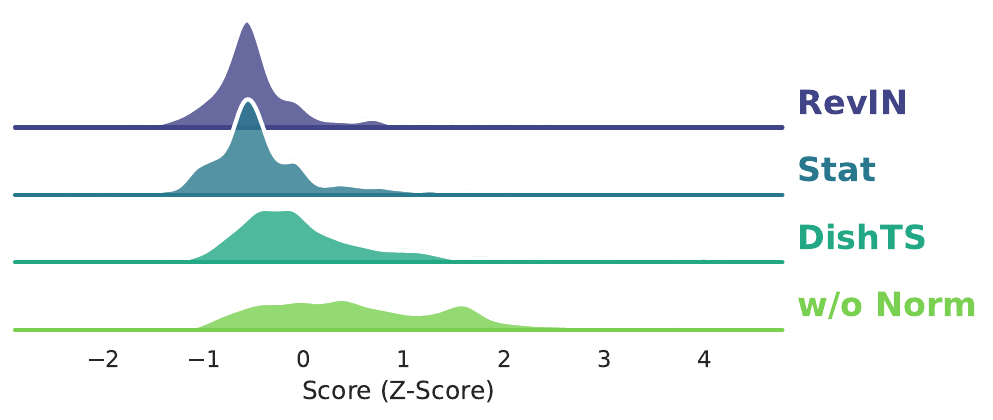}
    \caption{TSFM}
    \label{fig:appx_dist_gym_series_norm_TSFM}
  \end{subfigure}
  \caption{Performance Distributions for Series Normalization (Ridgeline Plots). This figure visualizes the performance distributions across different model architectures.}
  \label{fig:appx_dist_gym_series_norm}
\end{figure*}

\begin{figure*}[htbp]
  \centering
  \begin{subfigure}[t]{0.16\textwidth}
    \centering
    \includegraphics[width=\textwidth]{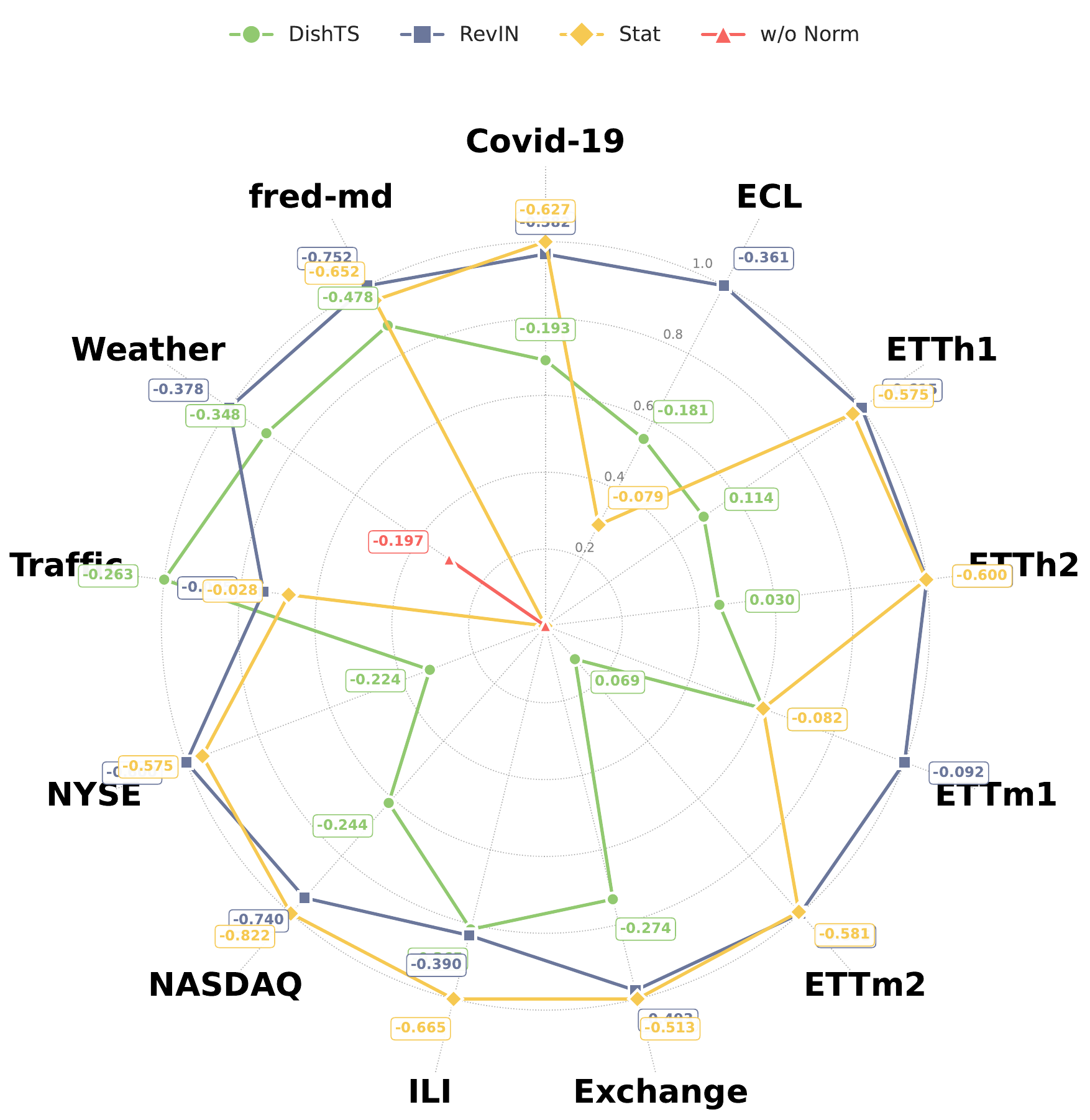}
    \caption{Global}
    \label{fig:appx_radar_gym_series_norm_Global}
  \end{subfigure}
  \hfill
  \begin{subfigure}[t]{0.16\textwidth}
    \centering
    \includegraphics[width=\textwidth]{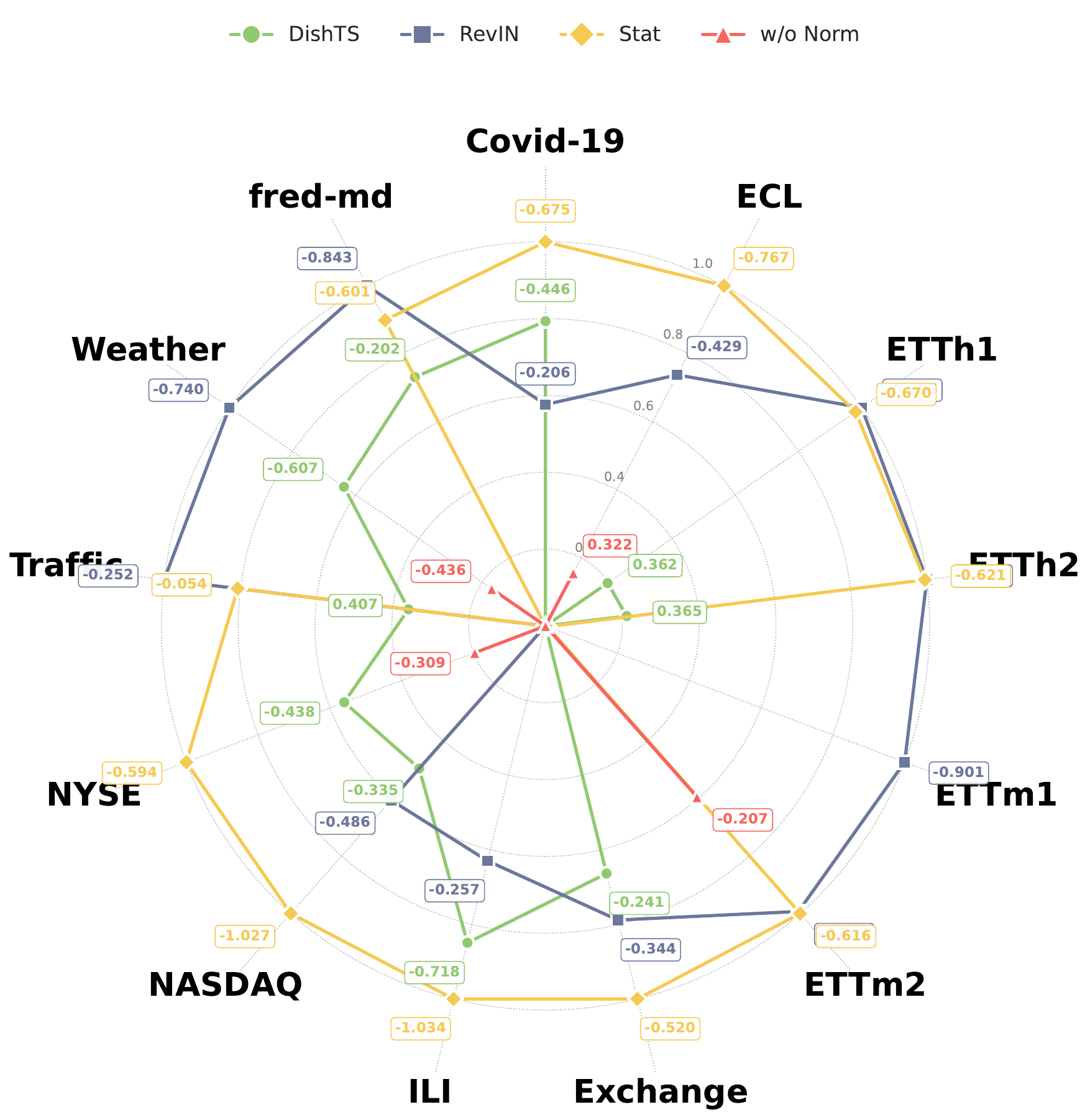}
    \caption{MLP}
    \label{fig:appx_radar_gym_series_norm_MLP}
  \end{subfigure}
  \hfill
  \begin{subfigure}[t]{0.16\textwidth}
    \centering
    \includegraphics[width=\textwidth]{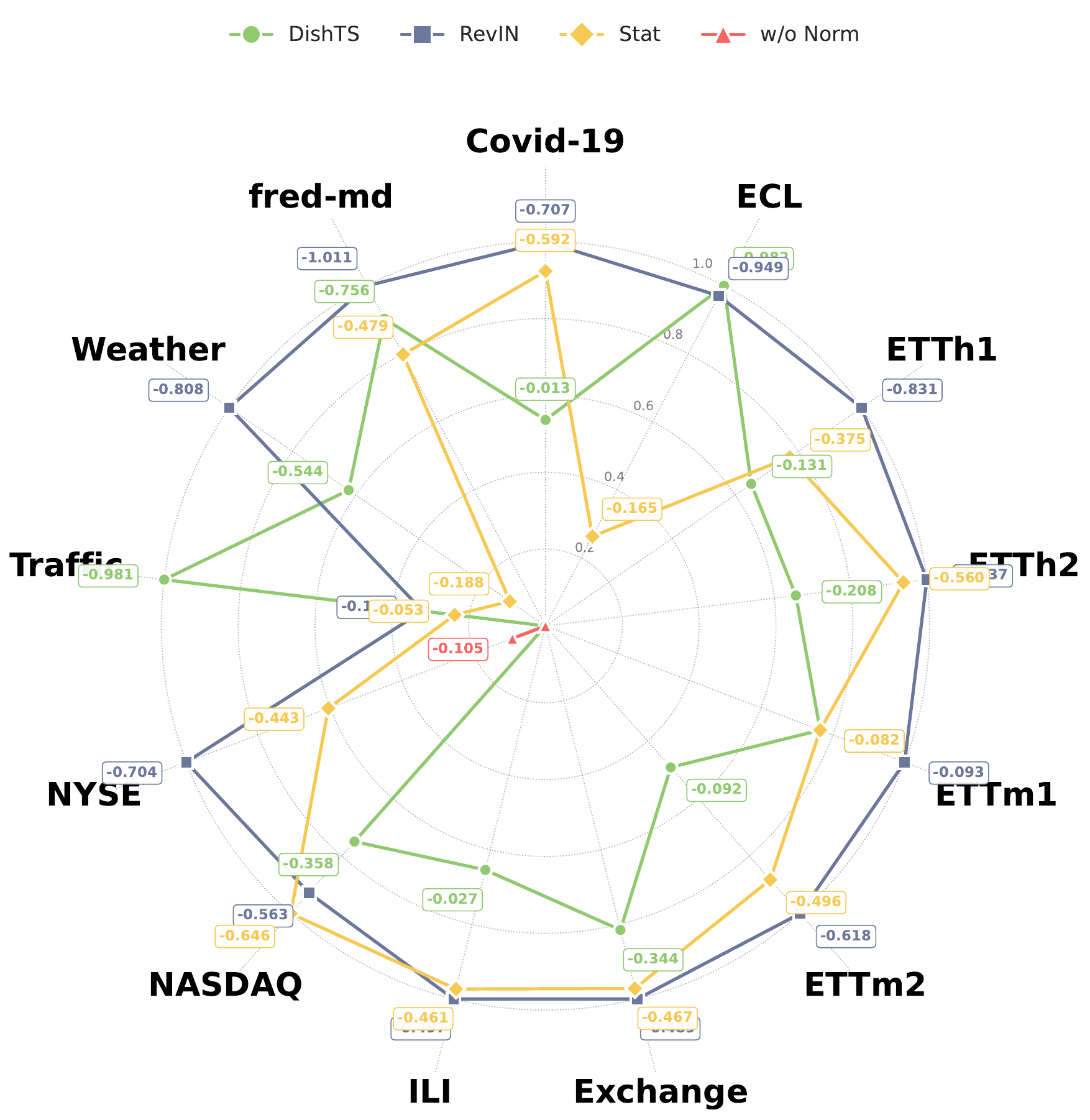}
    \caption{RNN}
    \label{fig:appx_radar_gym_series_norm_RNN}
  \end{subfigure}
  \begin{subfigure}[t]{0.16\textwidth}
    \centering
    \includegraphics[width=\textwidth]{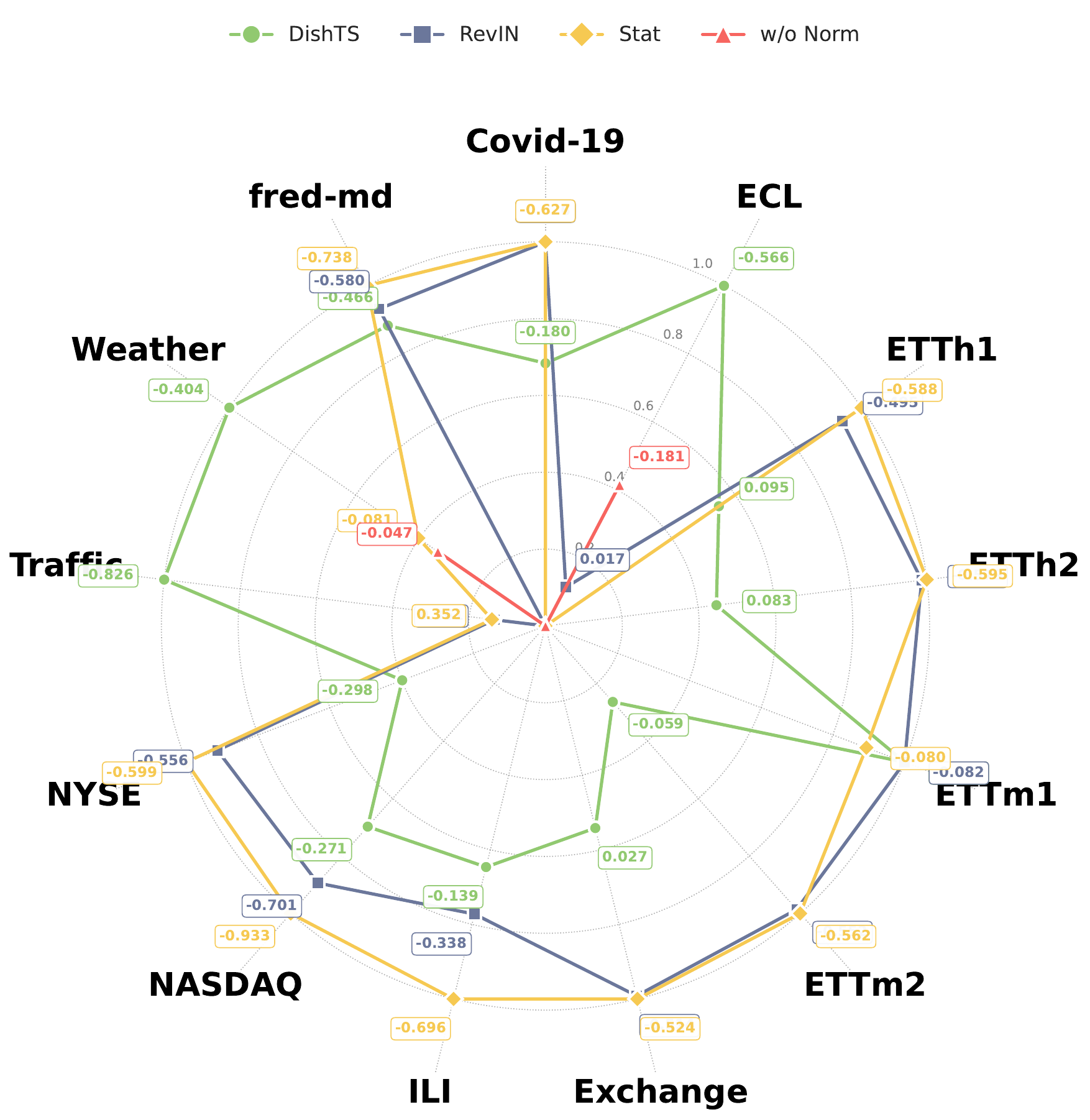}
    \caption{Transformer}
    \label{fig:appx_radar_gym_series_norm_Transformer}
  \end{subfigure}
  \hfill
  \begin{subfigure}[t]{0.16\textwidth}
    \centering
    \includegraphics[width=\textwidth]{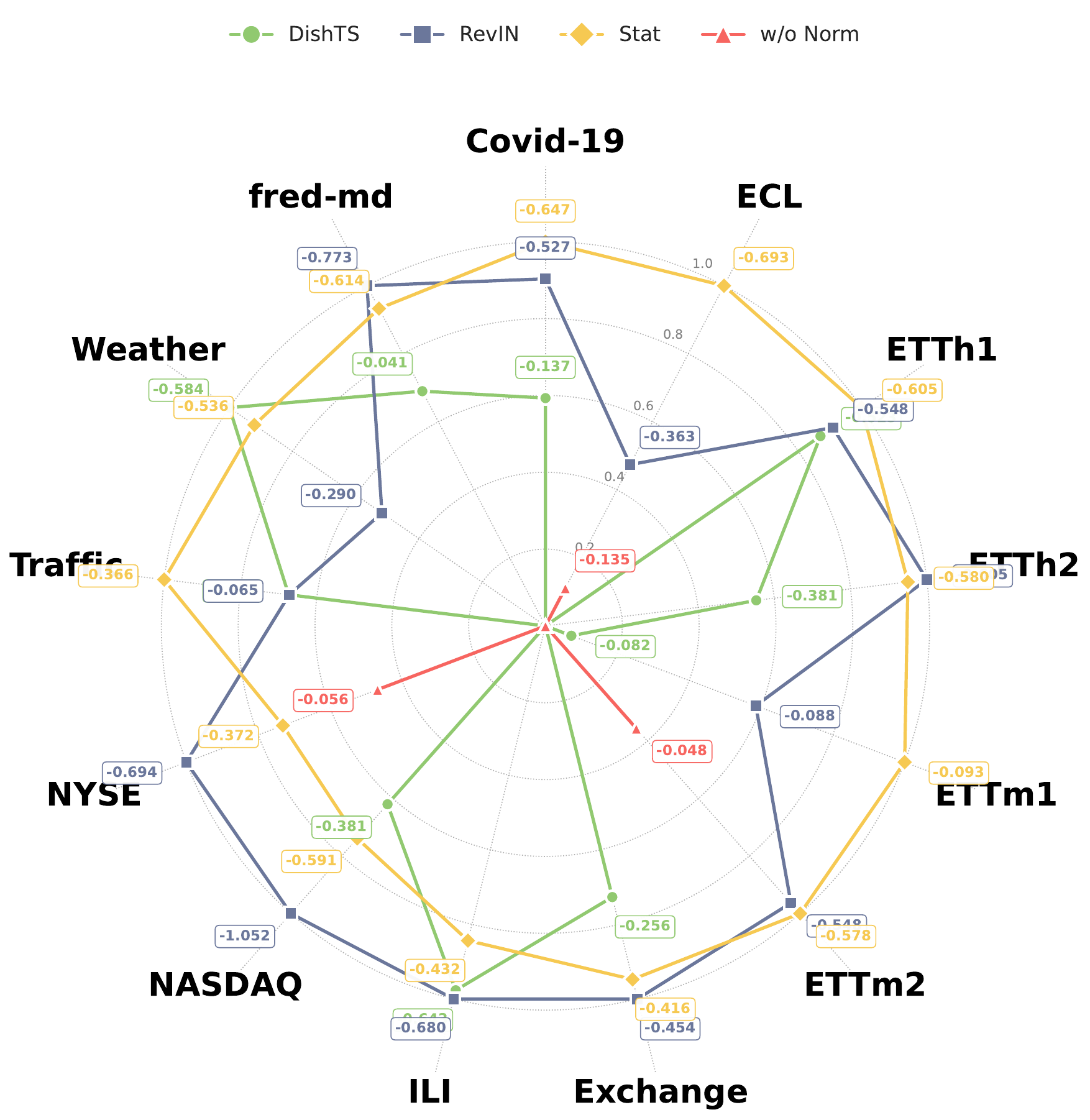}
    \caption{LLM}
    \label{fig:appx_radar_gym_series_norm_LLM}
  \end{subfigure}
  \hfill
  \begin{subfigure}[t]{0.16\textwidth}
    \centering
    \includegraphics[width=\textwidth]{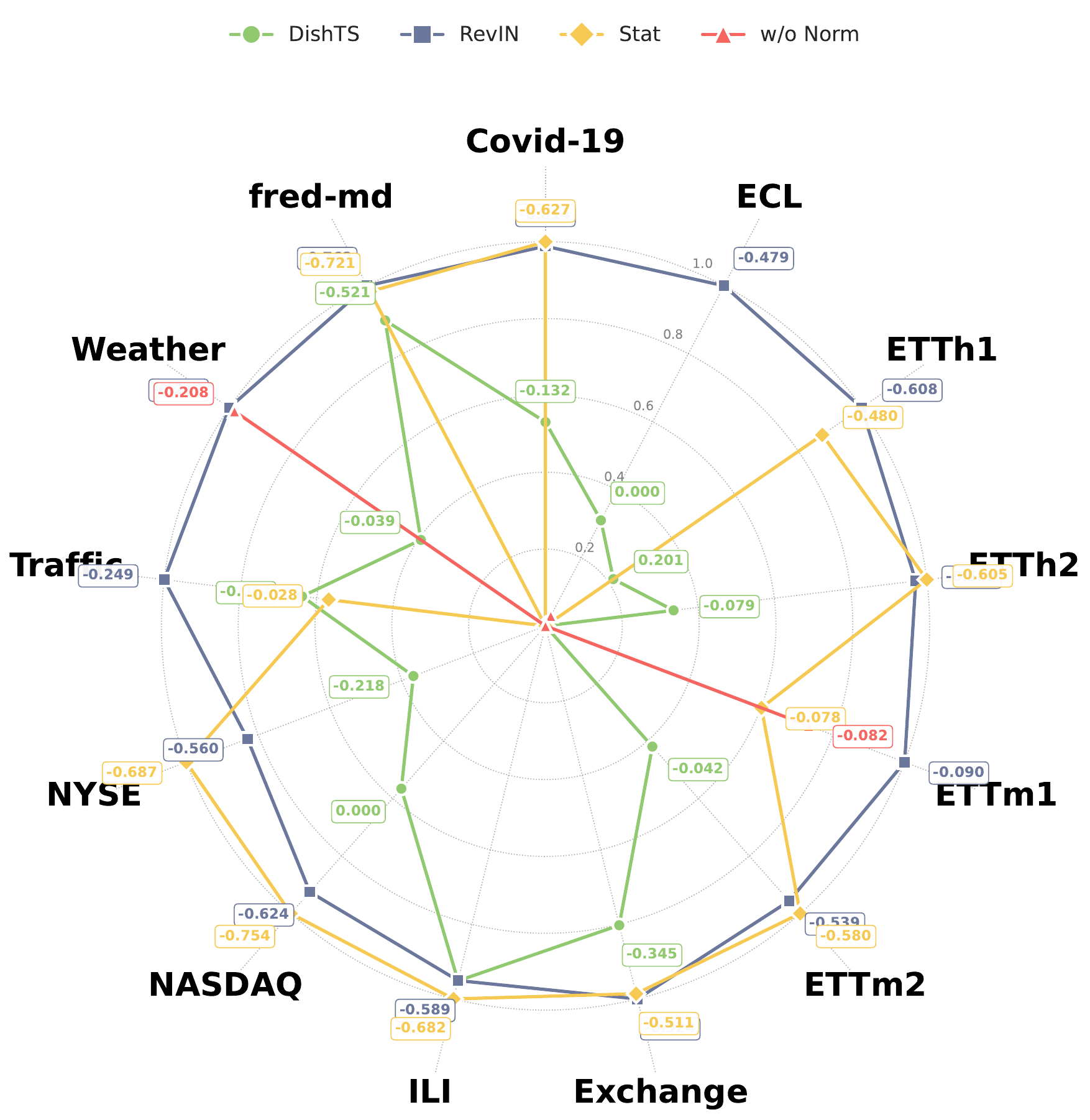}
    \caption{TSFM}
    \label{fig:appx_radar_gym_series_norm_TSFM}
  \end{subfigure}
  \caption{Dataset Adaptability (Radar Charts) for Series Normalization (Radar Plots). This figure visualizes the performance distributions across different model architectures.}
  \label{fig:appx_radar_gym_series_norm}
\end{figure*}

\begin{figure*}[htbp]
  \centering
  \begin{subfigure}[t]{0.16\textwidth}
    \centering
    \includegraphics[width=\textwidth]{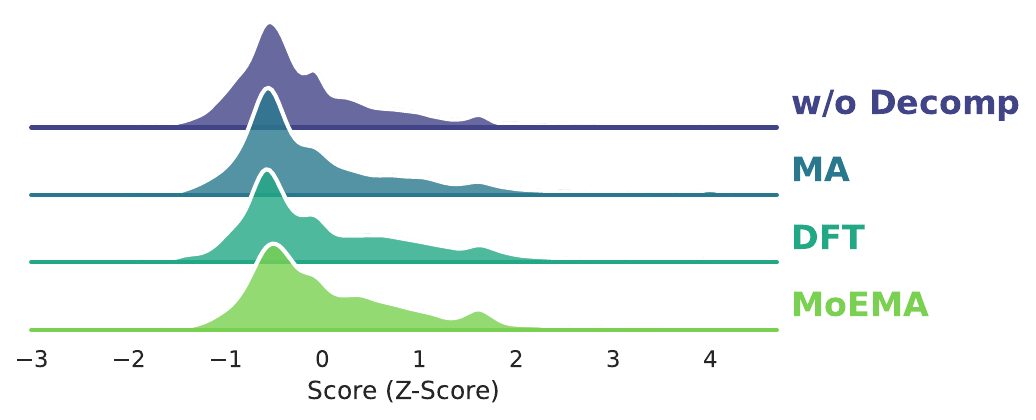}
    \caption{Global}
    \label{fig:appx_dist_gym_series_decomp_Global}
  \end{subfigure}
  \hfill
  \begin{subfigure}[t]{0.16\textwidth}
    \centering
    \includegraphics[width=\textwidth]{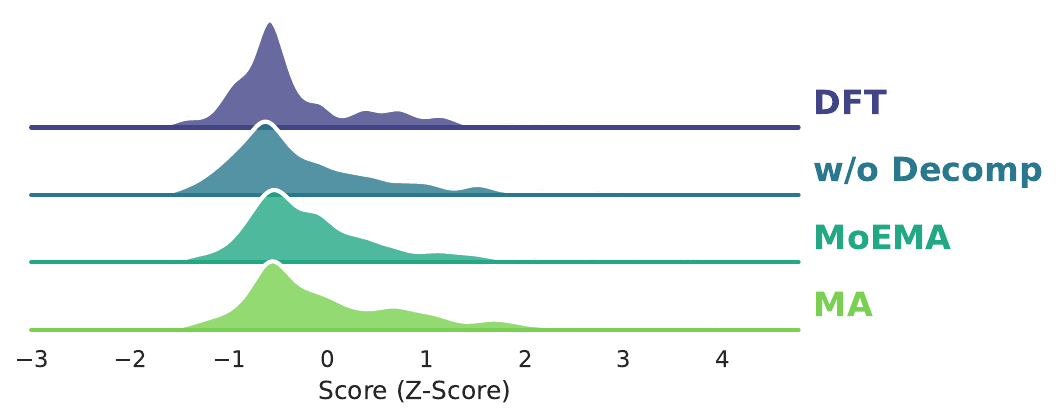}
    \caption{MLP}
    \label{fig:appx_dist_gym_series_decomp_MLP}
  \end{subfigure}
  \hfill
  \begin{subfigure}[t]{0.16\textwidth}
    \centering
    \includegraphics[width=\textwidth]{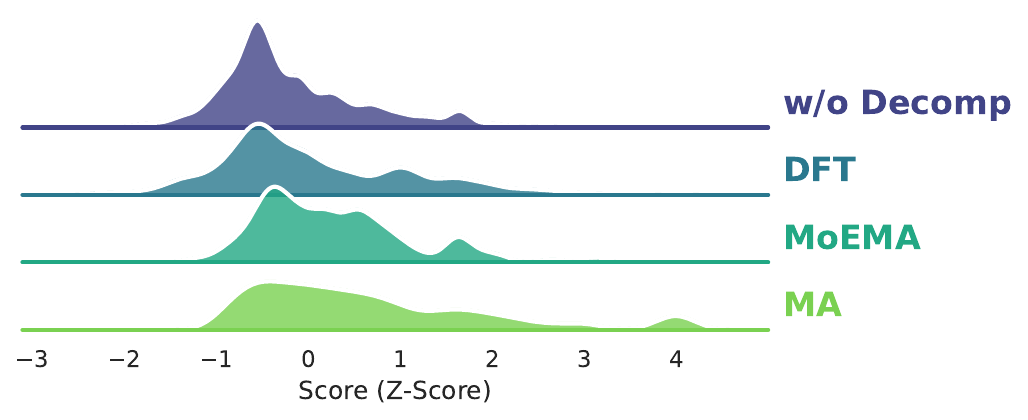}
    \caption{RNN}
    \label{fig:appx_dist_gym_series_decomp_RNN}
  \end{subfigure}
  \begin{subfigure}[t]{0.16\textwidth}
    \centering
    \includegraphics[width=\textwidth]{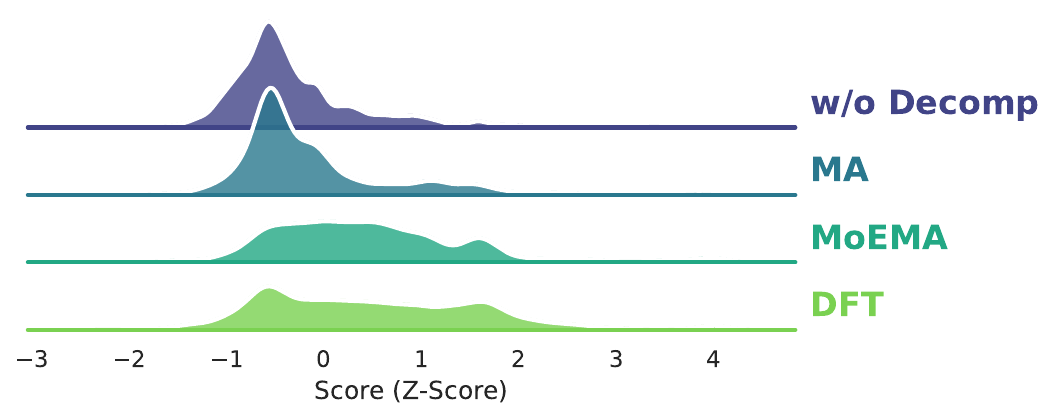}
    \caption{Transformer}
    \label{fig:appx_dist_gym_series_decomp_Transformer}
  \end{subfigure}
  \hfill
  \begin{subfigure}[t]{0.16\textwidth}
    \centering
    \includegraphics[width=\textwidth]{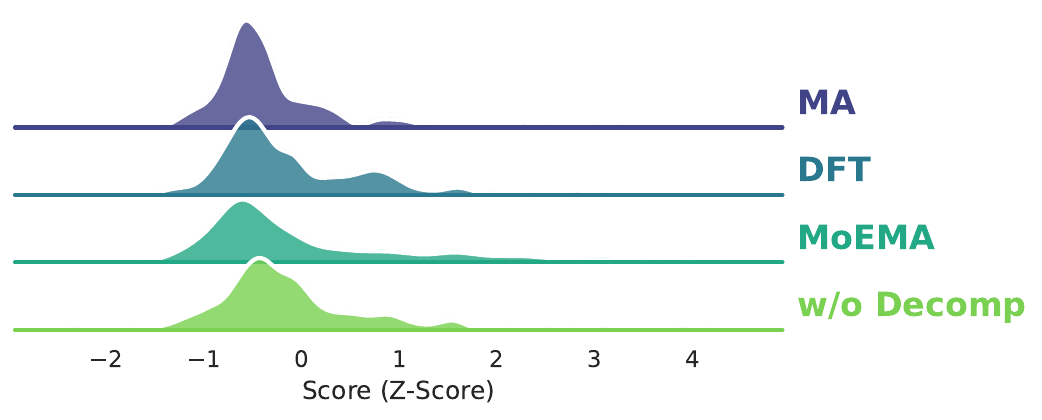}
    \caption{LLM}
    \label{fig:appx_dist_gym_series_decomp_LLM}
  \end{subfigure}
  \hfill
  \begin{subfigure}[t]{0.16\textwidth}
    \centering
    \includegraphics[width=\textwidth]{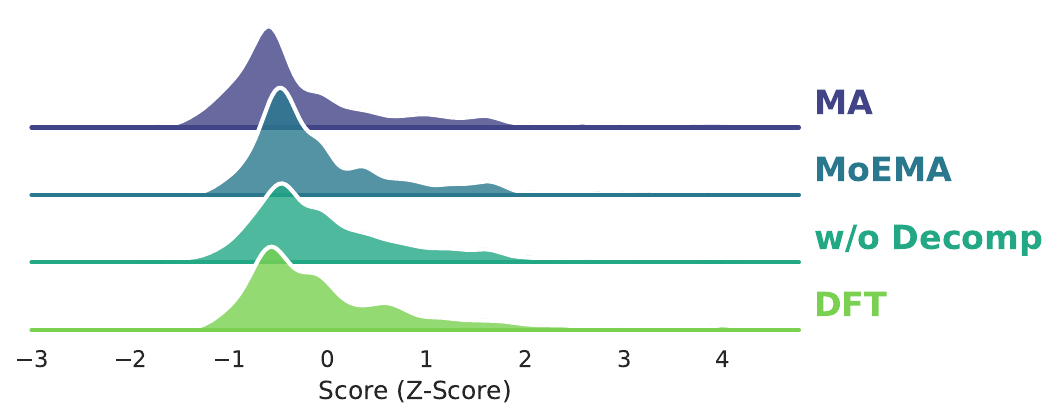}
    \caption{TSFM}
    \label{fig:appx_dist_gym_series_decomp_TSFM}
  \end{subfigure}
  \caption{Performance Distributions for Series Decomposition (Ridgeline Plots). This figure visualizes the performance distributions across different model architectures.}
  \label{fig:appx_dist_gym_series_decomp}
\end{figure*}

\begin{figure*}[htbp]
  \centering
  \begin{subfigure}[t]{0.16\textwidth}
    \centering
    \includegraphics[width=\textwidth]{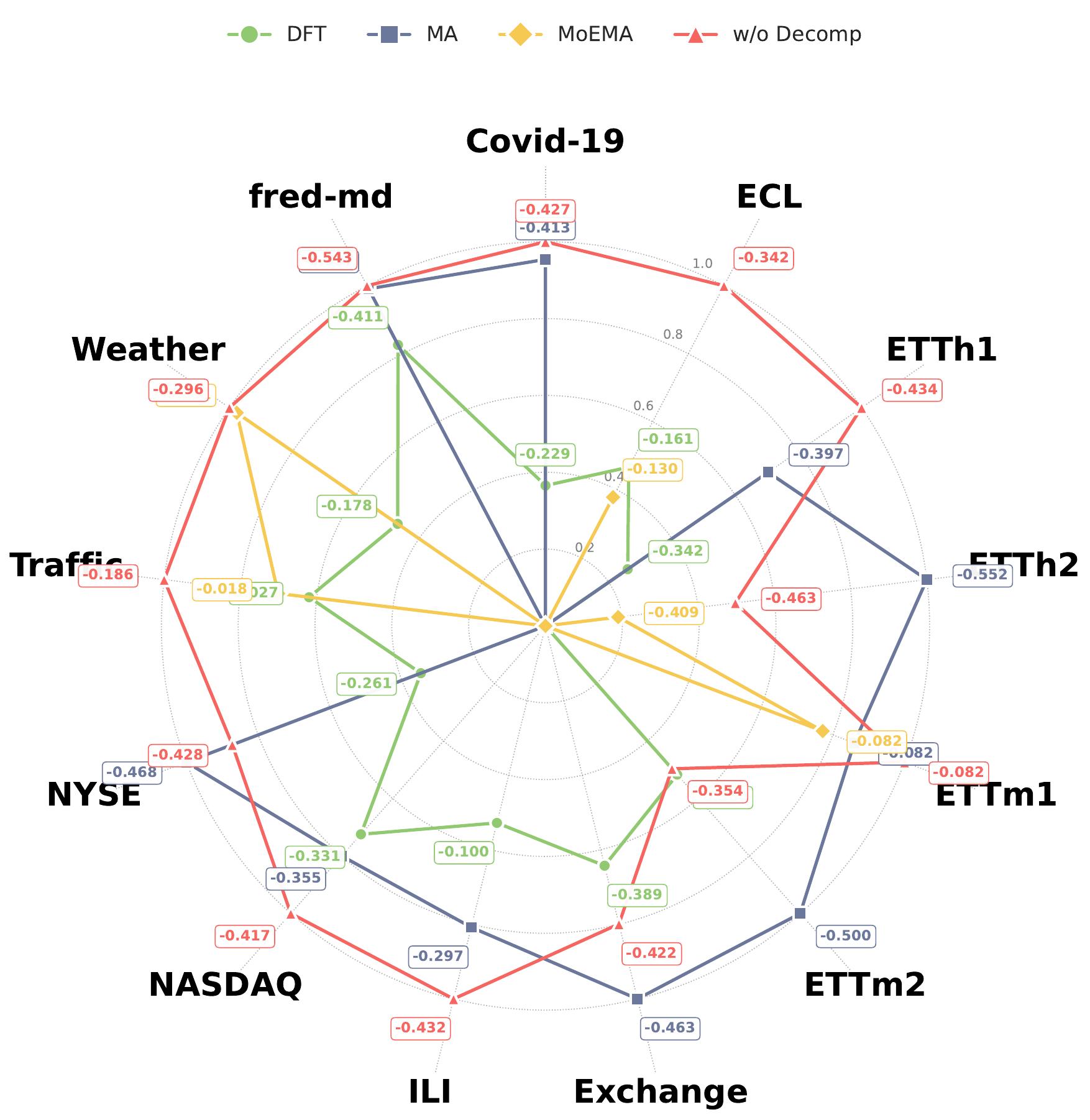}
    \caption{Global}
    \label{fig:appx_radar_gym_series_decomp_Global}
  \end{subfigure}
  \hfill
  \begin{subfigure}[t]{0.16\textwidth}
    \centering
    \includegraphics[width=\textwidth]{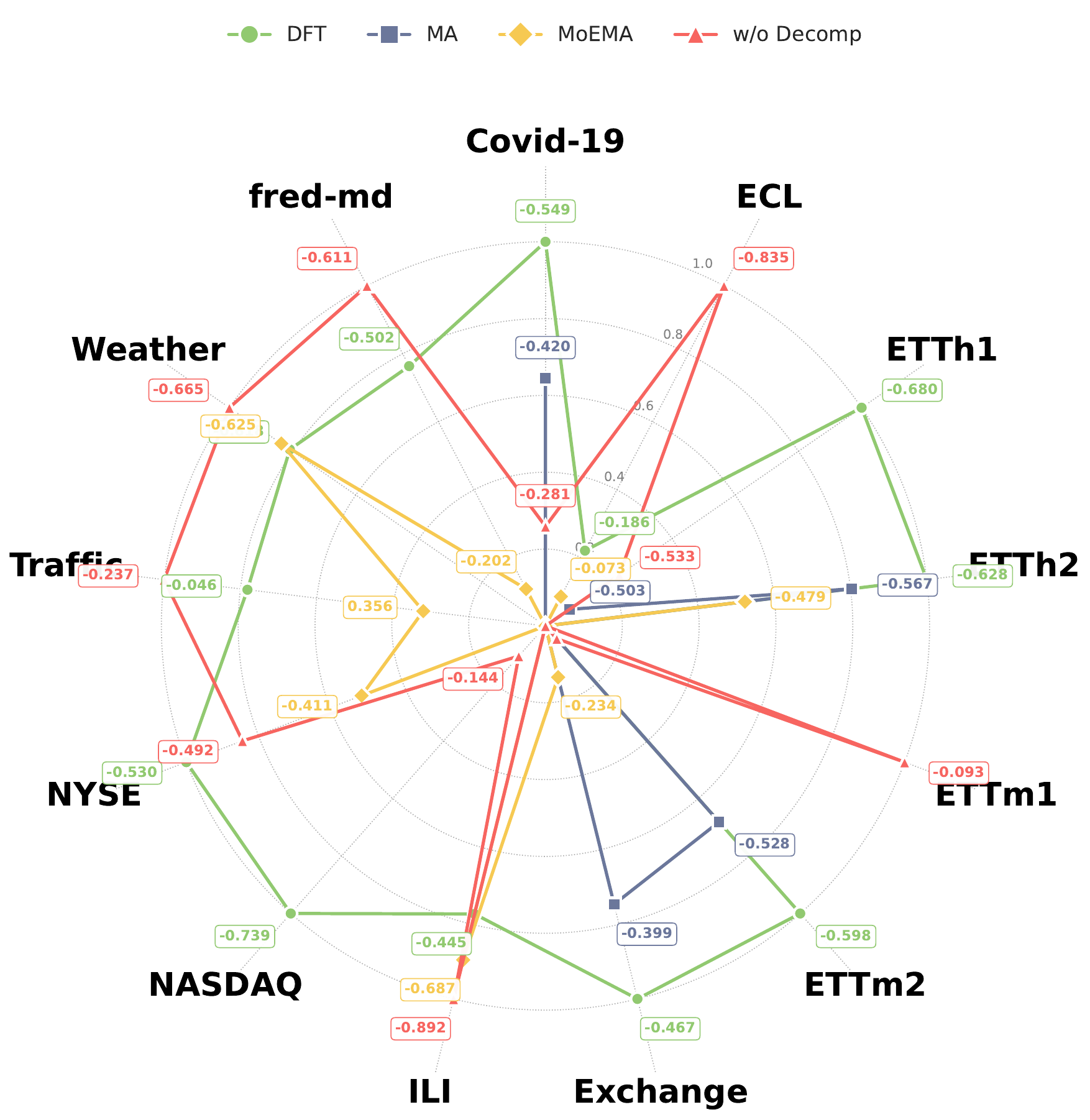}
    \caption{MLP}
    \label{fig:appx_radar_gym_series_decomp_MLP}
  \end{subfigure}
  \hfill
  \begin{subfigure}[t]{0.16\textwidth}
    \centering
    \includegraphics[width=\textwidth]{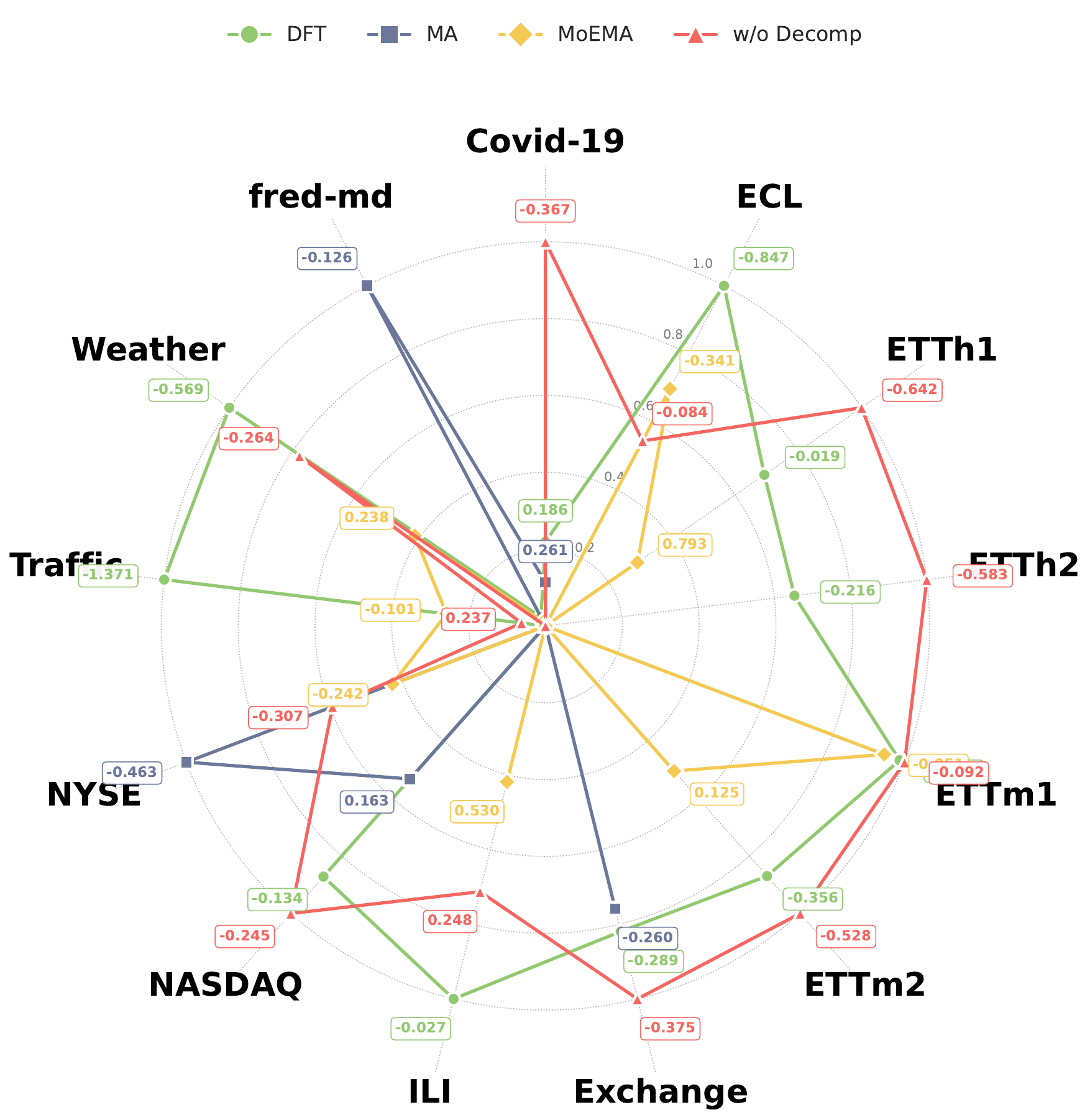}
    \caption{RNN}
    \label{fig:appx_radar_gym_series_decomp_RNN}
  \end{subfigure}
  \begin{subfigure}[t]{0.16\textwidth}
    \centering
    \includegraphics[width=\textwidth]{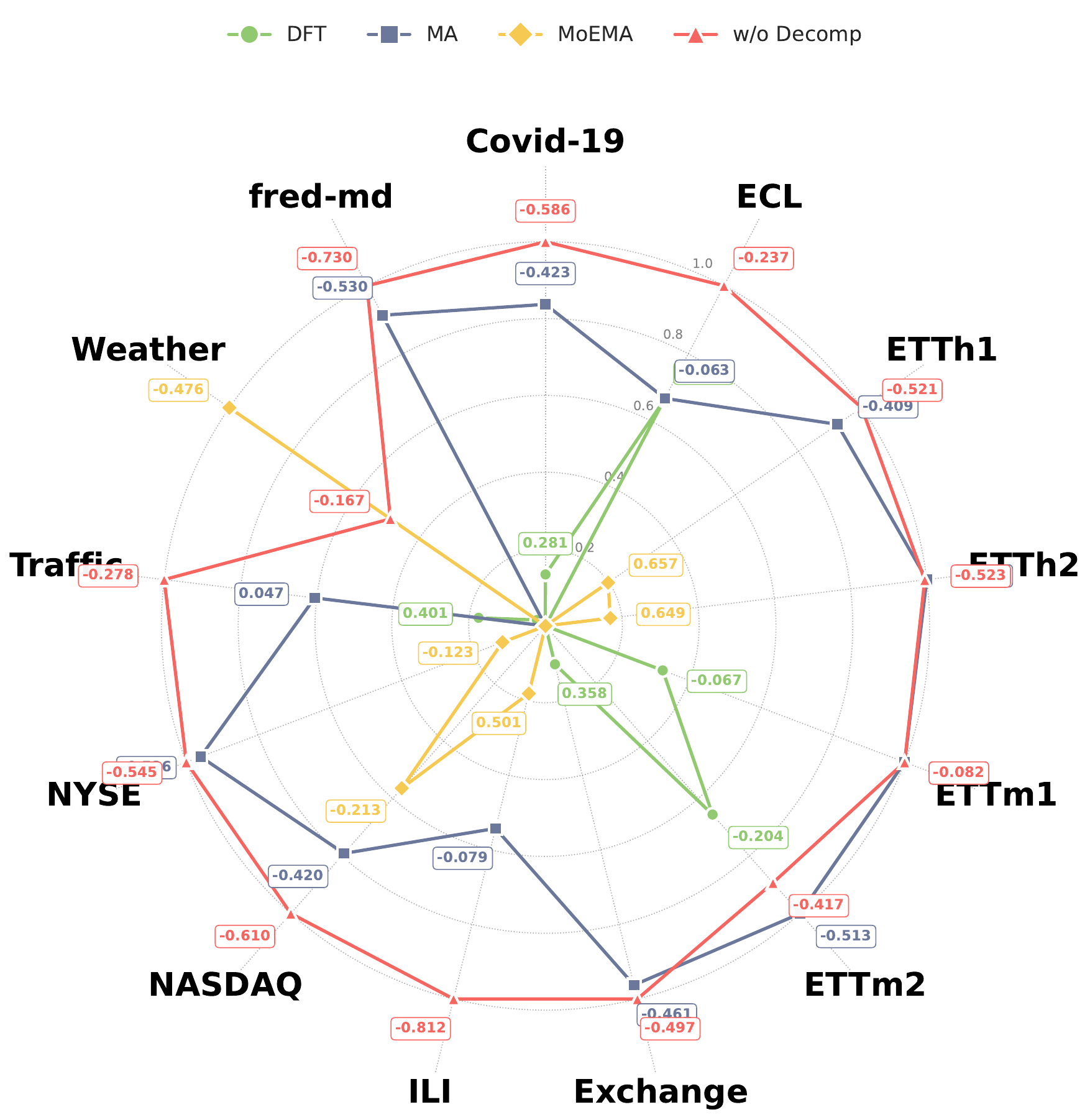}
    \caption{Transformer}
    \label{fig:appx_radar_gym_series_decomp_Transformer}
  \end{subfigure}
  \hfill
  \begin{subfigure}[t]{0.16\textwidth}
    \centering
    \includegraphics[width=\textwidth]{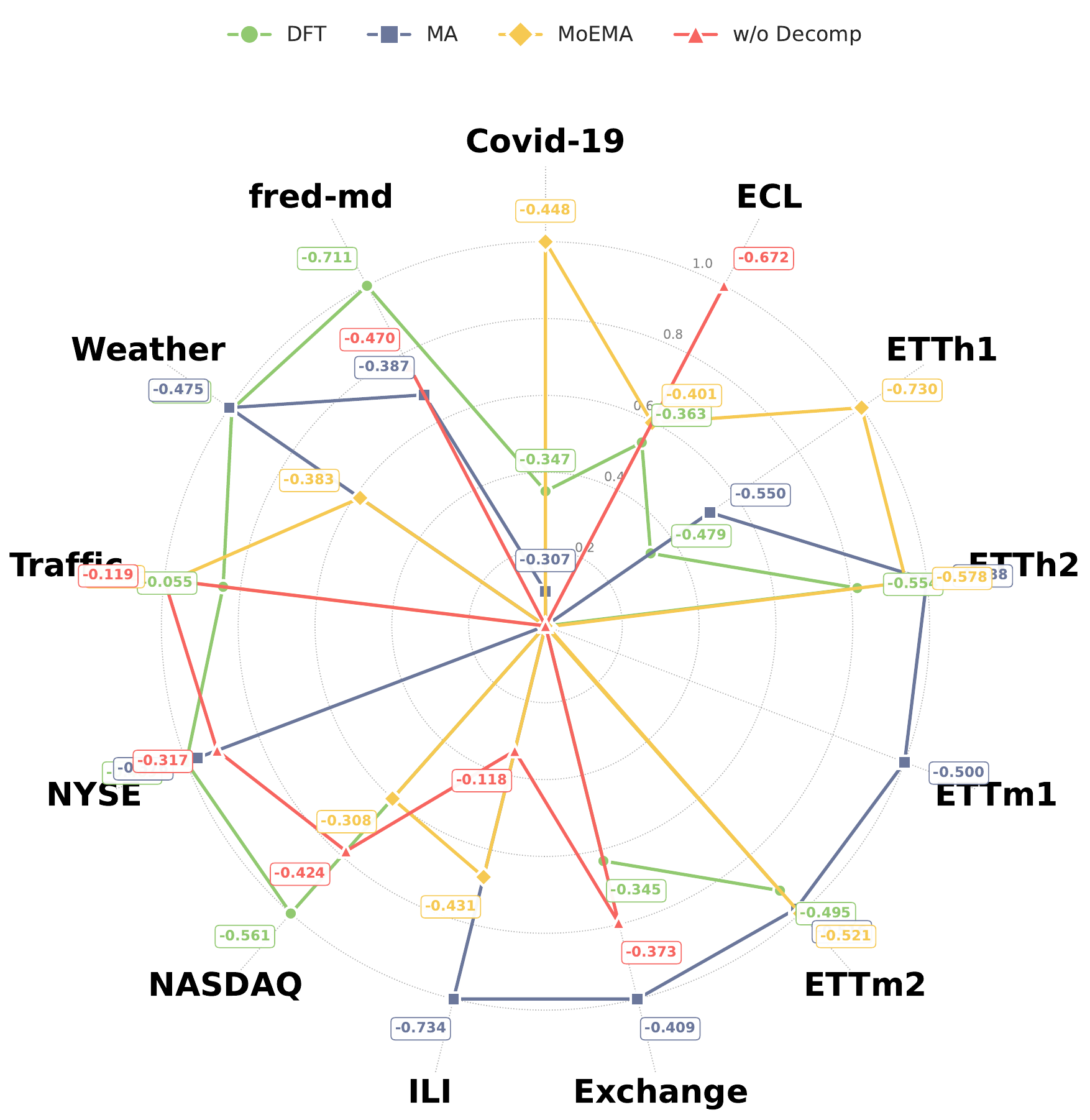}
    \caption{LLM}
    \label{fig:appx_radar_gym_series_decomp_LLM}
  \end{subfigure}
  \hfill
  \begin{subfigure}[t]{0.16\textwidth}
    \centering
    \includegraphics[width=\textwidth]{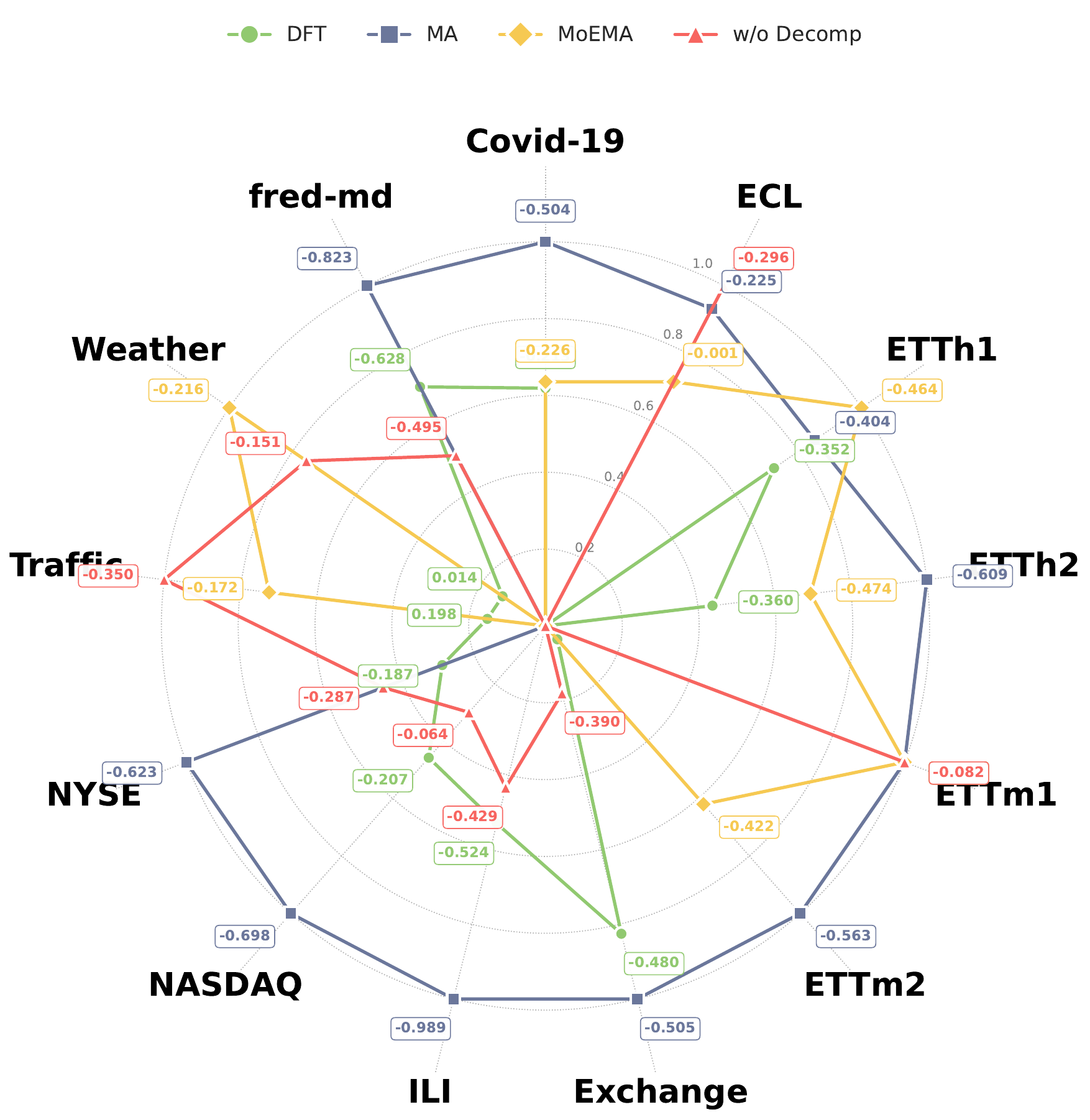}
    \caption{TSFM}
    \label{fig:appx_radar_gym_series_decomp_TSFM}
  \end{subfigure}
  \caption{Dataset Adaptability (Radar Charts) for Series Decomposition (Radar Plots). This figure visualizes the performance distributions across different model architectures.}
  \label{fig:appx_radar_gym_series_decomp}
\end{figure*}

\begin{figure*}[htbp]
  \centering
  \begin{subfigure}[t]{0.16\textwidth}
    \centering
    \includegraphics[width=\textwidth]{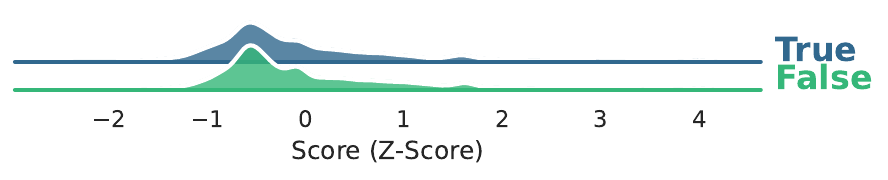}
    \caption{Global}
    \label{fig:appx_dist_series_sampling_Global}
  \end{subfigure}
  \hfill
  \begin{subfigure}[t]{0.16\textwidth}
    \centering
    \includegraphics[width=\textwidth]{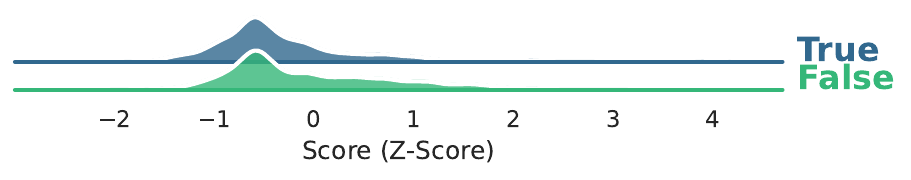}
    \caption{MLP}
    \label{fig:appx_dist_series_sampling_MLP}
  \end{subfigure}
  \hfill
  \begin{subfigure}[t]{0.16\textwidth}
    \centering
    \includegraphics[width=\textwidth]{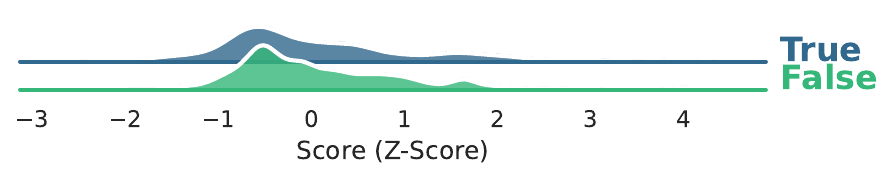}
    \caption{RNN}
    \label{fig:appx_dist_series_sampling_RNN}
  \end{subfigure}
  \begin{subfigure}[t]{0.16\textwidth}
    \centering
    \includegraphics[width=\textwidth]{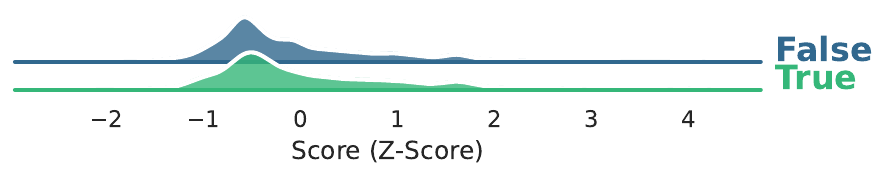}
    \caption{Transformer}
    \label{fig:appx_dist_series_sampling_Transformer}
  \end{subfigure}
  \hfill
  \begin{subfigure}[t]{0.16\textwidth}
    \centering
    \includegraphics[width=\textwidth]{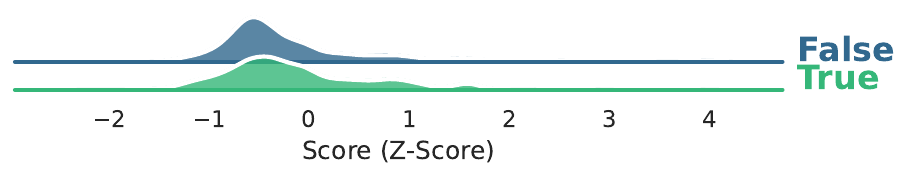}
    \caption{LLM}
    \label{fig:appx_dist_series_sampling_LLM}
  \end{subfigure}
  \hfill
  \begin{subfigure}[t]{0.16\textwidth}
    \centering
    \includegraphics[width=\textwidth]{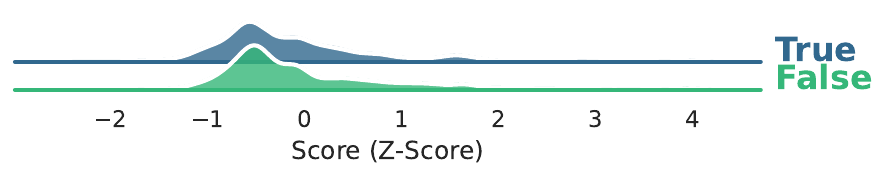}
    \caption{TSFM}
    \label{fig:appx_dist_series_sampling_TSFM}
  \end{subfigure}
  \caption{Performance Distributions for Series Sampling/Mixing (Ridgeline Plots). This figure visualizes the performance distributions across different model architectures.}
  \label{fig:appx_dist_series_sampling}
\end{figure*}

\begin{figure*}[htbp]
  \centering
  \begin{subfigure}[t]{0.16\textwidth}
    \centering
    \includegraphics[width=\textwidth]{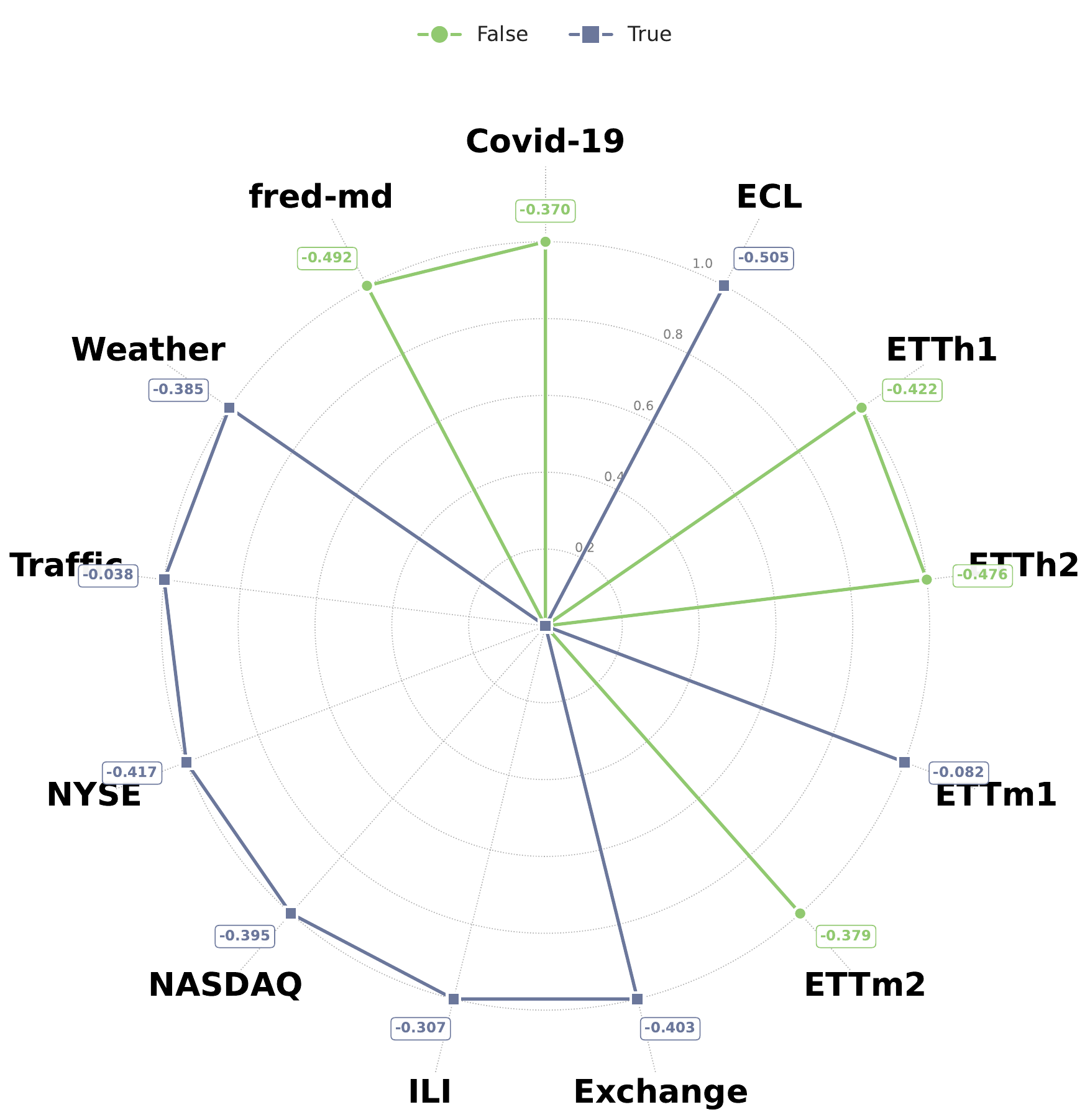}
    \caption{Global}
    \label{fig:appx_radar_series_sampling_Global}
  \end{subfigure}
  \hfill
  \begin{subfigure}[t]{0.16\textwidth}
    \centering
    \includegraphics[width=\textwidth]{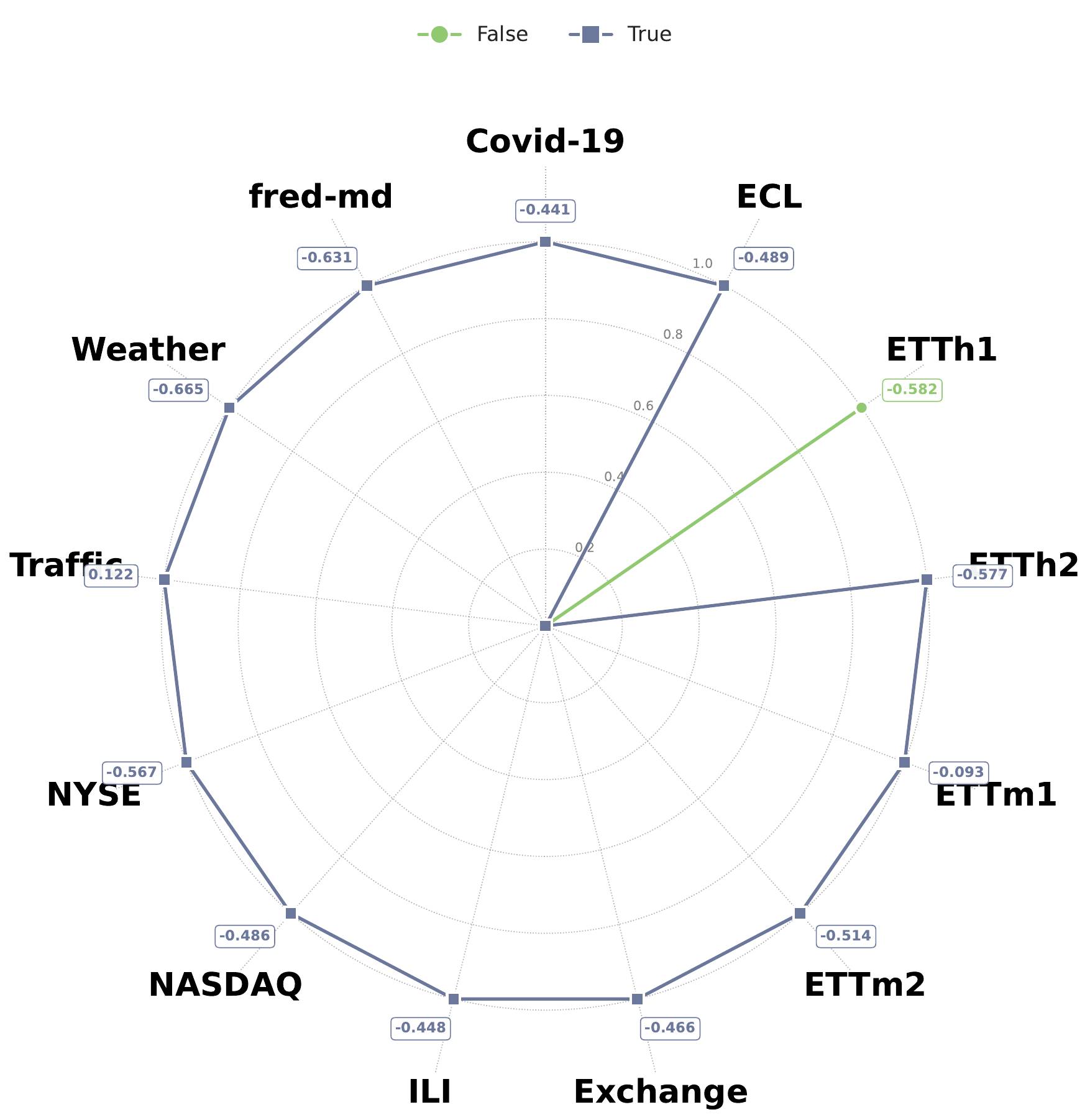}
    \caption{MLP}
    \label{fig:appx_radar_series_sampling_MLP}
  \end{subfigure}
  \hfill
  \begin{subfigure}[t]{0.16\textwidth}
    \centering
    \includegraphics[width=\textwidth]{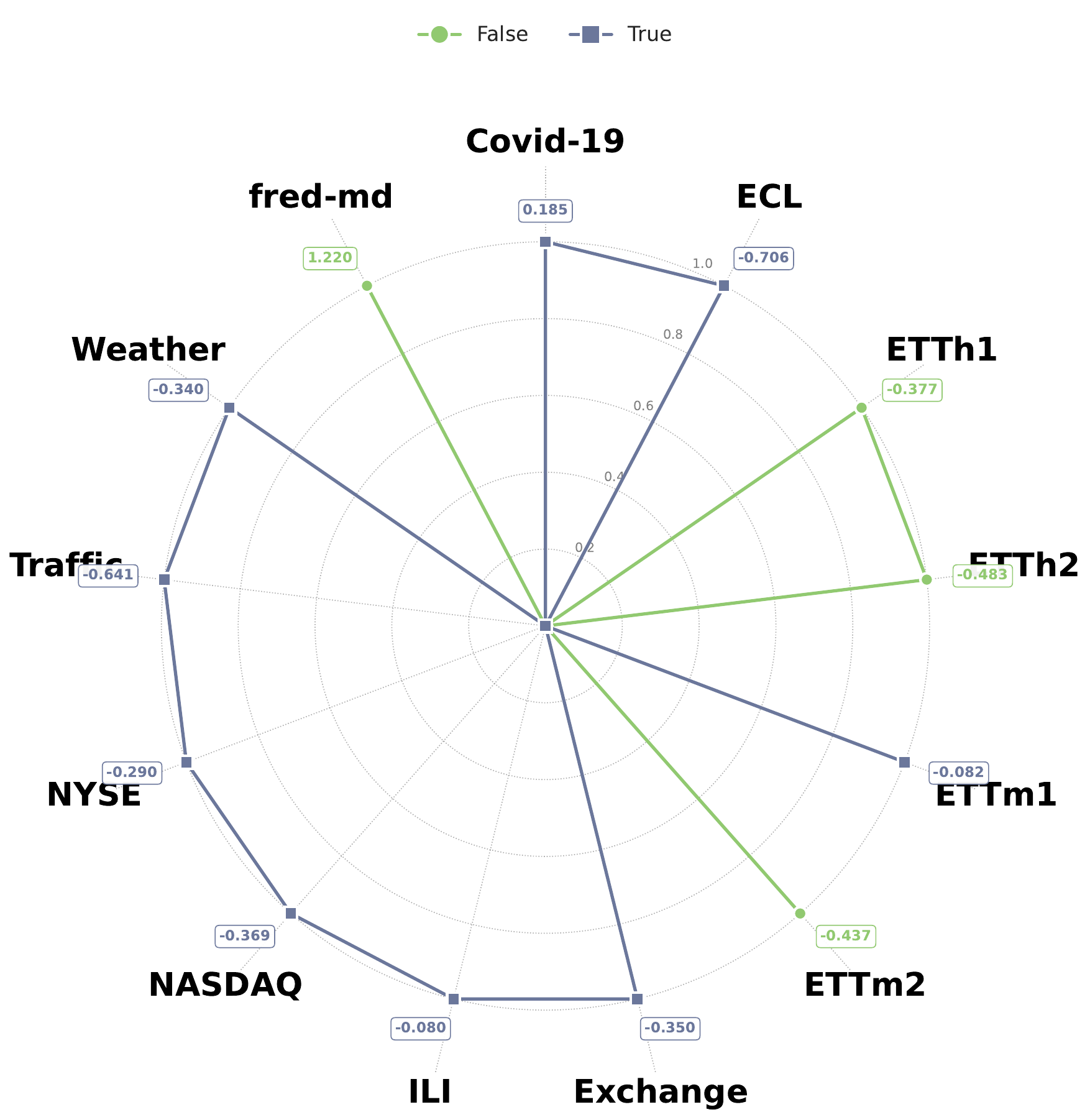}
    \caption{RNN}
    \label{fig:appx_radar_series_sampling_RNN}
  \end{subfigure}
  \begin{subfigure}[t]{0.16\textwidth}
    \centering
    \includegraphics[width=\textwidth]{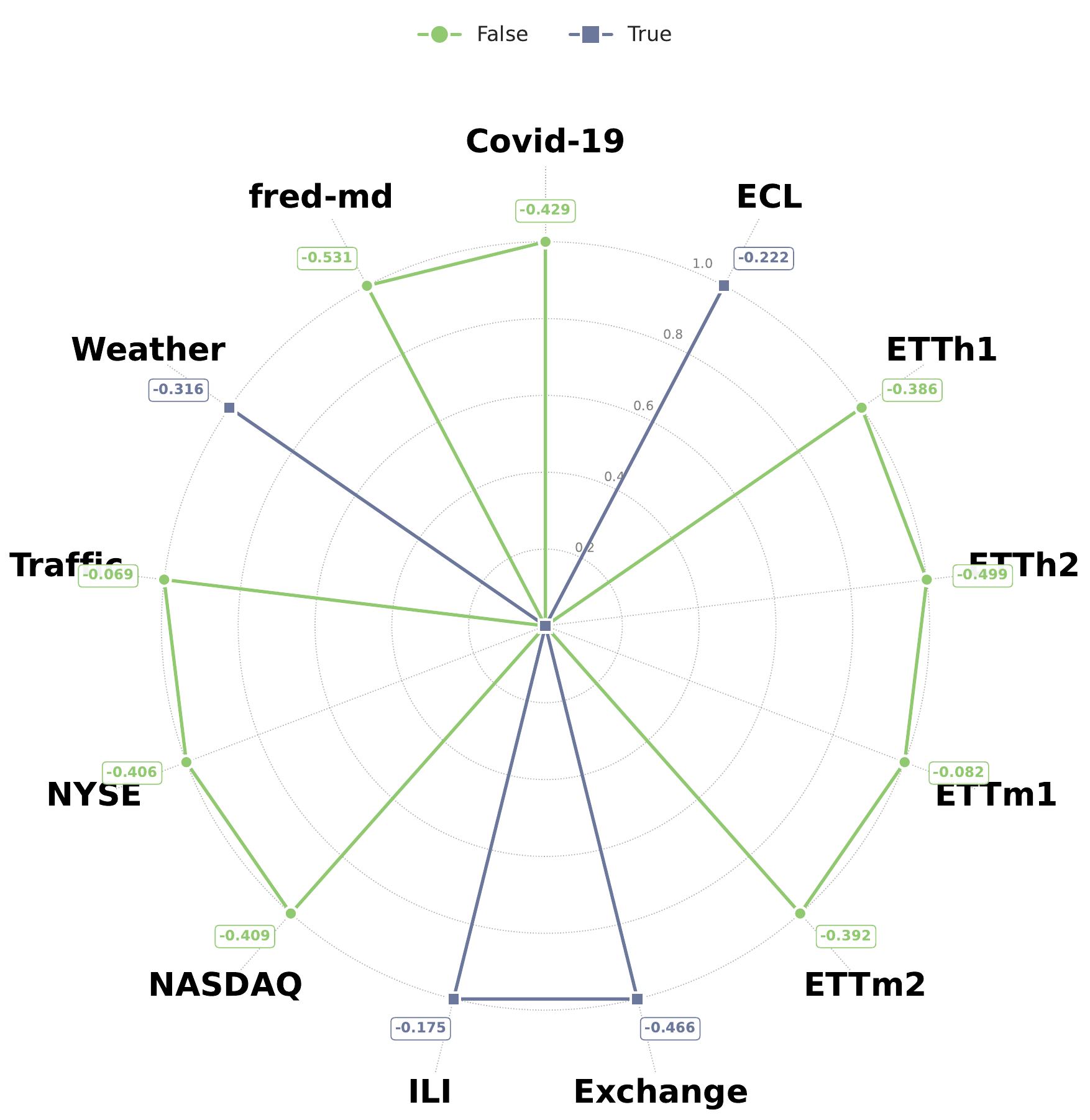}
    \caption{Transformer}
    \label{fig:appx_radar_series_sampling_Transformer}
  \end{subfigure}
  \hfill
  \begin{subfigure}[t]{0.16\textwidth}
    \centering
    \includegraphics[width=\textwidth]{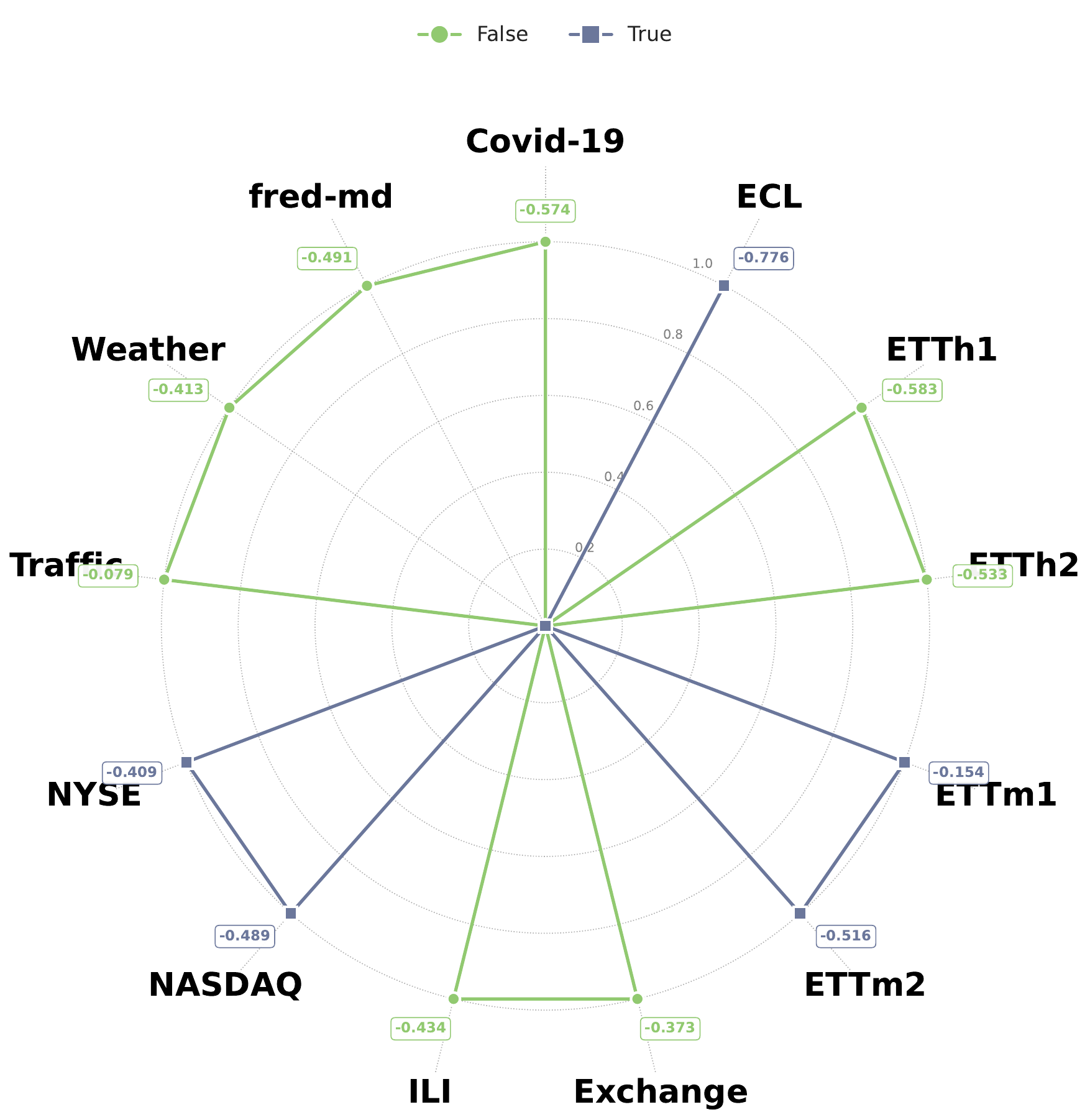}
    \caption{LLM}
    \label{fig:appx_radar_series_sampling_LLM}
  \end{subfigure}
  \hfill
  \begin{subfigure}[t]{0.16\textwidth}
    \centering
    \includegraphics[width=\textwidth]{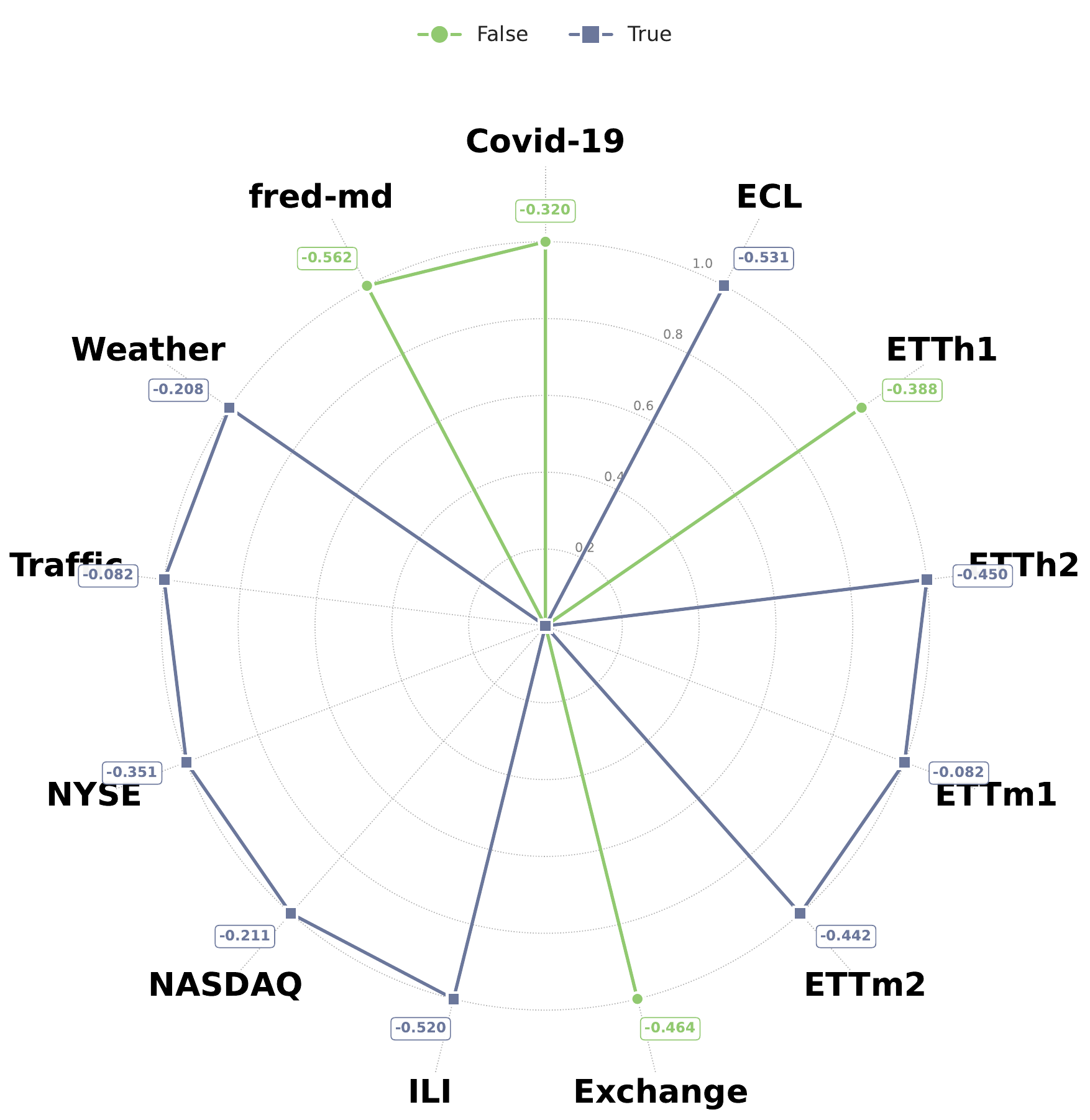}
    \caption{TSFM}
    \label{fig:appx_radar_series_sampling_TSFM}
  \end{subfigure}
  \caption{Dataset Adaptability (Radar Charts) for Series Sampling/Mixing (Radar Plots). This figure visualizes the performance distributions across different model architectures.}
  \label{fig:appx_radar_series_sampling}
\end{figure*}

We visualize the performance distributions and dataset adaptability for Series Normalization (Fig.~\ref{fig:appx_dist_gym_series_norm} and Fig.~\ref{fig:appx_radar_gym_series_norm}), Series Decomposition (Fig.~\ref{fig:appx_dist_gym_series_decomp} and Fig.~\ref{fig:appx_radar_gym_series_decomp}), and Series Sampling/Mixing (Fig.~\ref{fig:appx_dist_series_sampling} and Fig.~\ref{fig:appx_radar_series_sampling}).

\subsubsection{Series Encoding}
\begin{figure*}[htbp]
  \centering
  \begin{subfigure}[t]{0.16\textwidth}
    \centering
    \includegraphics[width=\textwidth]{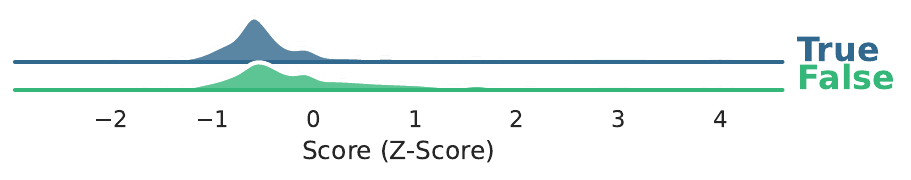}
    \caption{Global}
    \label{fig:appx_dist_channel_independent_Global}
  \end{subfigure}
  \hfill
  \begin{subfigure}[t]{0.16\textwidth}
    \centering
    \includegraphics[width=\textwidth]{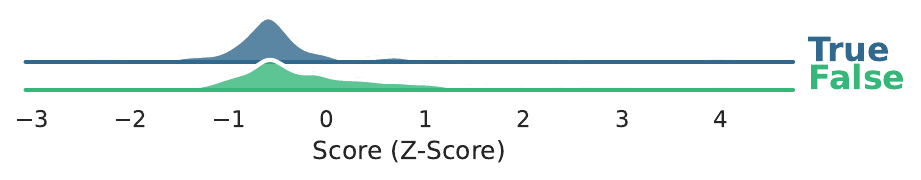}
    \caption{MLP}
    \label{fig:appx_dist_channel_independent_MLP}
  \end{subfigure}
  \hfill
  \begin{subfigure}[t]{0.16\textwidth}
    \centering
    \includegraphics[width=\textwidth]{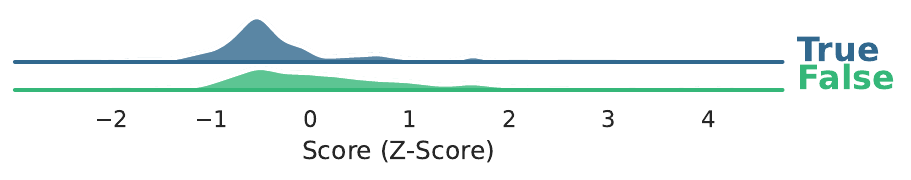}
    \caption{RNN}
    \label{fig:appx_dist_channel_independent_RNN}
  \end{subfigure}
  \begin{subfigure}[t]{0.16\textwidth}
    \centering
    \includegraphics[width=\textwidth]{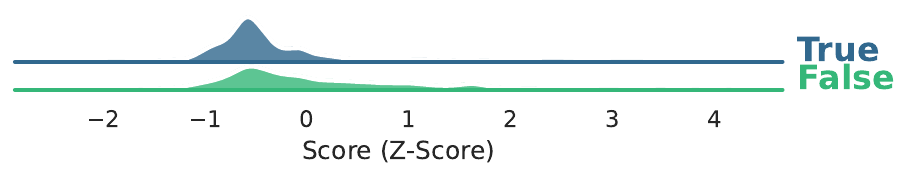}
    \caption{Transformer}
    \label{fig:appx_dist_channel_independent_Transformer}
  \end{subfigure}
  \hfill
  \begin{subfigure}[t]{0.16\textwidth}
    \centering
    \includegraphics[width=\textwidth]{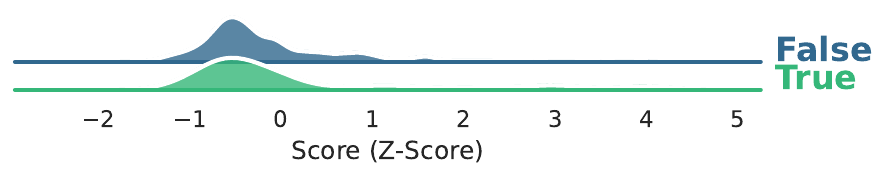}
    \caption{LLM}
    \label{fig:appx_dist_channel_independent_LLM}
  \end{subfigure}
  \hfill
  \begin{subfigure}[t]{0.16\textwidth}
    \centering
    \includegraphics[width=\textwidth]{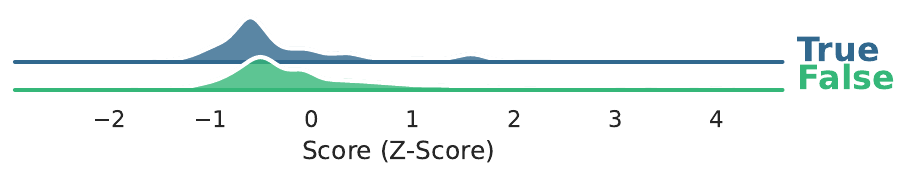}
    \caption{TSFM}
    \label{fig:appx_dist_channel_independent_TSFM}
  \end{subfigure}
  \caption{Performance Distributions for Channel Independence (Ridgeline Plots). This figure visualizes the performance distributions across different model architectures.}
  \label{fig:appx_dist_channel_independent}
\end{figure*}

\begin{figure*}[htbp]
  \centering
  \begin{subfigure}[t]{0.16\textwidth}
    \centering
    \includegraphics[width=\textwidth]{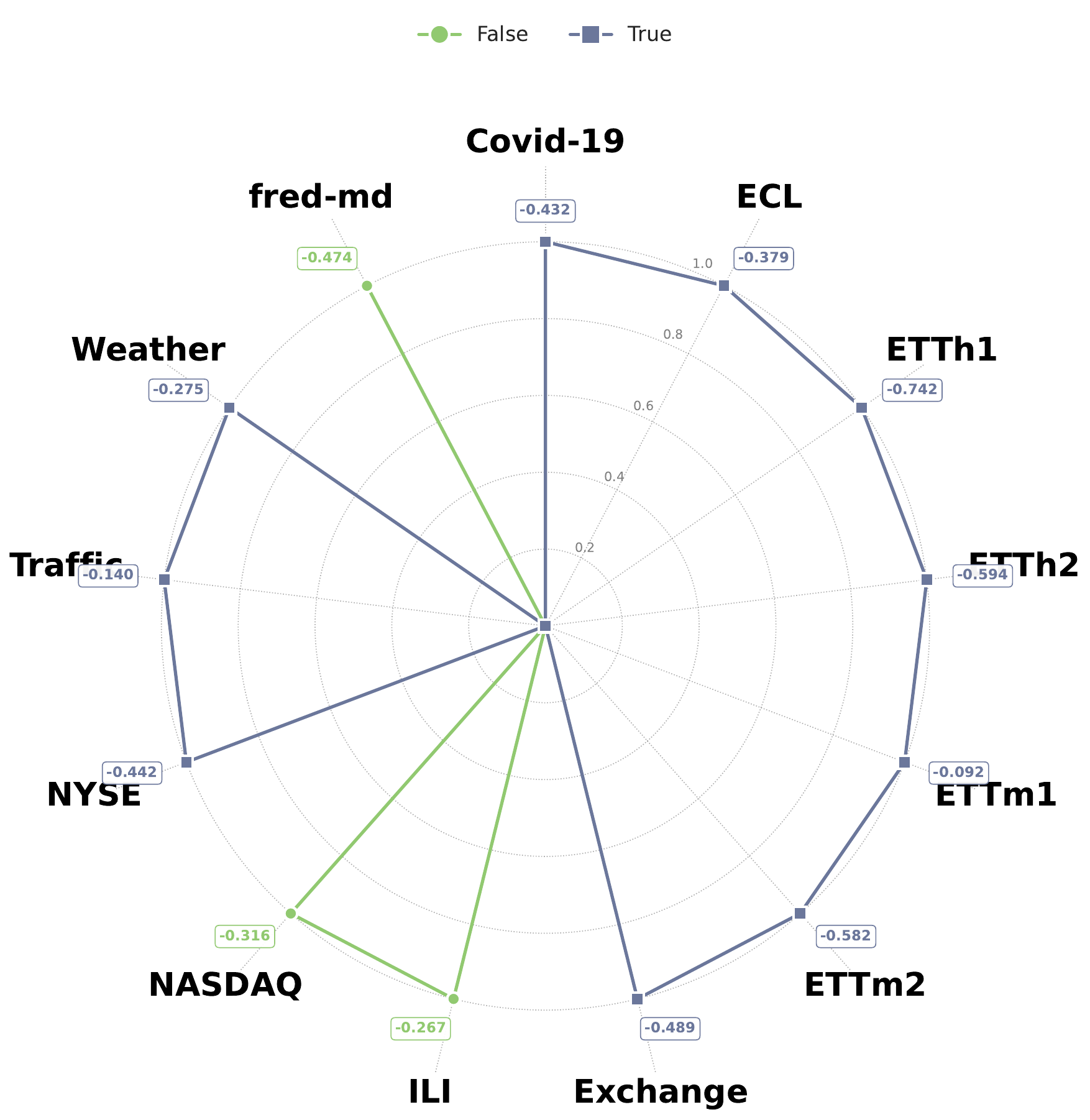}
    \caption{Global}
    \label{fig:appx_radar_channel_independent_Global}
  \end{subfigure}
  \hfill
  \begin{subfigure}[t]{0.16\textwidth}
    \centering
    \includegraphics[width=\textwidth]{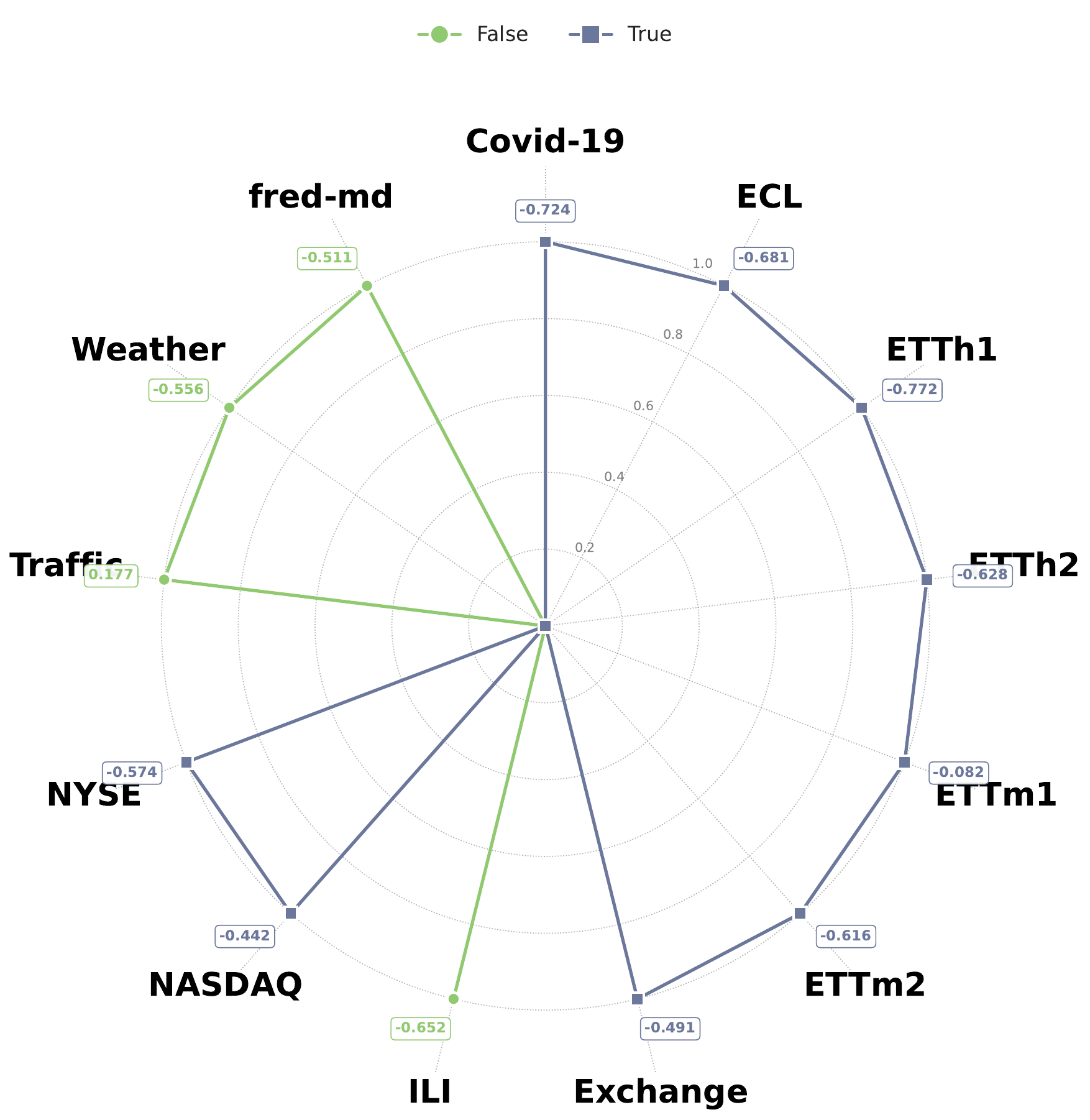}
    \caption{MLP}
    \label{fig:appx_radar_channel_independent_MLP}
  \end{subfigure}
  \hfill
  \begin{subfigure}[t]{0.16\textwidth}
    \centering
    \includegraphics[width=\textwidth]{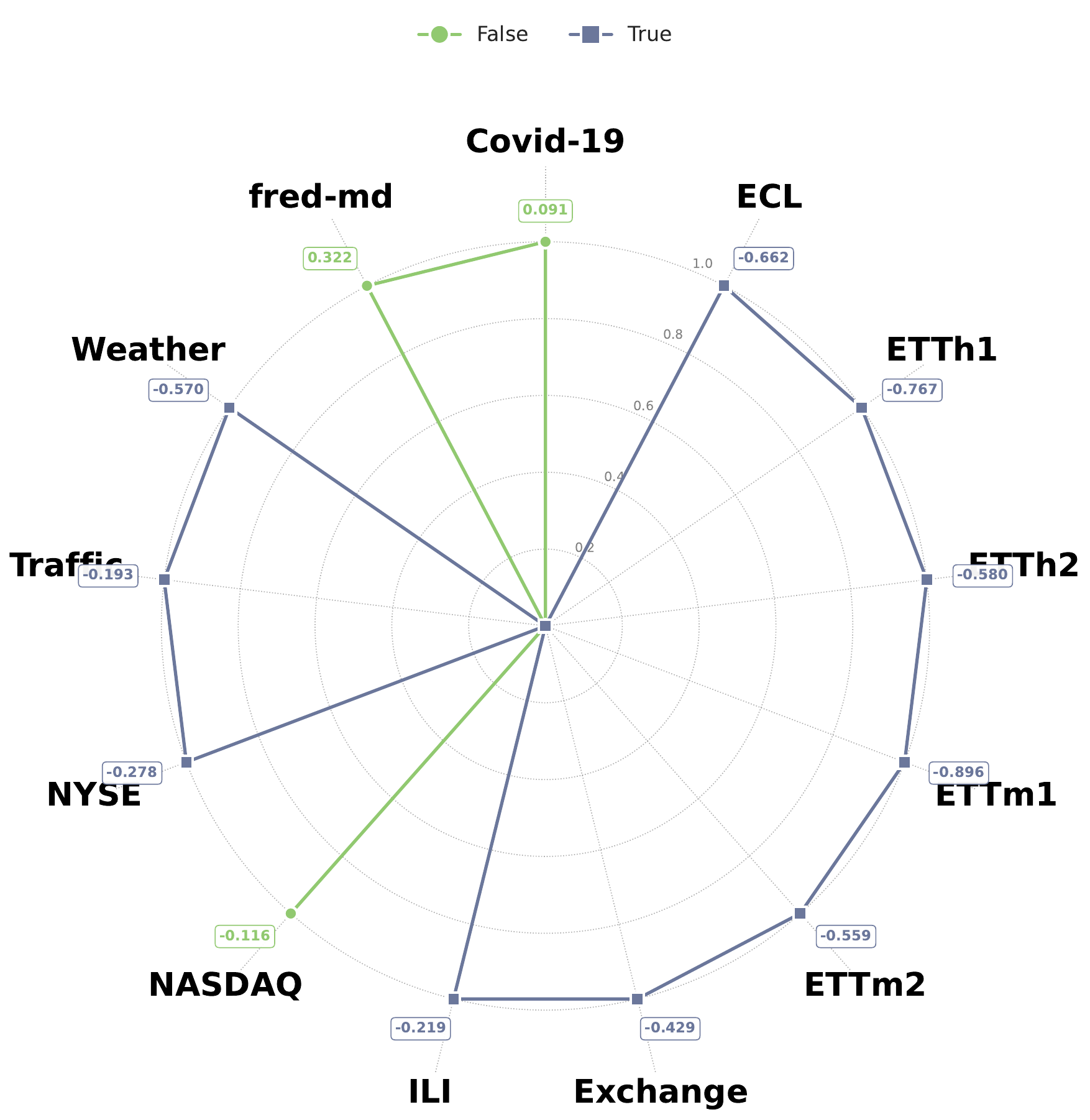}
    \caption{RNN}
    \label{fig:appx_radar_channel_independent_RNN}
  \end{subfigure}
  \begin{subfigure}[t]{0.16\textwidth}
    \centering
    \includegraphics[width=\textwidth]{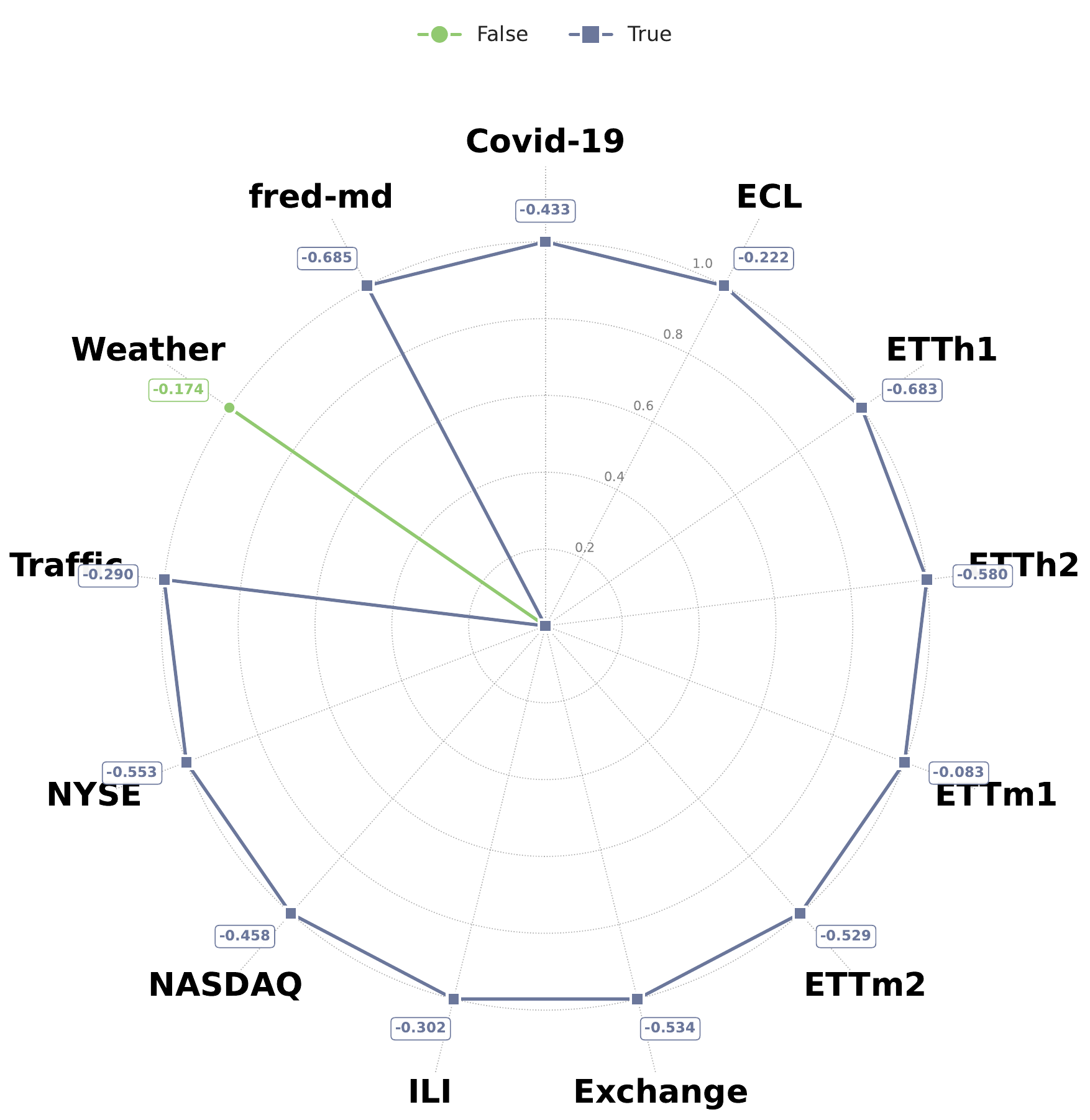}
    \caption{Transformer}
    \label{fig:appx_radar_channel_independent_Transformer}
  \end{subfigure}
  \hfill
  \begin{subfigure}[t]{0.16\textwidth}
    \centering
    \includegraphics[width=\textwidth]{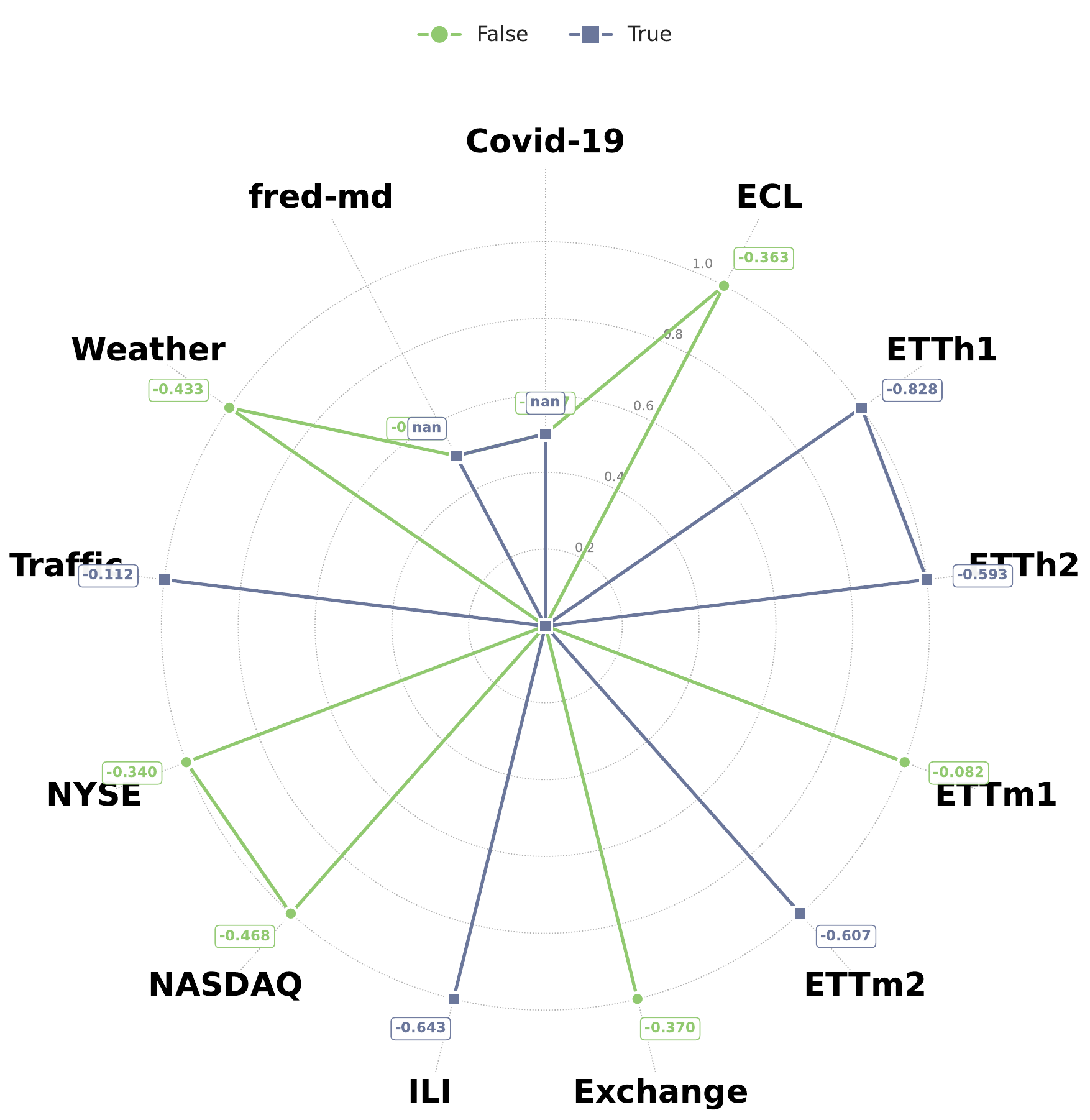}
    \caption{LLM}
    \label{fig:appx_radar_channel_independent_LLM}
  \end{subfigure}
  \hfill
  \begin{subfigure}[t]{0.16\textwidth}
    \centering
    \includegraphics[width=\textwidth]{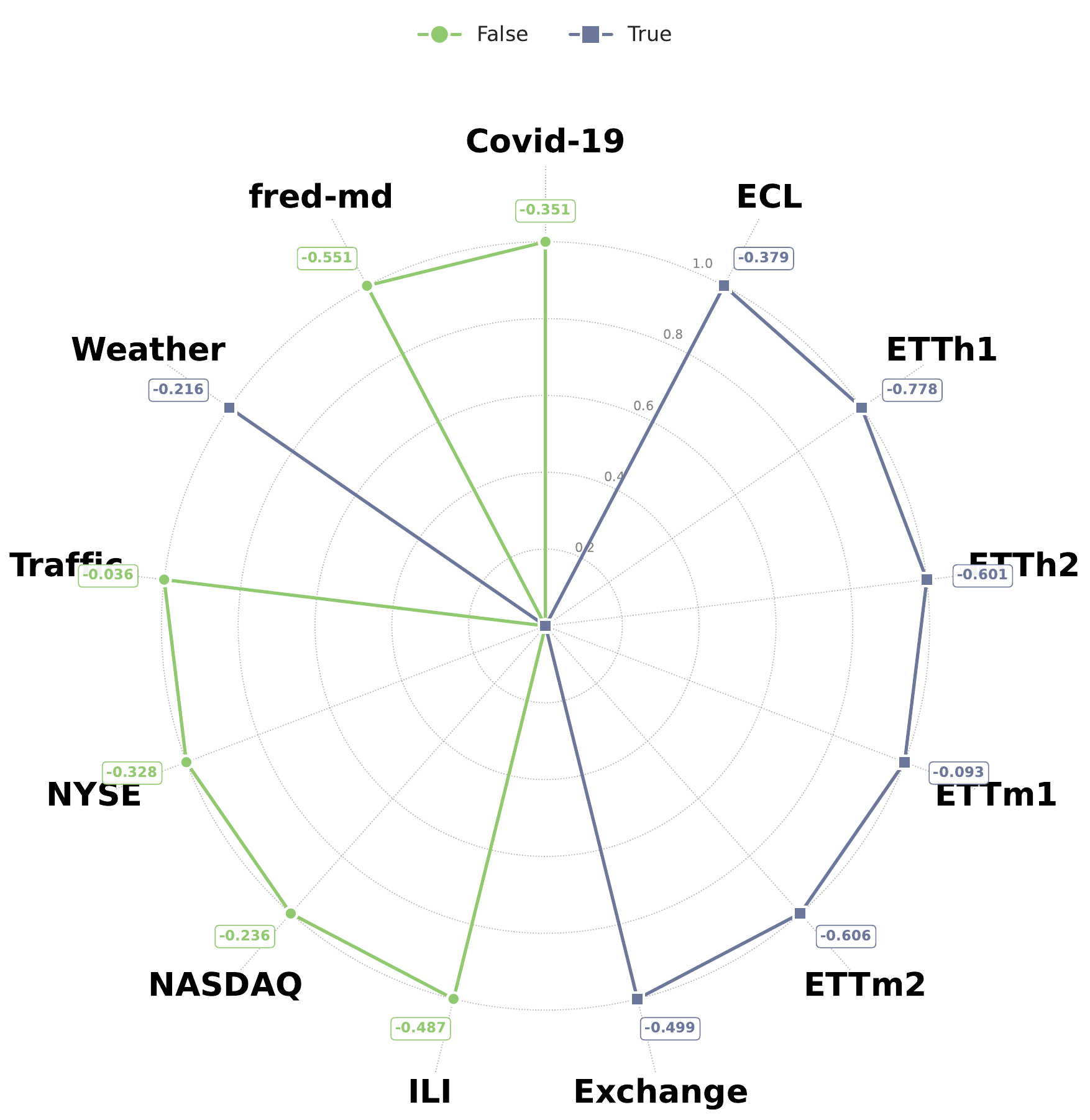}
    \caption{TSFM}
    \label{fig:appx_radar_channel_independent_TSFM}
  \end{subfigure}
  \caption{Dataset Adaptability (Radar Charts) for Channel Independence (Radar Plots). This figure visualizes the performance distributions across different model architectures.}
  \label{fig:appx_radar_channel_independent}
\end{figure*}

\begin{figure*}[htbp]
  \centering
  \begin{subfigure}[t]{0.16\textwidth}
    \centering
    \includegraphics[width=\textwidth]{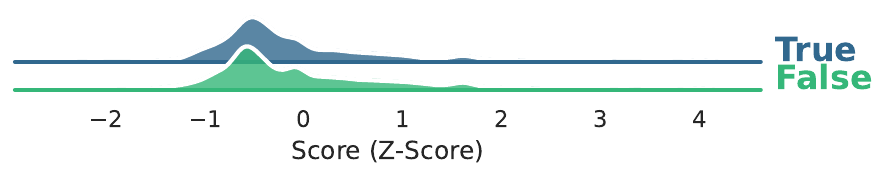}
    \caption{Global}
    \label{fig:appx_dist_gym_x_mark_Global}
  \end{subfigure}
  \hfill
  \begin{subfigure}[t]{0.16\textwidth}
    \centering
    \includegraphics[width=\textwidth]{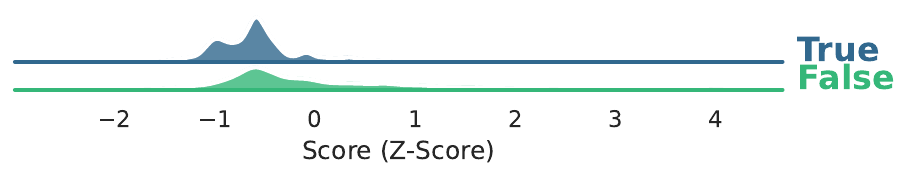}
    \caption{MLP}
    \label{fig:appx_dist_gym_x_mark_MLP}
  \end{subfigure}
  \hfill
  \begin{subfigure}[t]{0.16\textwidth}
    \centering
    \includegraphics[width=\textwidth]{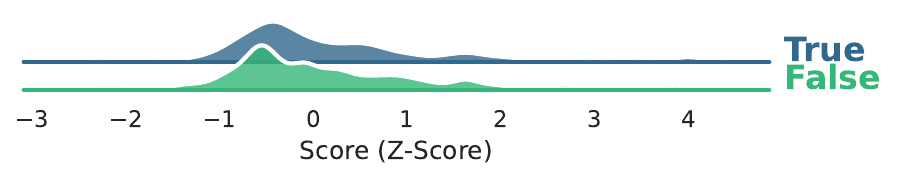}
    \caption{RNN}
    \label{fig:appx_dist_gym_x_mark_RNN}
  \end{subfigure}
  \begin{subfigure}[t]{0.16\textwidth}
    \centering
    \includegraphics[width=\textwidth]{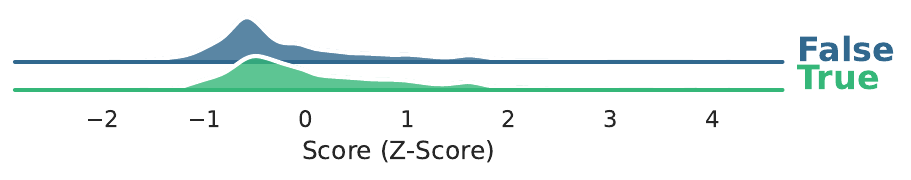}
    \caption{Transformer}
    \label{fig:appx_dist_gym_x_mark_Transformer}
  \end{subfigure}
  \hfill
  \begin{subfigure}[t]{0.16\textwidth}
    \centering
    \includegraphics[width=\textwidth]{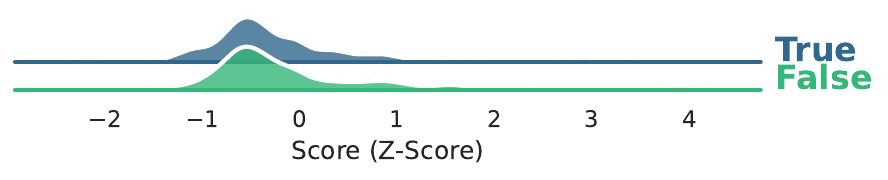}
    \caption{LLM}
    \label{fig:appx_dist_gym_x_mark_LLM}
  \end{subfigure}
  \hfill
  \begin{subfigure}[t]{0.16\textwidth}
    \centering
    \includegraphics[width=\textwidth]{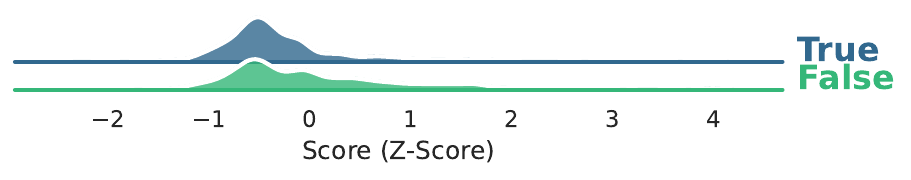}
    \caption{TSFM}
    \label{fig:appx_dist_gym_x_mark_TSFM}
  \end{subfigure}
  \caption{Performance Distributions for Timestamp Embedding (Ridgeline Plots). This figure visualizes the performance distributions across different model architectures.}
  \label{fig:appx_dist_gym_x_mark}
\end{figure*}

\begin{figure*}[htbp]
  \centering
  \begin{subfigure}[t]{0.16\textwidth}
    \centering
    \includegraphics[width=\textwidth]{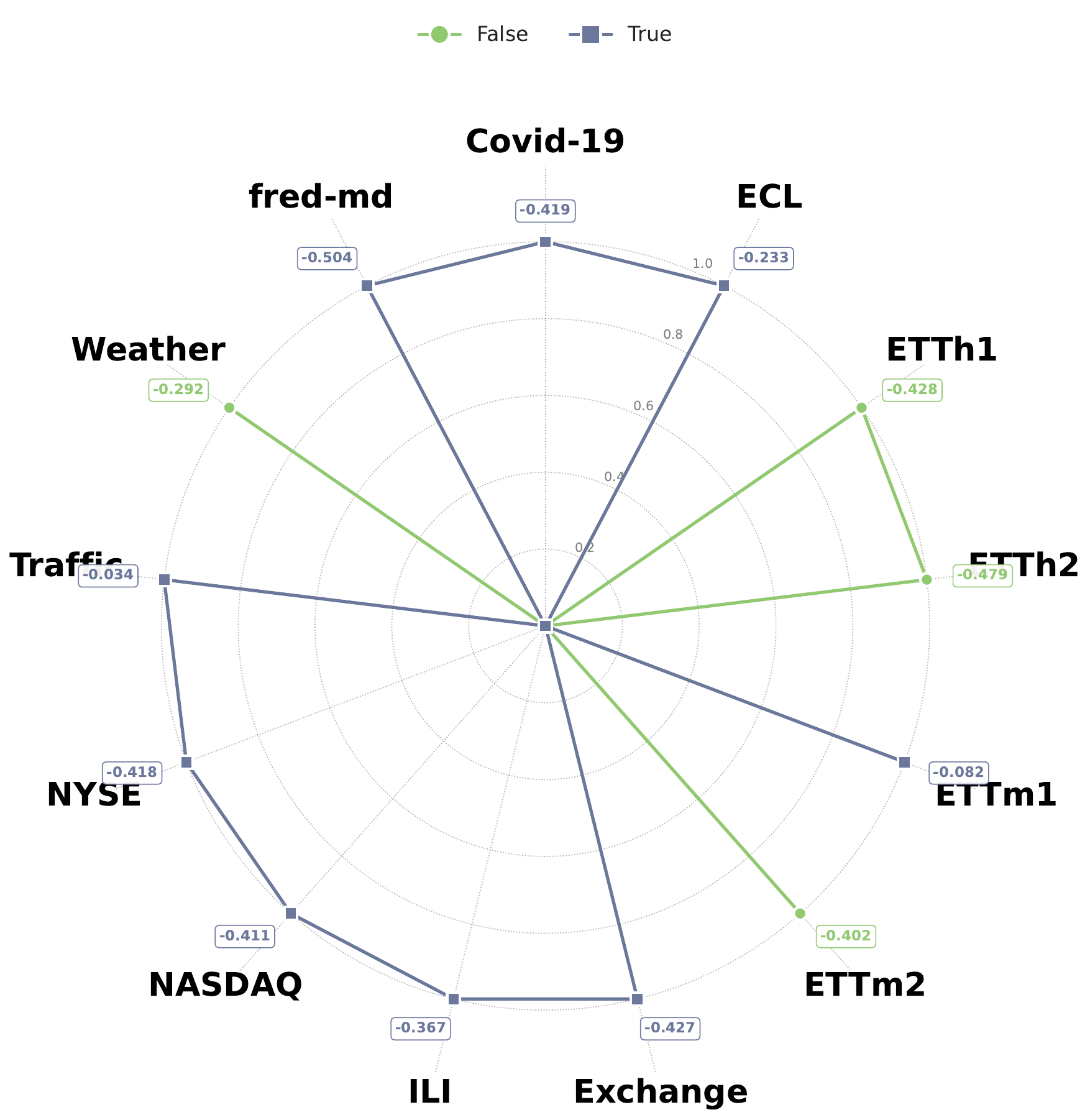}
    \caption{Global}
    \label{fig:appx_radar_gym_x_mark_Global}
  \end{subfigure}
  \hfill
  \begin{subfigure}[t]{0.16\textwidth}
    \centering
    \includegraphics[width=\textwidth]{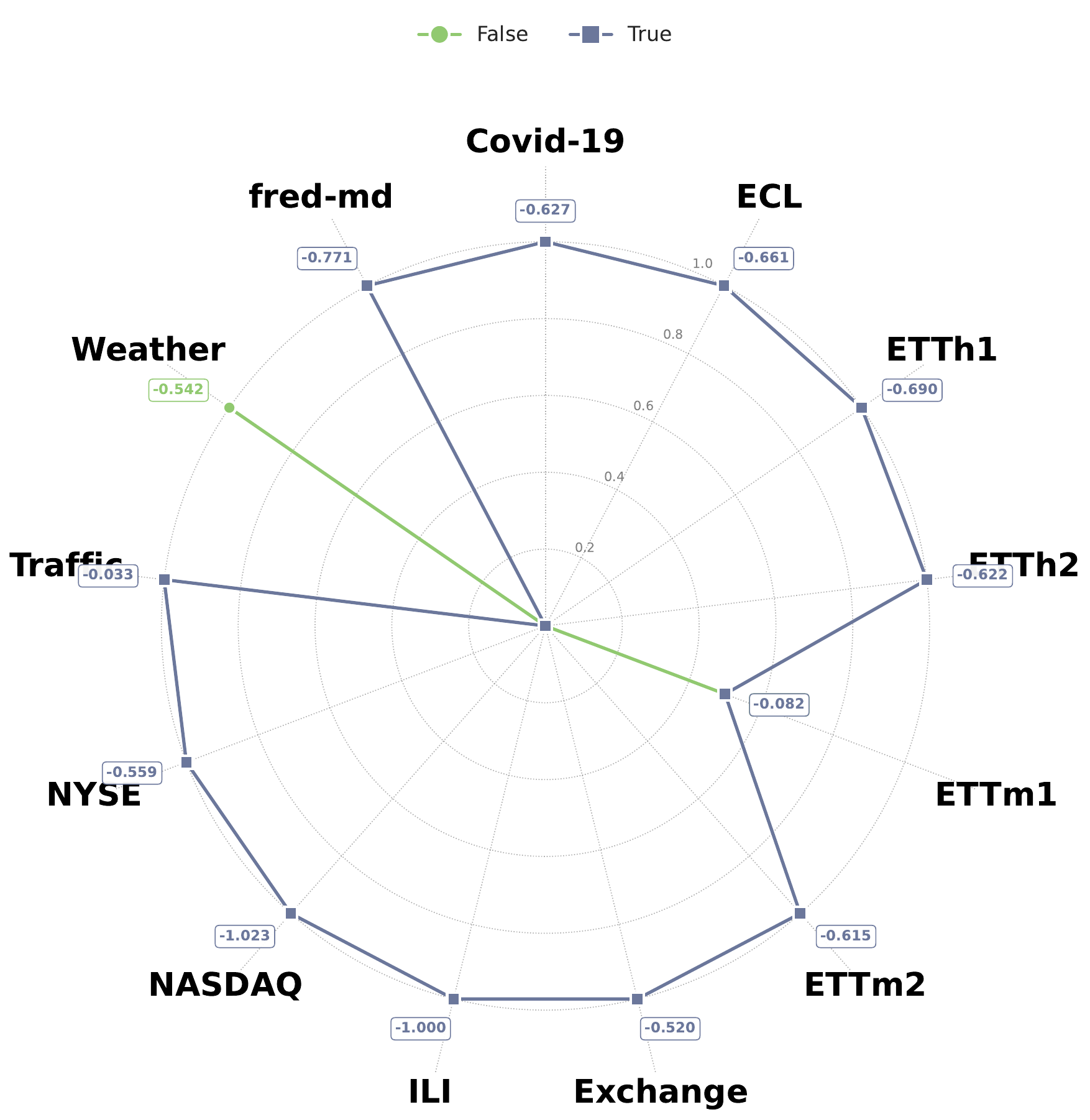}
    \caption{MLP}
    \label{fig:appx_radar_gym_x_mark_MLP}
  \end{subfigure}
  \hfill
  \begin{subfigure}[t]{0.16\textwidth}
    \centering
    \includegraphics[width=\textwidth]{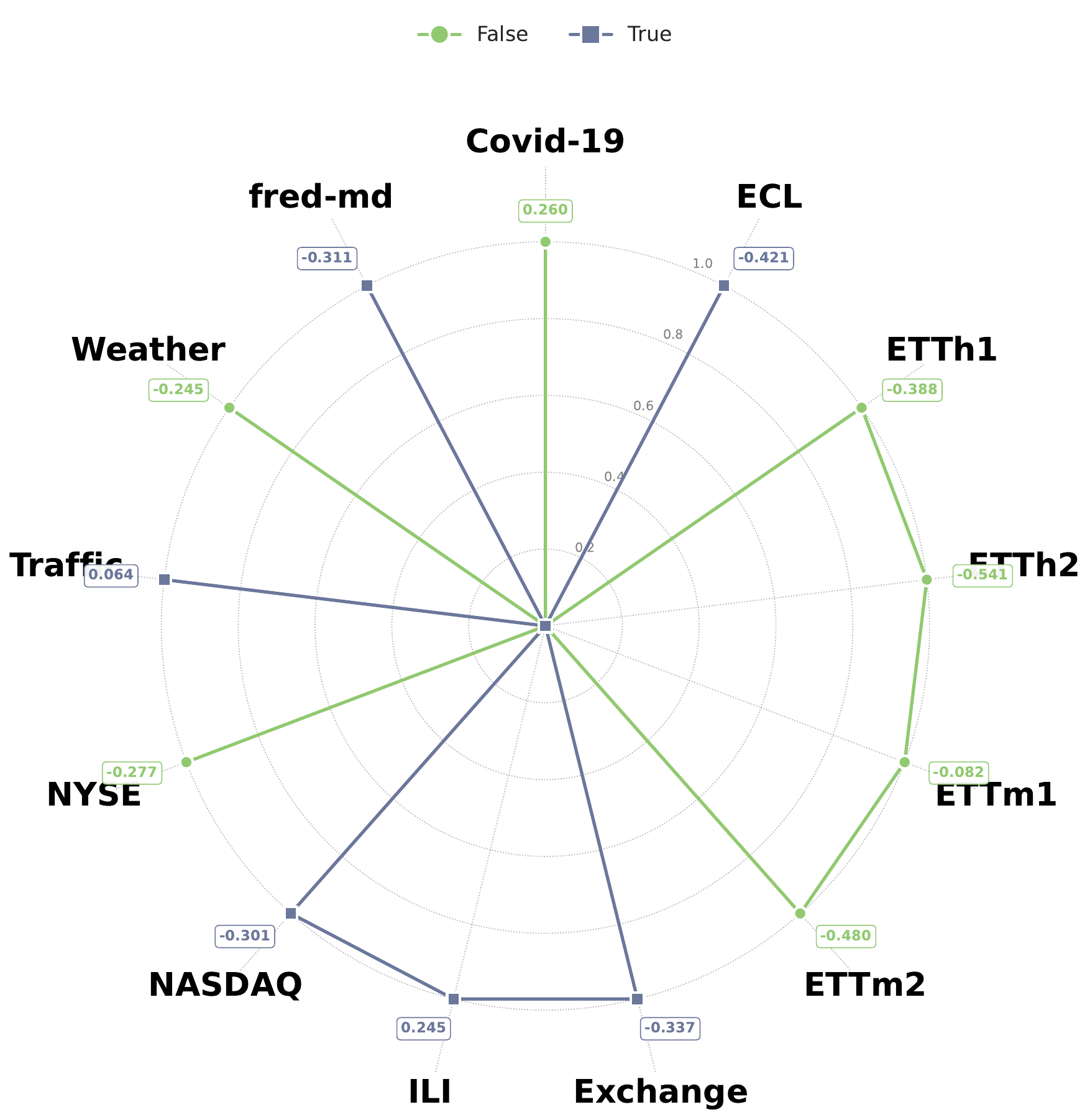}
    \caption{RNN}
    \label{fig:appx_radar_gym_x_mark_RNN}
  \end{subfigure}
  \begin{subfigure}[t]{0.16\textwidth}
    \centering
    \includegraphics[width=\textwidth]{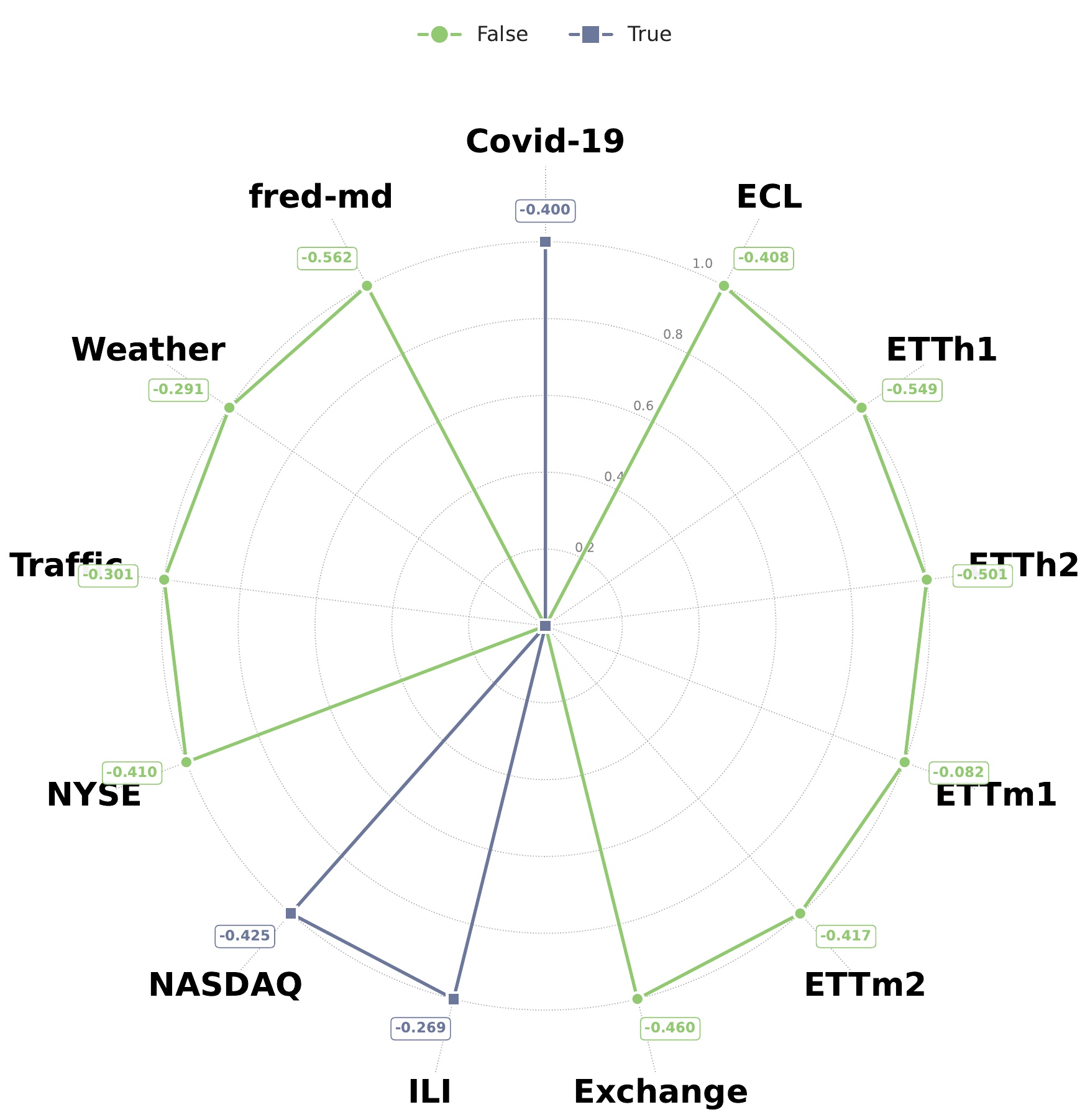}
    \caption{Transformer}
    \label{fig:appx_radar_gym_x_mark_Transformer}
  \end{subfigure}
  \hfill
  \begin{subfigure}[t]{0.16\textwidth}
    \centering
    \includegraphics[width=\textwidth]{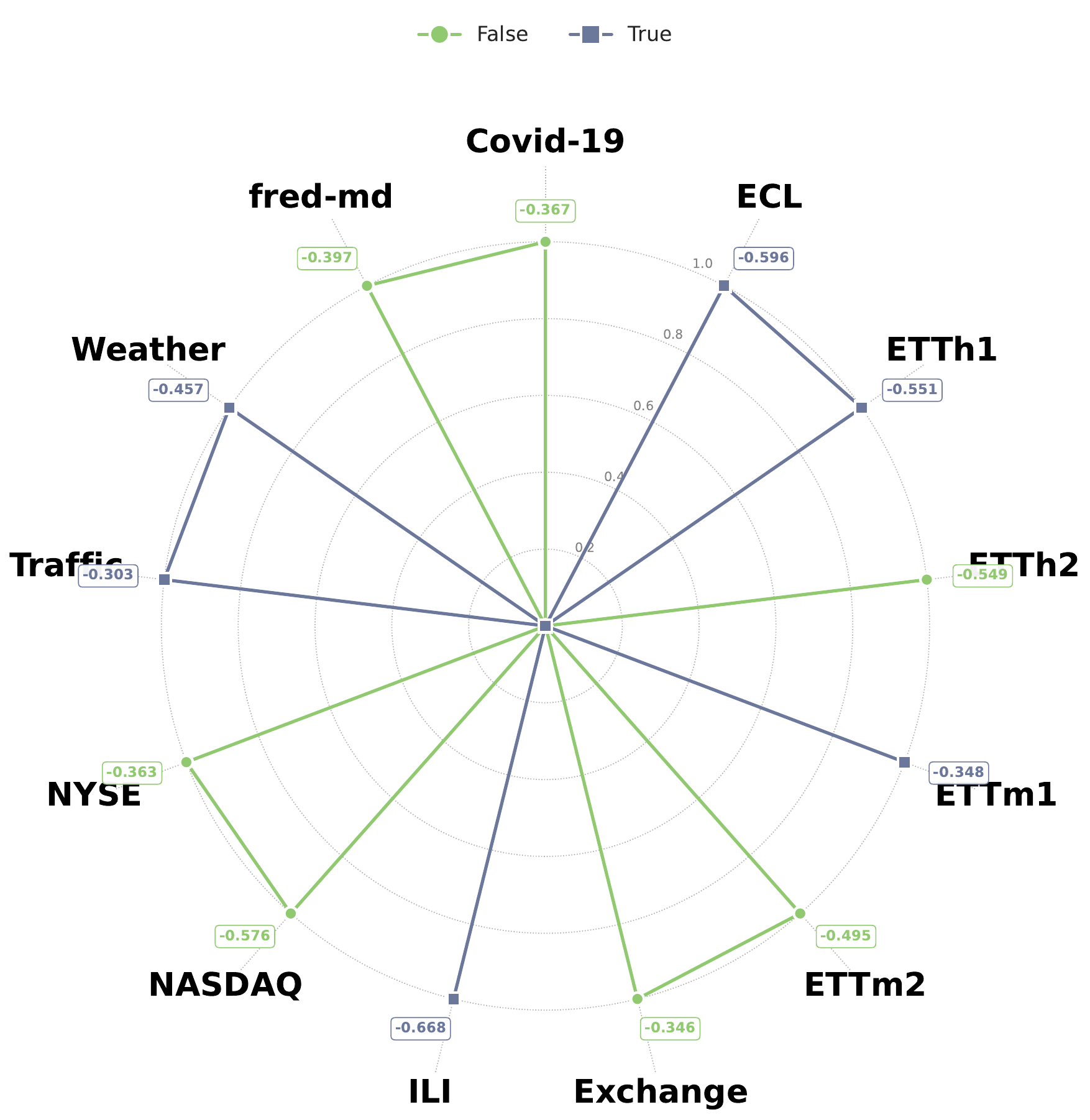}
    \caption{LLM}
    \label{fig:appx_radar_gym_x_mark_LLM}
  \end{subfigure}
  \hfill
  \begin{subfigure}[t]{0.16\textwidth}
    \centering
    \includegraphics[width=\textwidth]{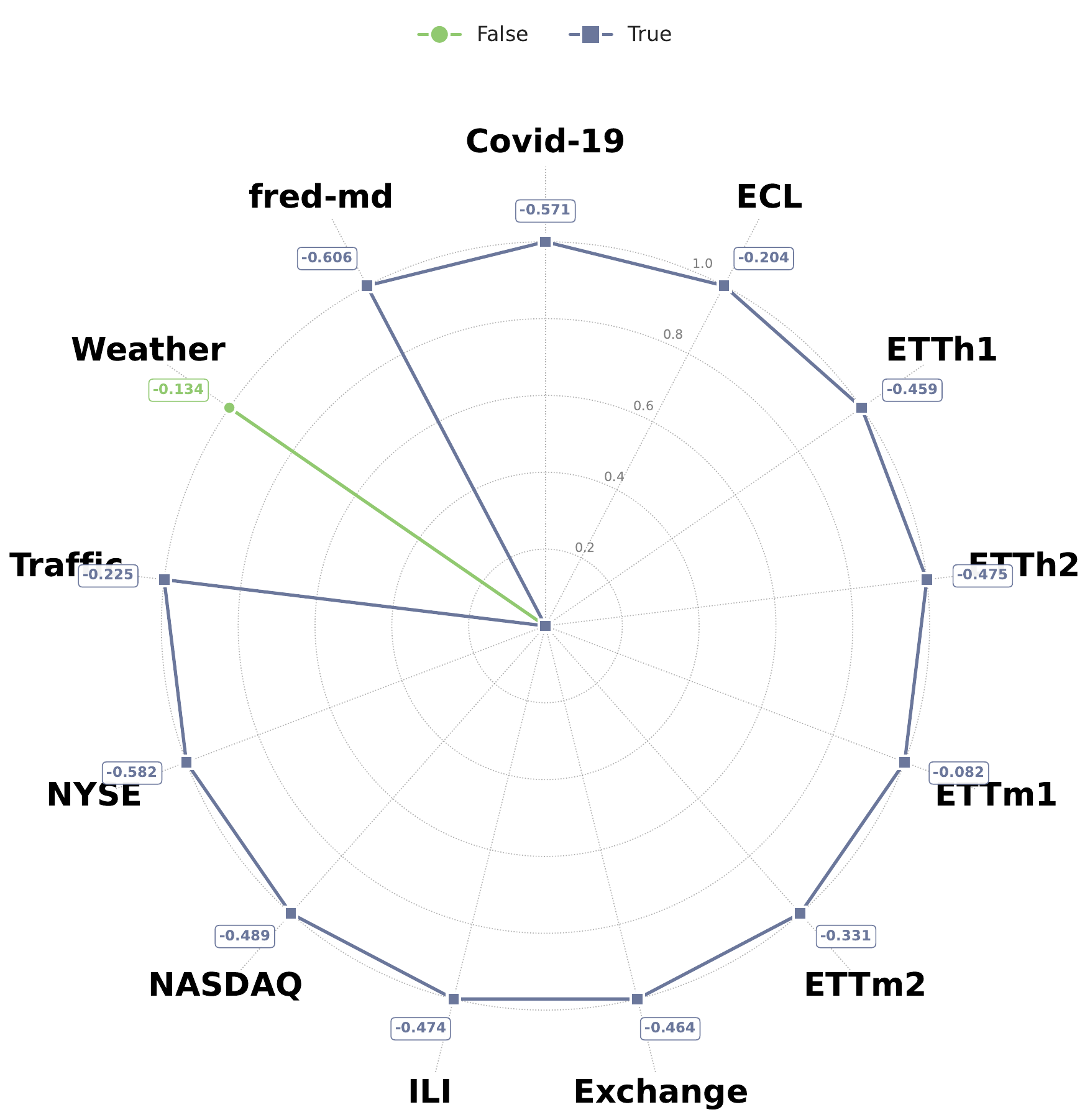}
    \caption{TSFM}
    \label{fig:appx_radar_gym_x_mark_TSFM}
  \end{subfigure}
  \caption{Dataset Adaptability (Radar Charts) for Timestamp Embedding (Radar Plots). This figure visualizes the performance distributions across different model architectures.}
  \label{fig:appx_radar_gym_x_mark}
\end{figure*}

\begin{figure*}[htbp]
  \centering
  \begin{subfigure}[t]{0.16\textwidth}
    \centering
    \includegraphics[width=\textwidth]{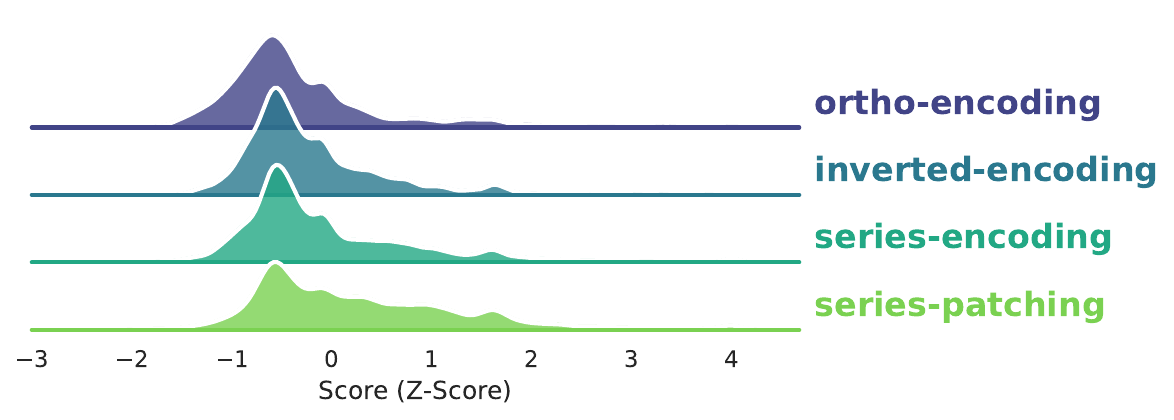}
    \caption{Global}
    \label{fig:appx_dist_gym_input_embed_Global}
  \end{subfigure}
  \hfill
  \begin{subfigure}[t]{0.16\textwidth}
    \centering
    \includegraphics[width=\textwidth]{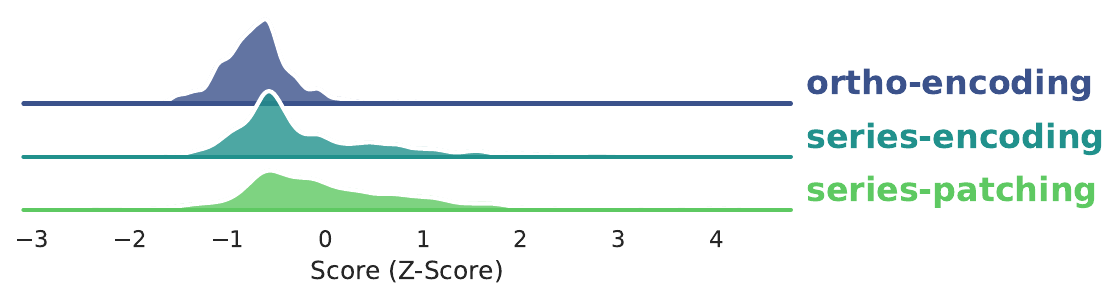}
    \caption{MLP}
    \label{fig:appx_dist_gym_input_embed_MLP}
  \end{subfigure}
  \hfill
  \begin{subfigure}[t]{0.16\textwidth}
    \centering
    \includegraphics[width=\textwidth]{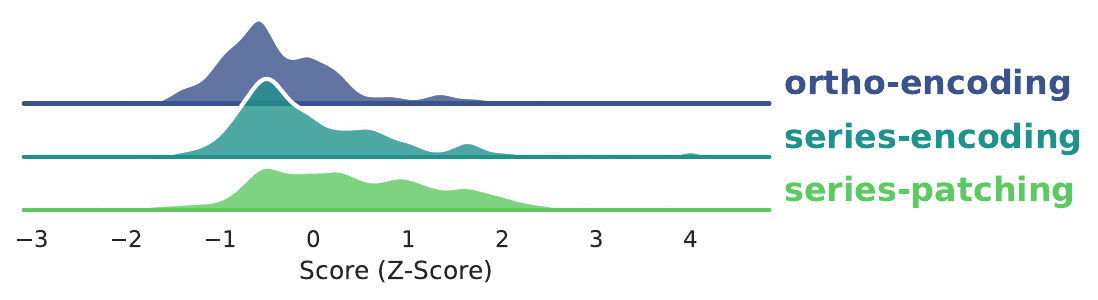}
    \caption{RNN}
    \label{fig:appx_dist_gym_input_embed_RNN}
  \end{subfigure}
  \begin{subfigure}[t]{0.16\textwidth}
    \centering
    \includegraphics[width=\textwidth]{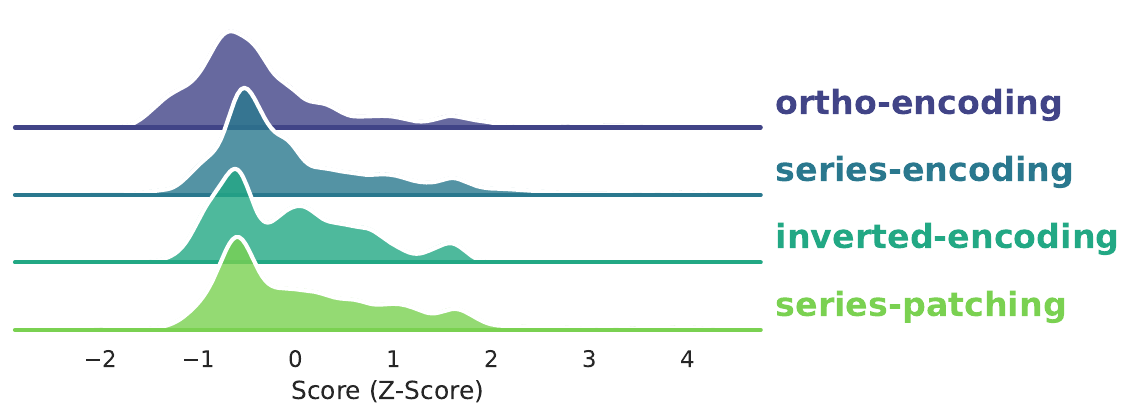}
    \caption{Transformer}
    \label{fig:appx_dist_gym_input_embed_Transformer}
  \end{subfigure}
  \hfill
  \begin{subfigure}[t]{0.16\textwidth}
    \centering
    \includegraphics[width=\textwidth]{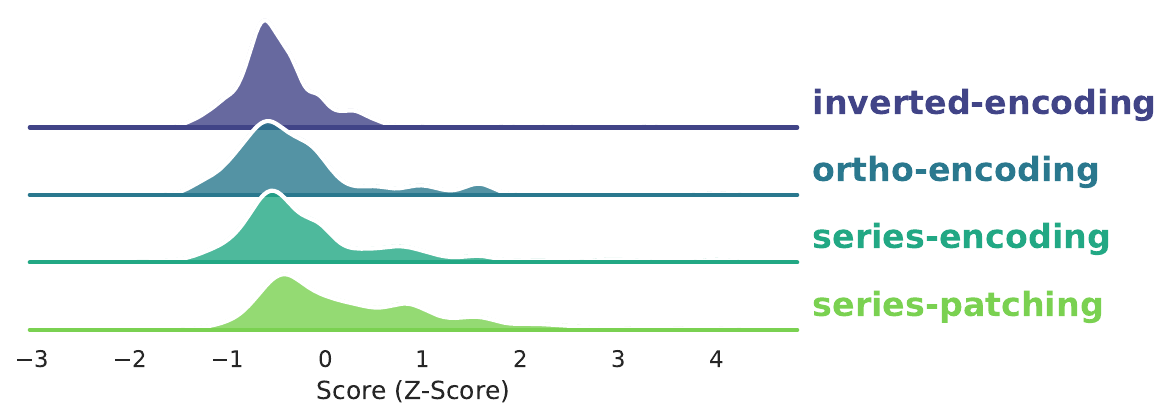}
    \caption{LLM}
    \label{fig:appx_dist_gym_input_embed_LLM}
  \end{subfigure}
  \hfill
  \begin{subfigure}[t]{0.16\textwidth}
    \centering
    \includegraphics[width=\textwidth]{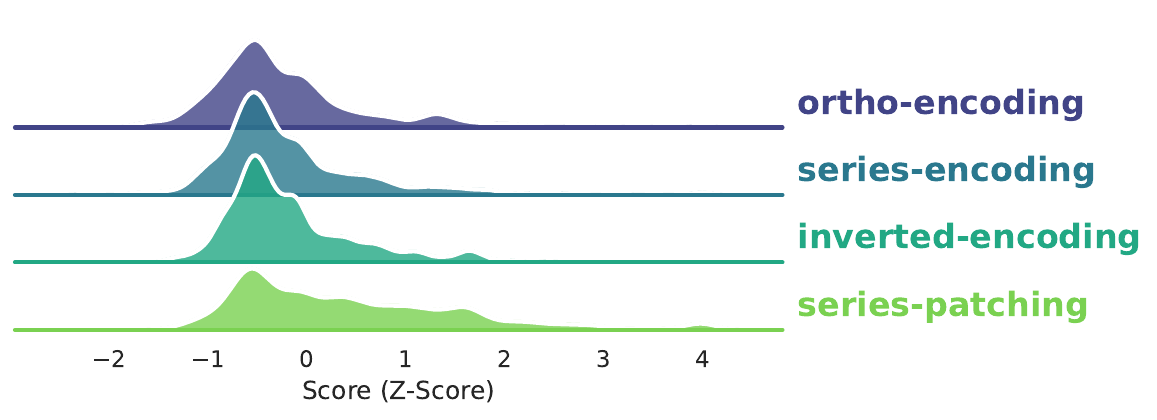}
    \caption{TSFM}
    \label{fig:appx_dist_gym_input_embed_TSFM}
  \end{subfigure}
  \caption{Performance Distributions for Series Tokenization (Ridgeline Plots). This figure visualizes the performance distributions across different model architectures.}
  \label{fig:appx_dist_gym_input_embed}
\end{figure*}

\begin{figure*}[htbp]
  \centering
  \begin{subfigure}[t]{0.16\textwidth}
    \centering
    \includegraphics[width=\textwidth]{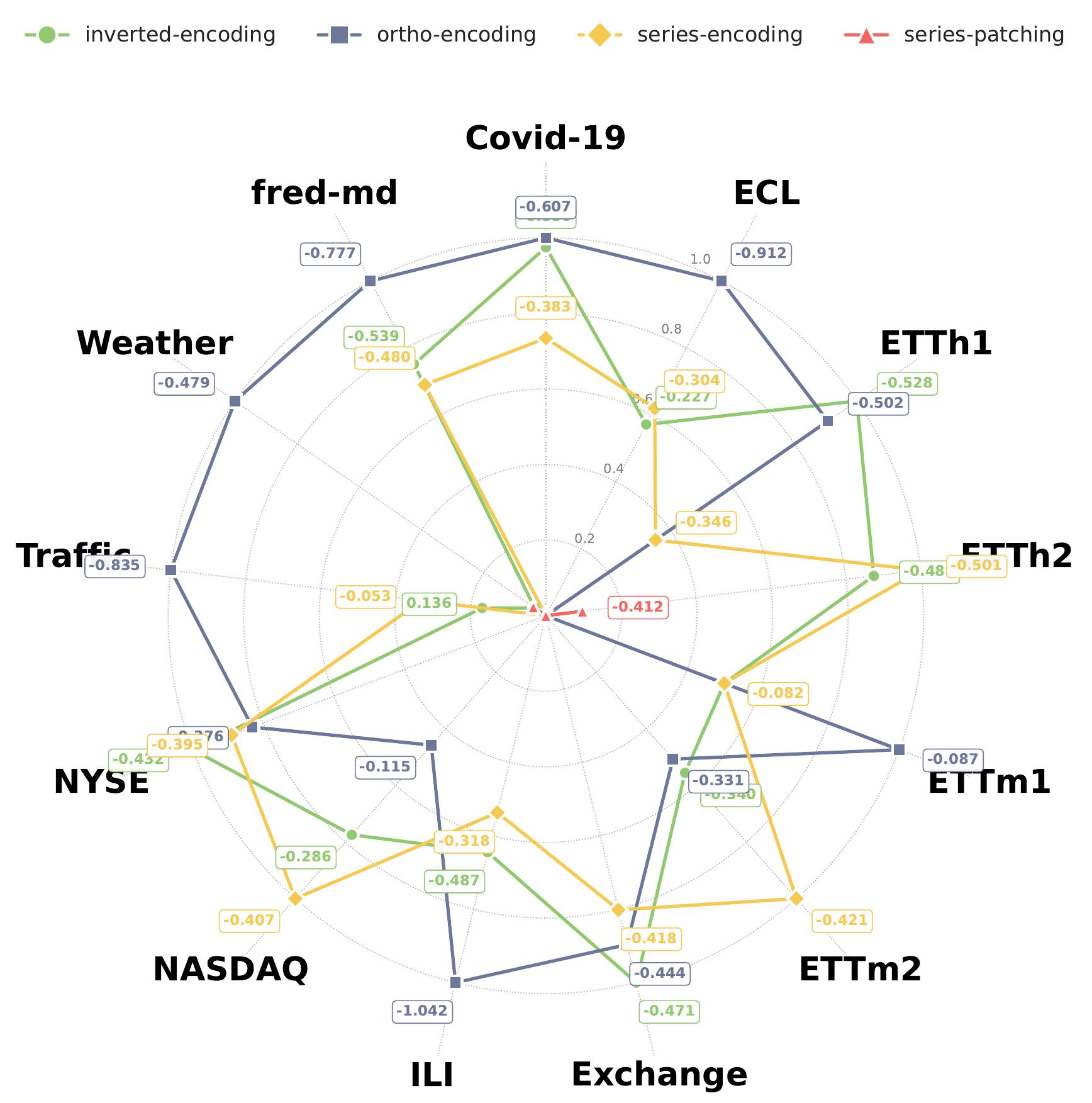}
    \caption{Global}
    \label{fig:appx_radar_gym_input_embed_Global}
  \end{subfigure}
  \hfill
  \begin{subfigure}[t]{0.16\textwidth}
    \centering
    \includegraphics[width=\textwidth]{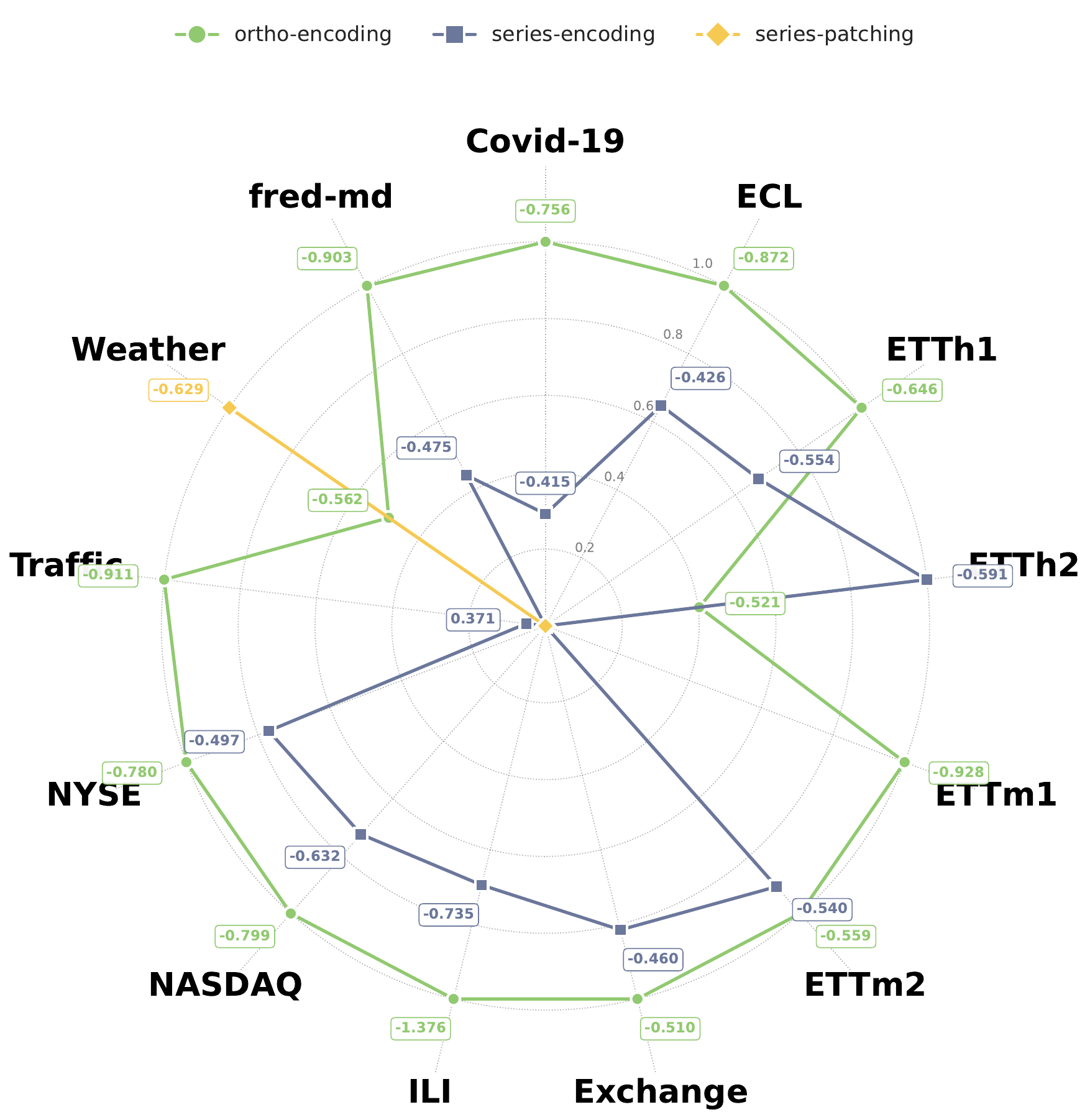}
    \caption{MLP}
    \label{fig:appx_radar_gym_input_embed_MLP}
  \end{subfigure}
  \hfill
  \begin{subfigure}[t]{0.16\textwidth}
    \centering
    \includegraphics[width=\textwidth]{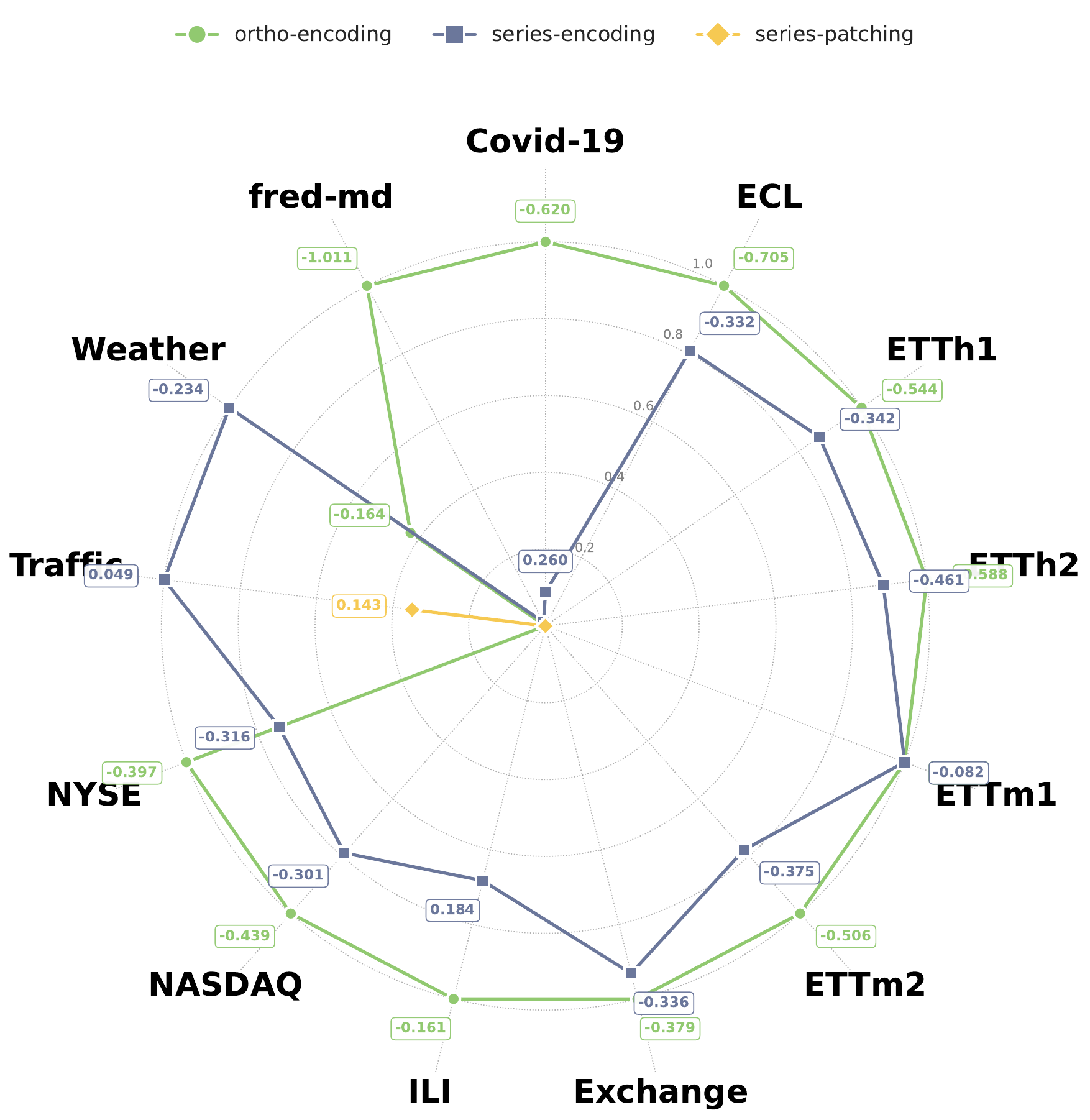}
    \caption{RNN}
    \label{fig:appx_radar_gym_input_embed_RNN}
  \end{subfigure}
  \begin{subfigure}[t]{0.16\textwidth}
    \centering
    \includegraphics[width=\textwidth]{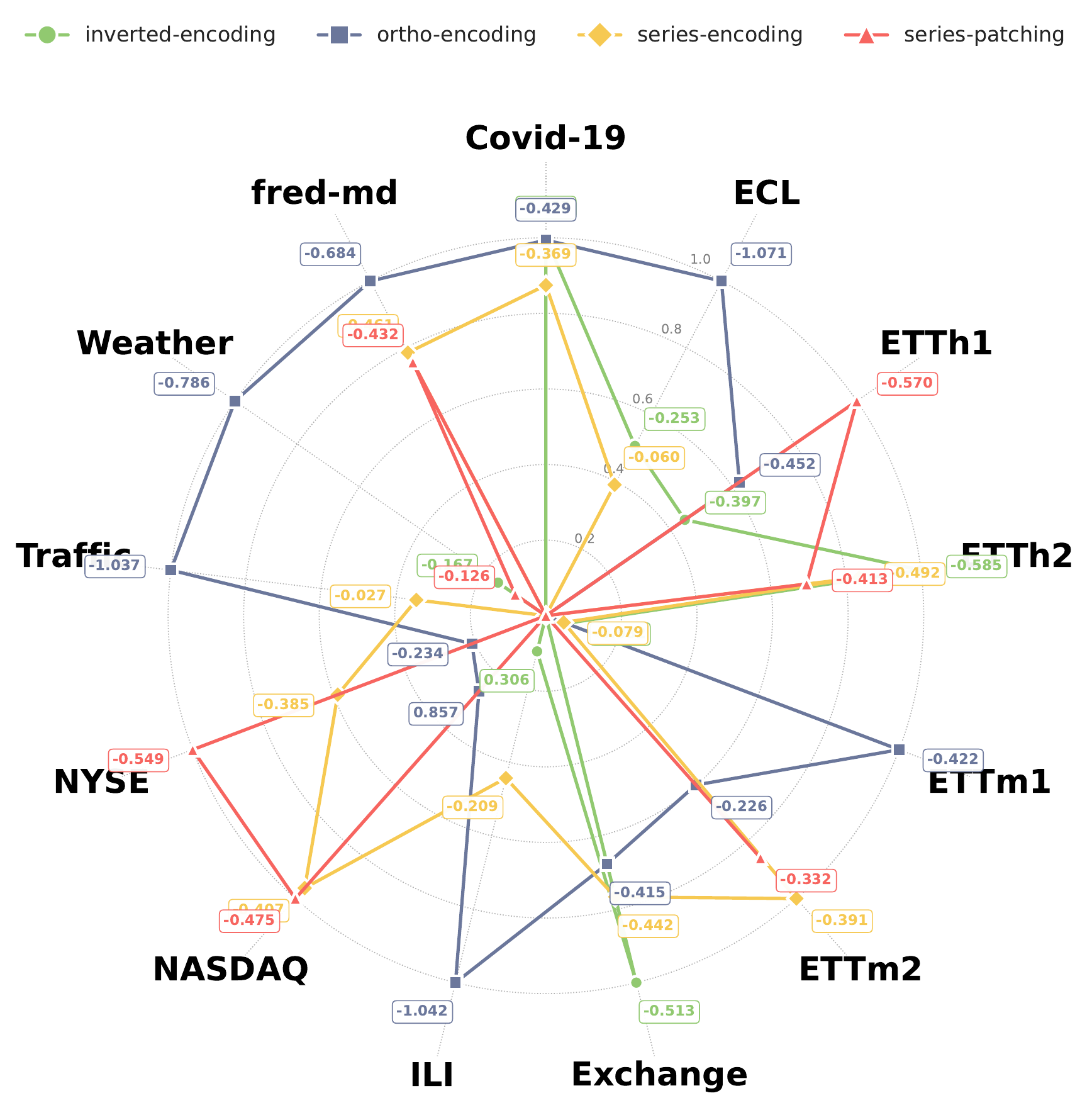}
    \caption{Transformer}
    \label{fig:appx_radar_gym_input_embed_Transformer}
  \end{subfigure}
  \hfill
  \begin{subfigure}[t]{0.16\textwidth}
    \centering
    \includegraphics[width=\textwidth]{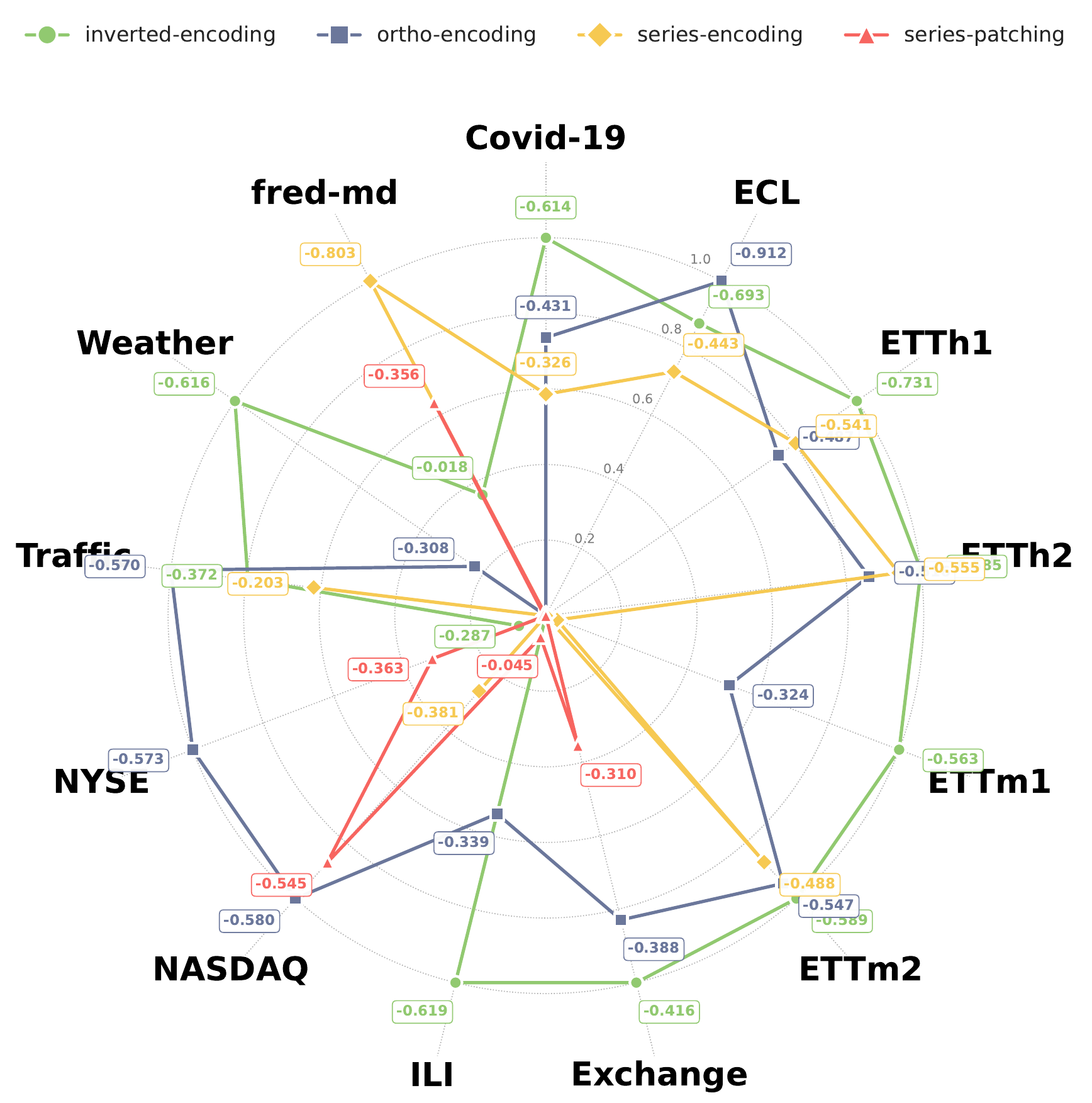}
    \caption{LLM}
    \label{fig:appx_radar_gym_input_embed_LLM}
  \end{subfigure}
  \hfill
  \begin{subfigure}[t]{0.16\textwidth}
    \centering
    \includegraphics[width=\textwidth]{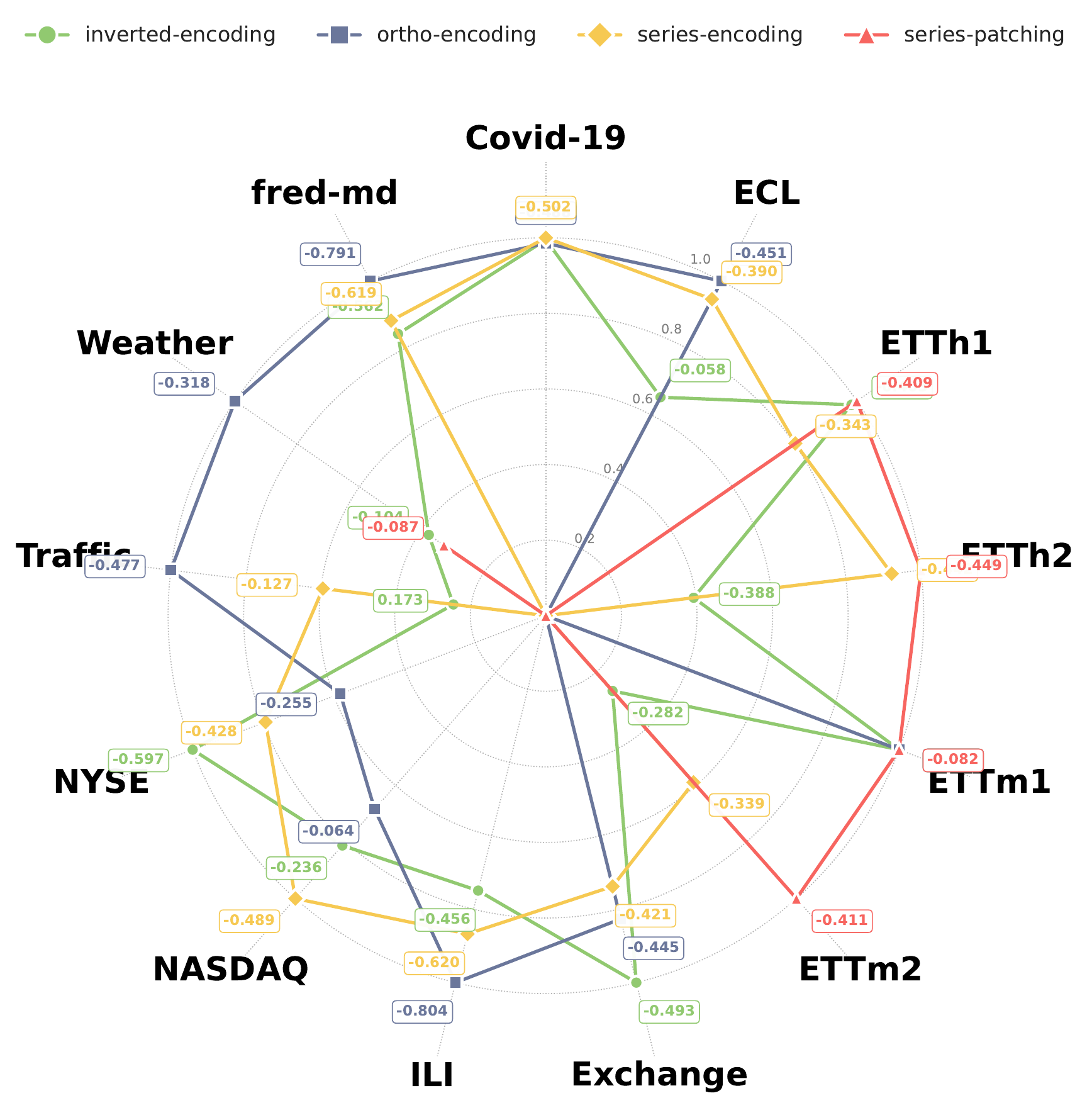}
    \caption{TSFM}
    \label{fig:appx_radar_gym_input_embed_TSFM}
  \end{subfigure}
  \caption{Dataset Adaptability (Radar Charts) for Series Tokenization (Radar Plots). This figure visualizes the performance distributions across different model architectures.}
  \label{fig:appx_radar_gym_input_embed}
\end{figure*}

We examine Channel Independence (Fig.~\ref{fig:appx_dist_channel_independent} and Fig.~\ref{fig:appx_radar_channel_independent}), Timestamp Embeddings (Fig.~\ref{fig:appx_dist_gym_x_mark} and Fig.~\ref{fig:appx_radar_gym_x_mark}), and Series Tokenization (Fig.~\ref{fig:appx_dist_gym_input_embed} and Fig.~\ref{fig:appx_radar_gym_input_embed}).

\subsubsection{Network Architecture}
\begin{figure*}[htbp]
  \centering
  \begin{minipage}[t]{0.48\textwidth}
    \centering
    \begin{subfigure}[t]{0.5\textwidth}
      \centering
      \includegraphics[width=\textwidth]{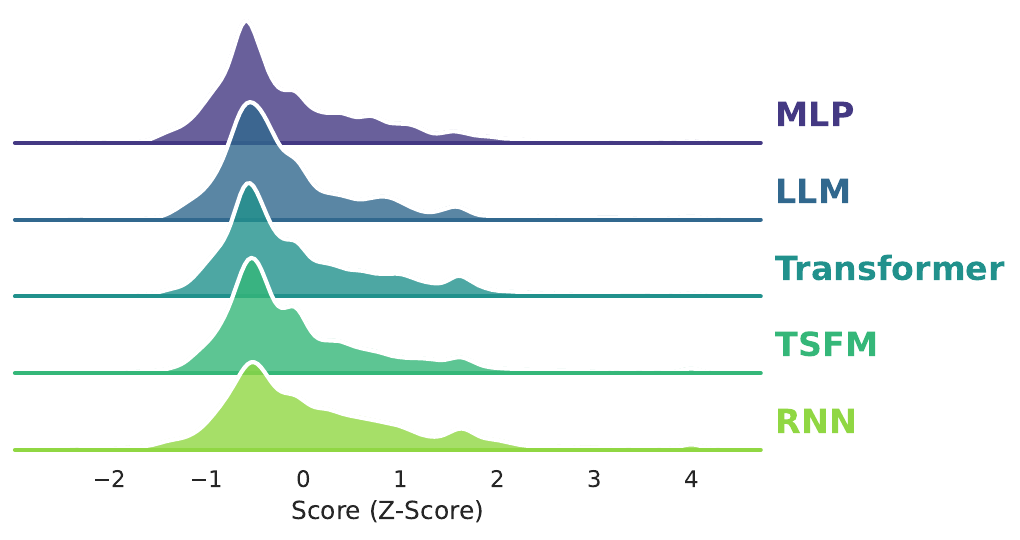}
      \label{fig:appx_dist_network_architecture_Global}
    \end{subfigure}
    \caption{Performance Distributions for Network Backbones (Ridgeline Plots). This figure visualizes the performance distributions across different model architectures.}
    \label{fig:appx_dist_network_architecture}
  \end{minipage}
  \hfill
  \begin{minipage}[t]{0.48\textwidth}
    \centering
    \begin{subfigure}[t]{0.5\textwidth}
      \centering
      \includegraphics[width=\textwidth]{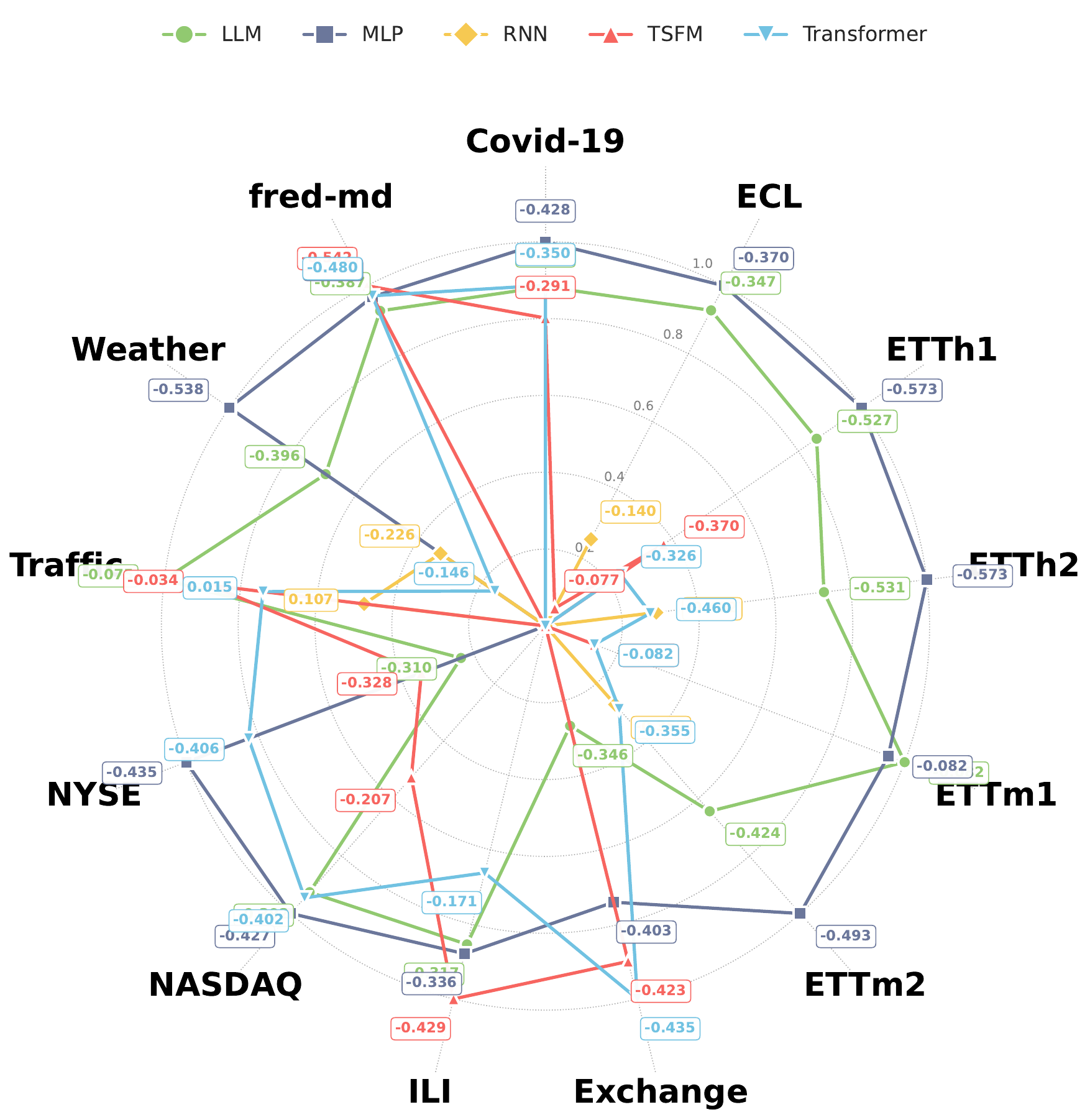}
      \label{fig:appx_radar_network_architecture_Global}
    \end{subfigure}
    \caption{Dataset Adaptability (Radar Charts) for Network Backbones (Radar Plots). This figure visualizes the performance distributions across different model architectures.}
    \label{fig:appx_radar_network_architecture}
  \end{minipage}
\end{figure*}

\begin{figure*}[htbp]
  \centering
  \begin{subfigure}[t]{0.16\textwidth}
    \centering
    \includegraphics[width=\textwidth]{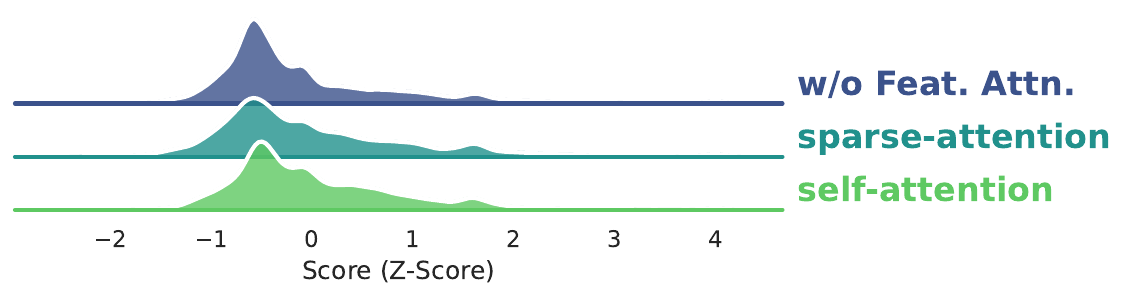}
    \caption{Global}
    \label{fig:appx_dist_feature_attn_Global}
  \end{subfigure}
  \hfill
  \begin{subfigure}[t]{0.16\textwidth}
    \centering
    \includegraphics[width=\textwidth]{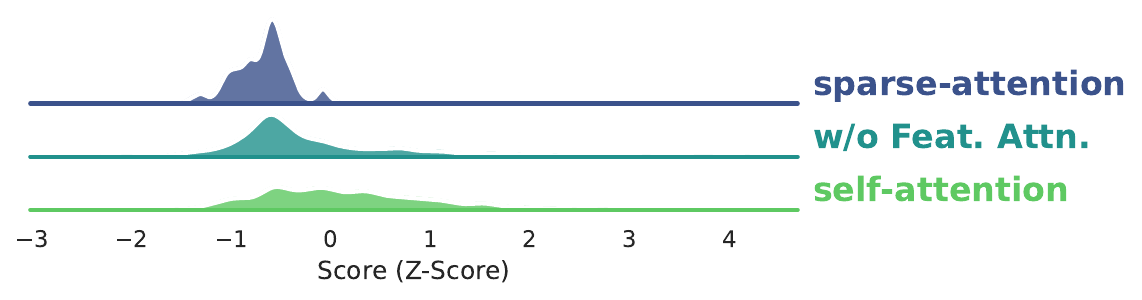}
    \caption{MLP}
    \label{fig:appx_dist_feature_attn_MLP}
  \end{subfigure}
  \hfill
  \begin{subfigure}[t]{0.16\textwidth}
    \centering
    \includegraphics[width=\textwidth]{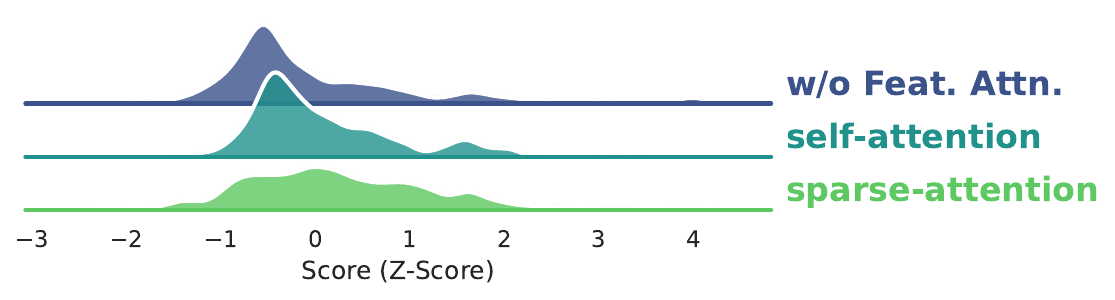}
    \caption{RNN}
    \label{fig:appx_dist_feature_attn_RNN}
  \end{subfigure}
  \begin{subfigure}[t]{0.16\textwidth}
    \centering
    \includegraphics[width=\textwidth]{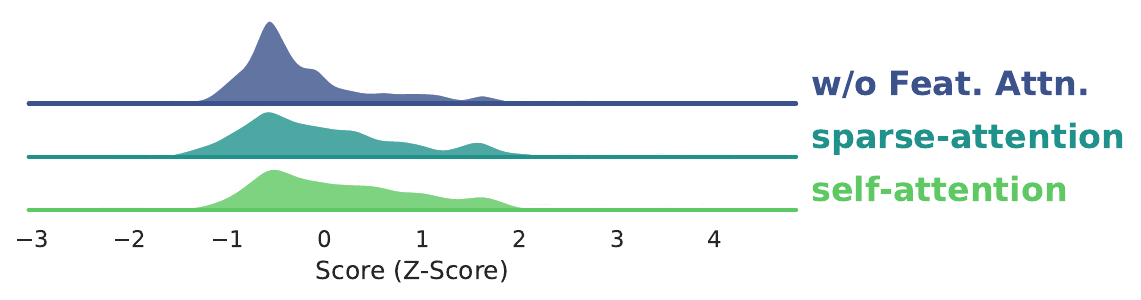}
    \caption{Transformer}
    \label{fig:appx_dist_feature_attn_Transformer}
  \end{subfigure}
  \hfill
  \begin{subfigure}[t]{0.16\textwidth}
    \centering
    \includegraphics[width=\textwidth]{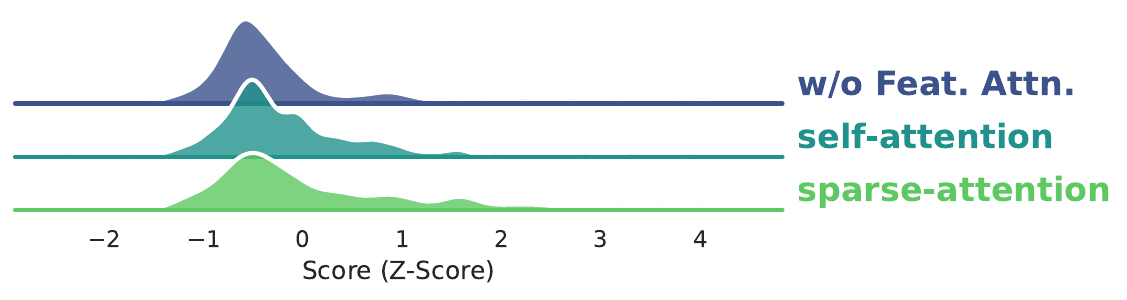}
    \caption{LLM}
    \label{fig:appx_dist_feature_attn_LLM}
  \end{subfigure}
  \hfill
  \begin{subfigure}[t]{0.16\textwidth}
    \centering
    \includegraphics[width=\textwidth]{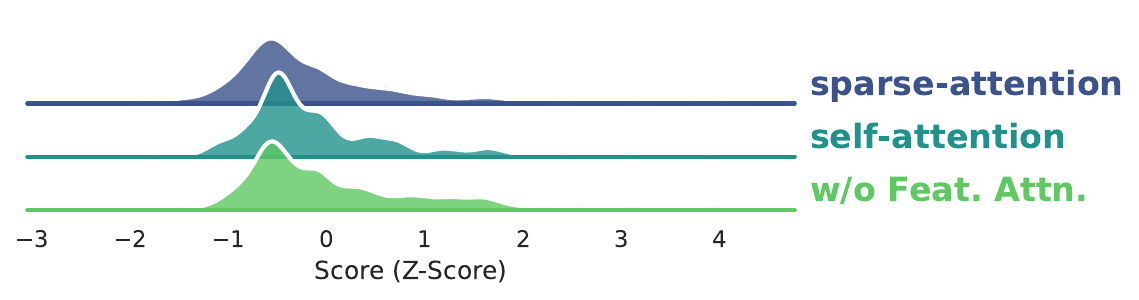}
    \caption{TSFM}
    \label{fig:appx_dist_feature_attn_TSFM}
  \end{subfigure}
  \caption{Performance Distributions for Feature Attention Mechanisms (Ridgeline Plots). This figure visualizes the performance distributions across different model architectures.}
  \label{fig:appx_dist_feature_attn}
\end{figure*}

\begin{figure*}[htbp]
  \centering
  \begin{subfigure}[t]{0.16\textwidth}
    \centering
    \includegraphics[width=\textwidth]{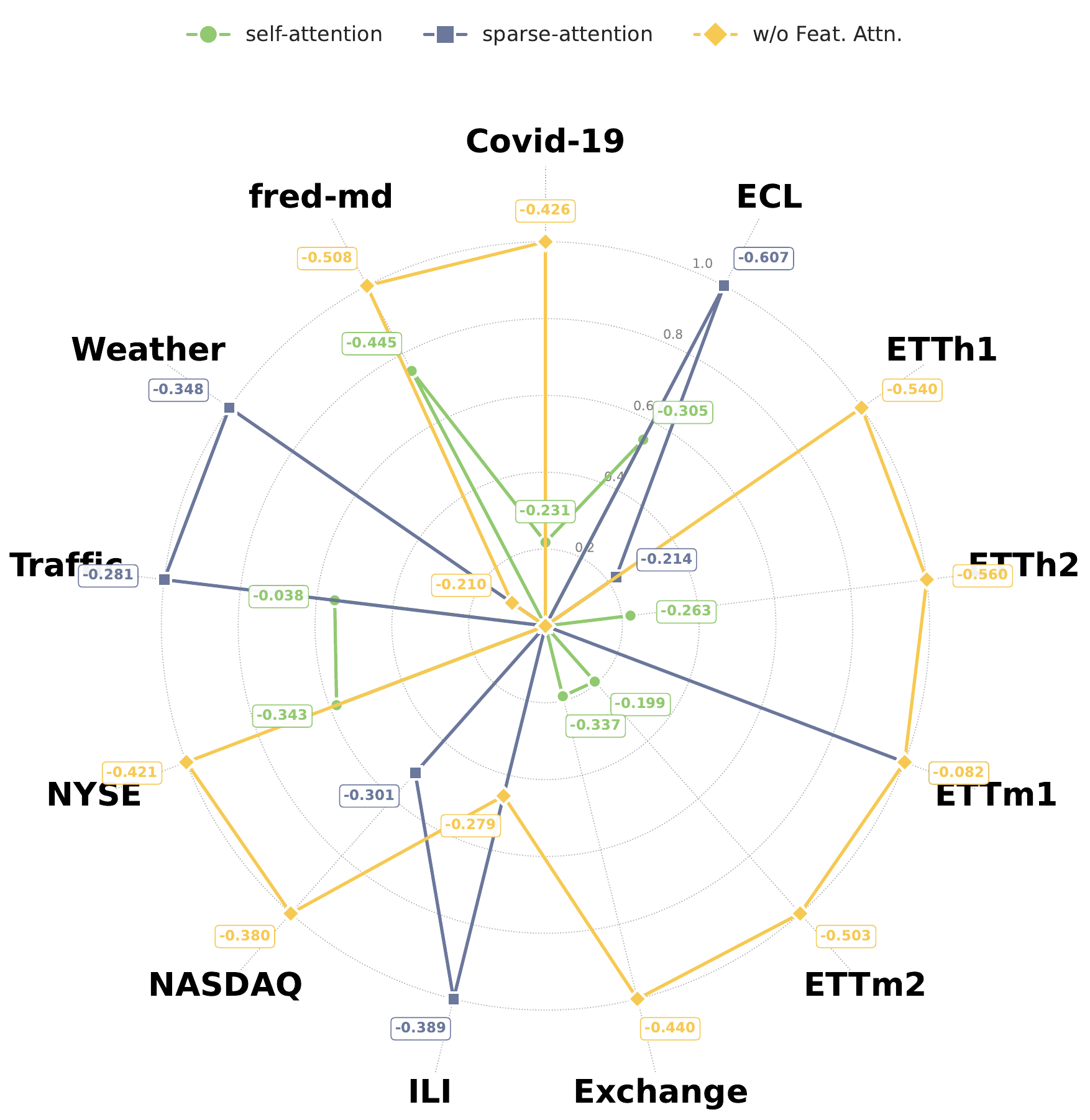}
    \caption{Global}
    \label{fig:appx_radar_feature_attn_Global}
  \end{subfigure}
  \hfill
  \begin{subfigure}[t]{0.16\textwidth}
    \centering
    \includegraphics[width=\textwidth]{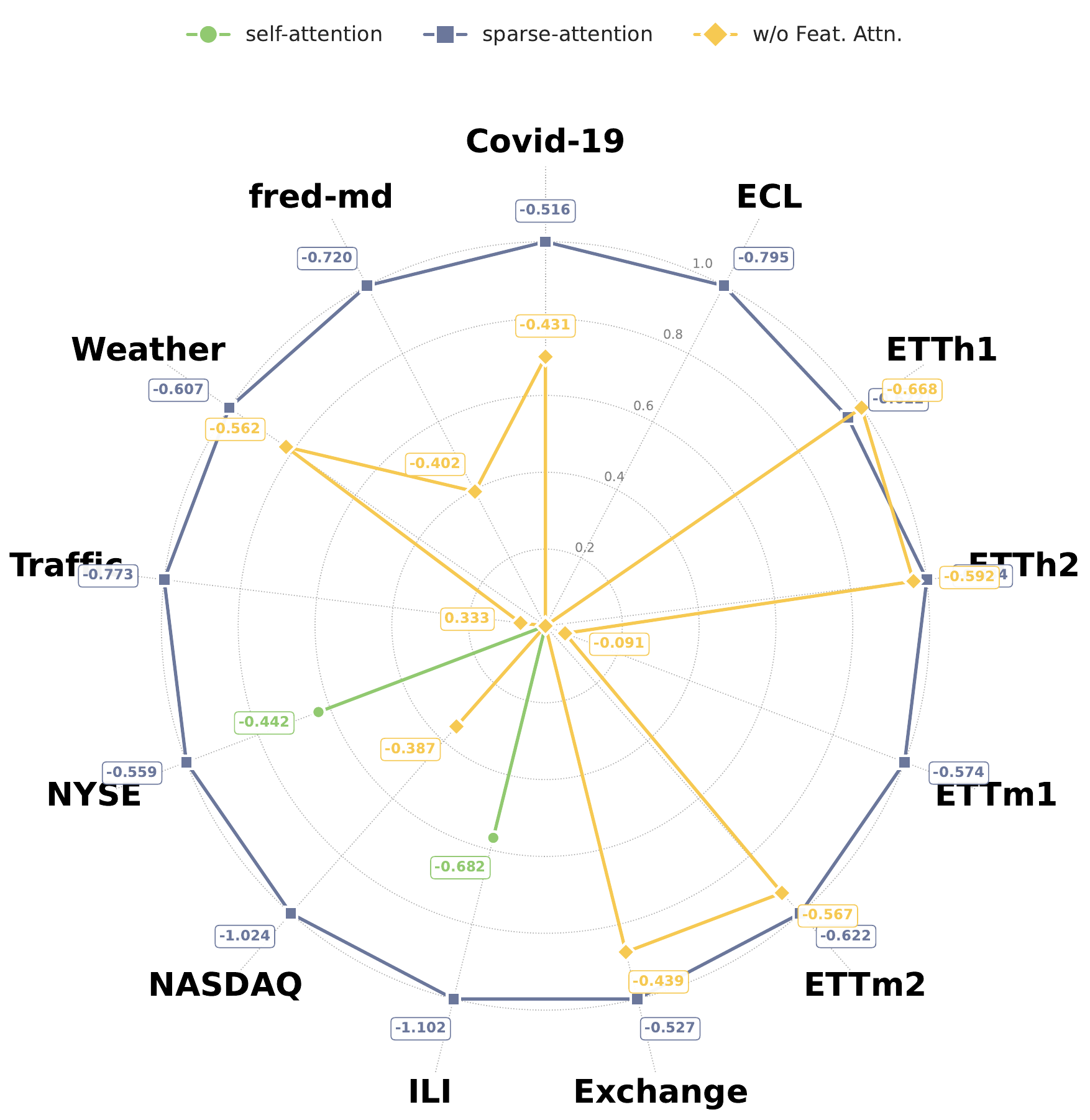}
    \caption{MLP}
    \label{fig:appx_radar_feature_attn_MLP}
  \end{subfigure}
  \hfill
  \begin{subfigure}[t]{0.16\textwidth}
    \centering
    \includegraphics[width=\textwidth]{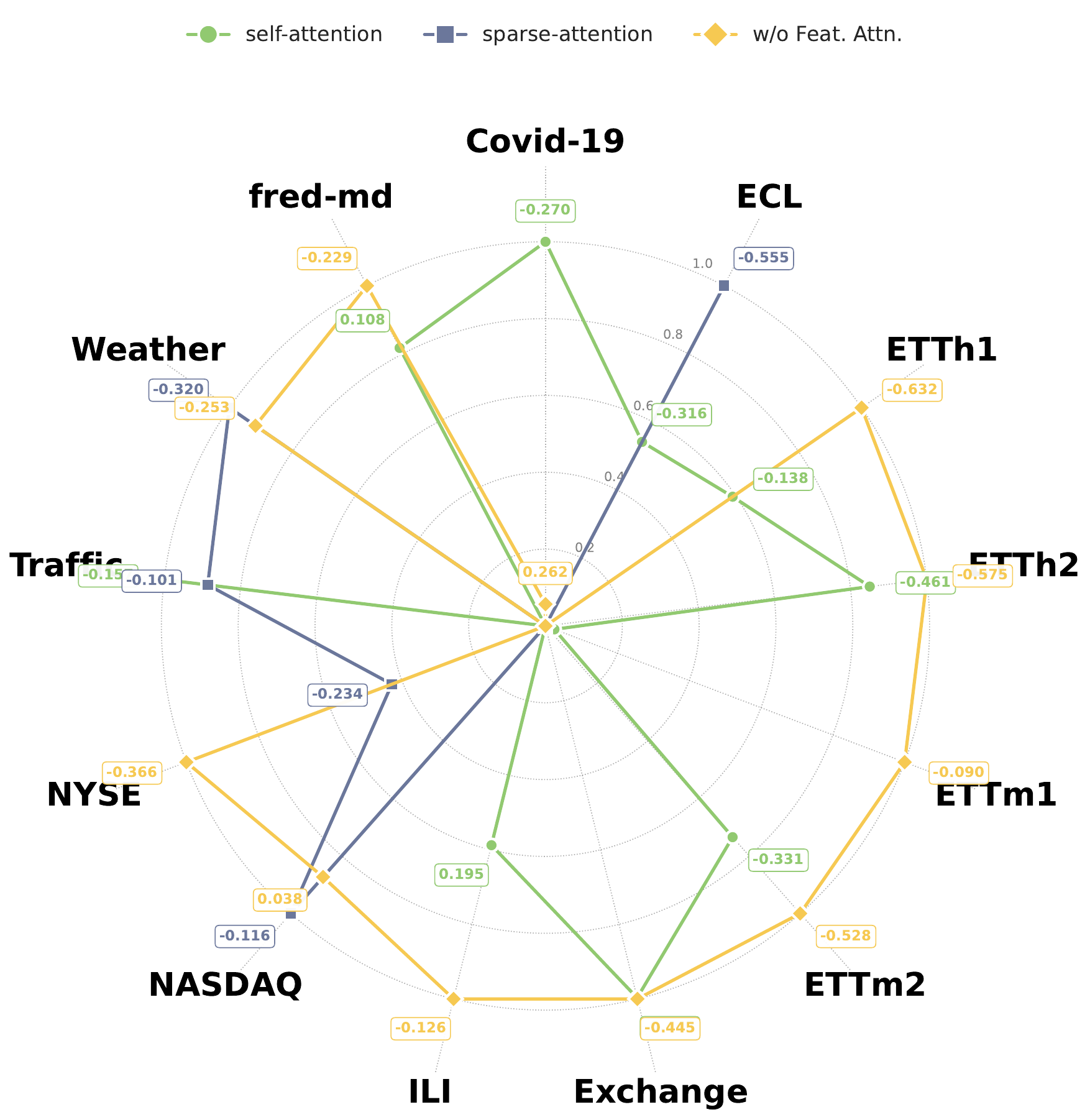}
    \caption{RNN}
    \label{fig:appx_radar_feature_attn_RNN}
  \end{subfigure}
  \begin{subfigure}[t]{0.16\textwidth}
    \centering
    \includegraphics[width=\textwidth]{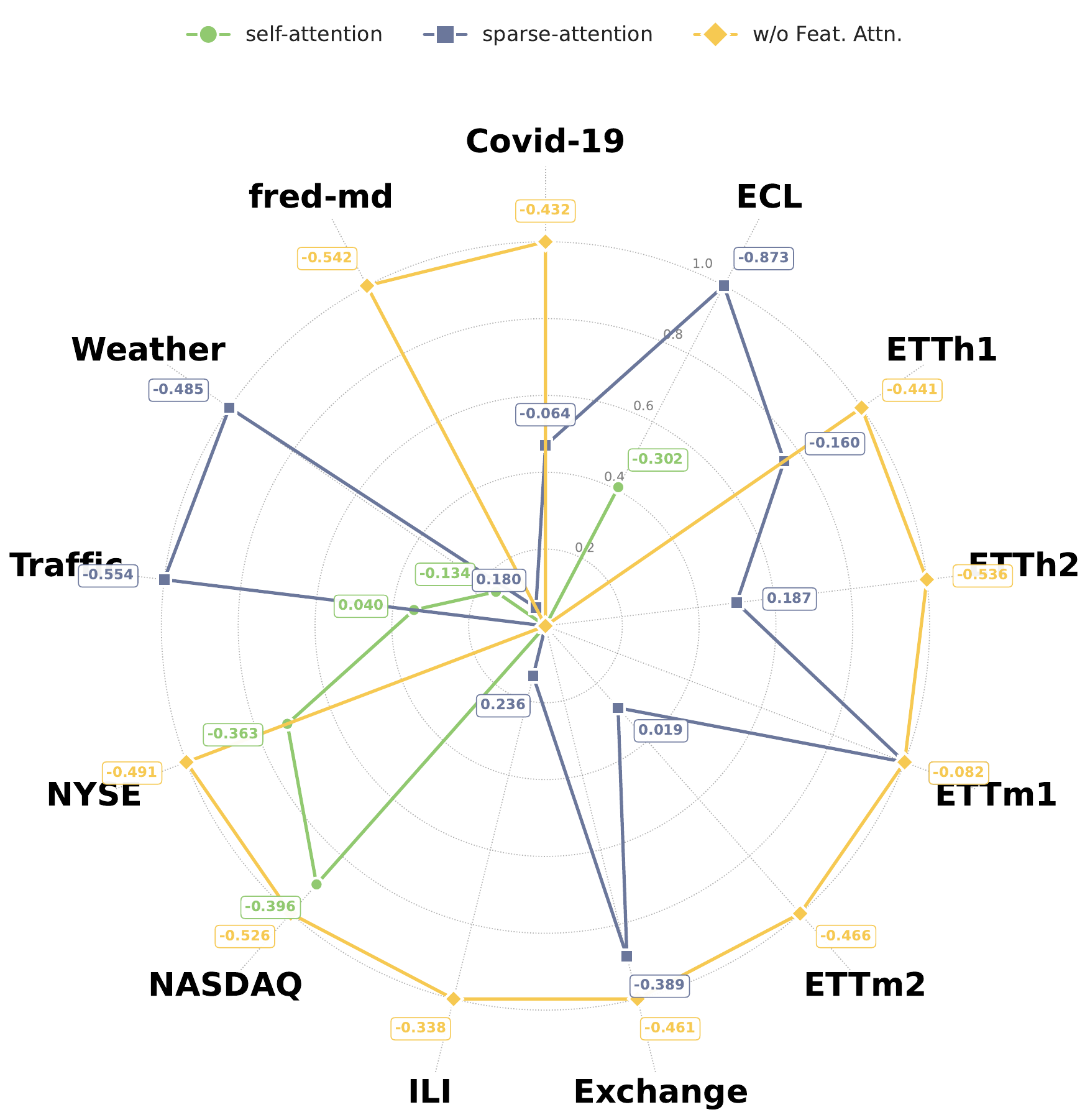}
    \caption{Transformer}
    \label{fig:appx_radar_feature_attn_Transformer}
  \end{subfigure}
  \hfill
  \begin{subfigure}[t]{0.16\textwidth}
    \centering
    \includegraphics[width=\textwidth]{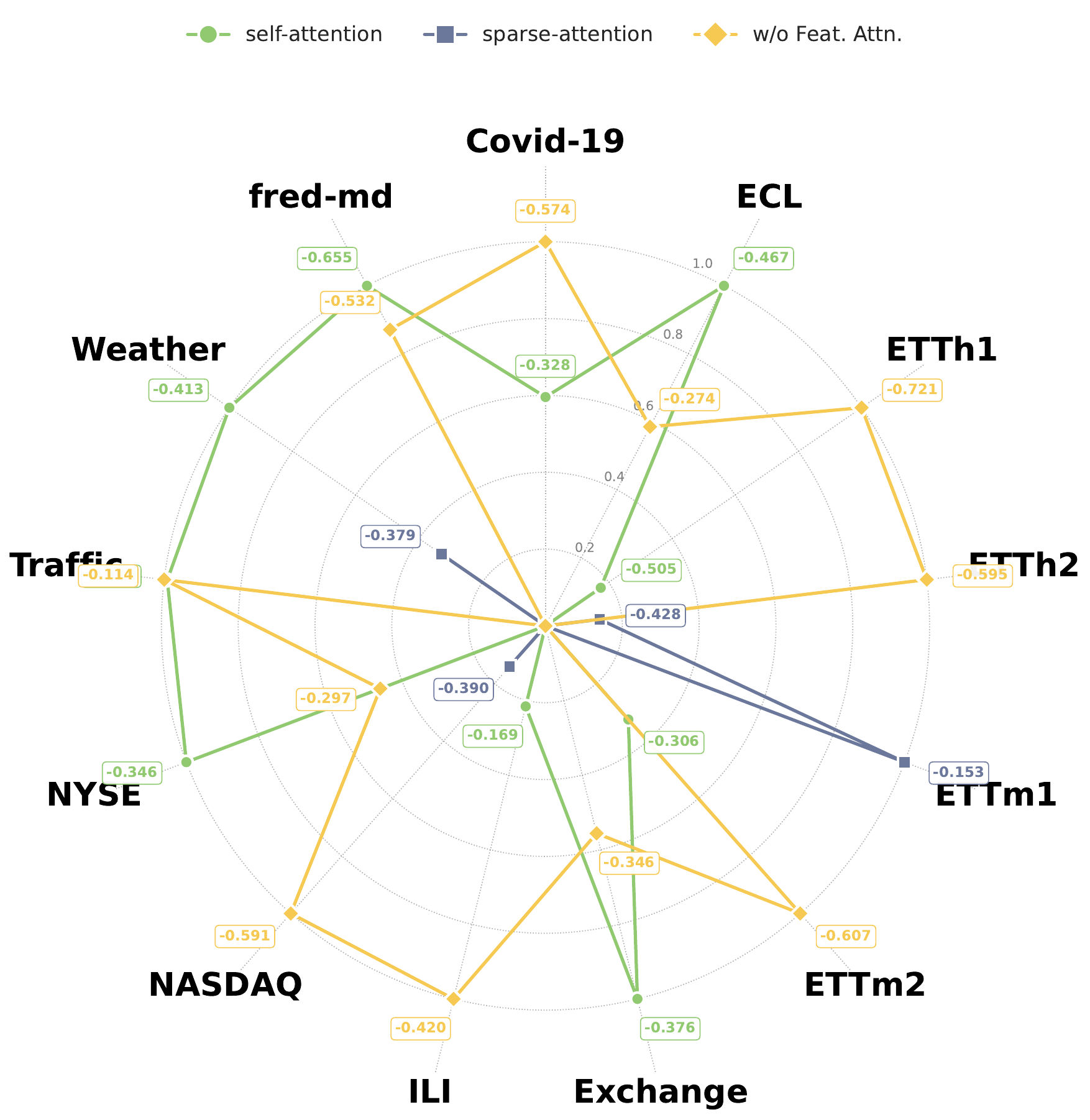}
    \caption{LLM}
    \label{fig:appx_radar_feature_attn_LLM}
  \end{subfigure}
  \hfill
  \begin{subfigure}[t]{0.16\textwidth}
    \centering
    \includegraphics[width=\textwidth]{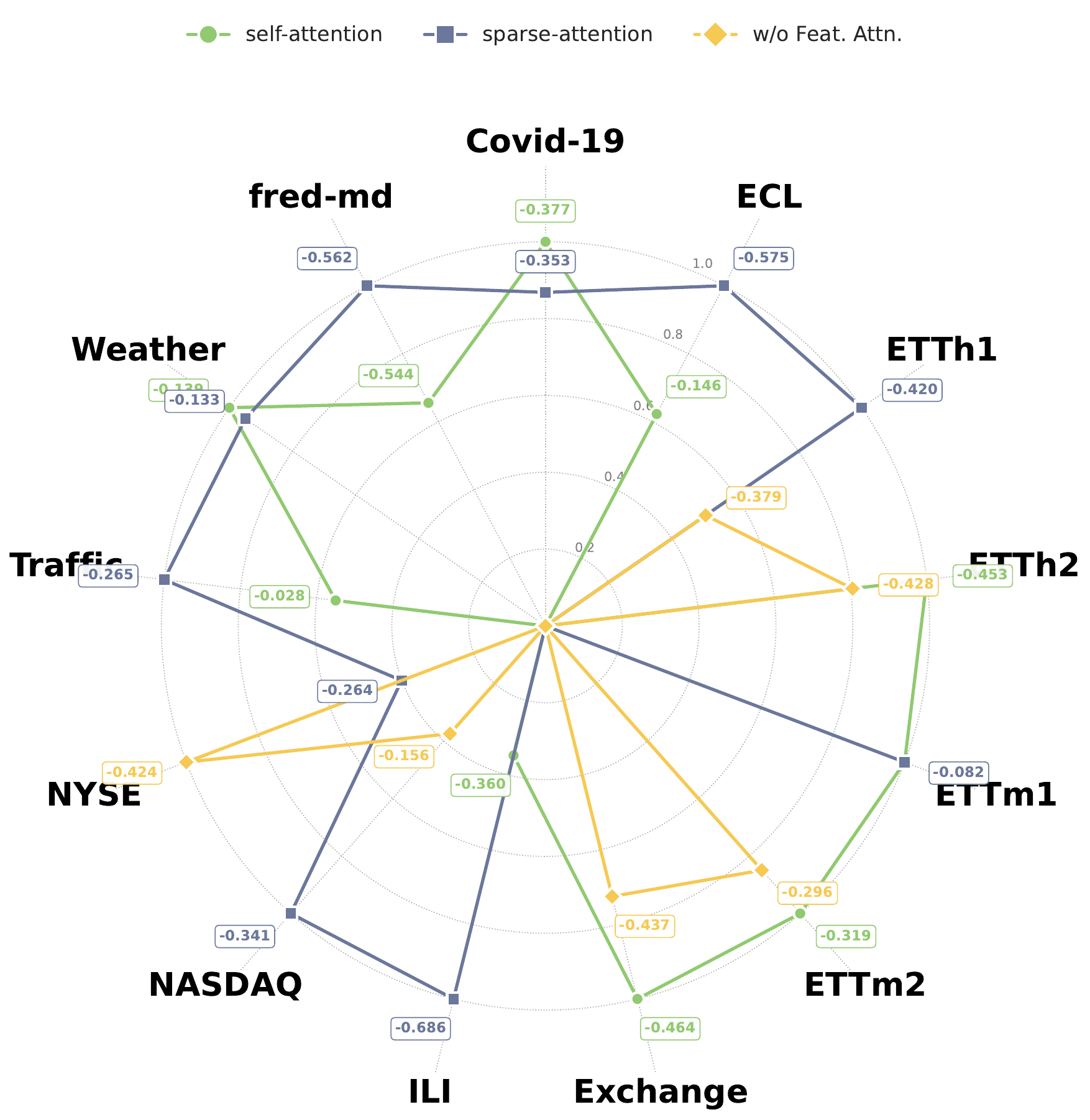}
    \caption{TSFM}
    \label{fig:appx_radar_feature_attn_TSFM}
  \end{subfigure}
  \caption{Dataset Adaptability (Radar Charts) for Feature Attention Mechanisms (Radar Plots). This figure visualizes the performance distributions across different model architectures.}
  \label{fig:appx_radar_feature_attn}
\end{figure*}

\begin{figure*}[htbp]
  \centering
  \begin{subfigure}[t]{0.16\textwidth}
    \centering
    \includegraphics[width=\textwidth]{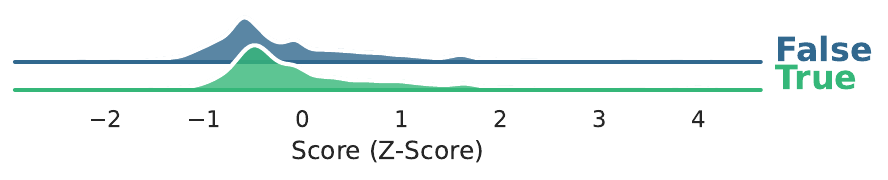}
    \caption{Global}
    \label{fig:appx_dist_gym_rag_Global}
  \end{subfigure}
  \hfill
  \begin{subfigure}[t]{0.16\textwidth}
    \centering
    \includegraphics[width=\textwidth]{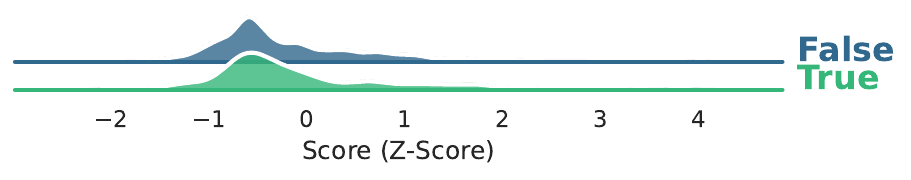}
    \caption{MLP}
    \label{fig:appx_dist_gym_rag_MLP}
  \end{subfigure}
  \hfill
  \begin{subfigure}[t]{0.16\textwidth}
    \centering
    \includegraphics[width=\textwidth]{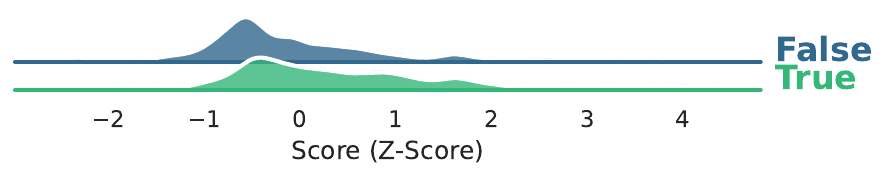}
    \caption{RNN}
    \label{fig:appx_dist_gym_rag_RNN}
  \end{subfigure}
  \begin{subfigure}[t]{0.16\textwidth}
    \centering
    \includegraphics[width=\textwidth]{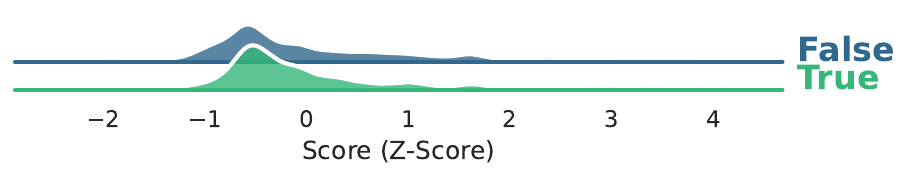}
    \caption{Transformer}
    \label{fig:appx_dist_gym_rag_Transformer}
  \end{subfigure}
  \hfill
  \begin{subfigure}[t]{0.16\textwidth}
    \centering
    \includegraphics[width=\textwidth]{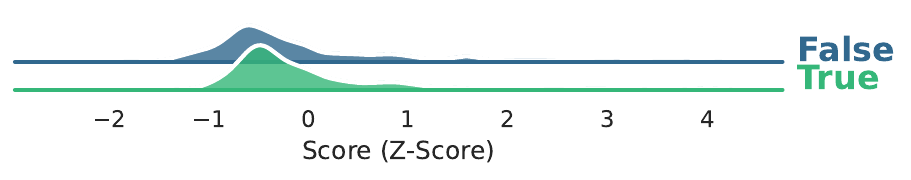}
    \caption{LLM}
    \label{fig:appx_dist_gym_rag_LLM}
  \end{subfigure}
  \hfill
  \begin{subfigure}[t]{0.16\textwidth}
    \centering
    \includegraphics[width=\textwidth]{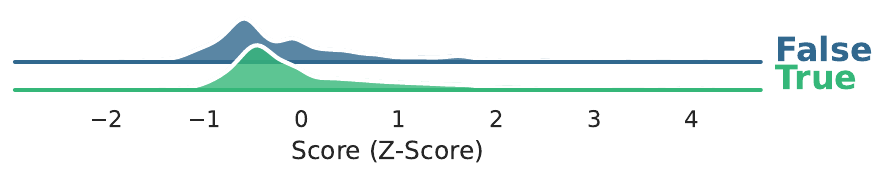}
    \caption{TSFM}
    \label{fig:appx_dist_gym_rag_TSFM}
  \end{subfigure}
  \caption{Performance Distributions for Retrieval Augmented Generation (Ridgeline Plots). This figure visualizes the performance distributions across different model architectures.}
  \label{fig:appx_dist_gym_rag}
\end{figure*}

\begin{figure*}[htbp]
  \centering
  \begin{subfigure}[t]{0.16\textwidth}
    \centering
    \includegraphics[width=\textwidth]{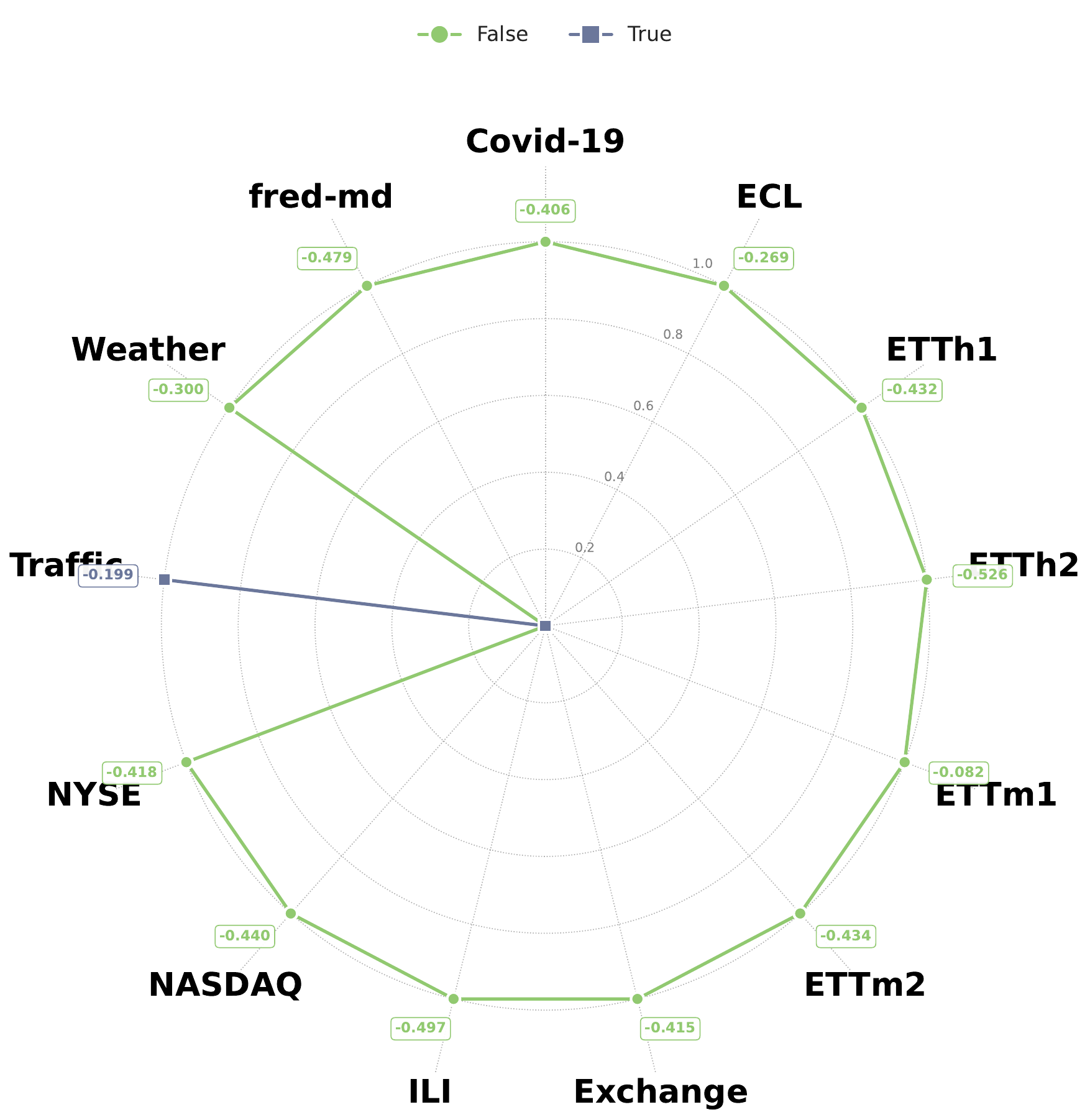}
    \caption{Global}
    \label{fig:appx_radar_gym_rag_Global}
  \end{subfigure}
  \hfill
  \begin{subfigure}[t]{0.16\textwidth}
    \centering
    \includegraphics[width=\textwidth]{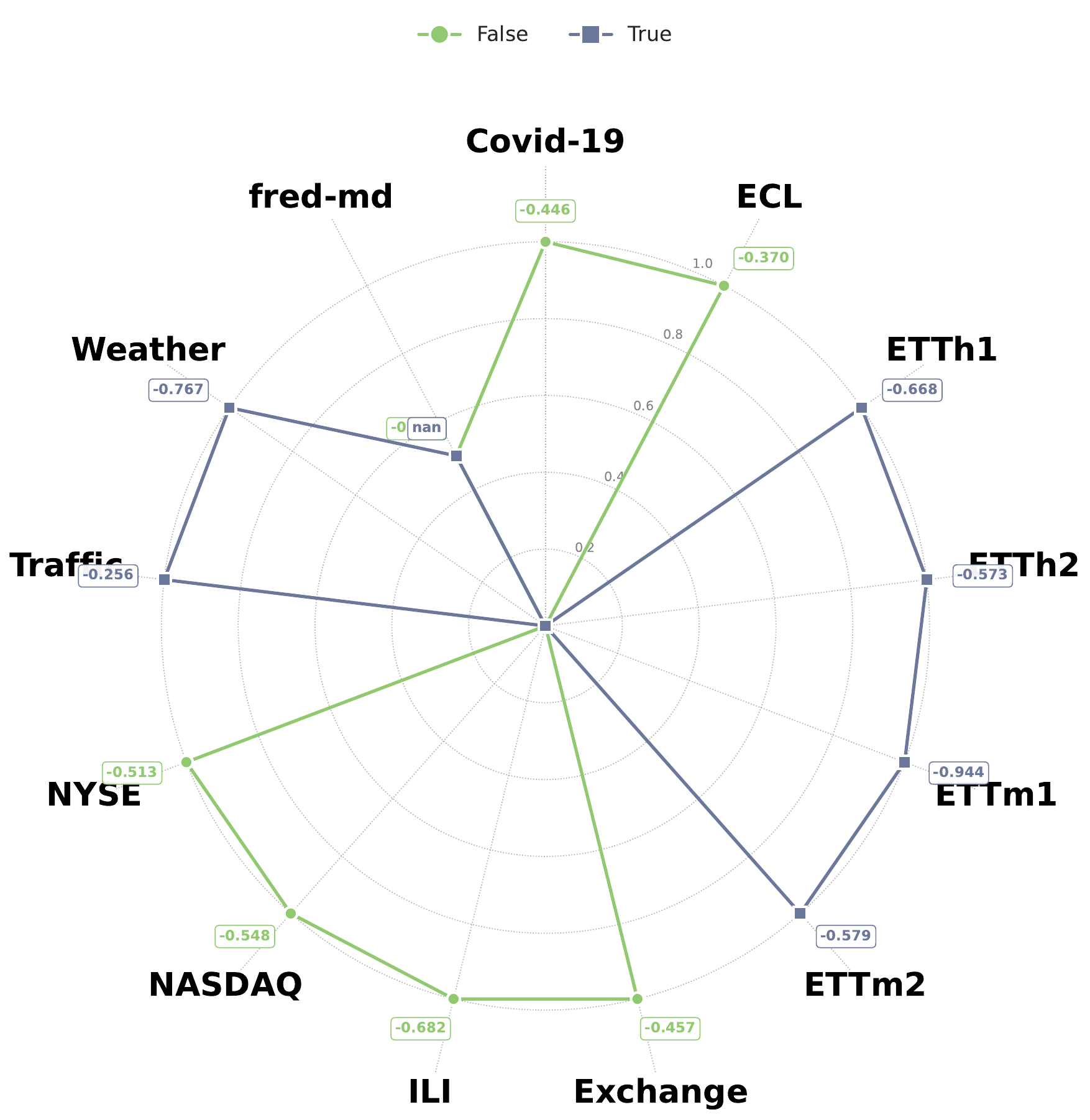}
    \caption{MLP}
    \label{fig:appx_radar_gym_rag_MLP}
  \end{subfigure}
  \hfill
  \begin{subfigure}[t]{0.16\textwidth}
    \centering
    \includegraphics[width=\textwidth]{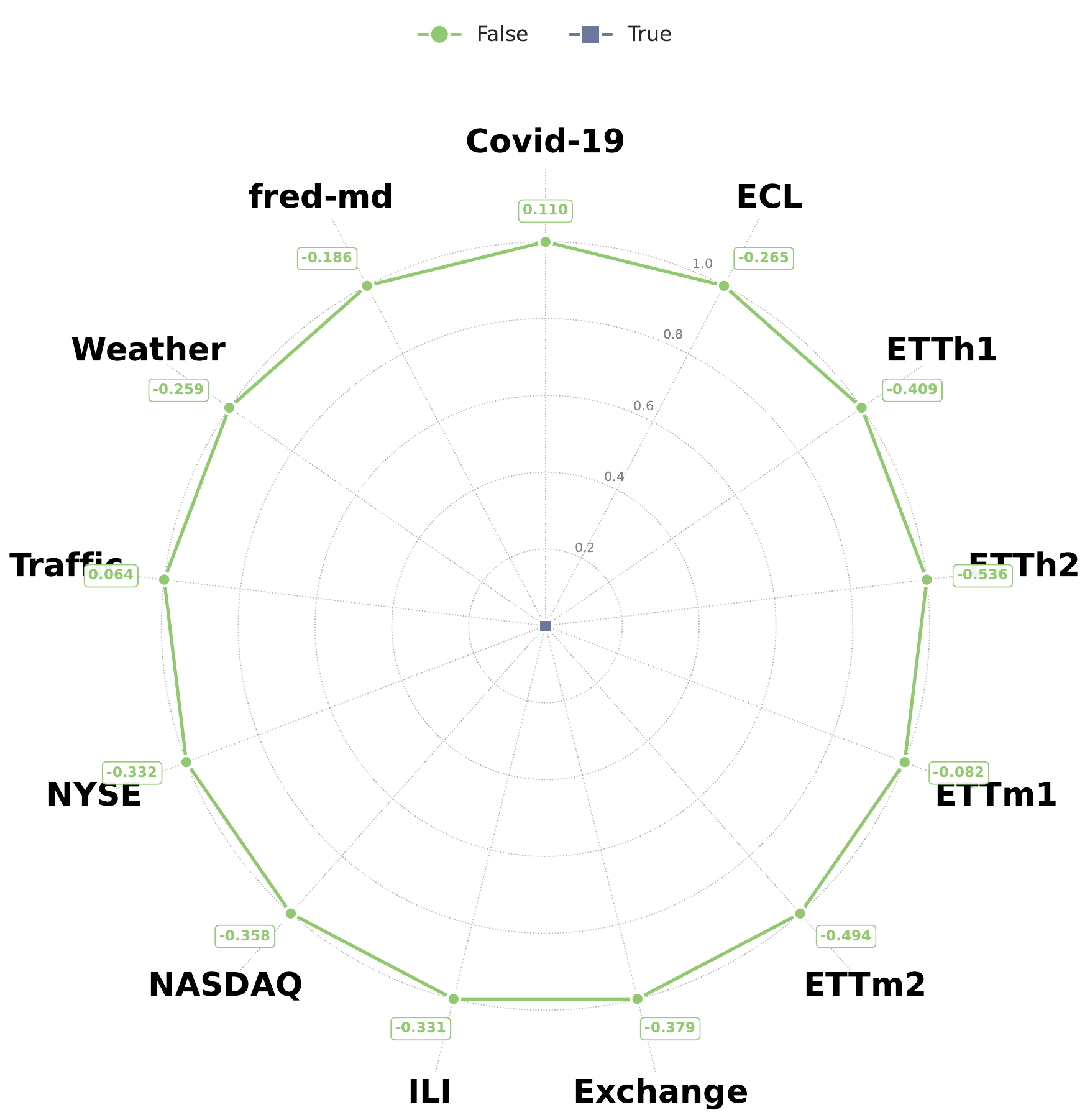}
    \caption{RNN}
    \label{fig:appx_radar_gym_rag_RNN}
  \end{subfigure}
  \begin{subfigure}[t]{0.16\textwidth}
    \centering
    \includegraphics[width=\textwidth]{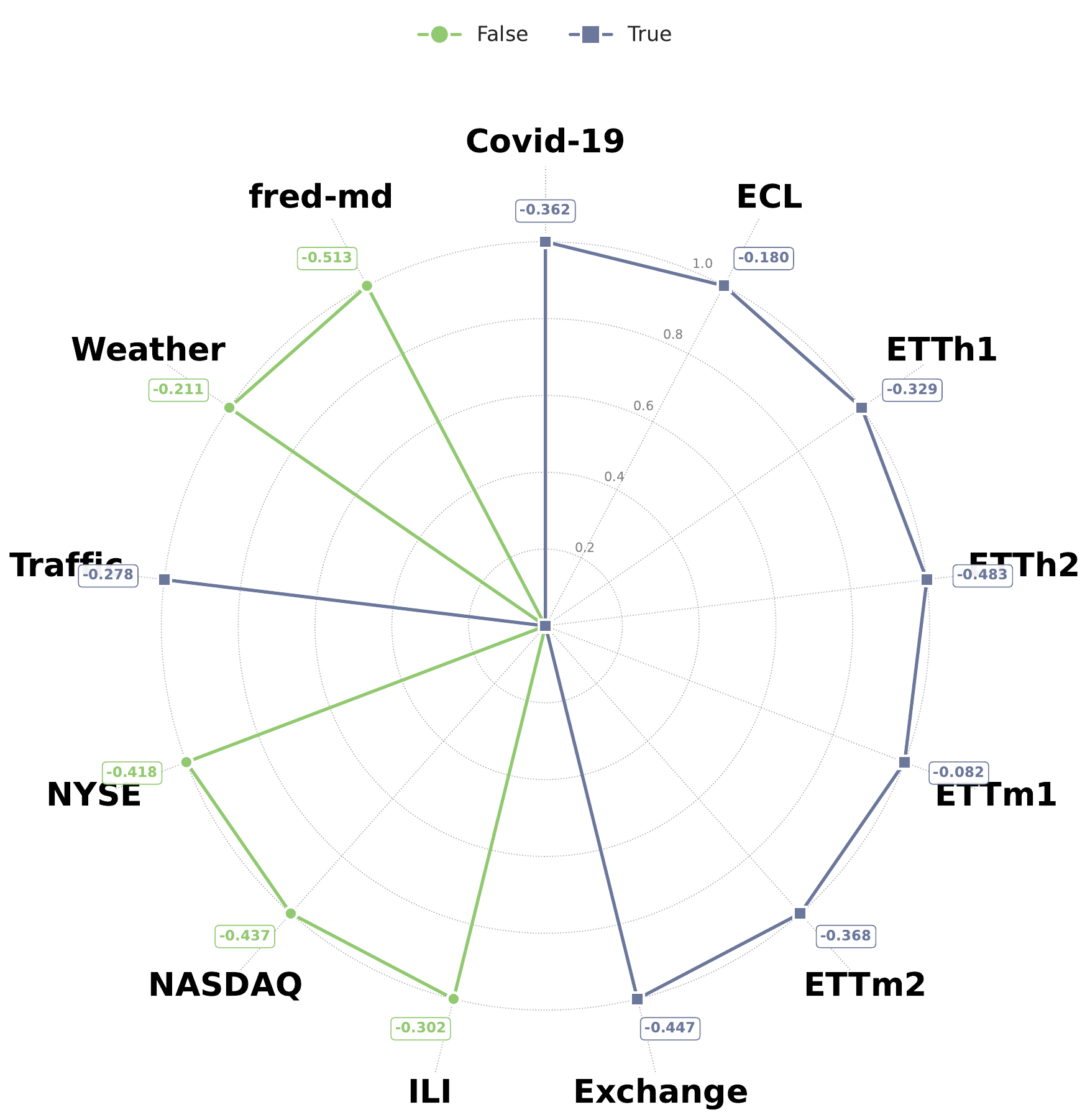}
    \caption{Transformer}
    \label{fig:appx_radar_gym_rag_Transformer}
  \end{subfigure}
  \hfill
  \begin{subfigure}[t]{0.16\textwidth}
    \centering
    \includegraphics[width=\textwidth]{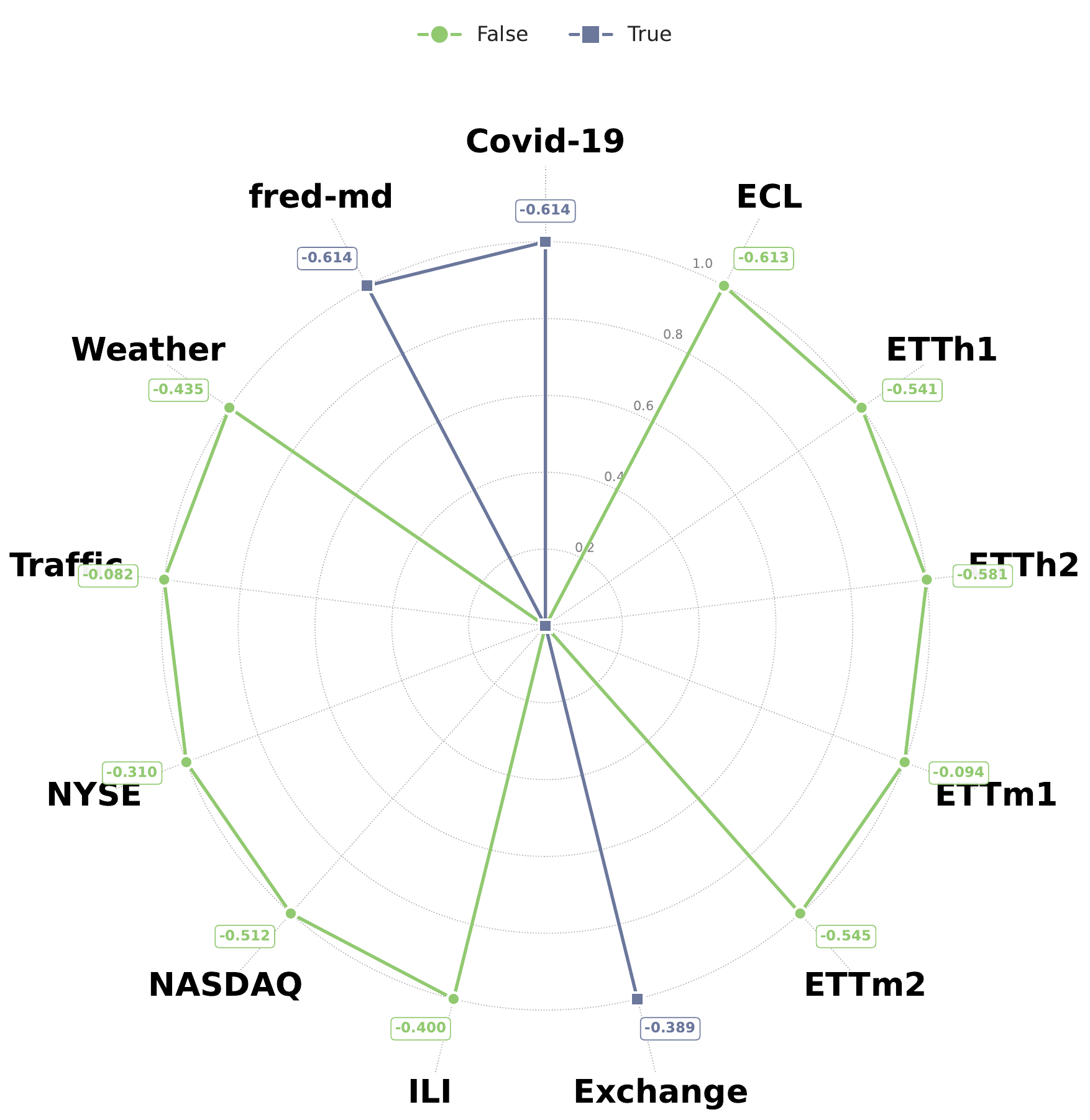}
    \caption{LLM}
    \label{fig:appx_radar_gym_rag_LLM}
  \end{subfigure}
  \hfill
  \begin{subfigure}[t]{0.16\textwidth}
    \centering
    \includegraphics[width=\textwidth]{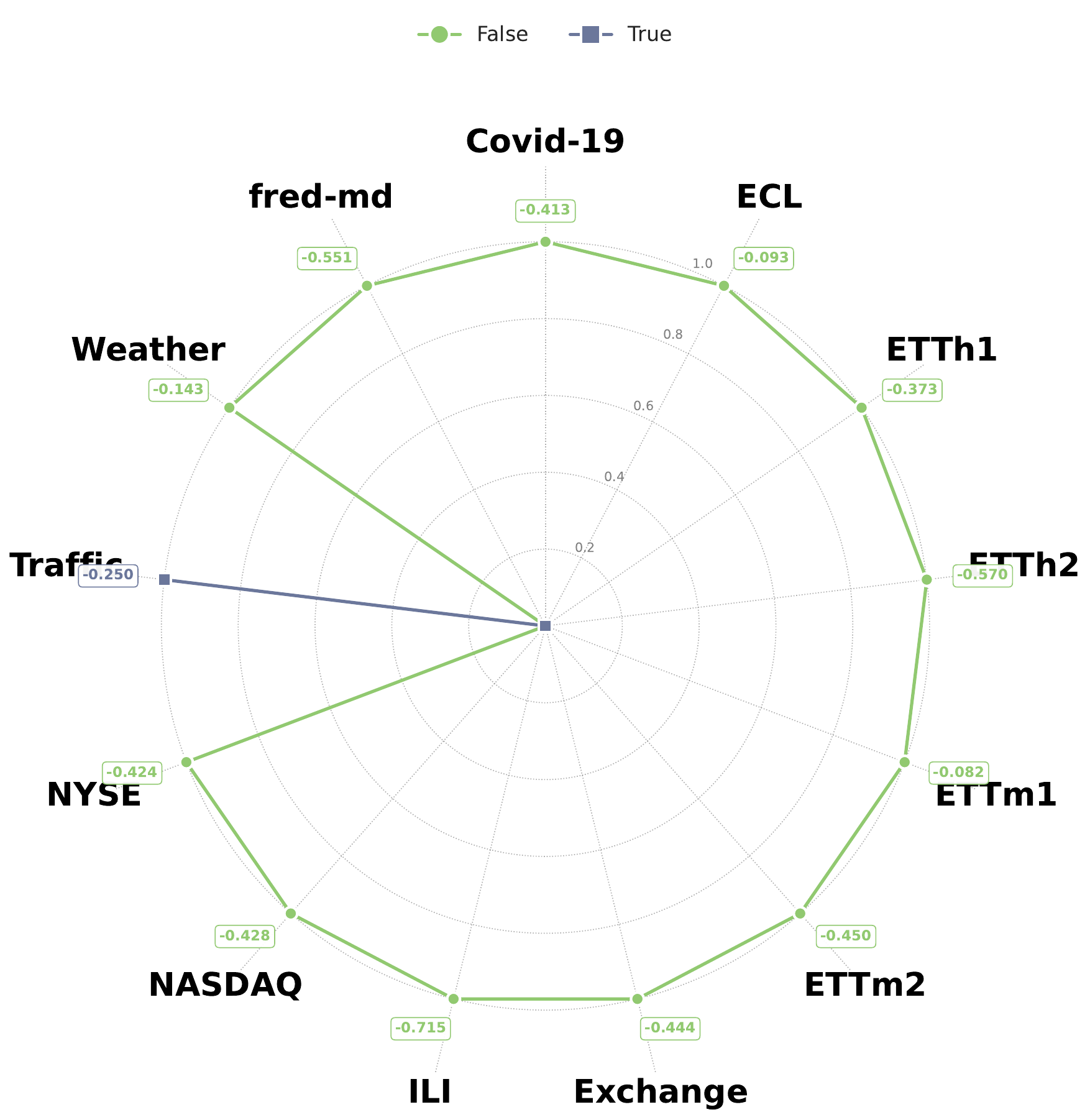}
    \caption{TSFM}
    \label{fig:appx_radar_gym_rag_TSFM}
  \end{subfigure}
  \caption{Dataset Adaptability (Radar Charts) for Retrieval Augmented Generation (Radar Plots). This figure visualizes the performance distributions across different model architectures.}
  \label{fig:appx_radar_gym_rag}
\end{figure*}

We analyze Network Backbones (Fig.~\ref{fig:appx_dist_network_architecture} and Fig.~\ref{fig:appx_radar_network_architecture}), Feature Attention Mechanisms (Fig.~\ref{fig:appx_dist_feature_attn} and Fig.~\ref{fig:appx_radar_feature_attn}), and Retrieval Augmented Generation (Fig.~\ref{fig:appx_dist_gym_rag} and Fig.~\ref{fig:appx_radar_gym_rag}).

\subsubsection{Network Optimization}
\begin{figure*}[htbp]
  \centering
  \begin{subfigure}[t]{0.16\textwidth}
    \centering
    \includegraphics[width=\textwidth]{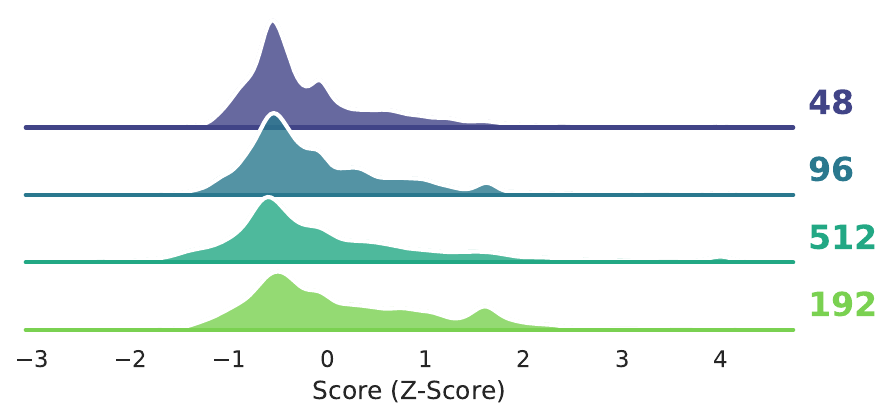}
    \caption{Global}
    \label{fig:appx_dist_seq_len_Global}
  \end{subfigure}
  \hfill
  \begin{subfigure}[t]{0.16\textwidth}
    \centering
    \includegraphics[width=\textwidth]{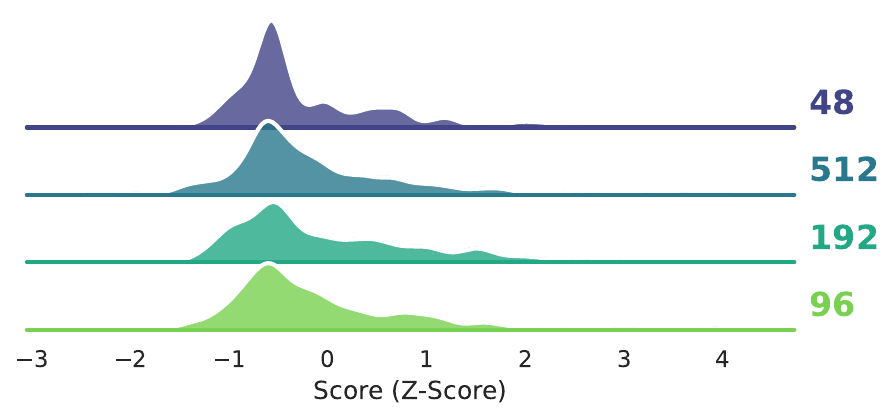}
    \caption{MLP}
    \label{fig:appx_dist_seq_len_MLP}
  \end{subfigure}
  \hfill
  \begin{subfigure}[t]{0.16\textwidth}
    \centering
    \includegraphics[width=\textwidth]{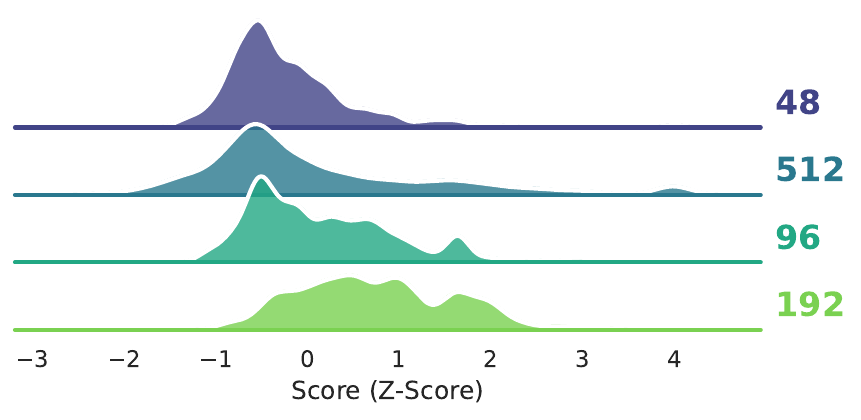}
    \caption{RNN}
    \label{fig:appx_dist_seq_len_RNN}
  \end{subfigure}
  \begin{subfigure}[t]{0.16\textwidth}
    \centering
    \includegraphics[width=\textwidth]{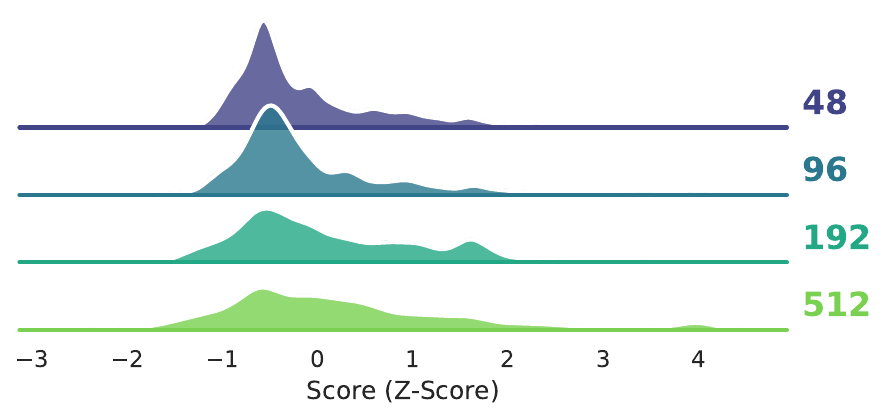}
    \caption{Transformer}
    \label{fig:appx_dist_seq_len_Transformer}
  \end{subfigure}
  \hfill
  \begin{subfigure}[t]{0.16\textwidth}
    \centering
    \includegraphics[width=\textwidth]{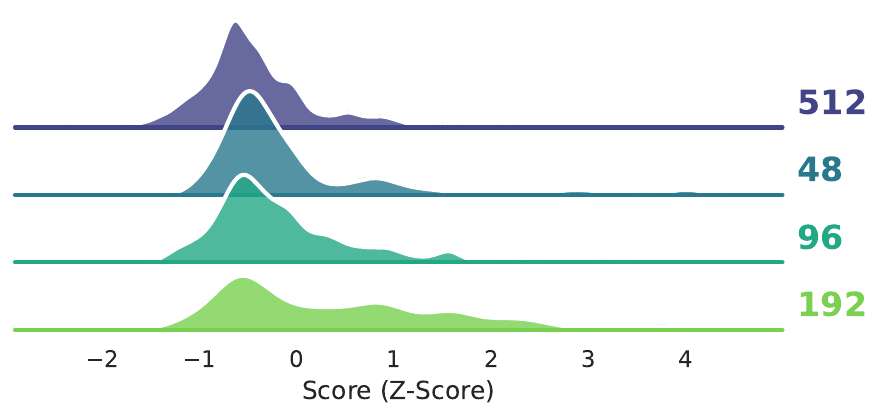}
    \caption{LLM}
    \label{fig:appx_dist_seq_len_LLM}
  \end{subfigure}
  \hfill
  \begin{subfigure}[t]{0.16\textwidth}
    \centering
    \includegraphics[width=\textwidth]{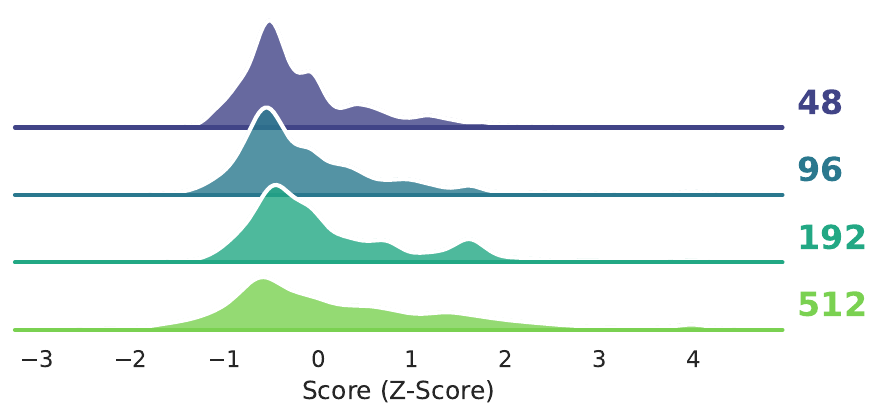}
    \caption{TSFM}
    \label{fig:appx_dist_seq_len_TSFM}
  \end{subfigure}
  \caption{Performance Distributions for Sequence Length Configurations (Ridgeline Plots). This figure visualizes the performance distributions across different model architectures.}
  \label{fig:appx_dist_seq_len}
\end{figure*}

\begin{figure*}[htbp]
  \centering
  \begin{subfigure}[t]{0.16\textwidth}
    \centering
    \includegraphics[width=\textwidth]{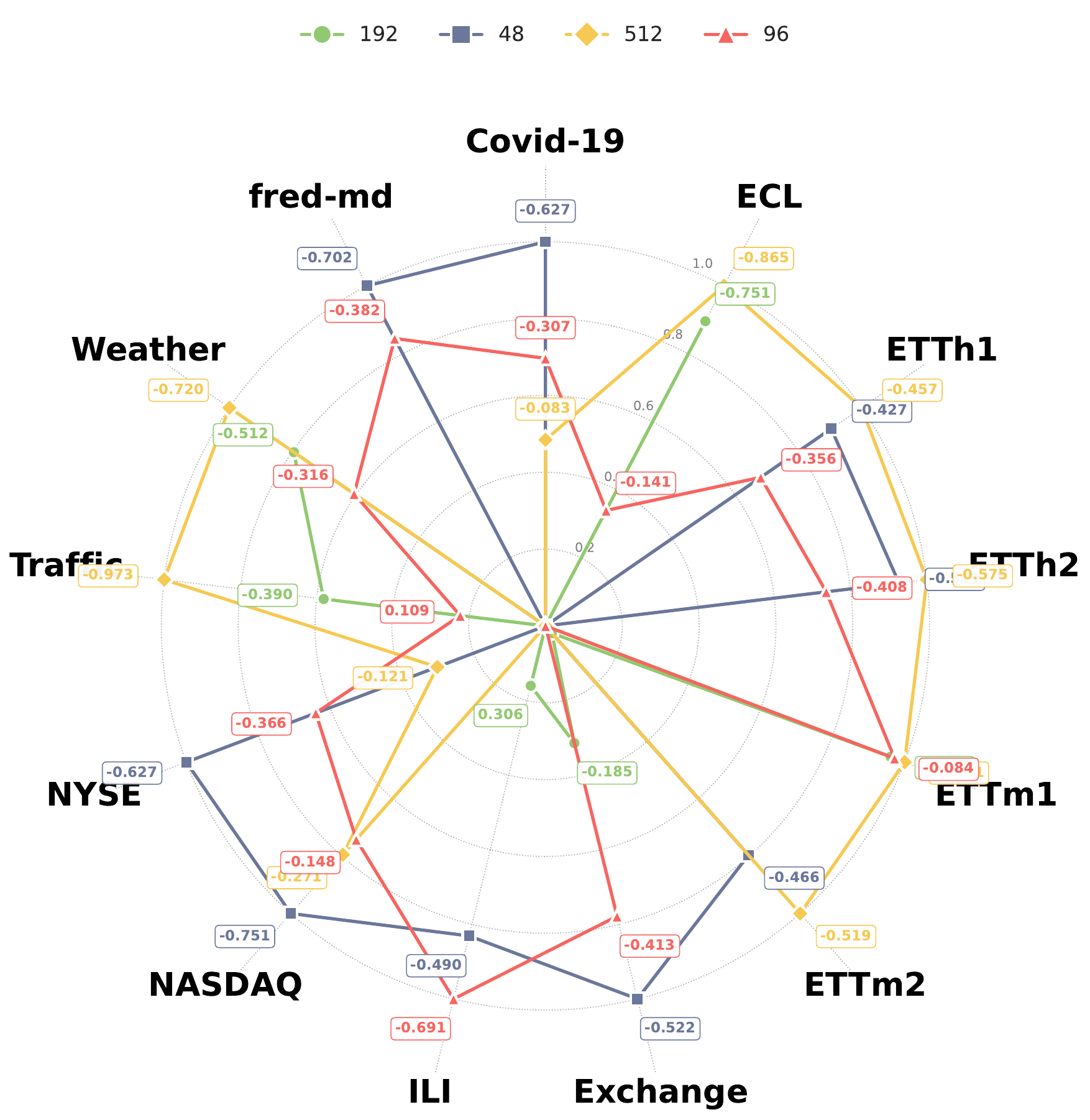}
    \caption{Global}
    \label{fig:appx_radar_seq_len_Global}
  \end{subfigure}
  \hfill
  \begin{subfigure}[t]{0.16\textwidth}
    \centering
    \includegraphics[width=\textwidth]{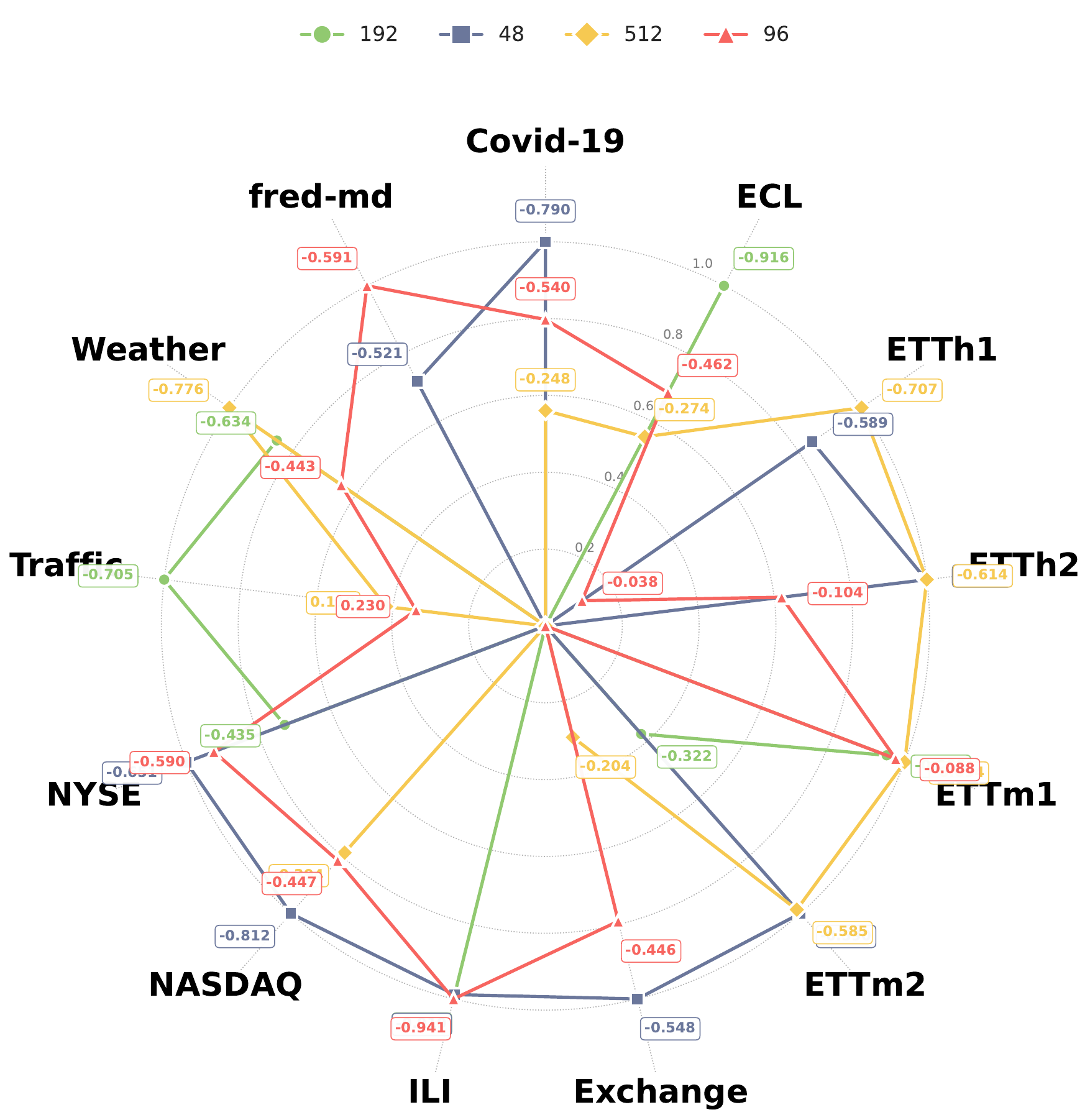}
    \caption{MLP}
    \label{fig:appx_radar_seq_len_MLP}
  \end{subfigure}
  \hfill
  \begin{subfigure}[t]{0.16\textwidth}
    \centering
    \includegraphics[width=\textwidth]{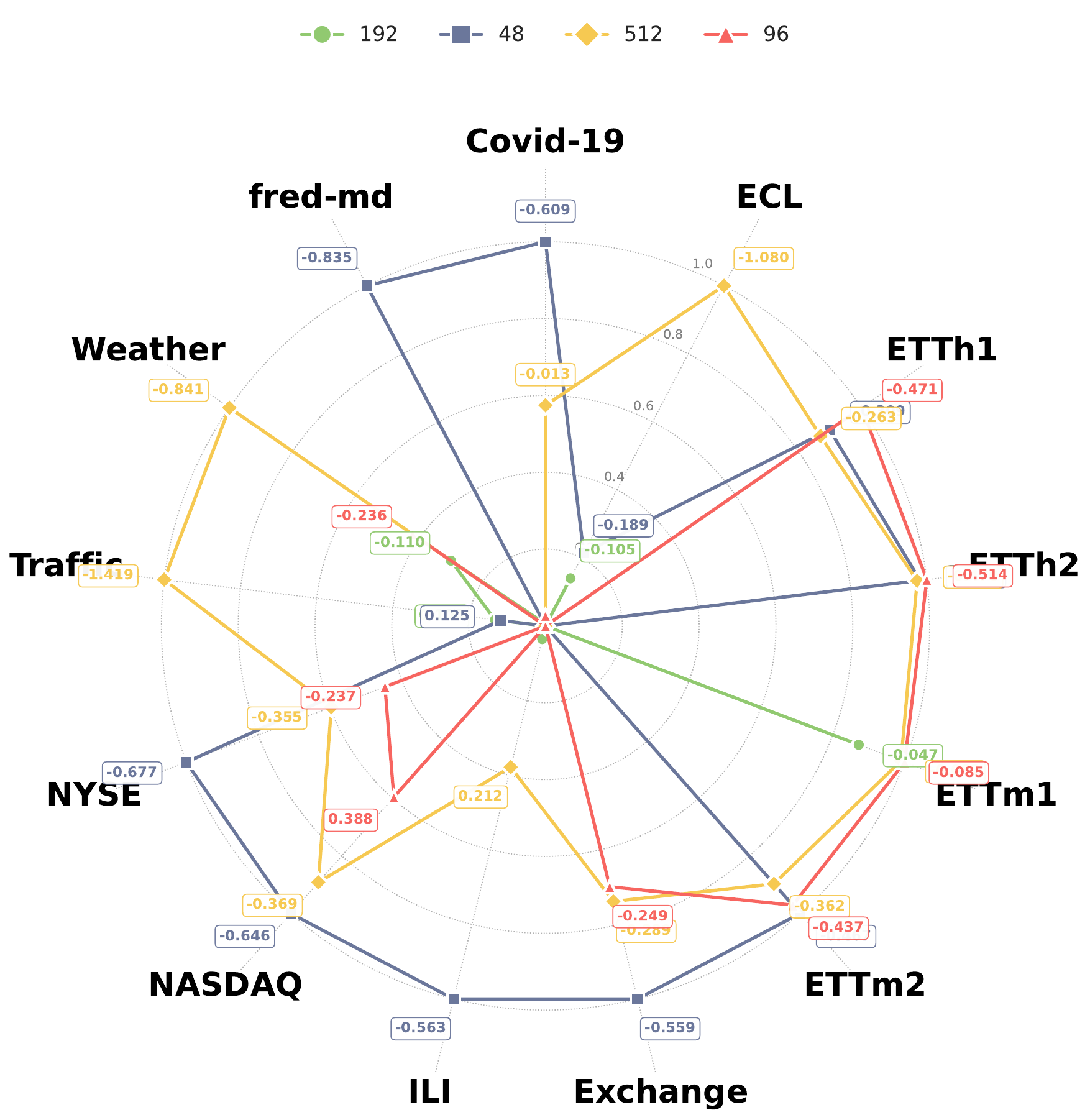}
    \caption{RNN}
    \label{fig:appx_radar_seq_len_RNN}
  \end{subfigure}
  \begin{subfigure}[t]{0.16\textwidth}
    \centering
    \includegraphics[width=\textwidth]{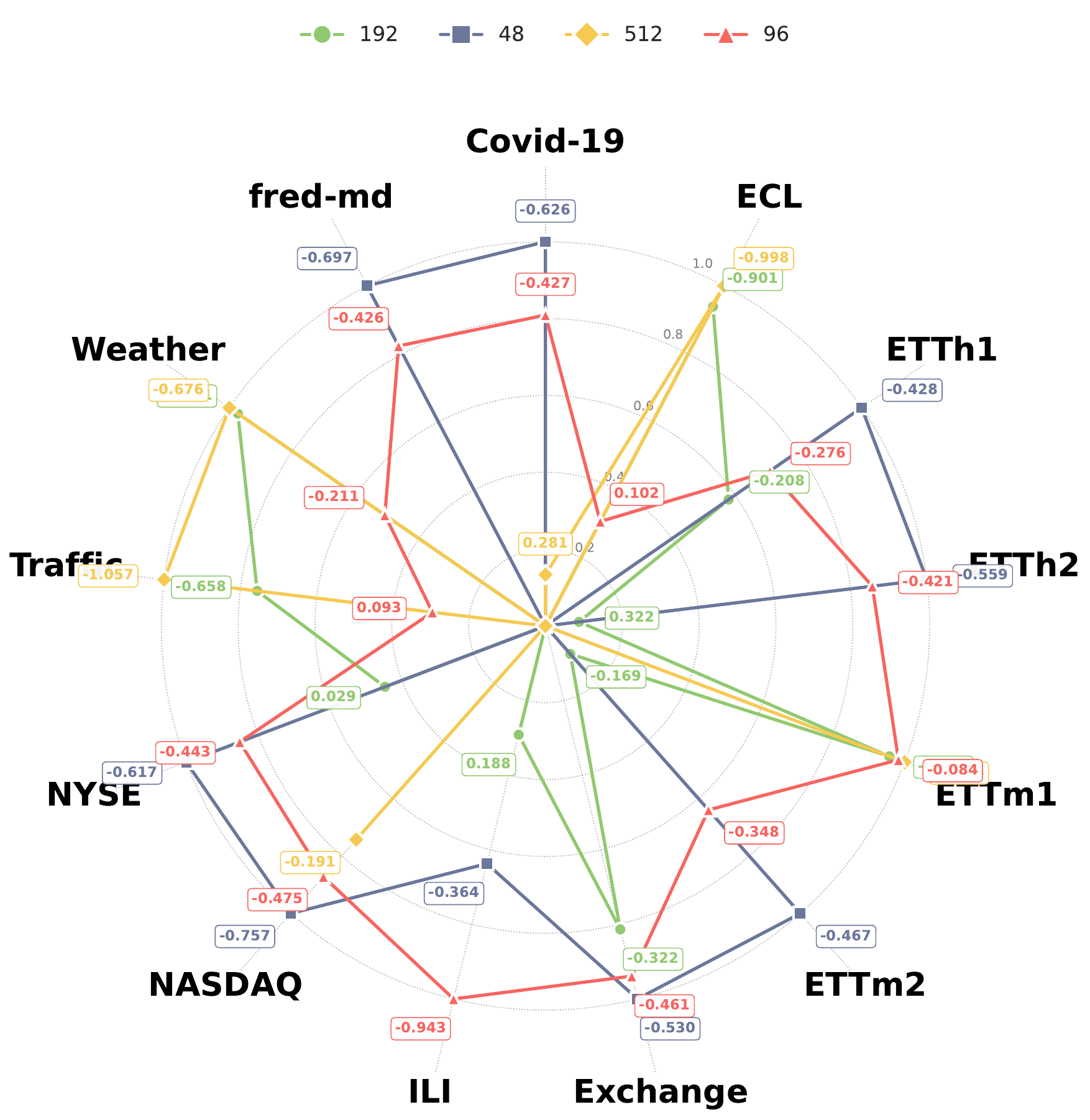}
    \caption{Transformer}
    \label{fig:appx_radar_seq_len_Transformer}
  \end{subfigure}
  \hfill
  \begin{subfigure}[t]{0.16\textwidth}
    \centering
    \includegraphics[width=\textwidth]{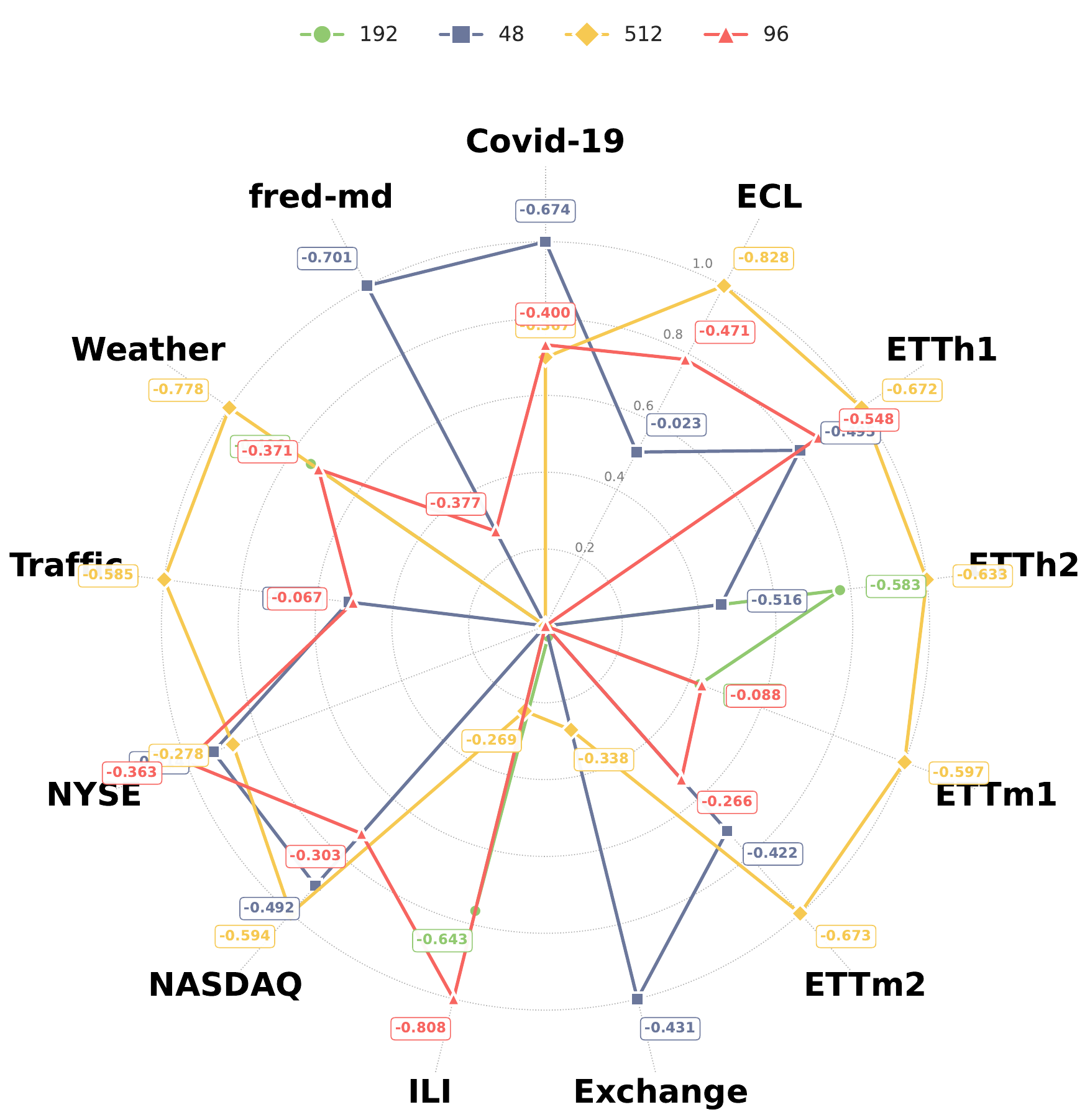}
    \caption{LLM}
    \label{fig:appx_radar_seq_len_LLM}
  \end{subfigure}
  \hfill
  \begin{subfigure}[t]{0.16\textwidth}
    \centering
    \includegraphics[width=\textwidth]{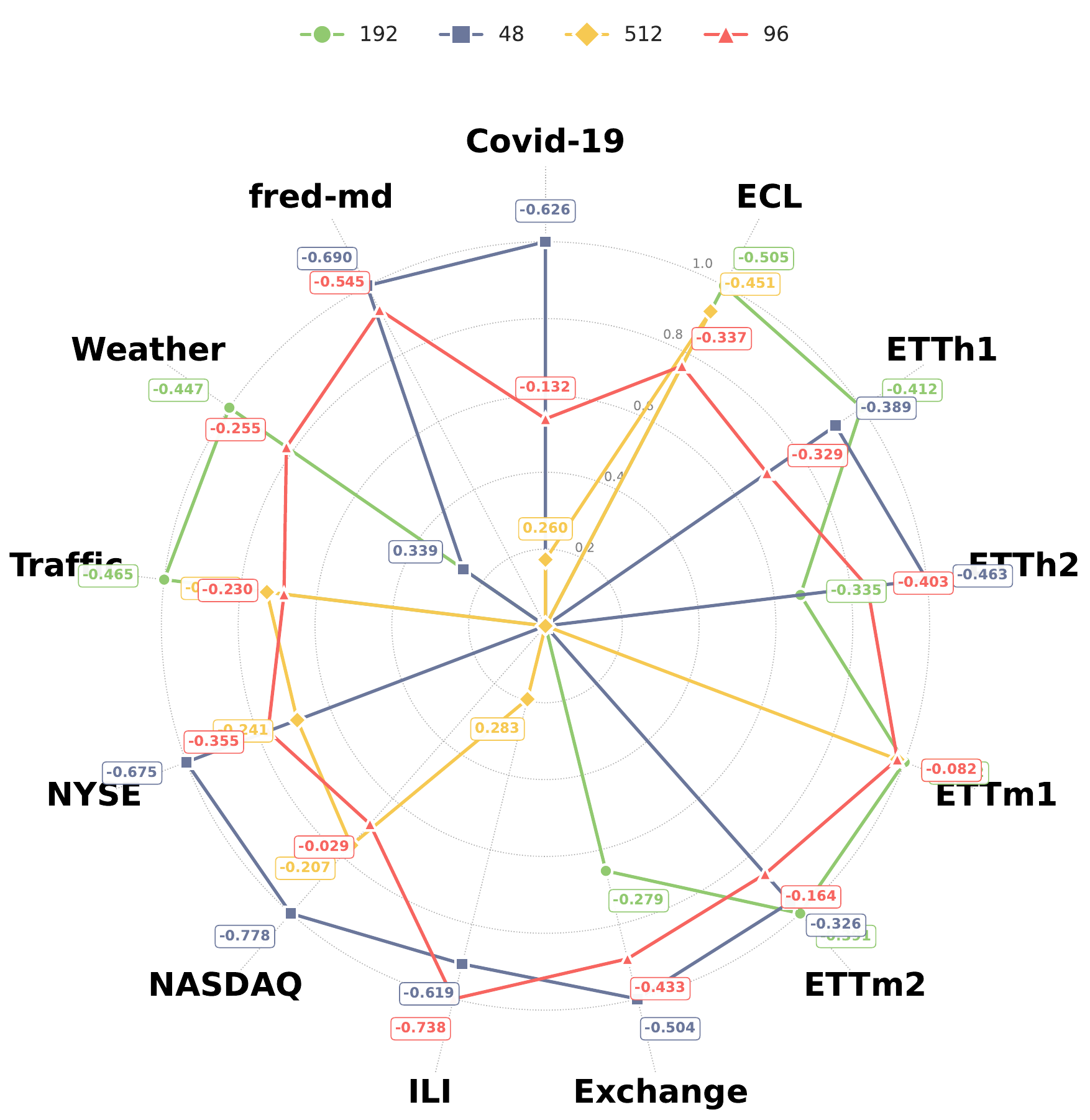}
    \caption{TSFM}
    \label{fig:appx_radar_seq_len_TSFM}
  \end{subfigure}
  \caption{Dataset Adaptability (Radar Charts) for Sequence Length Configurations (Radar Plots). This figure visualizes the performance distributions across different model architectures.}
  \label{fig:appx_radar_seq_len}
\end{figure*}

\begin{figure*}[htbp]
  \centering
  \begin{subfigure}[t]{0.16\textwidth}
    \centering
    \includegraphics[width=\textwidth]{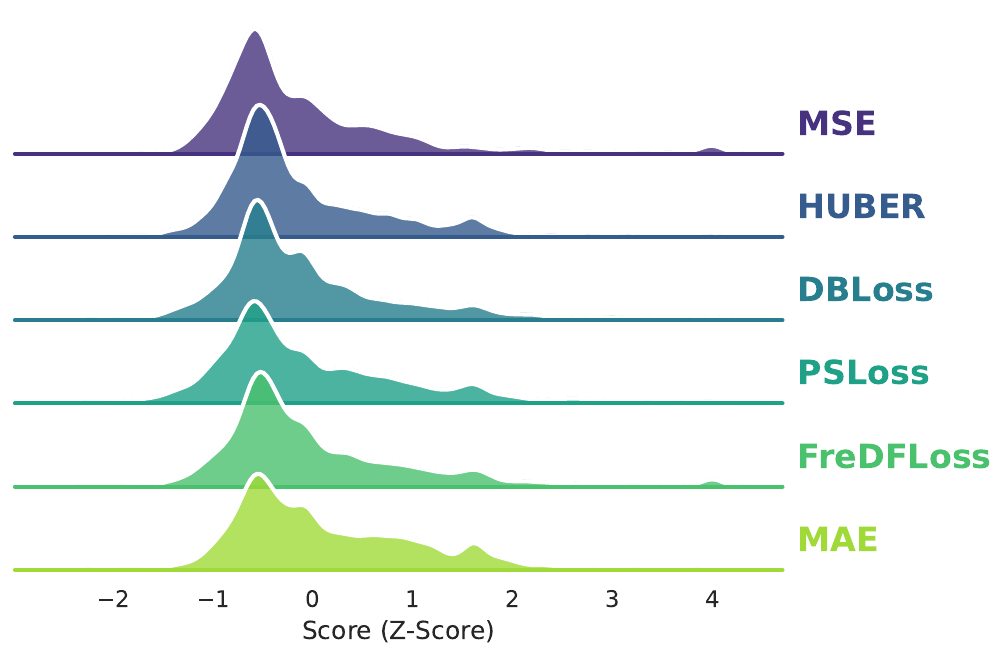}
    \caption{Global}
    \label{fig:appx_dist_loss_func_Global}
  \end{subfigure}
  \hfill
  \begin{subfigure}[t]{0.16\textwidth}
    \centering
    \includegraphics[width=\textwidth]{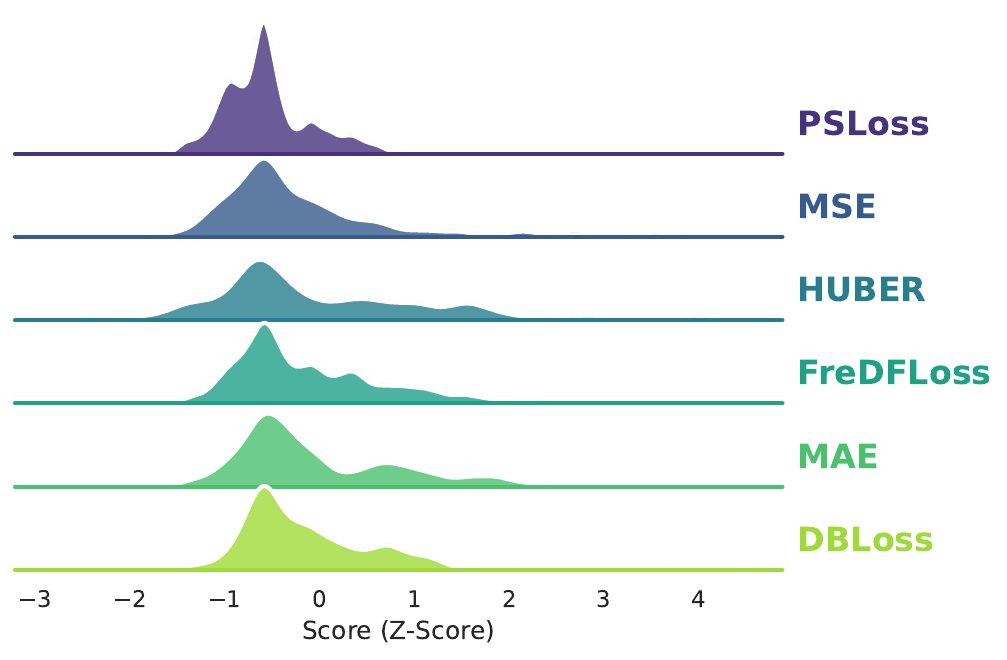}
    \caption{MLP}
    \label{fig:appx_dist_loss_func_MLP}
  \end{subfigure}
  \hfill
  \begin{subfigure}[t]{0.16\textwidth}
    \centering
    \includegraphics[width=\textwidth]{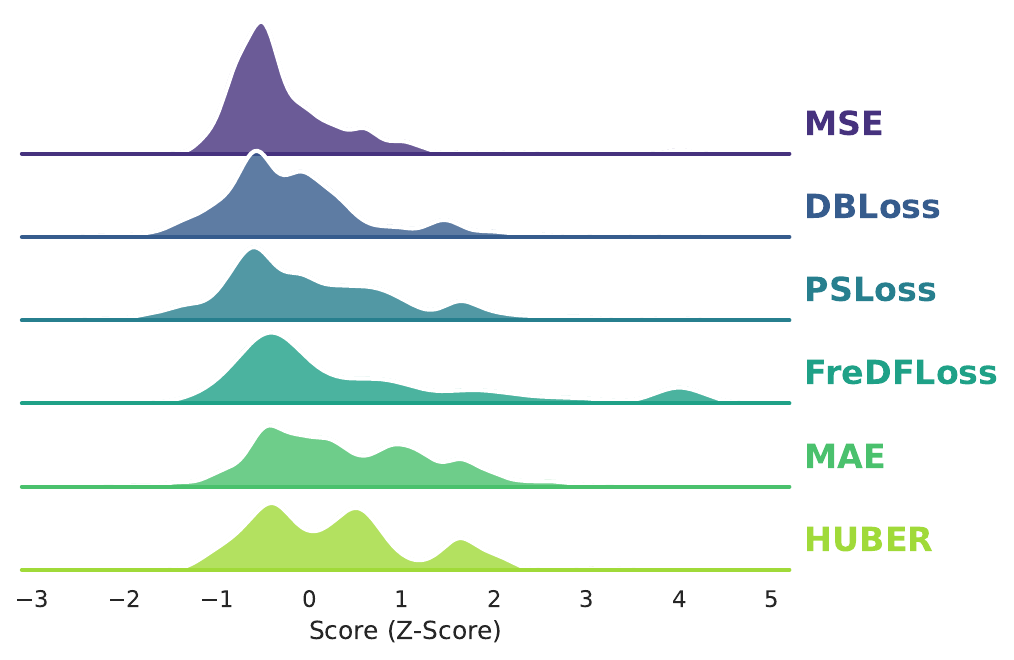}
    \caption{RNN}
    \label{fig:appx_dist_loss_func_RNN}
  \end{subfigure}
  \begin{subfigure}[t]{0.16\textwidth}
    \centering
    \includegraphics[width=\textwidth]{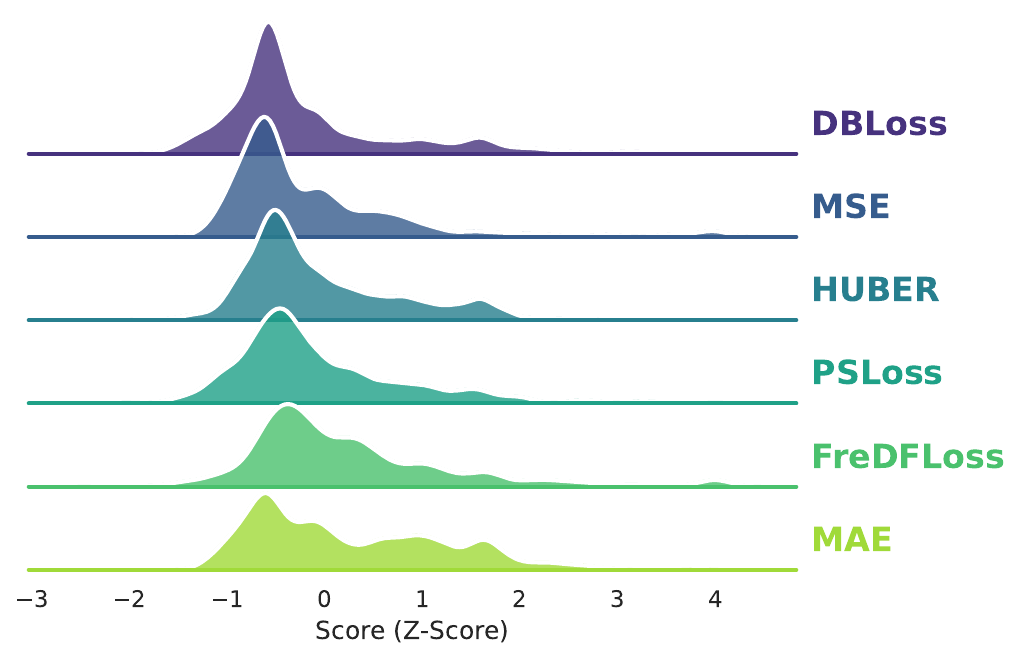}
    \caption{Transformer}
    \label{fig:appx_dist_loss_func_Transformer}
  \end{subfigure}
  \hfill
  \begin{subfigure}[t]{0.16\textwidth}
    \centering
    \includegraphics[width=\textwidth]{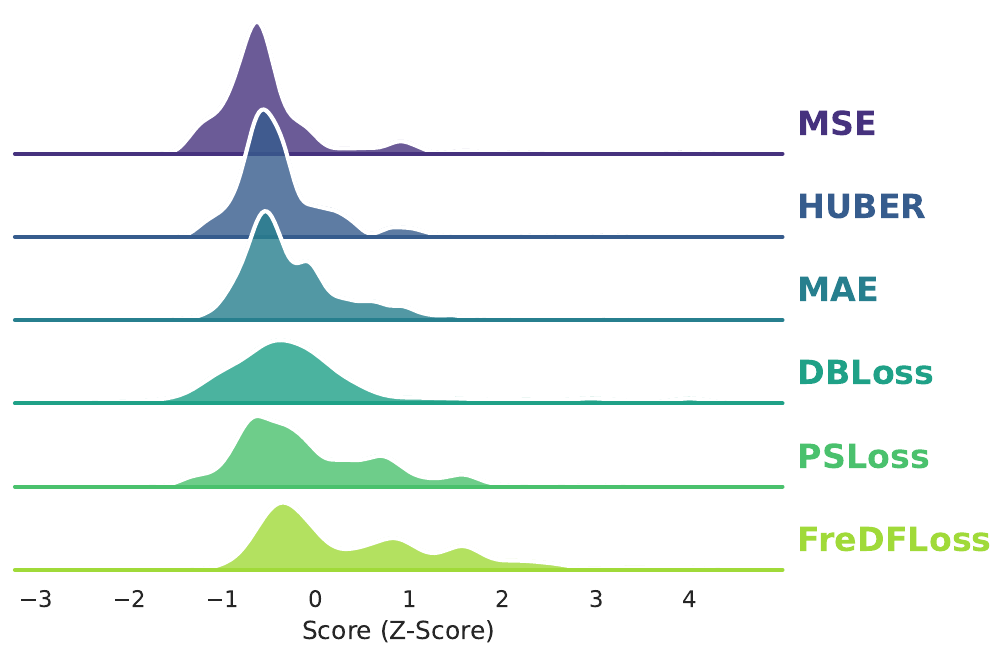}
    \caption{LLM}
    \label{fig:appx_dist_loss_func_LLM}
  \end{subfigure}
  \hfill
  \begin{subfigure}[t]{0.16\textwidth}
    \centering
    \includegraphics[width=\textwidth]{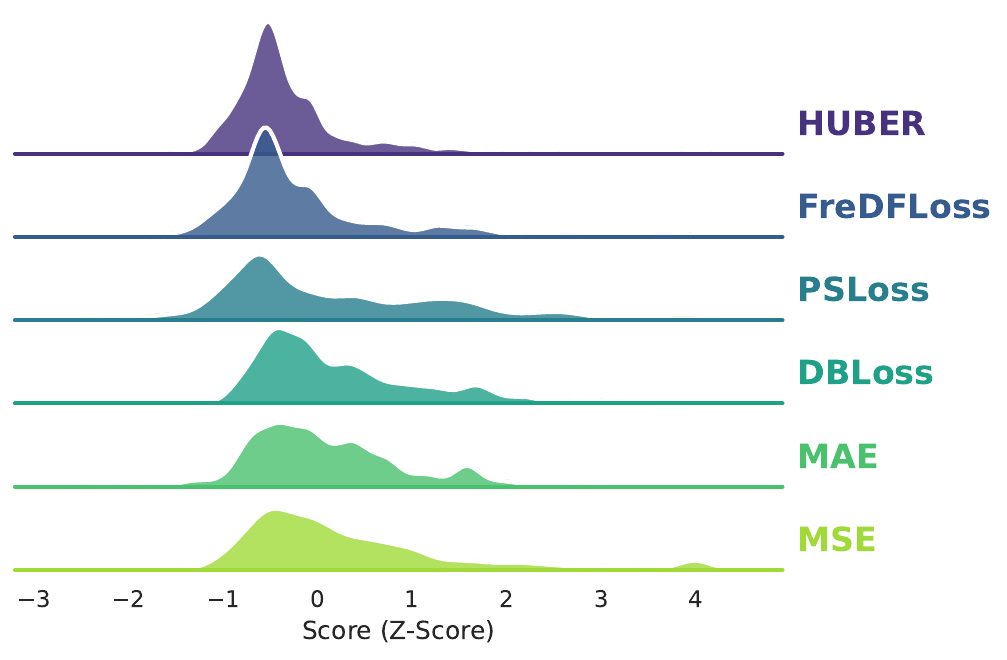}
    \caption{TSFM}
    \label{fig:appx_dist_loss_func_TSFM}
  \end{subfigure}
  \caption{Performance Distributions for Loss Functions (Ridgeline Plots). This figure visualizes the performance distributions across different model architectures.}
  \label{fig:appx_dist_loss_func}
\end{figure*}

\begin{figure*}[htbp]
  \centering
  \begin{subfigure}[t]{0.16\textwidth}
    \centering
    \includegraphics[width=\textwidth]{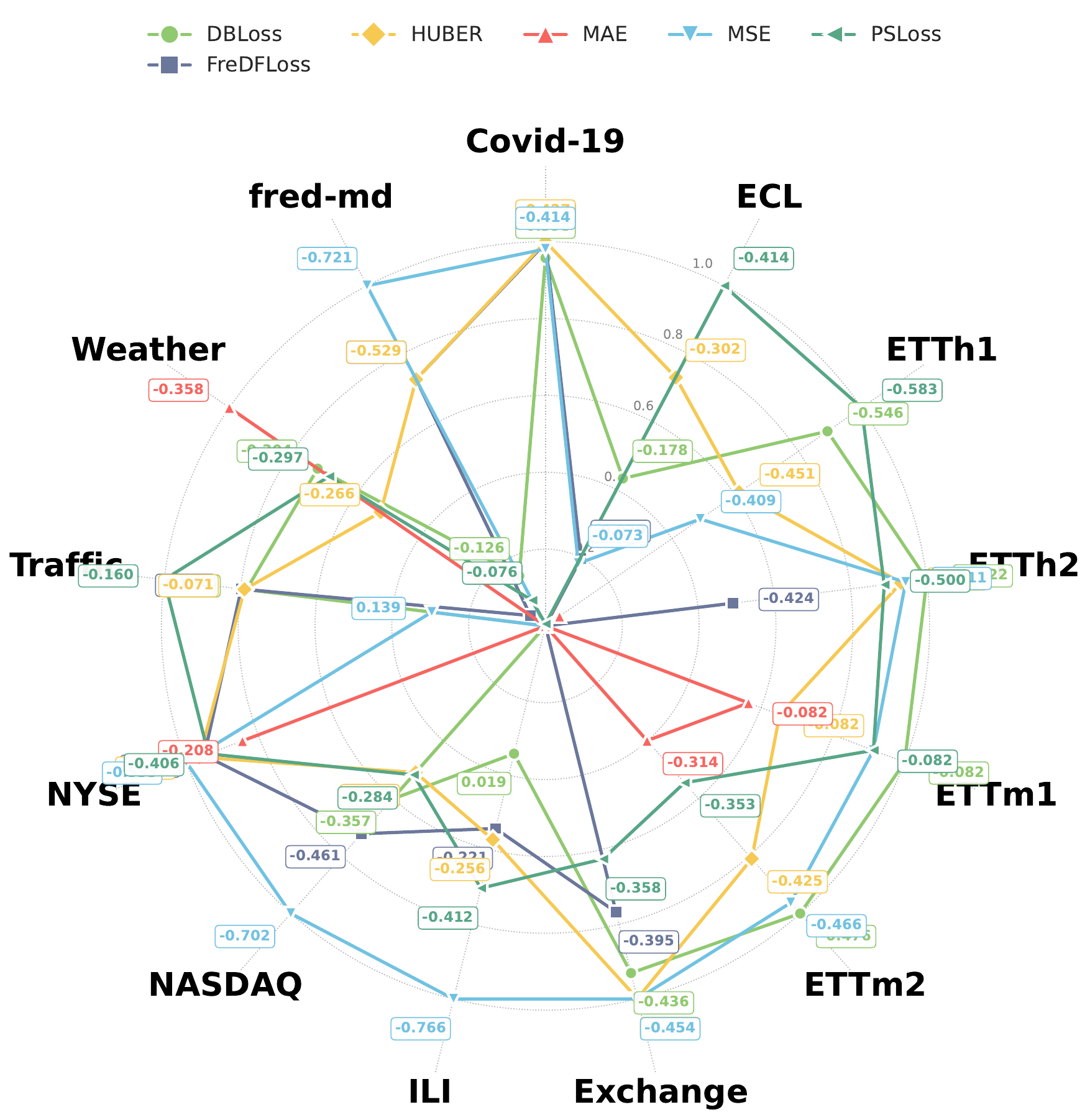}
    \caption{Global}
    \label{fig:appx_radar_loss_func_Global}
  \end{subfigure}
  \hfill
  \begin{subfigure}[t]{0.16\textwidth}
    \centering
    \includegraphics[width=\textwidth]{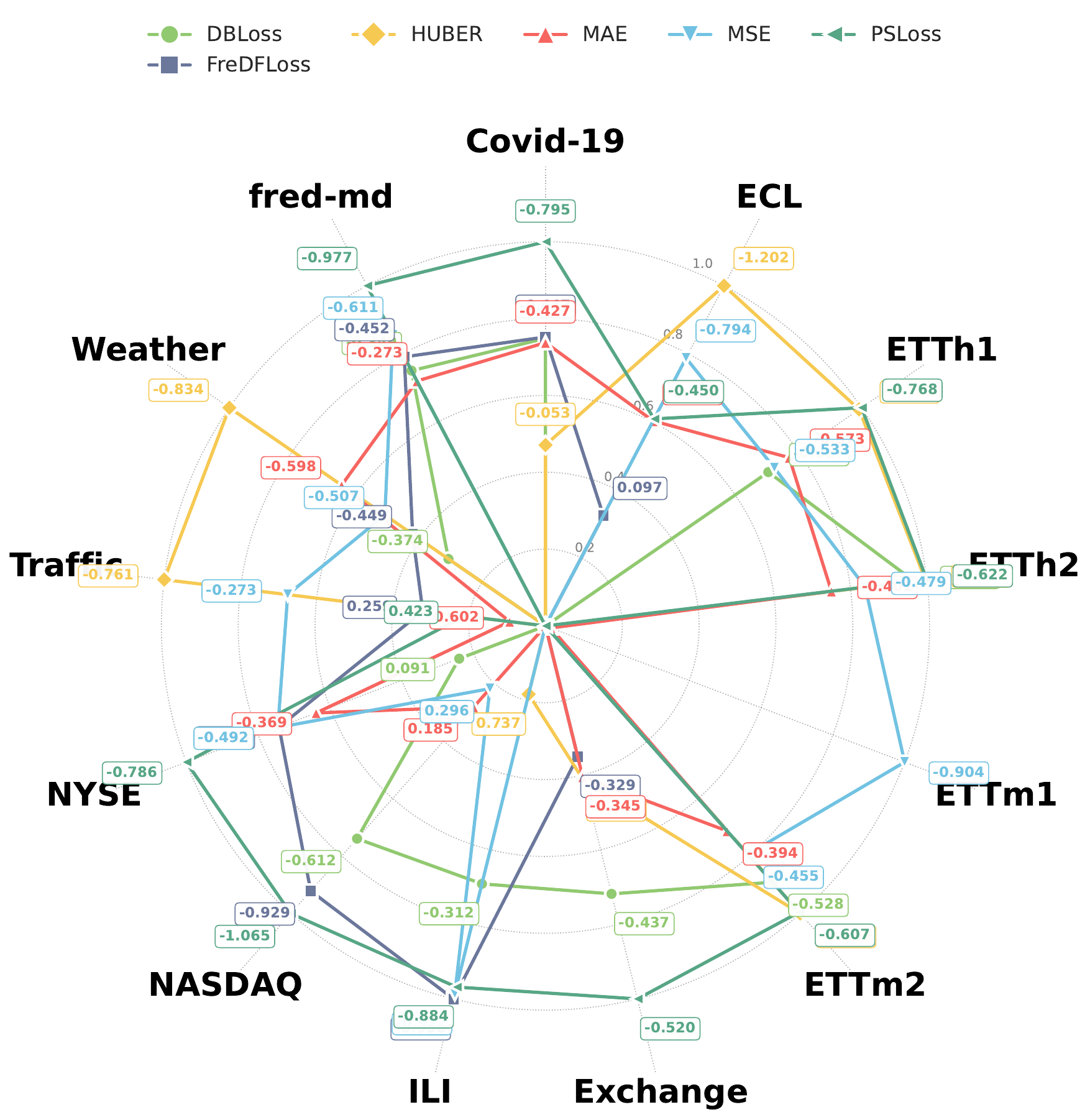}
    \caption{MLP}
    \label{fig:appx_radar_loss_func_MLP}
  \end{subfigure}
  \hfill
  \begin{subfigure}[t]{0.16\textwidth}
    \centering
    \includegraphics[width=\textwidth]{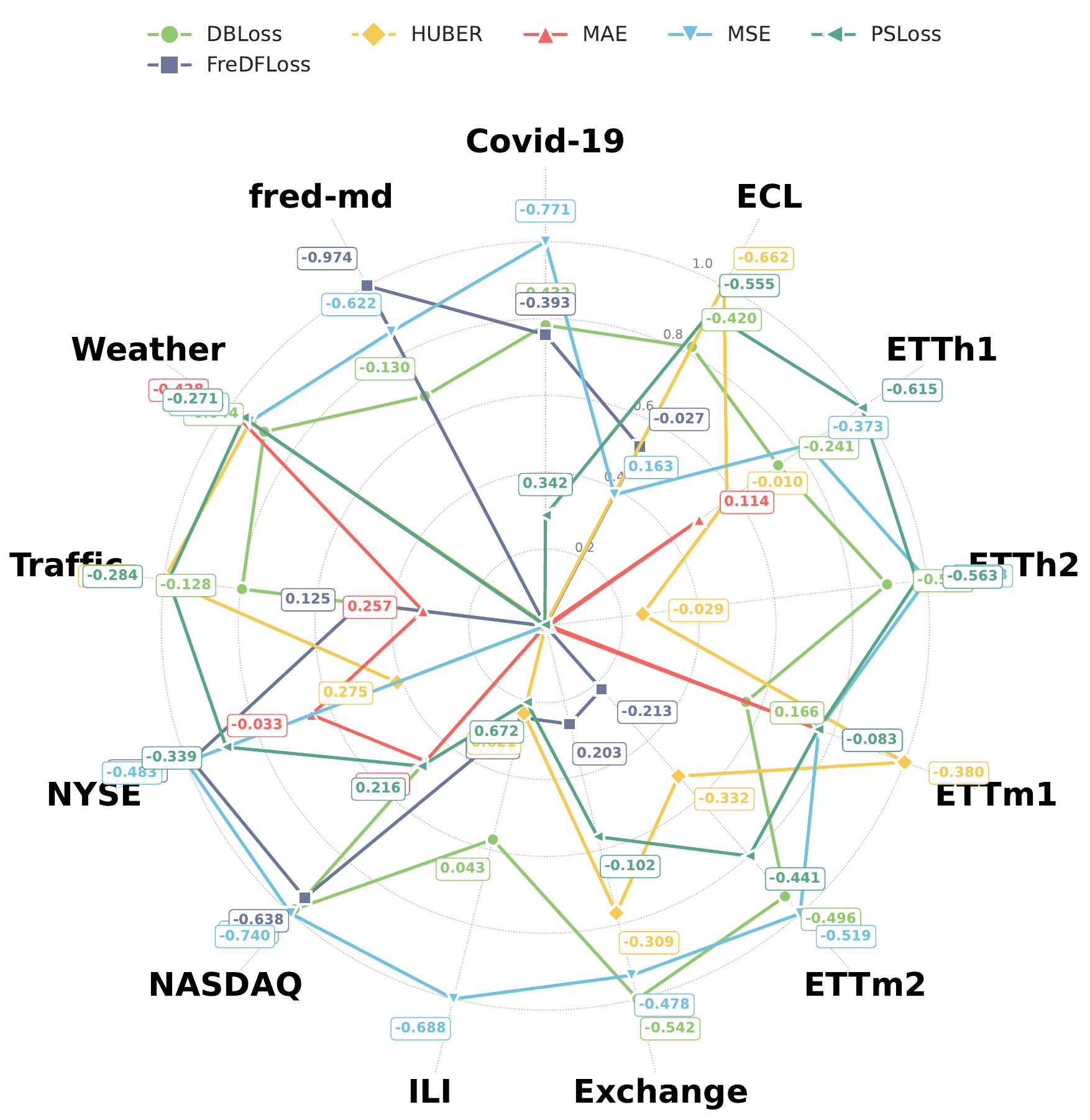}
    \caption{RNN}
    \label{fig:appx_radar_loss_func_RNN}
  \end{subfigure}
  \begin{subfigure}[t]{0.16\textwidth}
    \centering
    \includegraphics[width=\textwidth]{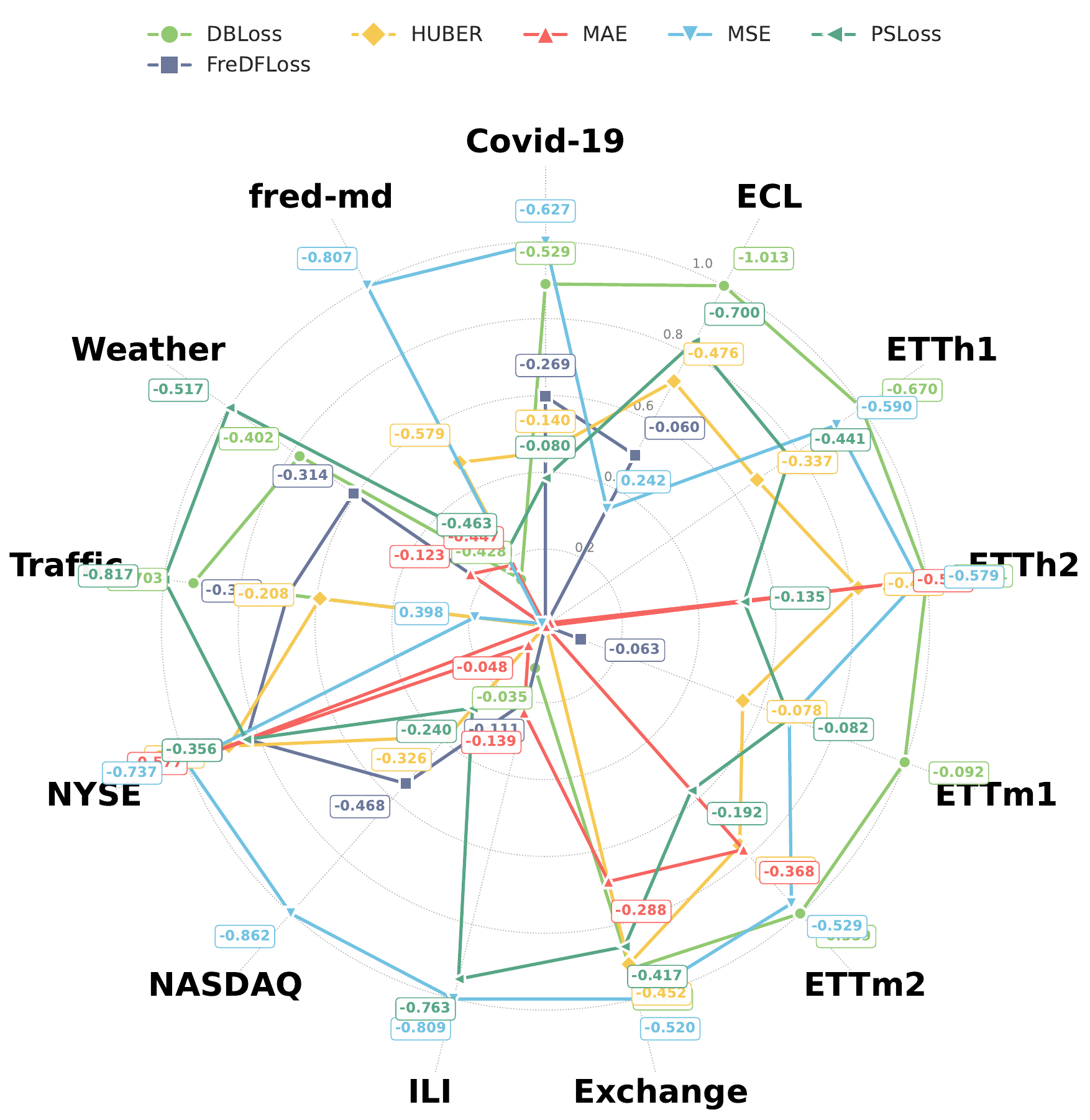}
    \caption{Transformer}
    \label{fig:appx_radar_loss_func_Transformer}
  \end{subfigure}
  \hfill
  \begin{subfigure}[t]{0.16\textwidth}
    \centering
    \includegraphics[width=\textwidth]{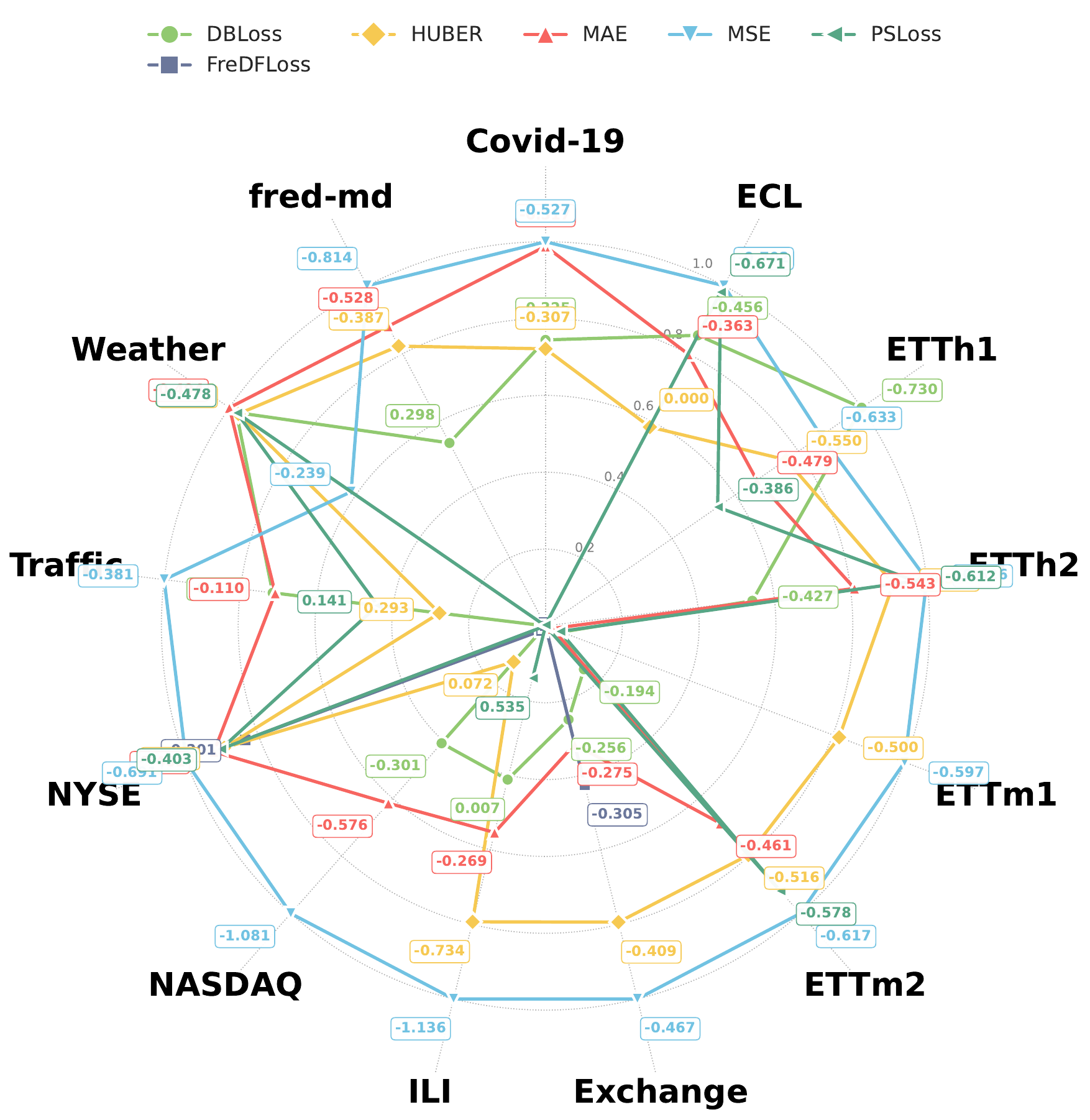}
    \caption{LLM}
    \label{fig:appx_radar_loss_func_LLM}
  \end{subfigure}
  \hfill
  \begin{subfigure}[t]{0.16\textwidth}
    \centering
    \includegraphics[width=\textwidth]{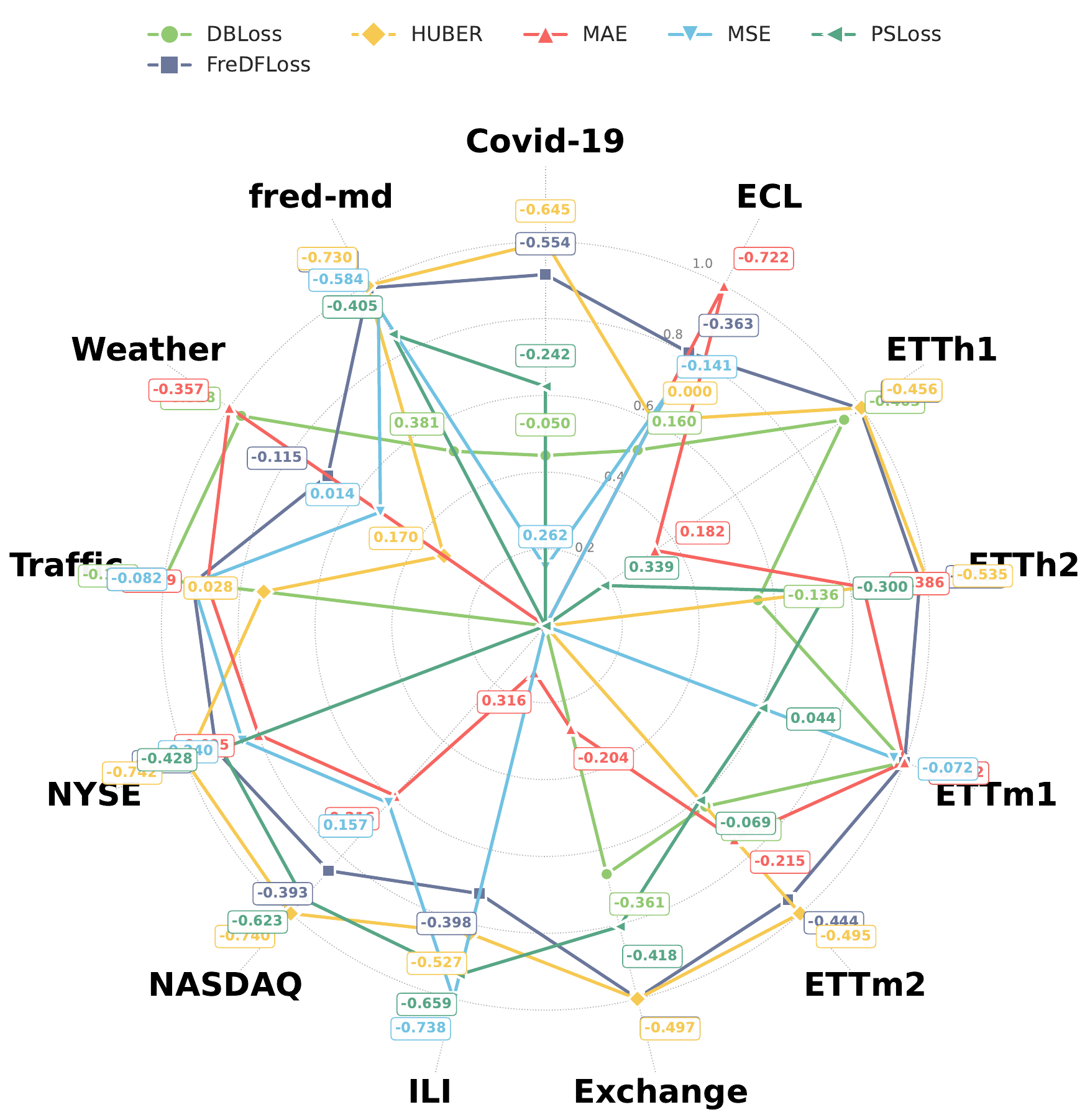}
    \caption{TSFM}
    \label{fig:appx_radar_loss_func_TSFM}
  \end{subfigure}
  \caption{Dataset Adaptability (Radar Charts) for Loss Functions (Radar Plots). This figure visualizes the performance distributions across different model architectures.}
  \label{fig:appx_radar_loss_func}
\end{figure*}

We evaluate Sequence Length Configurations (Fig.~\ref{fig:appx_dist_seq_len} and Fig.~\ref{fig:appx_radar_seq_len}) and Loss Functions (Fig.~\ref{fig:appx_dist_loss_func} and Fig.~\ref{fig:appx_radar_loss_func}).

